%% file: aos-SIMPLE-RC.tex
\documentclass[aos]{imsart}

\RequirePackage{amsthm,amsmath,amsfonts,amssymb,mathrsfs}
\RequirePackage[numbers]{natbib}
\RequirePackage{graphicx}%% uncomment this for including figures
\RequirePackage{booktabs}
\usepackage{hyperref}
\usepackage{bm}

\startlocaldefs

\input{macro_fan}

\newcommand{\var}{\mathrm{var}}

\newcommand{\cov}{\mathrm{cov}}

\newcommand{\tr}{\mathrm{tr}}
\newcommand{\diag}{\mathrm{diag}}

\newcommand{\sgn}{\mathrm{sgn}}
\newcommand{\supp}{\mathrm{supp}}

\newcommand{\ba}{\mbox{\bf a}}

\newcommand{\be}{\mbox{\bf e}}

\newcommand{\bu}{{\mathbf u}}
\newcommand{\bv}{{\mathbf v}}
\newcommand{\bw}{\mbox{\bf w}}
\newcommand{\bx}{\mbox{\bf x}}
\newcommand{\by}{\mbox{\bf y}}

\newcommand{\bA}{\mbox{\bf A}}
\newcommand{\bB}{\mbox{\bf B}}

\newcommand{\bD}{\mbox{\bf D}}

\newcommand{\bG}{\mbox{\bf G}}
\newcommand{\bH}{\mbox{\bf H}}
\newcommand{\bI}{\mbox{\bf I}}

\newcommand{\bR}{\mbox{\bf R}}
\newcommand{\bS}{\mbox{\bf S}}
\newcommand{\bP}{\mbox{\bf P}}
\newcommand{\bU}{\mbox{\bf U}}

\newcommand{\bV}{\mbox{\bf V}}
\newcommand{\bW}{\mbox{\bf W}}
\newcommand{\bX}{\mbox{\bf X}}

\newcommand{\bY}{\mbox{\bf Y}}

\newcommand{\bpi}{\mbox{\boldmath $\pi$}}

\newcommand{\bDelta}{\mbox{\boldmath $\Delta$}}

\newcommand{\bTheta}{\mbox{\boldmath $\Theta$}}

\newcommand{\bSig}{\mbox{\boldmath $\Sigma$}}

\newcommand{\bPi}{\mbox{\boldmath $\Pi$}}
\newcommand{\bcE}{\mbox{\boldmath $\mathcal E$}}

\newcommand{\vart}{\vartheta}

\newcommand{\wt}{\widetilde}
\newcommand{\wh}{\widehat}

\newcommand{\bbA}{{\bf A}}

\newcommand{\bbD}{{\bf D}}

\newcommand{\bbf}{{\bf f}}

\newcommand{\bbP}{{\bf P}}

\newcommand{\bbH}{{\bf H}}

\newcommand{\bbM}{{\bf M}}

\newcommand{\bbS}{{\bf S}}

\newcommand{\bbT}{{\bf T}}

\newcommand{\bbV}{{\bf V}}
\newcommand{\bbv}{{\bf v}}

\newcommand{\bbW}{{\bf W}}

\newcommand{\bbX}{{\bf X}}

\newcommand{\Pii}{\mbox{\boldmath $\Upsilon$}}
\newcommand{\Pie}{\Upsilon}

\allowdisplaybreaks

\usepackage[draft,inline,nomargin,index]{fixme}
\fxsetup{theme=color,mode=multiuser}
\FXRegisterAuthor{yf}{ayf}{\color{magenta} YF}
\renewcommand{\top}{T}
\renewcommand{\subseteq}{\subset}
\renewcommand{\hat}{\widehat}

\newtheorem{assu}{Condition}

\newtheorem{lemma}{Lemma}%[section]
\newtheorem{proposition}{Proposition}%[section]
\newtheorem{theorem}{Theorem}%[section]
\newtheorem{definition}{Definition}%[section]
\newtheorem{remark}{Remark}%[section]
\newtheorem{claim}{Claim}%[section]
\def\toD{\overset{\mathscr{D}}{\longrightarrow}}

\endlocaldefs

\begin{document}

\begin{frontmatter}
%%%%%%%%%%%%%%%%%%%%%%%%%%%%%%%%%%%%%%%%%%%%%%
%%                                          %%
%% Enter the title of your article here     %%
%%                                          %%
%%%%%%%%%%%%%%%%%%%%%%%%%%%%%%%%%%%%%%%%%%%%%%
\title{SIMPLE-RC: Group Network Inference with Non-Sharp Nulls and Weak Signals}
%\title{A sample article title with some additional note\thanksref{T1}}
\runtitle{SIMPLE-RC}
%\thankstext{T1}{A sample of additional note to the title.}

\begin{aug}
%%%%%%%%%%%%%%%%%%%%%%%%%%%%%%%%%%%%%%%%%%%%%%%
%% Only one address is permitted per author. %%
%% Only division, organization and e-mail is %%
%% included in the address.                  %%
%% Additional information can be included in %%
%% the Acknowledgments section if necessary. %%
%% ORCID can be inserted by command:         %%
%% \orcid{0000-0000-0000-0000}               %%
%%%%%%%%%%%%%%%%%%%%%%%%%%%%%%%%%%%%%%%%%%%%%%%
\author[A]{\fnms{Jianqing}~\snm{Fan}\ead[label=e1]{jqfan@princeton.edu}},
\author[B]{\fnms{Yingying}~\snm{Fan}\ead[label=e2]{fanyingy@marshall.usc.edu}}
\author[B]{\fnms{Jinchi}~\snm{Lv}\ead[label=e3]{jinchilv@marshall.usc.edu}}
\and
\author[C]{\fnms{Fan}~\snm{Yang}\ead[label=e4]{fyangmath@mail.tsinghua.edu.cn}}
%%%%%%%%%%%%%%%%%%%%%%%%%%%%%%%%%%%%%%%%%%%%%%
%% Addresses                                %%
%%%%%%%%%%%%%%%%%%%%%%%%%%%%%%%%%%%%%%%%%%%%%%
\address[A]{Princeton University, Princeton, NJ 08544, USA\printead[presep={,\ }]{e1}}

\address[B]{University of Southern California, Los Angeles, CA 90089, USA\printead[presep={,\ }]{e2,e3}}
\address[C]{Tsinghua University, Beijing 100084, China\printead[presep={,\ }]{e4}}
			\thankstext{A,B,C}{This work was supported by NSF Grants DMS-1953356, EF-2125142,  DMS-2052926, and DMS-2052964. Co-corresponding authors: Yingying Fan (fanyingy@marshall.usc.edu) and Fan Yang (fyangmath@mail.tsinghua.edu.cn) }
\end{aug}

\begin{abstract}
Large-scale network inference with uncertainty quantification has important applications in natural, social, and medical sciences. The recent work of Fan, Fan, Han and Lv (2022) introduced a general framework of statistical inference on membership profiles in large networks (SIMPLE) for testing the sharp null hypothesis that a pair of given nodes share the same membership profiles. In real applications, there are often groups of nodes under investigation that may share similar membership profiles at the presence of relatively weaker signals than the setting considered in SIMPLE. To address these practical challenges, in this paper we propose a SIMPLE method with random coupling (SIMPLE-RC) for testing the non-sharp null hypothesis that a group of given nodes share similar (not necessarily identical) membership profiles under weaker signals. Utilizing the idea of random coupling, we construct our test as the maximum of the SIMPLE tests for subsampled node pairs from the group. Such technique reduces significantly the correlation among individual SIMPLE tests while largely maintaining the power, enabling delicate analysis on the asymptotic distributions of the SIMPLE-RC test. Our method and theory cover both the cases with and without node degree heterogeneity. %Under the mixed membership models without degree heterogeneity, we establish a simple-to-use asymptotic null Gumbel distribution for the SIMPLE-RC test and a formal power analysis. We further extend the SIMPLE-RC method and theory to the degree-corrected mixed membership models for accommodating the practical issue of degree heterogeneity. 
These new theoretical developments are empowered by a second-order expansion of spiked eigenvectors under the $\ell_\infty$-norm, built upon our work for random matrices with weak spikes. Our theoretical results and the practical advantages of the newly suggested method are demonstrated through several simulation and real data examples.
\end{abstract}

\begin{keyword}[class=MSC]
\kwd[Primary ]{62F03}
%\kwd{Network inference}
\kwd[; secondary ]{62F05}
\end{keyword}

\begin{keyword}
\kwd{Large-scale network inference}
\kwd{SIMPLE}
\kwd{Group testing}
\kwd{Random matrix theory}
\kwd{Random coupling}
\end{keyword}

\end{frontmatter}
%%%%%%%%%%%%%%%%%%%%%%%%%%%%%%%%%%%%%%%%%%%%%%
%% Please use \tableofcontents for articles %%
%% with 50 pages and more                   %%
%%%%%%%%%%%%%%%%%%%%%%%%%%%%%%%%%%%%%%%%%%%%%%
%\tableofcontents

%%%%%%%%%%%%%%%%%%%%%%%%%%%%%%%%%%%%%%%%%%%%%%
%%%% Main text entry area:

%\newpage
\section{Introduction} \label{Sec.intro}

Statistical estimation and inference of large-scale network data have been important topics in statistics and related fields. Many existing works in the statistics literature have focused on learning the global structure of the networks, such as the community detection \citep{Abbe2018, Jin2015, Le2016,  Lei2015, Rohe2011}, rank estimation and inference \citep{airoldi2008,ba2017,gao2017,2019J,LeLevina2022,Mc2013}, and network parameter estimation \citep{Bickel2011,Levin2019,li2020network, ShanLevina2022}.  We refer to such works as \textit{global} network inference throughout the paper for the simplicity of presentation. Recently, \cite{SIMPLE} proposed a hypothesis testing approach to inferring the similarity of the membership profiles for a pre-selected pair of nodes in large networks, a brand new SIMPLE framework for \textit{local} network inference. Despite the innovative approach in \cite{SIMPLE}, their work has several key limitations, including the relatively strong signal-to-noise ratio and its narrow focus on testing only the sharp null hypothesis for a pair of pre-selected nodes. Our work aims at overcoming such limitations and extending the framework to broader settings with much weaker signal strength assumption and non-sharp nulls among a group of nodes. Another important goal of our study is to generalize some random matrix theory (RMT) results to the network setting so that they are more applicable to such practical applications.   

In this paper, we propose and investigate the framework of \emph{statistical inference on membership profiles in large networks with random coupling}, %abbreviated 
named as the SIMPLE-RC, for testing the closeness of the membership profiles for a group of pre-selected nodes. Here, the group size $m$ can diverge with the network size $n$. We study the problem under the broad setting of the degree-corrected mixed membership model, which allows for both heterogeneous node degrees and mixed memberships of nodes. The adjacency matrix, denoted as $\bX \in \mathbb R^{n\times n}$, under such model setting can be written as a deterministic low-rank mean matrix $\bbH$ plus a noise random matrix $\bbW$, where the eigenvalues and eigenvectors of the mean matrix $\bbH$ record the complete community membership information including the node degrees and membership profiles. For the ease of presentation, we refer to the nonzero eigenvalues and the corresponding eigenvectors of matrix $\bH$ as the spiked eigenvalues and spiked eigenvectors, respectively\footnote{We slightly misuse the term ``spiked" here because our theoretical results do not need the smallest nonzero eigenvalues of the mean matrix $\bH$ to be larger than the eigenvalues of the noise matrix.}. 

Motivated by the SIMPLE test for a pair of pre-selected nodes in \cite{SIMPLE}, we form our group test using the similar idea of pairwise eigenvector contrasting, that is, if two nodes have similar membership profiles, their corresponding (appropriately weighted) eigenvectors are also close. A natural idea is to construct the group test as the maximum of all pairwise SIMPLE tests formed from all node pairs in the group. However, this naive approach fails to work because of the  high correlations among the individual SIMPLE tests; such dependency makes it highly challenging to analyze the limiting null distribution of such a group test. Indeed, it is largely unclear whether the limiting distribution even exists when the correlation level is high. To overcome such difficulty, we propose the method of random coupling, which randomly couples one node with another without  multiple partners. We then form our SIMPLE-RC test as the maximum of the individual SIMPLE tests resulting from the random coupling procedure. This strategy reduces greatly the correlations among the individual SIMPLE tests used in constructing the group test, enabling us to derive its asymptotic null distribution. We show that under the null hypothesis that all nodes in the group have similar membership profiles, the distribution of the SIMPLE-RC test statistic converges asymptotically to the Gumbel distribution after an appropriate centering and rescaling. This result allows us to construct a rejection region with a valid asymptotic size.  

To facilitate a formal power analysis of the SIMPLE-RC test, we introduce a new measure of the closeness of pairwise node membership profiles. We formulate our alternative hypothesis using such  a measure. We note that our null hypothesis, although  not directly formulated using this measure, imposes an upper bound on the node closeness under such a measure. We show that under some regularity conditions and the alternative hypothesis, the SIMPLE-RC test statistic asymptotically diverges in probability, and hence achieves a high power. 

Compared to the original SIMPLE framework, our method of SIMPLE-RC has both important methodological and technical innovations. Instead of constructing a test using all the spiked eigenvectors (i.e., the empirical counterparts of those corresponding to the nonzero eigenvalues of the mean matrix $\bH$), the SIMPLE-RC advocates the use of only those that correspond to large enough spiked eigenvalues in magnitude. At a high level, this is related to the idea of eigen-selection, whose importance in high-dimensional clustering was revealed recently in \cite{HanTongFan2022}.  An added advantage of using only a subset of spiked eigenvectors is that our method and theory can accommodate a diverging number of network communities, denoted as $K$, without imposing any assumptions on the smallest nonzero eigenvalues of the mean matrix $\bH$. In other words, it is allowed in our framework that the smallest nonzero eigenvalues of the mean matrix $\bH$ fall below the noise level. 
%It is well-known in network analyses and random matrix theory that the difficulty in certain global network inference such as parameter estimation can be measured by the magnitudes of  the spiked eigenvalues  relative to the non-spiked ones (from the noise matrix $\bW$). %Our study suggests an interesting difference between certain global inference and our local inference on the required magnitudes of spiked eigenvalues---our local inference only needs conditions on the larger eigenvalues and their eigenvectors while some global inference imposes constraints on all spiked eigenvalues and eigenvectors. 
In fact, most of existing works in network analysis advocate the use of all the spiked eigenvalues and their eigenvectors corresponding to the mean matrix $\bH$, while our study reveals an interesting phenomenon that this may not be necessary, at least in \textit{local} network inference studied in the current paper. More formal and systematic study on when and how eigenvalue/eigenvector selection can be beneficial deserves a separate paper. 
%More formal comparisons between the global and local inferences including their detection boundaries are beyond the scope of our study and deserve a separate paper.  

The idea of using only a subset of spiked eigenvalues and eigenvectors naturally raises the question of which eigenvalues (and eigenvectors) one should use in constructing the SIMPLE-RC test. We adopt the same idea of eigenvalue thresholding as in \cite{SIMPLE} to decide on how many spiked eigenvalues to use.  Despite of using the identical estimator, we remark that the underlying mechanism is very different.  Indeed, we do not require that $\widehat K_0$, the estimated number of spiked eigenvectors, consistently estimates any deterministic population parameter. In fact, as long as $\widehat K_0$ satisfies some upper bound requirement almost surely, we prove that the SIMPLE-RC test statistic constructed with the $\widehat K_0$ spiked eigenvalues and their eigenvectors converges weakly to the Gumbel distribution asymptotically under the null hypothesis, ensuring easy and tuning-free application of our test with theoretical guarantees on the valid asymptotic size. We regard this as a major advantage of the SIMPLE-RC test. It is also worth mentioning that our theory can accommodate sparse networks whose average node degree is of higher order than $(\log n)^8$. Such result may be further improved at the cost of more tedious analyses, but we do not pursue such direction in this paper.

Our theoretical analysis is empowered by a new random matrix theory built in this paper. These new RMT results can be of independent interest to the statistics community. Our key theoretical result, Theorem \ref{asymp_evector},  in the Supplementary Material gives an asymptotic expansion for the empirical spiked eigenvectors of the random adjacency matrix. This result improves the corresponding ones in \cite{ATE} (i.e., their Theorem 2) and \cite{SIMPLE} (i.e., their Lemma 9) in the following important perspectives. In the current paper, we deal with sparser networks, weaker signal strengths, and possibly diverging number of communities. 
A key step in deriving the asymptotic expansion of the empirical spiked eigenvectors is deriving a sharper anisotropic local law for the resolvent of a general Wigner-type random matrix under weaker conditions on both the sparsity level (i.e., average node degree slightly larger than $(\log n)^8$) and the signal strengths (in terms of magnitude of the spiked eigenvalues). Our improved results are due to the use of the method developed in the recent RMT literature \citep{AjaErdKru2015,EKYY_ER1}. We refer readers to Section \ref{new.Sec4} for more details. 

The rest of the paper is organized as follows. Section \ref{Sec2.1} introduces the model setting and the non-sharp null hypothesis involving a group of network nodes. We suggest the framework of SIMPLE-RC for group network inference and present the asymptotic theory in Section \ref{new.Sec3}. Section \ref{new.Sec4} discusses some basic ideas in establishing 
the asymptotic expansion of empirical spiked eigenvectors and some related works. We present several simulation and real data examples in Sections \ref{new.Sec5} and \ref{new.Sec6}. %Section \ref{new.Sec7} discusses some implications and extensions of our work. 
All the  RMT results, proofs, and technical details are provided in the Supplementary Material.

%\newpage
\section{Model setting} \label{Sec2.1}

%We provide in this section the specific model setting for the problem of group network inference mentioned in the Introduction. Let us consider an undirected graph $\mathcal{N} = (V, E)$ with $n$ nodes, where $V = \{1,\cdots, n\}$ denotes the set of nodes and $E$ represents the set of links. 

Consider an undirected graph $\mathcal{N} = (V, E)$ with $n$ nodes, where $V = \{1,\cdots, n\}$ denotes the set of nodes and $E$ represents the set of links. For simplicity, we will use the shorthand notation $[n] = \{1,\cdots, n\}$ throughout the paper. Denote by $\bbX= (x_{ij})\in \mathbb{R}^{n\times n}$ the symmetric adjacency matrix representing the connectivity structure of graph $\mathcal N$, where $x_{ij}=1$ and 0 corresponds to the existence and lack of a link connecting nodes $i$ and $j$, respectively. To accommodate different practical applications, we will consider the general case when graph $\mathcal N$ may or may not admit self loops, where $x_{ii}=0$ over all $i\in[n]$ for the latter scenario. Under a probabilistic model, we will assume that the observed adjacency matrix $\bbX$ is an independent realization of a Bernoulli random matrix with independent entries modulo the symmetry.

Specifically, to model the connectivity pattern of the graph $\mathcal N$, let us consider a symmetric binary random matrix $\bX^*$ with the latent structure
\begin{equation} \label{eq: model}
	\bbX^*=\bbH+\bbW^*,
\end{equation}
where $\bbH=(h_{ij})\in \mathbb{R}^{n\times n}$ denotes the deterministic mean matrix (i.e., probability matrix) of low rank $K \geq 1$ (see (\ref{new.DCMM2}) later for a specification) and $\bbW^*=(w_{ij}^*)\in \mathbb{R}^{n\times n}$ is a symmetric noise random matrix with mean zero and independent entries on and above the diagonal. Assume that the observed adjacency matrix $\bbX$ is either $\bbX^*$ or $\bX^* - \diag(\bbX^*)$, corresponding to the cases with or without self loops, respectively. In view of (\ref{eq: model}), for either case we have the  decomposition for the adjacency matrix $\bbX$ given by
\begin{equation} \label{eq: model.general}
	\bX = \bH + \bbW,	
\end{equation}
where $\bbW = \bbW^*$ in the presence of self loops and $\bbW = \bbW^* - \diag(\bbX^*)$ in the absence of self loops. We see that for either case, matrix $\bW$ in the general model (\ref{eq: model.general}) above is symmetric with independent entries on and above the diagonal. Our study will cover both cases with and without self loops. To simplify the presentation, we will slightly abuse the notation by still referring to $\bbH$ as the mean matrix and $\bbW$ as the noise matrix hereafter.

For the model setting given in (\ref{eq: model.general}), we assume that there is an underlying latent community structure that the network $\mathcal N$ can be decomposed into $K$ disjoint latent communities
%\begin{equation*}
$\mathcal{C}_1, \cdots,  \mathcal{C}_K,$
%\end{equation*}
where each node $i$ has a community membership probability vector $\bpi_i = (\bpi_i(1),\cdots, \bpi_i(K))^T \in \mathbb{R}^K$  such that%within each community $V^{(i)}$ share some common characteristics.
\begin{equation} \label{new.eq004}
\mathbb{P} \left\{ \text{node $i$ belongs to community $\mathcal{C}_k$} \right\} = \bpi_i(k)
\end{equation}
for each $1 \leq k \leq K$. 
%This follows automatically that
%\begin{equation}\label{eq: mixed}
%\bpi_i(k)\in [0,1], \ \sum_{k=1}^K\bpi_i(k)=1.
%\end{equation} 
Throughout the paper, we assume that the number of latent communities $K$ is \textit{unknown} but finite or slowly diverging (e.g., with $K\sim (\log n)^c$ for some fixed constant $c>0$, where $\sim$ stands for the asymptotic order).

For a group of given network nodes  $\mathcal{M} \subset [ n ]$, we are interested in testing whether they share similar (but not necessarily identical) membership profiles. %, that is, whether their community membership probability vectors are close to each other despite possible discrepancies. 
Without loss of generality, let us assume that the group is given by $\mathcal M = \{1,\cdots, m\}$ for some $1 \leq m \leq n$.  We will focus  on the more interesting case when the group size $m = |\mathcal M|$ can diverge with the network size $n$. Our goal is to test the null hypothesis
\begin{align}\label{eq: hypothesis}
H_0: \max_{i,j\in \mathcal M} \left\|\bpi_i - \bpi_j\right\|\leq c_{1n},  %\quad \text{ versus } \quad H_a: 
%\max_{i,j\in \mathcal M}\|\bpi_i - \bpi_j\|
%> c_{2n},
\end{align}
where $c_{1n}$ is a positive sequence 
that converges slowly to zero as network size $n$ increases. 
% and satisfying that $c_{1n}\leq c_{2n}$. 
There are various ways to formulate the alternative hypothesis, and our theoretical analyses suggest the following formulation for the alternative hypothesis 
\begin{align}\label{eq: hypothesis-a}
%H_0: \max_{i,j\in \mathcal M}\|\bpi_i - \bpi_j\|\leq c_{1n}  \quad \text{ versus } \quad 
H_a: \max_{i,j\in \mathcal M}\lambda_{\min}^{1/2}\left\{ \left(\bpi_i , \bpi_j\right)^\top  \left(\bpi_i , \bpi_j\right)\right\}> c_{2n},
\end{align}
where $\lambda_{\min}\{\cdot\}$ denotes the smallest eigenvalue of a given matrix and $c_{2n}>c_{1n}$ is also a positive sequence slowly converging to zero. It is easy to see that there exists some absolute constant $C>0$ such that
$\lambda_{\min}  \{ \left(\bpi_i , \bpi_j\right)^\top  \left(\bpi_i , \bpi_j\right) \} \le C\|\bpi_i - \bpi_j\|^2.$
Thus, the null hypothesis $H_0$ in  (\ref{eq: hypothesis}) above gives an upper bound on $\max_{i,j\in \mathcal M}\lambda_{\min}^{1/2}\{ \left(\bpi_i , \bpi_j\right)^\top  \left(\bpi_i , \bpi_j\right)\}$, while the alternative hypothesis $H_a$ in (\ref{eq: hypothesis-a}) above implies a lower bound on $\max_{i,j\in \mathcal M}\|\bpi_i - \bpi_j\|$.

To make the problem of the group network inference given in (\ref{eq: hypothesis}) and (\ref{eq: hypothesis-a}) more explicit, we will consider the commonly used degree-corrected mixed membership (DCMM) model.  Following the formulation in \cite{JKL2017}, the probability of a link between nodes $i$ and $j$ with $1 \leq i\neq j \leq n$ under the DCMM model can be written as
\begin{equation} \label{new.DCMM}
\mathbb{P} \left\{x_{ij}=1\right\} = \vartheta_i\vartheta_j  \sum_{k=1}^K\sum_{l=1}^K\bpi_i(k)\bpi_j(l)p_{kl},
\end{equation}
where $\vart_i>0$ with $i\in [n]$ is a deterministic parameter measuring the degree heterogeneity of each node $i$. Here, the parameter $p_{kl}$ can be interpreted as the probability that a typical member (say $\vartheta_i=1$) in community $\mathcal{C}_k$ connects with a typical member (say $\vartheta_j=1$) in community $\mathcal{C}_l$, as in the specific case of the stochastic block model where  for each $i\in [n]$, $\bpi_i\in \{\be_1,\cdots,\be_K\}$ with $\be_k$'s the standard basis unit vectors in $\mathbb{R}^K$ and $1 \leq k \leq K$. Rewriting \eqref{new.DCMM} in the matrix form, we have the following representation for the DCMM model under the setting in (\ref{eq: model.general}),
\begin{equation} \label{new.DCMM2}
\bbH= \bTheta \bPi \bP \bPi^\top \bTheta,
\eeq
where $\bTheta:= \diag\{\vartheta_1,\cdots,\vartheta_n\}$ is the degree heterogeneity matrix, $\bPi = (\bpi_1,\cdots, \bpi_n)^T \in \mathbb{R}^{n\times K}$ denotes the matrix of community membership probability vectors, and $\bbP = (p_{kl})\in \mathbb{R}^{K\times K}$. Note that the deterministic mean matrix $\bbH$ in the DCMM model (\ref{new.DCMM2}) above is assumed to be of rank $K$ as in the general setting (\ref{eq: model.general}). 
% and $\bbP$ and $\bPi$ have the same meaning as in \eqref{MM}. 
For the special case when the degree heterogeneity matrix $\bTheta$ in \eqref{new.DCMM2} takes the form of $\bTheta^2 = \theta\bI_n$ with $\bI_n$ an $n\times n$ identity matrix and $\theta>0$ a scaling parameter, we have the following popularly used mixed membership (MM) model 
% \begin{equation} \label{new.eq006}
% \mathbb{P} \left\{x_{ij}=1\right\} = \theta\sum_{k=1}^K\sum_{l=1}^K\bpi_i(k)\bpi_j(l)p_{kl},
% \end{equation}
% which gives rise to 
\begin{equation} \label{MM}
\bbH = \theta\bPi\bbP\bPi^T.
\end{equation}

Observe that the column space spanned by the $n \times K$ matrix $\bPi$ in (\ref{new.DCMM2}) is the same as the eigenspace spanned by the top $K$ eigenvectors of matrix $\bH$ corresponding to its top $K$ eigenvalues in magnitude. In other words, the community structure information of the network is encoded naturally in the eigen-structure of the mean matrix $\bbH$ since matrix $\bPi$ consists of the community membership probability vectors for all the nodes.
Denote by 
\begin{equation} \label{new.eq.FL.eigendeco}
\bH=\bbV\bbD\bbV^T
\end{equation}
the eigen-decomposition of the mean matrix $\bH$,
where  $\bbD=\diag\{d_1,\cdots,d_K\}$ with {$|d_1|\geq \cdots \geq |d_K|>0$} is a diagonal matrix of all $K$ nonzero eigenvalues and $\bbV=(\bbv_1,\cdots,\bbv_k) \in \mathbb{R}^{n\times K}$ represents the corresponding orthonormal matrix of top $K$ eigenvectors. 
%Our model includes the popularly used network models such as the stochastic block model, the mixed membership model, and the degree corrected mixed membership  model. Throughout the paper, the number of communities $K$ is assumed to be known. We are interested in inferring the community structure from the adjacency matrix $\bbX$.
%It is well known that the matrices $\bV$ and $\bD$ carry all the information on the community structure in the network model.
In practice, one can estimate the population matrices $\bD$ and $\bV$ in (\ref{new.eq.FL.eigendeco}) above using the empirical counterparts based on the observed adjacency matrix $\bX$. Specifically, let $\widehat d_1, \cdots , \widehat d_n$ be the eigenvalues of the adjacency matrix $\bX$, and  $\widehat\bbv_1, \cdots, \widehat{\bbv}_n$ be the corresponding eigenvectors. Without loss of generality, let us assume that $|\widehat d_1| \geq \cdots \geq |\widehat d_n|$ and define  $\widehat\bbV=(\widehat \bbv_1,\cdots,\widehat \bbv_K)\in \mathbb{R}^{n\times K}$ consisting of the top $K$ empirical eigenvectors. 
%Due to technical reasons, we will investigate the cases when $m = |\mathcal M|=2$ and when $m = |\mathcal M|>2$ separately in Section \ref{new.Sec3} for testing the null hypothesis $H_0$ in (\ref{eq: hypothesis}) against the alternative hypothesis $H_a$ in (\ref{eq: hypothesis-a}).

To facilitate the technical presentation, let us introduce some necessary notation that will be used throughout the paper. We are interested in the asymptotic regime when the network size $n\to \infty$. Whenever we refer to a constant, it will not depend on the parameter $n$. We will use $C$ to denote a generic large positive constant, whose value may change from line to line. Similarly, we will use $\epsilon$ and $c$ to denote generic small positive constants. %If a constant depend on a quantity $a$, we use $C(a)$ or $C_a$ to indicate this dependence.
For any two sequences $a_n$ and $b_n$ depending on $n$, $a_n = \OO(b_n)$ means that $|a_n| \le C|b_n|$ for some constant $C>0$, whereas $a_n=\oo(b_n)$ or $|a_n|\ll |b_n|$ means that $|a_n| /|b_n| \to 0$ as $n\to \infty$. We say that $a_n \lesssim b_n$ if $a_n = \OO(b_n)$ and that $a_n \sim b_n$ if $a_n = \OO(b_n)$ and $b_n = \OO(a_n)$. Given a vector $\mathbf v=(v_i)_{i=1}^n$, $\|\mathbf v\|\equiv \|\mathbf v\|_2$ denotes the Euclidean norm and $\|\mathbf v\|_p$ denotes the $\ell_p$-norm. Given a matrix $\bbA = (A_{ij})$, denote by $\|\bbA\|$, $\|\bbA\|_F $, and $\|\bbA\|_{\max}:=\max_{i,j}|A_{ij}|$ the operator norm, Frobenius norm, and maximum norm. For notational simplicity, we write $\bbA=\OO(a_n)$ and $\bbA=\oo(a_n)$ to mean that $\|\bbA\|=\OO(a_n)$ and $\|\bbA\|=\oo(a_n)$, respectively. Moreover, we will use $A_{ij}$ and $\bbA(k)$ to denote the $(i,j)$th entry and $k$th \textit{row vector} of a given matrix $\bbA$, respectively, and $v(k)$ to denote the $k$th component of a given vector $\bv$. We will often write an identity matrix of appropriate size as $\bI$ without specifying the size in the subscript. 
%or $1$, 
% and will use $a\wedge b$ to mean $\min\{a,b\}$. 
% In this paper, we often write an identity matrix as $I$ or $1$ and use $\be_i$ to denote the $i$-th standard basis unit vector. 

%\newpage
\section{SIMPLE-RC for group network inference} \label{new.Sec3}

%In this section, we will formally introduce the two forms of the SIMPLE with random pairing (SIMPLE-RC) test for group network inference under non-sharp nulls and weak signals, and investigate their asymptotic properties in terms of both size and power.

\subsection{Testing a given pair of nodes without degree heterogeneity} \label{new.Sec3.1}
To motivate our method of SIMPLE-RC, we begin with the specific case of $m=2$ which corresponds to the problem of testing a pair of given network nodes. To simplify the technical presentation, let us first focus on the case of no degree heterogeneity based on the mixed membership model given in (\ref{MM}). The more general case of DCMM model (\ref{new.DCMM2}) will be investigated later in Section \ref{new.Sec3.4}. 

For a group of size two, assume that $\mathcal M=\{i,j\}$ with some $i \neq j \in \{1, \cdots, n\}$. Let $K_0$ be an integer satisfying $1\leq K_0\leq K$,  $\bV_{K_0}$ an $n\times K_0$ matrix formed by the first $K_0$ columns of matrix $\bV$, and $\bbD_{K_0}$ a $K_0\times K_0$ principal minor of the matrix $\bbD$ containing its first $K_0$ diagonal entries. We will make use of two simple observations. First, under the mixed membership model (\ref{MM}) with $\bH = \theta\bPi\bP\bPi^T = \bV\bD\bV^T$, it holds that
\begin{equation}\label{eq: pi-diff-weighted}
\theta(\bpi_i-\bpi_j)^T\bP (\bpi_i-\bpi_j) = \left[\bV(i)-\bV(j)\right]^T\bD\left[\bV(i)-\bV(j)\right].
\end{equation}
%With an additional assumption that the matrix $\bP$ has eigenvalues satisfying that $0< \lambda_{K}(\bP)\leq \cdots \leq \lambda_1(\bP)\leq C$ for some constant $C>0$, 
It follows that 
 \begin{align*}
  %\theta_{\min}\|\bpi_i-\bpi_j\|_2^2 \leq 
  \left[\bV(i)-\bV(j)\right]^T\bD\left[\bV(i)-\bV(j)\right]\leq \theta_{\max}\|\bpi_i-\bpi_j\|^2
 \end{align*}
with % $\theta_{\min} := \lambda_K(\bP)\theta$ and 
$\theta_{\max} := \lambda_1(\bP)\theta$. Then under the null hypothesis $H_0$ in (\ref{eq: hypothesis}), we have
\begin{align}\label{null_hyp}
&\left\|\bD_{K_0} \left[\bV_{K_0}(i)-\bV_{K_0}(j)\right]\right\| %\le \left\|\bD \left[\bV(i)-\bV(j)\right]\right\| 
\leq \sqrt{d_1} \big\|\bD^{1/2} \left[\bV(i)-\bV(j)\right] \big\|  
\leq c_{1n}\sqrt{d_1\theta_{\max}}.
\end{align}

Second, for the power analysis, we make the assumptions that matrix $\bP$ has positive eigenvalues and that for a constant $c>0$,
\beq\label{DV-V}
    \left\|\bD_{K_0}\left[ \bV_{K_0}(i)  -  \bV_{K_0}(j) \right] \right\| \ge c \left\|\bD \left[ \bV (i)  -  \bV (j) \right] \right\|.
\eeq
%The above can be equivalently written as $\left\| \bV_{K_0}^\top \bPi \bP\left( \bpi_i  -  \bpi_j \right) \right\| \ge c\left\|\bV^\top \bPi \bP \left( \bpi_i  -  \bpi_j \right) \right\|$, 
%i.e., using only $\bV_{K_0}$ can capture the difference between $\bpi_i$ and $\bpi_j$. 
In view of $\theta\bPi\bP\bPi^T = \bV\bD\bV^T$, we see that $\bV=\theta \bPi\bB$ with $\bB:=  \bP \bPi^\top \bV \bD^{-1}.$ Then, it holds that 
\beq\nonumber
 {\bV(i)}  -  {\bV(j)} =\theta \bB^\top \left(\bpi_i , \bpi_j\right)(1 , -1)^T,
\eeq
which entails that 
\begin{align*}
  \left\|\bD\left[ \bV(i)  -  \bV(j) \right] \right\| & \gtrsim \theta\lambda_{\min}^{1/2} \big\{ \left(\bpi_i , \bpi_j\right)^\top  \left(\bpi_i , \bpi_j\right) \big\} \lambda_{\min}^{1/2}\left(\bB \bD^2\bB^\top \right) \nonumber\\
&\ge \sqrt{\theta_{\min}}\lambda_{\min}^{1/2} \big\{ \left(\bpi_i , \bpi_j\right)^\top  \left(\bpi_i , \bpi_j\right) \big\} \lambda_{\min}^{1/2}\left(\bV^\top \bH\bV \right) \\
&\ge  \sqrt{d_{K}\theta_{\min}} \lambda_{\min}^{1/2} \big\{ \left(\bpi_i , \bpi_j\right)^\top  \left(\bpi_i , \bpi_j\right) \big\}
\end{align*}
with $\theta_{\min} := \lambda_K(\bP)\theta$. Thus, under \eqref{DV-V} and the alternative hypothesis $H_a$ in (\ref{eq: hypothesis-a}), we have 
\beq\label{alt_hyp}
\left\|\bD_{K_0}\left[ \bV_{K_0}(i)  -  \bV_{K_0}(j) \right] \right\| \gtrsim c_{2n}\sqrt{d_{K}\theta_{\min}}. % \sqrt{\theta_{\min}} c_{2n}.
\eeq
It is seen that assuming \eqref{DV-V} ensures that using only $\bV_{K_0}$ (instead of $\bV$) can still capture a significant fraction of difference between $\bpi_i$ and $\bpi_j$, which is important for achieving high power using SIMPLE-RC.

The two observations in (\ref{null_hyp}) and (\ref{alt_hyp}) above have motivated us to exploit the similar test statistic to that for the SIMPLE test proposed originally in \cite{SIMPLE} for assessing the membership profile information of a given pair of nodes $\{i, j\}$. Different from \cite{SIMPLE} focusing on the case of sharp nulls under relatively stronger signals, we will consider the more general setting with both non-sharp nulls and weak signals. Specifically, we define a test statistic 
\begin{align}\label{eq: T-test}
T_{ij}(K_0) := \left[\widehat\bbV_{K_0}(i)-\widehat\bbV_{K_0}(j)\right]^T \left[\bSig_{i,j}(K_0)\right]^{-1} \left[\widehat\bbV_{K_0}(i)-\widehat\bbV_{K_0}(j)\right], %\widehat{\lambda}_k(\widehat\bbv_k(i)-\widehat\bbv_k(j)).
\end{align}
where $1\le K_0\le K$ is some pre-determined number, $\wh \bV_{K_0}$ is the $n\times K_0$ matrix formed by the first $K_0$ columns of matrix $\wh\bV$, and  $\bSig_{i,j}(K_0)$ denotes the $K_0 \times K_0$ asymptotic covariance matrix of $\widehat\bbV_{K_0}(i)-\widehat\bbV_{K_0}(j)$ that is defined as 
\begin{equation} \label{new.FL.sig}
\bSig_{i,j}(K_0) :=  \cov[(\be_i-\be_j)^T\bW\bV_{K_0}\bbD_{K_0}^{-1}]
\end{equation}
with $\be_k$ being the $k$th standard basis unit vector in $\mathbb{R}^n$. As shown later, the choice of the parameter $1\le K_0\le K$ plays a key role in network inference under weak signals, which is one of the major distinctions from the work of \cite{SIMPLE}.

To facilitate our technical analysis, we will need to impose some basic regularity conditions under the mixed membership model setting (\ref{MM}).

\begin{assu} 
%For the mixed membership model (\ref{MM}), 
We assume that 
\begin{itemize}
\item[(i)] (Network sparsity) It holds that $q\gg (\log n)^{4}$ with $q=\sqrt{n\theta}$.

\item[(ii)] (Spiked eigenvalues) %There exists a fixed integer $1\le K_0\le K$ so that 
It holds that $|d_k|\ge q \log\log n$ for all $1\le k \le K_{0}$. 

\item[(iii)] (Eigengap) There exists some constant $\e_0>0$ such that 
\beq\label{eq_eigengap}
\min_{1\le k \le K_0} \frac{|d_k|}{|d_{k+1}|} >1+\e_0, %\quad \frac{|d_{K_0}|}{|d_{K_0+1}|} >1+c_0,.
\eeq
where we do not require eigengaps for smaller eigenvalues $|d_k|$ with $K_0+1\le k \le K$.

\item[(iv)] (Mean matrix) There exists some constant $\e_1>0$ such that $\max_{i,j \in [n]} h_{ij}\le 1-\e_1$ and $\max_{ i\in [n]}\sum_{ j\in [n]} h_{ij} \ge \e_1 n\theta$, and the eigenvalues of $\bP$ satisfy that $0< \lambda_{K}(\bP)\leq \cdots \leq \lambda_1(\bP)\leq C$ for some large constant $C>0$. 

\item[(v)] (Covariance matrix) There exists some constant $0<\e_2<1$ such that all the eigenvalues of $\theta^{-1}\bbD_{K_0'}\bSig_{i,j}(K_0')\bbD_{K_0'}$ are between $\e_2$ and $\e_2^{-1}$ %bounded away from 0 and $\infty$ 
for all $\{i,j\} \subset \cal M$ and $1\le K_0'\le K_{0}$.
\end{itemize}

\noindent Moreover, for fixed constants $\e_0$ and $\e_2$, denote by $K_{\max}\equiv K_{\max}(n,\e_0,\e_2)\le K$ the largest $K_0$ such that parts (ii), (iii), and (v) above hold.\label{main_assm} 
\end{assu}

% {\color{red} If we slightly modify the definition and let $K_0$ be the largest integer such that Condition 1(ii), (iii) and (v) hold. Then it seems that all $\tilde K\leq K_0$ would also satisfy the above conditions. } {\color{blue}If we fix $\e_0$, $\e_2$ and $d_k\ge C_nq$ for some specific parameter $C_n\to \infty$, then we can choose the largest integer $K_0$ such that Condition 1(ii), (iii) and (v) hold. But, there are some tricky points: $K_0$ chosen in this way may depend on $n$, the choice of $\e_0$, $\e_2$ and $C_n$ is arbitrary in some sense, and the condition $|d_k|\gg q $ will be replaced by stronger assumptions in the statements of the main results below, so different versions of $K_0$ need to be defined.}
% {\color{red} Comment on 15th: I suggest we define $K_{0,max}$ with some fixed $C_n = \log\log n$, say. This $K_{0,max}$ may depend on $n$, $\varepsilon_0$ and $\varepsilon_2$, but we can state our theorems using some fixed and finite $K_0 < K_{0,\max}$. Do you think this would work?}

The parameter $q$ in Condition \ref{main_assm} above is a key parameter for our technical study. It gives the typical size of the eigenvalues of the noise random matrix $\bW$. Specifically, we will see from Lemma \ref{lem_opbound} in Section \ref{new.Sec4.2} that such noise eignevalues are bounded as $\|\bbW\|\lesssim q$ with probability $1-\oo(1)$. On the other hand, the convergence rates for various asymptotic expansions developed in our paper will be expressed in terms of $q^{-1}$. By imposing a lower bound on $q$, part (i) of Condition \ref{main_assm} above gives a constraint on the network sparsity $\theta\gg (\log n)^8/n$, which greatly relaxes the assumption of  $\theta\ge n^{-1+\e}$ employed in \cite{SIMPLE,ATE}. Our technical arguments may be improved to relax the assumption further to the scenario of $q\gg (\log n)^C$ for some smaller constant $1<C<4$, but we will not pursue such improvement in the current paper. Part  (ii) of Condition \ref{main_assm} imposes a constraint on the signal-to-noise ratio and shows that the eigenvalues $d_k$ with $1\le k \le K_0$ are indeed spikes.  The factor $\log\log n$ is chosen for definiteness in defining $K_{\max}$ and can be replaced by a different factor $C_n\to \infty$. Parts (iii)--(v) of Condition \ref{main_assm} are essentially the same as those in \cite{SIMPLE}. %In particular, part (iii) of Condition \ref{main_assm} means that the supercritical spikes are well separated from each other. It is possible to consider extensions to a more general setting with degenerate spikes (i.e., with eigenvalue multiplicity), where we will need to establish an asymptotic expansion of the projection onto the subspace spanned by the corresponding empirical eigenvectors instead of an expansion of each individual empirical eigenvector separately. Note that the smaller eigenvalues $|d_i|$ with $K_{\max}+1\le i \le K$ can contain subcritical signals smaller than $\|\bbW\|$. We, however, do not place any constraint on such eigenvalues since we will construct test statistic $T_{ij}$ given in (\ref{eq: T-test}) by utilizing only relatively large spiked eigenvalues and their corresponding eigenvectors.

It would be interesting to study how many spikes $K_0$ one should use for the construction of the test statistic $T_{ij}$ for achieving the best size and power tradeoff; we will investigate such problem in the future work. It is also worth mentioning that the importance of eigen-selection was investigated recently in \cite{HanTongFan2022} for a different problem of high-dimensional clustering with the spectral method. 

%\begin{remark}
%From condition (iv), we immediately obtain that $|d_1|\lesssim n\theta = q^2$ and $|d_K|\gtrsim \lambda_K(\bbP) q^2$.
% \iffalse    
% More precisely, assume that $D$ has $K_0$ nonzero eigenvalues and $K$ spiked eigenvalues satisfying Condition \ref{main_assm}.2. We still assume that \eqref{eq_eigengap} holds and 
% $$\frac{|d_K|}{|d_{K+1}|} >1+c_0,$$
% but we do not have other constraints on eigenvalues $|d_i|$, $i\ge K+1$. Moreover, we need to strengthen Condition \ref{main_assm}.2 as $|d_k|\gg K_0^c\sqrt{\log n}q $, $1\le k \le K$, for some power $c>0$ that can be made explicit. Then Theorem \ref{thm:pair-null} holds with $\lambda_K(\bP)$ replaced by $\lambda_{K_0}(\bP)$ and a new covariance matrix $\bSig_{i,j}$ defined using the first $K$ spiked eigenvalues and eigenvectors only. 
% \fi
%\end{remark} 

Our first main theoretical result in the theorem below characterizes the asymptotic behaviors of the test statistic $T_{ij}$ under the null and alternative hypotheses $H_0$ and $H_a$, respectively.

\begin{theorem} \label{thm:pair-null}
Assume that Condition \ref{main_assm} holds, $K_0$ is a random variable such that $1 \leq K_0\le K_{\max}\wedge C_0 $ almost surely for some large constant $C_0>0$, $|d_{K_0}|\gg q\sqrt{\log n}$ almost surely, and 
\beq\label{eq: condK}
1\le K\ll \frac{q}{(\sqrt{n}\|\bV\|_{\max})\log n}\wedge \frac{|d_{K_0}|^2}{(\sqrt{n}\|\bV\|_{\max})^2 q^2}
\eeq
with $\wedge$ standing for the minimum of two given numbers. Then the test statistic $T_{ij}(K_0)$ defined in (\ref{eq: T-test}) satisfies that 
\begin{enumerate}
    \item If $c_{1n} \ll [d_{1}\lambda_1(\bP)]^{-\frac12}$, it holds that under the null hypothesis $H_0$ in \eqref{eq: hypothesis}, 
  \beq\label{eq_null}
    \lim_{n\to \infty}\sup_{x\in \R}\left| \P\left\{T_{ij}(K_0)\le x\right\}- F_{K_0}(x)\right|= 0,
  \eeq
    where conditional on $K_0$, $F_{K_0}$ is the chi-square distribution with $K_0$ degrees of freedom.

    \item If  $c_{2n} \gg [d_{K}\lambda_K(\bP)]^{-\frac12}$, it holds that under \eqref{DV-V} and the alternative hypothesis $H_a$ in \eqref{eq: hypothesis-a},
   \beq\label{eq_power_pair}
    \lim_{n\to \infty}\P\left\{T_{ij}(K_0)>C\right\} = 1 % \quad \text{ as } n\rightarrow \infty.
    \eeq
  for each arbitrarily large constant $C>0$. 
\iffalse    In particular, when $K=K_0$, \eqref{alt_hyp} holds  under $H_a$. Thus, if $c_{2n}\gg q^{-1} \lambda_K(\bP)^{-1}$,
    % \beq\label{eq_c2n} 
    % c_{2n}\gg q^{-1} \lambda_K(\bP)^{-1/2} \bigg[ \lambda_{K_0}(\bP)^{-1/2} + \bigg(\sum_{k=K_0+1}^K |d_k|\bigg)^{1/2}\bigg], 
    % \eeq
    then \eqref{eq_power_pair} holds under $H_a$. \fi
%     for any large constant $C>0$,
%   \beq\label{eq_power_pair2}
%     P(T_{ij}>C) \rightarrow 1, \quad \text{ as } n\rightarrow \infty.
%     \eeq
\end{enumerate}
\end{theorem}

%This theorem will be proved in Section \ref{subsec:cal-thm-pair} of the supplement. 
Theorem \ref{thm:pair-null} above improves the corresponding result in Theorem 1 of \cite{SIMPLE} in several important aspects. First, it considers hypothesis testing with non-sharp nulls. Second, it allows for a slowly diverging number of communities $K$. %(as we will discuss in more details below) 
Third, it relaxes the lower bound on the parameter $q$ from $q\ge n^\e$ in \cite{SIMPLE} to $q\gg (\log n)^{4}$, that is, much sparser networks are accommodated in our setting. Fourth, it relaxes the lower bound on the signal-to-noise ratio $|d_{K_0}|/q$ from $n^\e$ in \cite{SIMPLE} to $\sqrt{\log n}$. As mentioned in the Introduction, these extensions are crucially based on the new asymptotic expansion of the empirical spiked eigenvectors under relaxed conditions on the network sparsity and signal-to-noise ratio, which will be presented in Section \ref{new.Sec4.3} of the Supplementary Material. 

We emphasize that we regard $K_0$ as a random variable in Theorem \ref{thm:pair-null}. This is because later on it will be replaced with a sample counterpart learned from the data which is naturally random. For the special case when $K_0\le K_{\max}$ is a fixed integer, the distributional bound in \eqref{eq_null} means that $T_{ij}$ converges in distribution to the chi-square distribution $\chi_{K_0}^2$ as the network size $n$ increases. It is possible to remove the constraint $K_0\le C_0$ and extend Theorem \ref{thm:pair-null} to the case with a slowly diverging $K_0$ if we are willing to impose stronger assumptions on $|d_k|$ and $K$. However, we will not pursue such extension in the current paper. We also remark that if we are willing to make the following assumption (see, e.g., equation (2.14) of \cite{JKL2017}) 
 \beq\label{cond:Pi} \|(\bPi^\top \bPi)^{-1}\| \lesssim K/n,
 \eeq 
then it can be derived that constraint \eqref{eq: condK} is satisfied if $K\lesssim \left(q/\log n\right)^{2/3}\wedge (|d_{K_0}|/q)$. This shows that our framework allows for diverging $K$ depending on both the network sparsity and the signal-to-noise ratio. %We would like to point out that the eigengap assumption \eqref{eq_eigengap} in Condition \ref{main_assm} actually leads to a constraint that $K_0\lesssim \log (|d_1|/q) \lesssim \log q$. 

% Let us gain some insights into the assumption \eqref{eq: condK} above on the parameter $K$ through examining the upper bound
% $$
% \mathbb K :=  \frac{q}{(\sqrt{n}\|\bV\|_{\max})\log n}\wedge \frac{|d_{K_0}|^2}{(\sqrt{n}\|\bV\|_{\max})^2 q^2}. 
% $$
% Since the spiked eigenvectors $\bv_k$ with $1\le k \le K$ are normalized, it holds that $\sqrt{n}\|\bV\|_{\max}\ge 1$, which entails that  
% $$
% \mathbb K \lesssim \frac{q}{\log n}\wedge \frac{|d_{K_0}|^2}{q^2}.
% $$ 
% On the other hand, as shown by equations (B.17) and (B.18) of \cite{SIMPLE}, we have the upper bound 
% \beq\label{Vmax0}\|\bV\|_{\max} \le \|(\bPi^\top \bPi)^{-1}\|^{1/2}.\eeq
% If we further assume that (see, e.g., equation (2.14) of \cite{JKL2017}) 
% \beq\label{cond:Pi} \|(\bPi^\top \bPi)^{-1}\| \lesssim K/n,
% \eeq
% %For example, this holds when each community contains at least $cn/K$ pure nodes for a small constant $c>0$. Under \eqref{cond:Pi}, 
% it follows that $\sqrt{n}\|\bV\|_{\max}\lesssim \sqrt{K}$ and 
% \beq\label{eq: condKs} \mathbb K\gtrsim \left(\frac{q}{\log n}\right)^{2/3}\wedge \frac{|d_{K_0}|}{q}.\eeq
% In light of the above bounds on $\mathbb K$, our technical analysis indeed allows for diverging $K$ depending on both the network sparsity and the signal-to-noise ratio. 

%For generality, we do not make the assumption \eqref{cond:Pi} for the mixed membership model. However, observe that the assumption \eqref{eq: condK} does impose a constraint on $\|\bV\|_{\max}$ through 
Observe that the assumption \eqref{eq: condK} imposes a constraint on $\|\bV\|_{\max}$ through 
\beq\label{upper-Vmax0}
\|\bV\|_{\max} \ll \frac{\sqrt{\theta}}{\log n}\wedge \frac{|d_{K_0}|}{n\sqrt{\theta}},
\eeq
which restricts implicitly the number of pure nodes. For example, denote by $n_{\min}$ the minimum number of pure nodes among all communities. Then it holds that $\bPi^T\bPi = \sum_{i\in [n]} \bpi_i \bpi_i^T \ge n_{\min} \bI$. As shown by (B.17) and (B.18) of \cite{SIMPLE}, we have the upper bound 
 \beq\label{Vmax0}\|\bV\|_{\max} \le \|(\bPi^\top \bPi)^{-1}\|^{1/2}.\eeq
 Hence, from \eqref{Vmax0} we see that the bound \eqref{upper-Vmax0} can be satisfied when  
$$ \sqrt {\frac{n}{n_{\min}}}\ll \frac{ {q}}{\log n}\wedge \frac{|d_{K_0}|}{ {q}}.$$
Since $|d_{K_0}|\gg q\sqrt{\log n}$, the right-hand side of the above expression diverges to infinity as the network size $n$ increases, meaning that our SIMPLE-RC test can accommodate vanishing proportion of pure nodes with $n_{\min}/n\ll 1$.

\subsection{Test statistic in group setting without degree heterogeneity} \label{sec:group-test}

We now consider the group testing for the case of diverging $m$ without degree heterogeneity. Without loss of generality, let us assume that $m\in 2\N$. A natural idea for testing the null hypothesis $H_0$ in \eqref{eq: hypothesis} would be to investigate the test statistic
$ \max_{\{i, j\}\subset {\cal M}} T_{ij}
$
with $T_{ij}$ given in (\ref{eq: T-test}). Yet, doing so is rather challenging because of potentially high correlations among all the individual $T_{ij}$'s. To deal with such a challenging issue, we will suggest a random coupling strategy for group network inference. Specifically, we randomly pick pairs of nodes in $\cal M$ without replacement until all nodes are coupled. Denote by $\cal P$ the resulting pairs of such random coupling. 

Given $\cal P$, we formally define our SIMPLE-RC test statistic $T$ as 
\beq \label{eq: T-test_group}
T= \max_{\{i,j\}\in {\cal P} } T_{ij}.
\eeq
We will show in the theorem below that under a suitable centering and rescaling, the test statistic $T$ in (\ref{eq: T-test_group}) converges to a Gumbel distribution under the null hypothesis.
% and the assumption that $1\ll m\ll \sqrt{\theta}/\|\bV_{K_0}\|_{\max}$. %{\color{red} should the upper bound be $\sqrt{\theta}/\|\bV_{K_0}\|_{\max}$} 

\begin{theorem}\label{thm:group-null}
Assume that Condition \ref{main_assm} holds, $K_0$ is a random variable such that $1 \leq K_0\le K_{\max}\wedge C_0 $ almost surely for some large constant $C_0>0$, $|d_{K_0}|\gg q {\log n}$ almost surely,   
\beq\label{eq: condK2}
1\le K\ll \frac{q}{(\sqrt{n}\|\bV\|_{\max})(\log n)^{3/2}}\wedge \frac{|d_{K_0}|^2}{(\sqrt{n}\|\bV\|_{\max})^2 q^2\log n},
\eeq
%$1\le K\ll [q/(\log n)^{3/2}]\wedge [|d_{K_0}|/(q\sqrt{\log n})]$
\beq\label{cond_m} 
 1\ll m \ll {\sqrt{\theta}}/{\|\bbV_{K_0}\|_{\max}}, \ \ \max_{1\le k \le K_0} \sum_{l\in \cal M}|v_k(l)|^2 \ll  (\log n)^{-2}, %\max_{1\le k \le K_0}\max_{l\in \cal M} \sum_{l\in \cal M}|\bv_k(l)|^2 \ll  (\log n)^{-2}. %\min\left\{\frac{n}{\theta(\log n)^2},\ \sqrt{{n\theta}\right\}.
\eeq
and $c_{1n} \ll [d_{1}\lambda_1(\bP)]^{-1/2}(\log n)^{-1/2}$. Then the SIMPLE-RC test statistic $T$ in (\ref{eq: T-test_group}) satisfies that under the null hypothesis $H_0$ in \eqref{eq: hypothesis}, 
\beq\label{eq:conv_Gumbel}
\lim_{n\to \infty}\sup_{x\in \R}\left| \P\left\{\frac{T(K_0)-b_m(K_0)}{2}\le x\right\}-\cal G(x) \right| = 0,
\eeq
%as $n \rightarrow \infty$, 
where  $\cal G(x)=\exp(-e^{-x})$ denotes the Gumbel distribution and
\beq\label{eq_ambm}
%a_m = 2, \ \ 
b_m(K_0)= 2\log \frac{m}{2} + (K_0 -2)\log \log \frac{m}{2} - 2 \log \Gamma\left(\frac{K_0}2\right)
\eeq
with $\Gamma(\cdot)$ representing the gamma function. 
\end{theorem}

In the proof of Theorem \ref{thm:group-null} given in Section \ref{Sec.newA.2} of the Supplementary Material, we will show that the individual test statistics $T_{ij}$ based on the random coupling are asymptotically independent. Such a result entails that when the group size $m$ is bounded, an application of Theorem \ref{thm:pair-null} yields that the asymptotic distribution of the SIMPLE-RC test statistic $T$ becomes the maximum of $m/2$ independent $\chi_{K_0}^2$ random variables under the null hypothesis $H_0$. Nevertheless, we focus on the more interesting case of diverging $m$ here. It is interesting to see that the limiting null distribution is free of the random variable $K_0$, which is helpful in deriving the asymptotic null distribution when we replace $K_0$ with its sample counterpart later on. To the best of our knowledge, both the SIMPLE-RC test statistic $T$ based on empirical spiked eigenvectors and the idea of random coupling, and the theoretical result established above regarding its asymptotic Gumbel null distribution are new to the network 
literature.

Compared to Theorem \ref{thm:pair-null}, our Theorem \ref{thm:group-null} above assumes slightly stronger assumptions on $c_{1n}$, $d_{K_0}$, and $K$ by a factor of $(\log n)^{1/2}$. This is due to the fact that when taking the union bound over $m/2$ random variables $T_{ij}$'s, we need to ensure that the sum of their Gaussian tails is asymptotically negligible. %, i.e., each tail probability is of order $\oo(m^{-1})$. 
Our result can handle only group size satisfying the first assumption in \eqref{cond_m}. Such a restriction is because each individual test statistic $T_{ij}$ is not an exact %Gaussian 
chi-square random variable, and thus we will need to uniformly bound the difference between its distribution and the $\chi_{K_0}^2$ distribution using the Berry--Esseen inequality (see \eqref{BerryEsseen} in the Supplementary Material), which leads to a constraint on the group size $m$. %Ensuring these differences are negligible when calculating $T$ leads to the first condition in \eqref{cond_m}.
If we are willing to make the  stronger  assumption \eqref{cond:Pi},
then it holds that $\|\bbV_{K_0}\|_{\max} \lesssim \sqrt{K/n}$ by \eqref{Vmax0}, and the first assumption in \eqref{cond_m} can be satisfied as long as $1\ll m \ll q/\sqrt{K}$.

%\begin{remark}

The second assumption in \eqref{cond_m} is a delocalization condition meaning that the $\ell_2$-mass of each eigenvector $\bv_k$ with $1\le k \le K_0$ is not concentrated on the \textit{small} set $\cal M$. In fact, the correlations between different $T_{ij}$'s are measured by $ \sum_{l\in \cal M}|v_k(l)|^2$ (see \eqref{Bern_vw} in the Supplementary Material), which indicates that such condition ensures that random variables $T_{ij}$'s with $\{i,j\}\in {\cal P}$ are asymptotically independent of each other. It is worth mentioning that the second assumption in \eqref{cond_m} is satisfied automatically for networks with $\theta\le (\log n)^{-1/2}$. To understand this, observe that \eqref{eq: condK2} entails that 
\beq\label{upper-Vmax}
 \|\bV\|_{\max}\ll \frac{\sqrt{\theta}}{(\log n)^{3/2}}\wedge \frac{|d_{K_0}|}{q\sqrt{n\log n}}.
\eeq
Then combining (\ref{upper-Vmax}) and the first assumption in \eqref{cond_m}, we can obtain that  
\beq\label{cond_m222}
\begin{split}
\max_{1\le k \le K_0} \sum_{l\in \cal M}|v_k(l)|^2 & \le  m \|\bbV_{K_0}\|_{\max}^2 \ll \sqrt{\theta} \|\bbV_{K_0}\|_{\max} \ll \frac{\theta}{(\log n)^{3/2}}\le (\log n)^{-2},
\end{split}
\eeq
which establishes the second assumption in \eqref{cond_m}.

% It is actually a weak condition imposed on dense networks only. 

%It is actually a weak condition imposed on dense networks only. For sparse networks, we notice that \eqref{eq: condK2} implies 
%\beq\label{upper-Vmax}
% \|\bV\|_{\max}\ll \frac{\sqrt{\theta}}{(\log n)^{3/2}}\wedge \frac{|d_{K_0}|}{q\sqrt{n\log n}}.
%\eeq
%Together with the the first condition of \eqref{cond_m}, it gives that 
%\beq\label{cond_m222}
%\max_{1\le k \le K_0}\max_{l\in \cal M} \sum_{l\in \cal M}|\bv_k(l)|^2 \le  m \|\bbV_{K_0}\|_{\max}^2 \ll \sqrt{\theta} \|\bbV_{K_0}\|_{\max} \ll \frac{\theta}{(\log n)^{3/2}} .
%\eeq
%Hence, the second condition of \eqref{cond_m} is automatically satisfied if $\theta\le (\log n)^{-1/2}$. 
%\end{remark}

%\begin{remark}\label{rmk_power}

In addition to the simple-to-use asymptotic null distribution, the SIMPLE-RC test (\ref{eq: T-test_group}) also has appealing power under the alternative hypothesis $H_a$. To understand this, recall that from \eqref{alt_hyp}, we know that the signal strength is measured through the quantity 
\beq\label{eq:max_sig}
\max_{\{i, j\} \subset \cal M} \left\|\bD_{K_0} \left[\bV_{K_0}(i)-\bV_{K_0}(j)\right]\right\|.
\eeq
As shown in Section \ref{Sec.new.RPpf} of the Supplementary Material, we have a useful bound related to the power analysis in the lemma below.

 %\iffalse
%%alternative hypothesis in the $K=K_0$ case. 
%First, it is seen that under the alternative %hypothesis in \eqref{eq: hypothesis},
%$$
%\max_{i\neq j\in \cal %M}\|\bD^{1/2}(\bV(i)-\bV(j))\|_2 \geq %\sqrt{\theta_{\min}}\max_{i\neq j\in \cal %M}\|\bpi(i)-\bpi(j)\|_2^2\geq \sqrt{\theta_{\min}} %c_{2n}.
%$$
%\fi

\begin{lemma}\label{lem-signal}
      If $m\rightarrow \infty$, then it holds with probability $1-\oo(1)$ that  
\begin{equation}\label{eq:signal-after-pairing}
    \max_{\{i,j\}\in \cal P} \left\|\bD_{K_0} \left[\bV_{K_0}(i)-\bV_{K_0}(j)\right]\right\|\ge \frac{1}{3}\max_{\{i, j\} \subset \cal M} \left\|\bD_{K_0} \left[\bV_{K_0}(i)-\bV_{K_0}(j)\right]\right\|.
\end{equation}
\end{lemma}

% We claim that if $m\gg 1$, then with probability $1-\oo(1)$, 
% \begin{equation}\label{eq:signal-after-pairing}
%     \max_{\{i,j\}\in \cal P}\|\bD_{K_0} (\bV_{K_0}(i)-\bV_{K_0}(j))\|_2\ge \frac{1}{3}\max_{i\ne j \in \cal M}\|\bD_{K_0} (\bV_{K_0}(i)-\bV_{K_0}(j))\|_2.
% \end{equation}
% That is, with asymptotic probability one, the uniform random paring strategy preserves the signal strength at the same order. 
%With \eqref{power_pair}, 

With the aid of Lemma \ref{lem-signal}, we can show in the theorem below that the SIMPLE-RC test can admit asymptotic power one under the null hypothesis $H_a$.

\begin{theorem}\label{thm:group-alt}
Assume that all conditions of Theorem \ref{thm:group-null} hold and 
\beq\label{DV-Vgp}
\max_{\{i, j\} \subset \cal M} \left\|\bD_{K_0} \left[\bV_{K_0}(i)-\bV_{K_0}(j)\right]\right\| \ge c\max_{\{i, j\} \subset \cal M} \left\|\bD \left[\bV(i)-\bV(j)\right]\right\|
\eeq
% \beq\label{DV-Vgp}
% \max_{i\ne j \in \cal M}\left\| \bV_{K_0}^\top \bPi \bP\left( \bpi_i  -  \bpi_j \right) \right\| \ge c\max_{i\ne j \in \cal M}\left\| \bV^\top \bPi \bP \left( \bpi_i  -  \bpi_j \right) \right\|
% \eeq
for a constant $c>0$ almost surely. If $c_{2n}\gg [d_{K}\lambda_{K}(\bbP)]^{-1/2}\sqrt{\log n}$, then the SIMPLE-RC test statistic $T$ defined in (\ref{eq: T-test_group}) satisfies that under the alternative hypothesis $H_a$ in \eqref{eq: hypothesis-a}, for each arbitrarily large constant $C>0$,
\beq\label{eq_power_group}
    \lim_{n\to \infty}\P\left\{\frac{T(K_0)-b_m(K_0)}{2}>C\right\} =1.
    \eeq

%If $K=K_0$ and $c_{2n}$ satisfies $c_{2n}\gg q^{-1} \lambda_K(\bP)^{-1}\sqrt{\log n}$, then \eqref{eq_power_group} holds under $H_a$.
\end{theorem}

%\end{remark}

% \begin{remark}
% It may be more natural to use the following statistics  
% $$
% T' = \max_{i,j \in \mathcal M} T_{ij},\quad \text{or}\quad T''=\max_{i \in \mathcal M\setminus \{1\}} T_{1i}.
% $$
% However, we do not find a way to derive the asymptotic distribution (if there is one) for $T'$ or $T''$ due to the correlations among $T_{ij}$'s. %For $T''$, we can condition on the entries in the first row of $\bbW$ and then do the calculation, but it seems that $T''$ may not have a definite asymptotic distribution.
% \end{remark}

\begin{remark}
For the case of $m\gg {\sqrt{\theta}}/{\|\bbV_{K_0}\|_{\max}}$, we can design a new test statistic as follows. Denote by $\wh m:= \hat\theta^{1/2}/\|\wh\bV_{K_0}\|_{\max}$ with $\hat \theta$ an estimator of the network sparsity and $\|\wh\bbV_{K_0}\|_{\max}$ a good estimator of $\|\bbV_{K_0}\|_{\max}$ by our asymptotic expansion \eqref{expansion_evector2} in the technical analysis. Then we  randomly choose a subset $B$ of $m_0\ll \wh m$
many nodes out of set $\cal M$ and form a new test statistic    
$T= \max_{\{i,j\}\in {\cal P} } T_{ij},$
where $\cal P$ now denotes a random coupling of the nodes in $B$. Such test statistic is powerful when there are two clusters of nodes whose membership profile vectors are \textit{separated}. More precisely, assume that there exist some subsets $A_1, A_2\subset \cal M$ such that $\|A_1\|\gg m/m_0$, $\|A_2\|\gg m/m_0$, and
$$ \min_{i\in A_1,\, j \in A_2}\left\|\bD_{K_0}\left[\bV_{K_0}(i)-\bV_{K_0}(j)\right]\right\|=\ell.$$
Then we see that with probability $1-\oo(1)$, subset $B$ contains at least one node in subset $A_1$ and one node in subset $A_2$. Hence, it follows from Lemma \ref{lem-signal}
%\eqref{power_pair} in Section \ref{Sec.new.RPpf} of the Supplementary Material 
that with probability $1-\oo(1)$,
$$ \max_{\{i,j\}\in \cal P} \left\|\bD_{K_0} \left[\bV_{K_0}(i)-\bV_{K_0}(j)\right]\right\|\ge \frac{\ell}{3}. $$
%However, the above statistic cannot cover all cases. For example, it cannot handle the case where there is only one vertex that has a different membership profile from other vertices.  
\end{remark}

\subsection{Estimation of covariance matrices} \label{new.Sec3.3}

%The test statistics $T_{ij}$ in \eqref{eq: T-test} and $T$ in \eqref{eq: T-test_group} are not applicable directly because the covariance matrices $\bSig_{i,j}$ is still unkonwn and it is not clear how to choose $K_0$.

For the practical implementation of both test statistics $T_{ij}$ and $T$ introduced in \eqref{eq: T-test} and \eqref{eq: T-test_group}, respectively, we will need to provide an estimate of the covariance matrix $\bSig_{i,j}$ given in (\ref{new.FL.sig}) and specify the choice of  $K_0$. 

For a specification of $K_0$, we suggest a consistent estimator \smash{$\wh\bSig_{i,j}(K_0)$} of the covariance matrix $\bSig_{i,j}(K_0)$. Denote by  $\sigma^2_{kl}:= \var(W_{kl})$ with $  k,l \in [n]$. Some standard calculations yield that for each $1\le a,b \le  K_0$, 
\beq\label{Sigmaij}
\left(\bSig_{i,j}( K_0)\right)_{ab}=\frac{1}{d_a d_b} \sum_{l=1}^n \left(\sigma^2_{il}+\sigma^2_{jl}\right)v_a(l)v_b(l) - \frac{1}{d_a d_b}\sigma^2_{ij}\left[v_a(i)v_b(j)+ v_a(j)v_b(i)\right].
\eeq
A natural estimator of $\sigma^2_{ab}$ is $\wh w_{ab}^2$ with $\wh w_{ab}$ being the $(a, b)$th entry of the residual matrix 
\beq\label{whW}
\wh \bbW:= \bbX - \sum_{k=1}^{ K_0} \wh d_k \wh \bv_k \wh \bv_k^\top.
\eeq
Thus, in light of (\ref{Sigmaij}), we define the empirical counterpart of $\bSig_{i,j}( K_0)$ as $\wh \bSig_{i,j}( K_0)$ with entries given by
\beq\label{whSigmaij}
 (\wh \bSig_{i,j}( K_0) )_{ab}:=\frac{1}{\wh d_a \wh d_b} \sum_{l=1}^n \left(\wh w^2_{il}+\wh w^2_{jl}\right)\wh v_a(l)\wh v_b(l) - \frac{1}{\wh d_a \wh d_b}\wh w^2_{ij} \left[\wh v_a(i)\wh v_b(j)+ \wh v_a(j)\wh v_b(i)\right]
\eeq
for $1 \leq a, b \leq K_0$.

From (\ref{whW}), we see that the estimator $\wh \bSig_{i,j}( K_0)$ in (\ref{whSigmaij}) above disregards completely the weak signals \smash{$\wh d_k$} with $ K_0+1\le k \le d$. Such a feature will lead to an error that involves quantity $\theta \cal E( K_0)$ with
\beq\label{defn_En}  
\cal E( K_0):= \theta^{-1}\max_{ i,j\in [n]}\Big|\sum_{k= K_0+1}^{K} d_k  v_k(i)  v_k (j)\Big|
\eeq
as unveiled in the theorem below.

\begin{theorem}\label{thm_consist_Sigma}
Assume that all conditions of Theorem \ref{thm:pair-null} hold and $\cal E(K_0)=\OO(1)$ almost surely. %{\color{red} shall we say a.s. because $K_0$ is random?}. 
Then the covariance matrix estimator $\wh \bSig_{i,j}$ defined in (\ref{whSigmaij}) satisfies that for each large constant $C>1$, there exists some constant $\wt C$ such that
\beq\label{consist_Sigma}
\begin{split}
    &\P\left\{\max_{i \ne j \in {\cal M} } \left\|\bbD_{K_0} \left[\wh \bSig_{i,j}(K_0)-\bSig_{i,j}(K_0)\right] \bbD_{K_0}\right\| > \wt C \theta\left[\mathcal E( K_0)+\wt{\mathcal E}(K_0)\right] \right\}\le n^{-C},
\end{split}
%\P\left( \left\{ \theta^{-1}\|\bbD_{K_0} (\wh \bSig_{i,j}-\bSig_{i,j})\bbD_{K_0}\| > \wt C(\mathcal E(n)+\wt{\mathcal E}(n)) \right\} \cap \Omega\right) \le n^{-C},
\eeq
where $\wt{\cal E}(K_0)$ is defined as  
\beq\label{wtEn}
\wt{\cal E}(K_0):=  \frac{q\sqrt{ \log n }}{|d_{K_0}|} + \frac{\|\bbV\|_{\max}\sqrt{\log n}}{\sqrt{\theta}}.
\eeq
\end{theorem}

Theorem \ref{thm_consist_Sigma} above shows that $\wh \bSig_{i,j}(K_0)$ in (\ref{whSigmaij}) provides a good estimate for the covariance matrix $ \bSig_{i,j}(K_0)$ under the setting of Theorem \ref{thm:pair-null} (and hence under stronger conditions in Theorems \ref{thm:group-null} and \ref{thm:group-alt}) as long as the error \smash{$\mathcal E(K_0)+\wt{\mathcal E}(K_0)$} is sufficiently small. The proof of Theorem \ref{thm_consist_Sigma} is also rooted on the asymptotic expansion of the empirical spiked eigenvectors detailed in Section \ref{new.Sec4.3}. We also would like to point out that the assumption of $\cal E(K_0)=\OO(1)$ is mild since it holds that $\sum_{k=1}^{K} d_k  v_k(i)  v_k (j)=\OO(\theta)$. 

For the pairwise test with test statistic $T_{ij}$ in \eqref{eq: T-test} and the group test with test statistic $T$ in \eqref{eq: T-test_group}, we will suggest different choices of parameter $K_0$. Specifically, for testing a given pair of nodes, we propose to use the estimate
\beq\label{whK}
\wh K_0:=\max\left\{ k \in [n]: |\wh d_k| \ge \check{q}(\log n)^{1/2} \cdot \log\log n \right\},
\eeq
where $\check{q}>0$ and $\check{q}^2:=\max_{j \in [n]} \sum_{l=1}^n X_{lj}$ is the maximum node degree of the network. For the group test, we suggest the use of the estimate
\beq\label{whK_group}
\wh K_0:=\max\left\{ k \in [n]: |\wh d_k| \ge \check{q}(\log n)^{3/2} \cdot \log\log n \right\}.
\eeq
Note that it follows from a simple concentration inequality that with probability $1-\oo(1)$,% (see e.g., Bernstein's inequality in Lemma \ref{lemma_bern} below), 
\beq\label{checkq_q}
\check{q}^2 = \left[1+\oo(1)\right]\max_{ j \in [n]} \sum_{l=1}^n \E X_{lj} \sim n\theta = q^2.
\eeq
We have used the factor of $\log\log n$ for definiteness in both \eqref{whK} and \eqref{whK_group} above, but one can replace it with another sequence $C_n\to \infty$ as $n\to \infty$. See also, e.g., the recent work \cite{HanYangFan2022} for rank inference in the network setting.

Let us gain some insights into the estimates $\wh K_0$ introduced in \eqref{whK} and \eqref{whK_group}. Assume that assumption \eqref{cond:Pi} holds. Denote by $K_1:=\max\{1 \leq k\le K: d_k \ge q[(\log n) \cdot (\log\log n)]^{1/2}\}$ and ${ K_2:=\max\{1 \leq k\le K: d_k \ge q (\log n)^{3/2} \cdot (\log\log n)^{1/2}\} }$. 
In view of \eqref{checkq_q}, $K_1$ and $K_2$ are essentially the upper bounds for $\wh K_0$ defined in \eqref{whK} and \eqref{whK_group}, respectively. In the definitions of $K_1$ and $K_2$, the choice of the factor $(\log\log n)^{1/2}$ is arbitrary and can be replaced by another factor $1\le C_n \ll \log\log n$. Let us further assume that 
\beq\label{eq: condK-app}
1\le K\ll \left(\frac{q}{\log n}\right)^{1/2}\wedge \frac{|d_{K_1}|}{q}
\eeq
and choose parameter $K_0$ as $\wh K_0$ in \eqref{whK}. Then from \eqref{Vmax0} and \eqref{checkq_q}, we can obtain that with probability $1-\oo(1)$,
\beq\label{eq:small_E}
\begin{split}
    \cal E(\wh K_0)+ \wt{\cal E}(\wh K_0) &\lesssim \frac{K^2}{n\theta}q(\log n)^{1/2} \cdot \log\log n+ \frac{q\sqrt{ \log n }}{q(\log n)^{1/2} \cdot \log\log n} + \frac{ \sqrt{K\log n}}{q} \\
    &= \frac{K^2}{q}(\log n)^{1/2} \cdot \log\log n  + \frac{ \sqrt{K\log n}}{q} +\oo(1)\ll 1.
\end{split}
\eeq
Similarly, if we assume that 
\beq\label{eq: condK2-app}
1\le K\ll \frac{q^{1/2}}{ (\log n)^{3/2} }\wedge \frac{|d_{K_2}| }{ q \sqrt{\log n} }
\eeq
%\end{enumerate} 
and choose parameter $K_0$ as $\wh K_0$ in \eqref{whK_group}, then it holds that 
\beq\label{eq:small_E2}
{ \cal E(\wh K_0)+ \wt{\cal E}(\wh K_0)  \ll (\log n)^{-1}.}
\eeq

With the aid of \eqref{eq:small_E}, \eqref{eq:small_E2}, and Theorem \ref{thm_consist_Sigma}, we can establish in the theorem below the counterparts of Theorems \ref{thm:pair-null}, \ref{thm:group-null}, and \ref{thm:group-alt} with corresponding new test statistics $\wh T_{ij}$ and $\wh T$ constructed using estimates $\wh K_0$ and $\wh \bSig_{i,j}$.

\begin{theorem}\label{thm:pg-sample}
%Assume that $\Omega$ holds with probability $1-\oo(1)$. 
%Denote by $K_1:=\max\{k\le K: d_k \ge q(\log n \cdot \log\log n)^{1/2}\}$ and $K_2:=\max\{k\le K: d_k \ge q \log n \cdot (\log\log n)^{1/2}\}$. 
Assume that Condition \ref{main_assm} and \eqref{cond:Pi} hold, and $K_1 \le K_{\max}\wedge C_0 $. Consider the test statistics $\wh T_{ij} $ and $\wh T $ constructed by replacing $\bSig_{i,j}$ with $\wh\bSig_{i,j}$ in the definitions of $T_{ij} $ and $T $ in \eqref{eq: T-test} and \eqref{eq: T-test_group}, respectively. Then 

\begin{enumerate}
\item Under assumption \eqref{eq: condK-app}, with $K_0$ replaced with $\wh K_0$ given in \eqref{whK}, we have a) under $H_0$, \eqref{eq_null} in Theorem \ref{thm:pair-null} holds for $\wh T_{ij}(\wh K_0)$ if $c_{1n} \ll [d_{1}\lambda_1(\bP)]^{-\frac12}$, and b) under $H_a$, \eqref{eq_power_pair} in Theorem \ref{thm:pair-null} holds for $\wh T_{ij}(\wh K_0)$ if  $c_{2n} \gg [d_{K}\lambda_K(\bP)]^{-\frac12}$ and \eqref{DV-V} holds almost surely. %Theorem \ref{thm:pair-null} holds for $\wh T_{ij}(\wh K_0)$  with $\wh K_0$ given in \eqref{whK} under assumption \eqref{eq: condK-app}. %(recall \eqref{eq: condKs})

\item {Under assumptions $1\ll m \ll q/\sqrt{K}$ and \eqref{eq: condK2-app},  replacing $K_0$ with $\wh K_0$ given in \eqref{whK_group}, %for $\wh T(\wh K_0)$ with $\wh K_0$, 
we have a) under $H_0$, \smash{$\wh T(\wh K_0)$} satisfies \eqref{eq:conv_Gumbel} in Theorem \ref{thm:group-null} if $c_{1n} \ll [d_{1}\lambda_1(\bP)]^{-1/2}(\log n)^{-1/2}$, and (b) under $H_a$, $\wh T(\wh K_0)$ satisfies \eqref{eq_power_group} in Theorem \ref{thm:group-alt} if $c_{2n}\gg [d_{K}\lambda_{K}(\bbP)]^{-1/2}\sqrt{\log n}$ and \eqref{DV-Vgp} holds almost surely.}

\end{enumerate}
\end{theorem} 

We emphasize that the analyses and results above suggest that $\wh K_0$ does not need to be a consistent estimator of any population parameter for the asymptotic null and alternative distributions to remain valid. This is another major distinction of our results here from those in \cite{SIMPLE}.  Theorem \ref{thm:pg-sample} also suggests that for testing the null hypothesis $H_0$ in (\ref{eq: hypothesis}) with each prespecified significance level $\alpha\in (0,1)$, we can construct the rejection region 
$$
\left\{\widehat T_{ij}(\widehat K_0) \geq F^{-1}_{\hat K_0}(1-\alpha)\right\}
$$
for a pair of given nodes, and the rejection region 
$$
\left\{\widehat T(\widehat K_0) \geq 2\mathcal G^{-1}(1-\alpha) + b_m(\wh K_0)\right\}
$$
for a group of given nodes. Then Theorem  \ref{thm:pg-sample} guarantees that both forms of the SIMPLE-RC test have asymptotic size $\alpha$ and asymptotic power one under their respective conditions.

\subsection{Test statistics under degree heterogeneity} \label{new.Sec3.4}

We further investigate the more general case with degree heterogeneity. To make this form of our SIMPLE-RC test concrete, we will focus on the degree-corrected mixed membership (DCMM) model given in (\ref{new.DCMM2}). The key identity \eqref{eq: pi-diff-weighted} now takes the form of 
\begin{equation}\label{eq: pi-diff-weighted2}
\left[\bV(i)-\bV(j)\right]^T\bD\left[\bV(i)-\bV(j)\right]= \left(\vartheta_i \bpi_i-\vartheta_j \bpi_j\right)^T\bP\left(\vartheta_i \bpi_i-\vartheta_j \bpi_j\right).
\end{equation}

In light of (\ref{eq: pi-diff-weighted2}) and (\ref{eq: pi-diff-weighted}), we see that the network inference procedures developed in Sections \ref{new.Sec3.1} and \ref{sec:group-test}  cannot be applied directly here because of degree heterogeneity. 
%On the other hand, in the current paper, we focus on a special yet important case of \eqref{eq: hypothesis}, that is, the hypothesis testing for the DCMM model with sharp null $c_{1n}=c_{2n}=0$. In this case, 
%We adopt the idea in \cite{SIMPLE} and consider statistics formed from ratios of eigenvectors (through columnwise division). 
To motivate the second form of the SIMPLE-RC method, let us make a useful observation. It follows from the representation $\bbH= \bTheta \bPi \bP \bPi^\top \bTheta=\bV\bD\bV^\top$ that 
\beq\label{VthetaB}
\bV=\bTheta \bPi\bB
\eeq
where $\bB:= \bP \bPi^\top \bTheta \bV \bD^{-1}$. Then from (\ref{VthetaB}), we can show that 
\beq\label{diff_ratios}
\begin{split}
\frac{\bV_{K_0}(i)}{v_1(i)} - \frac{\bV_{K_0}(j)}{v_1(j)} %&= \frac{\bB_{K_0}^\top \bpi_i }{\bpi_i^{\top} \bB \be_1}- \frac{\bB_{K_0}^\top \bpi_j}{\bpi_j^{\top} \bB\be_1}\\
= \frac{\vartheta_i\vartheta_j}{v_1(i)v_1(j)}\left[\bB_{K_0}^\top \left(\bpi_i \bpi_j^\top- \bpi_j \bpi_i^\top\right) \bB\be_1 \right],
\end{split}
\eeq
where $\bV_{K_0}(i)$ denotes the vector obtained by taking the transpose of the $i$th row of matrix $\bV_{K_0}$ and $\bB_{K_0}$ represents the submatrix formed by the first $K_0$ columns of matrix $\bB$. 
%$\be_1$ is the standard unit vector along the first coordinate axis in $\R^{K_0}$ and $\bI_{K\times K_0} $ and 
% $\bB_{K_0}\in \R^{K\times K_0}$ is defined as
% \beq\label{defBK0}
%  \bB_{K_0}:=\bB\bI_{K\times K_0}= \bP \bPi^\top \bTheta \bV \bD^{-1}\bI_{K\times K_0}= \bP \bPi^\top \bTheta \bV_{K_0}\bD_{K_0}^{-1},
% \eeq
% with $\bI_{K, K_0}:=(\bI_{K_0} , \mathbf 0 )^\top \in \R^{K\times K_0}$. 

%stopped here
Under the null hypothesis $H_0$ in \eqref{eq: hypothesis} and Condition \ref{main_assm_DCMM} to be introduced later, it holds that 
\begin{align}
\left\|\bD_{K_0}\left[\frac{\bV_{K_0}(i)}{v_1(i)} - \frac{\bV_{K_0}(j)}{v_1(j)}\right] \right\|  \le  q \sqrt{Kd_1\lambda_1(\bP)}c_{1n}; \label{diff_ratios_null}
\end{align}
see Section \ref{sec:proof-diff_ratios_null} of the Supplementary Material for details. On the other hand, we have another representation 
\beq\label{diff_ratios2}
\frac{\bV(i)}{v_1(i)} - \frac{\bV(j)}{v_1(j)} =\bB^\top \left(\bpi_i , \bpi_j\right)\left( \frac{\vartheta_i}{v_1(i)}, -\frac{\vartheta_j}{v_1(j)}\right)^\top ,
\eeq
with which we can obtain that 
\begin{align}
 \left\|\bD\left[\frac{\bV(i)}{v_1(i)} - \frac{\bV(j)}{v_1(j)}\right] \right\| \gtrsim  q \sqrt{d_{K}\lambda_K(\bP)} \lambda_{\min}^{1/2}\left\{ \left(\bpi_i , \bpi_j\right)^\top  \left(\bpi_i , \bpi_j\right)\right\}; \label{diff_ratios_alt}
\end{align}
see Section \ref{sec:proof-diff_ratios_null} for detailed derivations. Similar to \eqref{DV-Vgp}, for power analysis we assume that 
\beq\label{DV-Vgp_het}
\begin{split}
\max_{\{i, j\} \subset \cal M}\left\|\bD_{K_0}\left[\frac{\bV_{K_0}(i)}{v_1(i)} - \frac{\bV_{K_0}(j)}{v_1(j)}\right]\right\| \ge c\max_{\{i, j\} \subset \cal M}\left\|\bD \left[\frac{\bV (i)}{v_1(i)} - \frac{\bV (j)}{v_1(j)}\right]\right\|
\end{split}
\eeq
for a constant $c>0$. The observations in (\ref{diff_ratios_null}) and (\ref{diff_ratios_alt}) above have motivated us to construct test statistics based on the ratio statistic on the left-hand side of \eqref{diff_ratios}. Such idea has been investigated in \cite{Jin2015} for a different goal of community detection and \cite{SIMPLE} for testing the sharp nulls for a pair of given nodes under strong signals.

Specifically, for given $1 \leq K_0\le K$,  we define vectors $\bY_i(K_0)\in \R^{K_0-1}$ with components
\beq\label{ratio_ev}
Y_i(k):=\frac{\wh v_k(i)}{\wh v_1(i)}
\eeq
for $2\le k \le K_0$ and $ i \in [n]$, where we adopt the convention of defining $0/0$ as $1$. Based on vectors $\bY_i$'s, we formally define the second form of our SIMPLE-RC test statistic $\mathcal T(K_0)$ as 
\begin{align}\label{eq: T-test DCMM}
\mathcal T(K_0)= \max_{\{i,j\}\in {\cal P} } \mathcal T_{ij}(K_0),
\end{align}
where $\cal P$ denotes the random coupling collection as in (\ref{eq: T-test_group}), we define the test statistics \smash{$\mathcal T_{ij} (K_0)= \left(\bY_i -\bY_j\right)^T[\bSig^{(2)}_{i,j}(K_0)]^{-1}\left(\bY_i -\bY_j\right)$} as in \cite{SIMPLE}, and \smash{$\bSig^{(2)}_{i,j}(K_0)$} represents the asymptotic covariance matrix of $\bY_i - \bY_j$. In particular, we have  $\bSig^{(2)}_{i,j}(K_0):=\cov(\bbf^{(i,j)}(K_0))$, where   {$\bbf^{(i,j)}(K_0):=(f_2^{(i,j)}, \cdots, f_{K_0}^{(i,j)})\in \R^{K_0-1}$} has components
\beq\label{asymp-variance2}
f_k^{(i,j)}:= \frac{\be_i^\top \bW\bv_{k}}{t_k v_1(i)}- \frac{\be_j^\top \bW\bv_{k}}{t_k v_1(j)}- \frac{v_k(i)\be_i^\top \bW\bv_{1}}{t_1 v_1^2(i)} + \frac{v_k(j)\be_j^\top \bW\bv_{1}}{t_1 v_1^2(j)}
\eeq
for $2 \leq k \leq K_0$, and the population quantities $t_k$ with $1\le k\le K_0$ will be defined later in \eqref{eq_evalue} of the Supplementary Material and they are in fact the asymptotic limits of the empirical spiked eigenvalues $\wh d_k$ (see Theorem \ref{asymp_evalue} in Section \ref{new.Sec4.3} for details).

To simplify the technical presentation, let us introduce some additional notation
$$\vartheta_{\min}:=\min_{i\in [n]}\vartheta_i, \ \  \vartheta_{\max}:=\max_{i\in [n]}\vartheta_i, \ \  \theta:=\frac{1}{n}\sum_{i\in [n]} \vartheta_{i}^2, 
\ \ \text{and} \ \ q:=\sqrt{n\theta }.$$
To facilitate our technical analysis, we will need to introduce some basic regularity conditions. In particular, in addition to Condition \ref{main_assm}, we also require several additional assumptions for dealing with the more challenging case of degree heterogeneity as in \cite{SIMPLE}.

\begin{assu}\label{main_assm_DCMM}
Assume that parts (i)--(iv) of Condition \ref{main_assm} hold and there exists a constant $\e_3>0$ such that
\begin{itemize}
\item[(i)] (Degree heterogeneity) It holds that $\vart_{\min}\ge \e_3\vart_{\max}$.

\item[(ii)] (Covariance matrix) All the eigenvalues of $q^{-2}\wt\bbD_{K_0'}\bSig^{(2)}_{i,j}(K_0')\wt\bbD_{K_0'}$ are between $\e_3$ and $\e_3^{-1}$ for all $\{i,j\} \subset \cal M$ and $2\le K_0'\le K_0$, where $\wt\bbD_{K_0'}:=\diag\{d_2,\cdots ,d_{K_0'}\}$.

\item[(iii)] (Membership matrix) The bound \eqref{cond:Pi} holds as $\|(\bPi^\top \bPi)^{-1}\| \le \e_3^{-1} K/n$.

\item[(iv)] (Leading eigenvalue and eigenvector) It holds that $d_1\ge \e_3n\theta$, and all components of $\bv_1$ are positive satisfying that $\min_{i\in [n]} v_1(i)\ge \e_3/\sqrt{n}$. 
\end{itemize}

\noindent Moreover, for fixed constants $\e_0$ and $\e_3$, denote by $K_{\max}\equiv K_{\max}(n,\e_0,\e_3)\le K$ the largest $K_0$ such that parts (ii) and (iii) of Condition \ref{main_assm} and part (ii) above hold.
\end{assu}

Part (i) of Condition \ref{main_assm_DCMM} above means that the degrees of all nodes are of the same order.  Similar to part (v) of Condition \ref{main_assm}, part (ii) of Condition \ref{main_assm_DCMM} requires that the covariance matrix of $\bbf^{(i,j)}$ is well-behaved asymptotically. 
%\begin{remark}
Part (iv) of Condition \ref{main_assm_DCMM} is also natural as discussed in \cite{JKL2017}. %By the Perron–Frobenius theorem, we can always choose the direction of $\bv_1$ so that all its components are non-negative. Furthermore, as explained in \cite{JKL2017}, the condition $\min_i v_1(i)\ge \e_3/\sqrt{n}$ is guaranteed by the condition (iii) and that 
%$$ \frac{\max_k \eta_1(k)}{\min_k \eta_k(i)}\le C$$
%for a constant $C>0$, where $\bleta_1$ is the first right singular vector of $\bbP\bPi^\top \bTheta^2\bPi$. The above condition is mild and satisfied in several natural scenarios; we refer the reader to the discussions in Section 2 of \cite{JKL2017}. 
We expect that parts (i), (iii), and (iv) of Condition \ref{main_assm_DCMM} above can be relaxed to certain extent, but for simplicity of the technical presentation, we do not pursue such direction in the current paper. 
With the aid of Condition \ref{main_assm_DCMM},  \eqref{diff_ratios_null}, and \eqref{diff_ratios_alt}, we can establish in the theorem below the counterpart of Theorem \ref{thm:pair-null} for the case with degree heterogeneity.  %\ref{thm:group-null} and \ref{thm:group-alt} for the DCMM model.  

\begin{theorem}\label{thm:pair-null-het}
Assume that Condition \ref{main_assm_DCMM} holds, $K_0$ is a random variable such that $1 \leq K_0\le K_{\max}\wedge C_0 $ for some large constant $C_0>0$, $|d_{K_0}|\gg q\sqrt{\log n}$, and 
\beq\label{eq: condK02}
1\le K \ll \left(\frac{q}{\log n}\right)^{1/2}\wedge \frac{|d_{K_0}|}{q}
\eeq
almost surely. Then the test statistic $\mathcal T_{ij}(K_0)$ defined below (\ref{eq: T-test DCMM}) satisfies that 
\begin{enumerate}
    \item If $c_{1n} \ll [Kd_1\lambda_1(\bP)]^{-\frac12}$, it holds that under the null hypothesis $H_0$ in \eqref{eq: hypothesis},
    \beq \label{eq_null_het}
    \lim_{n\to \infty}\sup_{x\in \R}\left| \P\{\cal T_{ij}(K_0)\le x\}- F_{K_0-1}(x)\right|\to 0,
  \eeq
     where conditional on $K_0$, $F_{K_0-1}$ is the chi-square distribution with $K_0-1$ degrees of freedom.

    \item  { If  $c_{2n} \gg [d_{K}\lambda_K(\bP)]^{-\frac12}$ and \eqref{DV-Vgp_het} is satisfied almost surely}, it holds that under the alternative hypothesis $H_a$ in \eqref{eq: hypothesis-a}, for each arbitrarily large constant $C>0$,
% \beq\label{hypothesis1_DCMM}
% \lambda_{\min}\left( \left(\bpi_i , \bpi_j\right)^\top  \left(\bpi_i , \bpi_j\right)\right) \gg  |d_{K_0}|^{-1},
% \eeq
\beq\label{eq_power_pair_DCMM}
\lim_{n\to \infty}P\{\cal T_{ij}(K_0)>C\} = 1. % \quad \text{ as } n\rightarrow \infty.
\eeq
\end{enumerate}
\end{theorem}

%Similar to our discussion below Theorem \ref{thm:pair-null}, the above theorem improves the corresponding result, Theorem 3, of \cite{SIMPLE} for the DCMM model in the following senses: it considers the hypothesis testing with non-sharp nulls; it allows for a diverging number of communities $K$; it relaxes the lower bound on $q$; it relaxes the lower bound on the signal-to-noise ratio. Note that, compared with Theorem \ref{thm:pair-null}, the condition \eqref{eq: condK02} is slightly weaker than \eqref{eq: condKs}, and we require a better upper bound on $c_{1n}$ by a factor $K^{-1/2}$. 

Since we have used the first empirical spiked eigenvector as a reference point in defining the ratio $Y_i(k)$ in \eqref{ratio_ev}, it is natural that one degree of freedom will be lost, which explains the asymptotic null distribution of $\chi_{K_0 -1}^2$ revealed in Theorem \ref{thm:pair-null-het} above. In comparison to Theorem \ref{thm:pair-null}, %assumption \eqref{eq: condK02} is slightly weaker than constraint \eqref{eq: condKs}, and 
we require a better upper bound on $c_{1n}$ by a factor of \smash{$K^{-1/2}$}. 

Based on Theorem \ref{thm:pair-null-het} and exploiting the ideas introduced in Section \ref{sec:group-test}, we can further extend Theorems \ref{thm:group-null} and \ref{thm:group-alt} to the case with degree heterogeneity in the theorem below.

\begin{theorem}\label{thm:group-het}
Assume that Condition \ref{main_assm_DCMM} holds, $K_0$ is a random variable such that $1 \leq K_0\le K_{\max}\wedge C_0 $ almost surely for some large constant $C_0>0$, $|d_{K_0}|\gg q {\log n}$ almost surely, 
\beq\label{eq: condK0222}
1\le K\ll \left(\frac{q}{(\log n)^{3/2}}\right)^{1/2}\wedge \frac{|d_{K_0}|}{q\sqrt{\log n}}, \ \text{ and } \ 1\ll m \ll \frac{q}{ {K}^{3/2}}.
\eeq
Then the SIMPLE-RC test statistic $\mathcal T$ defined in (\ref{eq: T-test DCMM}) satisfies that 
\begin{enumerate}
    \item If $c_{1n} \ll [Kd_1\lambda_1(\bP)]^{-1/2} (\log n)^{-1/2}$, it holds that under the null hypothesis $H_0$ in \eqref{eq: hypothesis}, 
  \beq\label{eq:conv_Gumbel_het}
\lim_{n\to \infty}\sup_{x\in \R}\left| \P\left\{\frac{\cal T(K_0)-  b_m(K_0-1)}{2}\le x\right\}-\cal G(x) \right|\to 0,
\eeq
%as $n \rightarrow \infty$, 
where $\cal G(x)$ denotes the Gumbel distribution and $b_m(K_0-1)$ is as given in (\ref{eq_ambm}). 

\item 
% If  
% \beq\label{hypothesis1_DCMM_alt} \max_{i\ne j}\lambda_{\min}\left( \left(\bpi_i , \bpi_j\right)^\top  \left(\bpi_i , \bpi_j\right)\right) \gg   {\log n}/{|d_{K_0}|}, \eeq
{If $c_{2n}\gg [d_{K}\lambda_{K}(\bbP)]^{-1/2}\sqrt{\log n} $ and \eqref{DV-Vgp_het} is satisfied almost surely}, it holds that under the alternative hypothesis $H_a$ in \eqref{eq: hypothesis-a}, 
\beq\label{eq_power_group-het}
    \lim_{n\to \infty}\P\left\{\frac{\cal T(K_0)-  b_m(K_0-1)}{2}>C\right\} =1
    \eeq
    for each arbitrarily large constant $C>0$. 
    %{\cob It is not necessary for $K_0$ to be independent of other sources of randomness. Let $\Omega(K_0)$ be an event depending on $K_0$ that appears in the proof. Then we decompose it as $\P(\Omega(K_0); K_0=a)=\P(\Omega(a); K_0=a)=\P(\Omega(a))\P( K_0=a)$ {\color{red}Comment on 30th: do we need the event $\Omega(a)$ and $K_0$ to be independent for the last step to hold?} {\color{cyan} Yes, we need $\Omega(a)$ and $K_0$ to be independent. Since our proofs only use Theorems \ref{asymp_evalue} and \ref{asymp_evector}, where \eqref{eq_evalue_conv} and \eqref{expansion_evector2} hold for any $k\le K_{max}$. Hence, $\Omega(a)$ and $K_0$ are independent as long as $a\le K_{\max}$.} for $a\le K_{\max}\wedge C_0$, because after taking $K_0$ as a fixed integer $a$ in an event we are interested in gives an event that does not depend on $K_0$ anymore (it is crucial that we have assumed $K_0\le K_{\max}$ for this to be true). Will add this discussion to Section \ref{subsec:cal-thm-pair}. }
\end{enumerate}
\end{theorem}

% By taking the ratio \eqref{ratio_ev}, we lose one degree of freedom, so we have used $b_m(K_0-1)$ in the above statement. Note that, compared with Theorem \ref{thm:group-null}, the condition \eqref{eq: condK0222} is slightly weaker than \eqref{eq: condK2} and \eqref{cond_m}, and we require a better upper bound on $c_{1n}$ by a factor $K^{-1/2}$. 

%From Theorem \ref{thm:group-het} above, we see that the use of componentwise eigenvector ratios in \eqref{ratio_ev} results in the loss of one degree of freedom as reflected in the parameter $b_m(K_0-1)$ involved in the asymptotic null distribution. 

As discussed in Section \ref{new.Sec3.3}, for the practical implementation of the degree heterogeneity form of our SIMPLE-RC test statistics $\mathcal T_{ij}$ and $\mathcal T$ introduced in (\ref{eq: T-test DCMM}), we will require a consistent estimate of the covariance matrix \smash{$\bSig^{(2)}_{i,j}$}. Some direct calculations show that for each $1\le a,b \le K_0-1$, 
\beq\label{Sigmaij2}
\begin{split}
 \left(\bSig^{(2)}_{i,j}\right)_{ab} &= \sum_{ l\in [n]\setminus \{j\}} \sigma^2_{il} \left[\frac{v_{a+1}(l)}{t_{a+1}v_1(i)}-\frac{v_{a+1}(i)v_1(l)}{t_{1}v_1^2(i)} \right]\left[\frac{v_{b+1}(l)}{t_{b+1}v_1(i)}-\frac{v_{b+1}(i)v_1(l)}{t_{1}v_1^2(i)} \right]   \\
& +\sum_{ l\in [n]\setminus \{i\}} \sigma^2_{jl} \left[\frac{v_{a+1}(l)}{t_{a+1}v_1(j)}-\frac{v_{a+1}(j)v_1(l)}{t_{1}v_1^2(j)} \right]\left[\frac{v_{b+1}(l)}{t_{b+1}v_1(j)}-\frac{v_{b+1}(j)v_1(l)}{t_{1}v_1^2(j)} \right]\\
& + \sigma_{ij}^2 \left[ \frac{v_{a+1}(j)}{t_{a+1}v_1(i)}-\frac{v_{a+1}(i)v_1(j)}{t_{1}v_1^2(i)}-\frac{v_{a+1}(i)}{t_{a+1}v_1(j)}+\frac{v_{a+1}(j)v_1(i)}{t_{1}v_1^2(j)}\right] \\
& \quad \times \left[ \frac{v_{b+1}(j)}{t_{b+1}v_1(i)}-\frac{v_{b+1}(i)v_1(j)}{t_{1}v_1^2(i)}-\frac{v_{b+1}(i)}{t_{b+1}v_1(j)}+\frac{v_{b+1}(j)v_1(i)}{t_{1}v_1^2(j)}\right].
\end{split}
\eeq

We again define estimate {$\wh K_0$} as in \eqref{whK} or \eqref{whK_group} accordingly and suggest the use of covariance matrix estimator  \smash{$\wh{\bSig}^{(2)}_{i,j}(K_0)$} obtained by replacing $t_k$, $\bv$, and $\sigma^2_{ab}$, respectively, with $\wh d_k$, $\wh \bv$, and $\wh w^2_{ab}$ in \eqref{Sigmaij2} (recall \eqref{whW}). Then similar to Theorem \ref{thm:pg-sample}, we can establish the results in the theorem below.

\begin{theorem}\label{thm:group-het-sample}
Assume that Condition \ref{main_assm_DCMM} holds and $K_1 \le K_{\max}\wedge C_0 $. Consider the test statistics $\wh {\cal T}_{ij}$ and $\wh {\cal T}$ constructed by replacing $\bSig^{(2)}_{i,j}$ with  $\wh\bSig^{(2)}_{i,j}$ in the definitions of $\cal T_{ij}$ and $\cal T$ in (\ref{eq: T-test DCMM}), respectively. Then 

{
\begin{enumerate}
   % \item Theorem \ref{thm:pair-null-het} holds for $\wh {\cal T}_{ij}(\wh K_0)$ with $\wh K_0$ given in \eqref{whK} if \eqref{eq: condK-app} holds. 
   
 \item    Under assumption \eqref{eq: condK-app}, replacing $K_0$ with $\wh K_0$ given in \eqref{whK}, we have a) under $H_0$, $\wh{\cal T}_{ij}(\wh K_0)$ satisfies \eqref{eq_null_het} in Theorem \ref{thm:pair-null-het} if  $c_{1n} \ll [Kd_1\lambda_1(\bP)]^{-\frac12}$, and b) under $H_a$, $\wh{\cal T}_{ij}(\wh K_0)$ satisfies  \eqref{eq_power_pair_DCMM} in Theorem \ref{thm:pair-null-het} if $c_{2n} \gg [d_{K}\lambda_K(\bP)]^{-\frac12}$ and \eqref{DV-Vgp_het} holds almost surely. % for $K_0=\wh K_0$.

\item Under assumptions $1\ll m \ll  {q}/{ {K}^{3/2}}$ and \eqref{eq: condK2-app}, replacing $K_0$ with $\wh K_0$ given in \eqref{whK_group},  we have a) under $H_0$, $\wh{\cal T}(\wh K_0)$ satisfies \eqref{eq:conv_Gumbel_het} in Theorem \ref{thm:group-het} if $c_{1n} \ll [Kd_1\lambda_1(\bP)]^{-1/2} (\log n)^{-1/2}$, and b) under $H_a$, $\wh{\cal T}(\wh K_0)$ satisfies \eqref{eq_power_group-het} in Theorem \ref{thm:group-het} if $c_{2n}\gg [d_{K}\lambda_{K}(\bbP)]^{-1/2}\sqrt{\log n} $ and \eqref{DV-Vgp_het} holds almost surely. % for $K_0 = \wh K_0$.

%Under $H_0$, \eqref{eq_null_het} in Theorem \ref{thm:group-het} holds  \eqref{eq:conv_Gumbel_het} in Theorem \ref{thm:group-het} holds Under assumptions  $1\ll m \ll  {q}/{ {K}^{3/2}}$ and \eqref{eq: condK2-app}, 

%Theorem \ref{thm:group-het} holds for $\wh {\cal T}(\wh K_0)$ with $\wh K_0$ given in \eqref{whK_group} if {\cor $1\ll m \ll  {q}/{ {K}^{3/2}}$} and \eqref{eq: condK2-app} holds.
\end{enumerate}
} 
\end{theorem}

%The proofs of Theorems \ref{thm:pair-null-het}--\ref{thm:group-het-sample} will be given in Sections \ref{appd_pair_het}--\ref{appd_group_het_sample}. 
Similar to the remark after Theorem \ref{thm:pg-sample}, $\wh K_0$ does not need to consistently estimate any population parameter for the asymptotic distributions to remain valid.  We can similarly construct an asymptotic level $\alpha$ test with the rejection region
$$
\left\{\wh T_{ij}(\wh K_0) \geq F^{-1}_{\wh K_0-1}(1-\alpha) \right\}
$$
for testing a pair of given nodes, and an asymptotic level $\alpha$ test with the rejection region 
$$
\left\{\wh T_{ij}(\wh K_0) \geq  2 \mathcal G^{-1}(1-\alpha) + b_m(\wh K_0-1)\right\}
$$
for testing a group of given nodes, where $\alpha \in (0, 1)$ and the null hypothesis $H_0$ is given in (\ref{eq: hypothesis}). The corresponding powers under the alternative hypothesis $H_a$ in \eqref{eq: hypothesis-a} are asymptotically one under the conditions of Theorem \ref{thm:pair-null-het}(ii) and Theorem \ref{thm:group-het}(ii), respectively.

%\newpage
%\section{A general theory on asymptotic expansions of empirical spiked eigenvectors} \label{new.Sec4}
\section{Roadmap of proofs and related works}\label{new.Sec4}
%In this section, we will lay a general theoretical foundation of the asymptotic theory of empirical spiked eigenvectors that is key to establishing the theoretical properties of the SIMPLE-RC presented in Section \ref{new.Sec3}. % for group network inference under non-sharp nulls and weak signals.

%\subsection{Roadmap of proofs and related works} \label{new.Sec4.4}

%To ease the reading, we will begin with a high-level description of our proof strategies and explain the new mathematical challenges in relation to existing works on network inference and random matrix theory (RMT). 
%As mentioned before, our theory presented in Section \ref{new.Sec3} is enpowered by the new asymptotic expansions of empirical spiked eigenvectors that we establish in this paper.  To better appreciate our technical contribution, we provide a high-level description of our proof strategies in establishing the empirical eigenvector explanations, and explain the new mathematical challenges in relation to existing works on network inference and random matrix theory (RMT). Formal results and proofs can be found in Supplementary file. 

As mentioned in the Introduction, our technical analyses are empowered by the asymptotic expansion for empirical spiked eigenvectors presented in Theorem \ref{asymp_evector} in Section \ref{new.Sec4.3} of the Supplementary Material.  To better appreciate our technical contribution, we provide a high-level description of our proof strategies in establishing the empirical eigenvector expansions, and explain the new mathematical challenges in relation to existing works on network inference and random matrix theory (RMT). Formal results and proofs can be found in the Supplementary Material. 

Hereafter, we will \emph{rescale} the $n \times n$ random matrix $\bbX$ as
\beq\label{scaling_eq}
\bbX \to \bbX/q,
\eeq
which  has been  commonly used in the RMT literature. As a consequence, the eigenvalues of the noise matrix $\bbW$ will be of order $O(1)$ with high probability under the rescaling in \eqref{scaling_eq}; see Lemma \ref{lem_opbound} in Section \ref{new.Sec4.2} for details. With a slight abuse of notation, we still denote the rescaled random matrix $\bbX$ as \eqref{eq: model.general}. Observe that the noise matrix $\bbW$ now satisfies that 
\beq\label{support-W}
\max_{i,j}|W_{ij}|\le \frac{1}{q} \ \text{ and } \ s_{ij}:= \E|W_{ij}|^2 %= \frac{h_{ij} (1-h_{ij})}{q^2}
\lesssim \frac1n.
\eeq
In this paper, we say that an event $\Omega$ holds \emph{with high probability} (w.h.p.) if for each constant $D>0$, we have 
%\beq\label{WHP}
$\P(\Omega^c)\le n^{-D}$
%\eeq
as long as $n$ is sufficiently large. 

%{\cob need revisions} Our theoretical analysis of the test statistics is en-powered by a new random matrix theory built in our paper, and can be of independent interests to the statistical community. Our key theoretical result, Theorem \ref{asymp_evector}, gives an asymptotic expansion of the spiked empirical eigenvectors of the random adjacency matrix. This result improves the corresponding ones in \cite{ATE} (i.e., Theorem 2) and \cite{SIMPLE} (i.e., Lemma 9) in the following senses: in the current paper, we deal with sparser networks, weaker signal strengths, and possibly diverging number of signals. To give more details, we consider an undirected random graph with $n$ nodes, a low-rank latent structure and sparsity level $0<\theta\ll 1$. Then, its adjacency matrix $\bX$ is an $n\times n$ symmetric random matrix whose upper triangular entries are independent Bernoulli random variables, with about $\theta$ fraction of non-zero entries in each row. We can decompose it into a deterministic mean matrix $\bH=\E \bX $ plus a random noise matrix $\bW=\bX-\E\bX$. The noise matrix $\bbW$ is called a Wigner type random matrix in the literature, meaning that it is a symmetric random matrix with centered independent (up to symmetry) entries. Suppose the mean matrix $\bH$ is of low rank $K$ and has an eigen-decomposition $\bH=\bV\bD\bV^\top$, where $\bbD=\diag(d_1,\cdots,d_K)$ is the diagonal matrix of eigenvalues and $\bbV=(\bbv_1,\cdots,\bbv_k) \in \mathbb{R}^{n\times K}$ is the corresponding matrix of eigenvectors. 

 Similar to \cite{ATE} and many other RMT works on empirical eigenvectors, our starting point is the use of the Cauchy integral formula from complex analysis. Specifically, denote by \smash{$\wh d_k$} a non-degenerate empirical outlier (i.e., spiked) eigenvalue and $\wh \bv_k$ the corresponding empirical spiked eigenvector. Then it holds that  
\beq\label{eq:Cauchy1}
\wh\bv_k \wh\bv_k^\top =  -\frac{1}{2\pi \ii}\oint_{\cal C_k} \frac{1}{\bbX-z}  \dd z,\eeq
where $\cal C_k$ represents a contour in the complex plane $\mathbb{C}$ that encloses only eigenvalue $\wh d_k$ and no other eigenvalues of the random matrix $\bX$. The integration formula (\ref{eq:Cauchy1}) above will allow us to calculate the asymptotic expansion of the bilinear form $\bx^\top \wh\bv_k \wh\bv_k^\top \by$ for arbitrary deterministic unit vectors $\bx,\by\in \R^n$, which can be further used to derive the limiting distribution of the linear form $\bx^\top \wh \bv_k$ for any deterministic unit vector $\bx\in \R^n$ (modulo the sign change). With an application of the Woodbury matrix identity to representation \eqref{eq:Cauchy1}, we can obtain that 
\beq\label{eq:Cauchy2}
\bx^\top \wh\bv_k \wh\bv_k^\top \by =  -\frac{1}{2\pi \ii}\oint_{\cal C_k} \left[ \bx^\top  \bG(z) \by- \bx^\top \bG(z)  \bV \frac{1}{\bbD^{-1} + \bV^\top \bG(z)\bV} \bV^\top \bG(z) \by\right] \dd z,
\eeq
where %$\bG(z):=(\bW-z)^{-1}$ 
\beq\label{eqn_defG}\bG(z):= (\bbW-z\bI)^{-1},\quad z \in \mathbb{C},\eeq
denotes the Green's function (i.e., the resolvent) of the noise random matrix $\bW$. To calculate the right-hand side (RHS) of \eqref{eq:Cauchy2} above, we will need to characterize the asymptotic behavior of $\bx^\top \bG(z) \by$ for any deterministic unit vectors $\bx,\by\in \R^n$. We expect that $\bx^\top \bG(z) \by$ converges to a deterministic limit as the network size $n$ increases, which is referred to as the \emph{anisotropic local law} in the RMT literature; see, e.g., \cite{isotropic,KY_isotropic,Anisotropic}.    

A major mathematical contribution of our paper is deriving a sharper anisotropic local law for the resolvent $\bG(z)$ under weaker conditions on the sparsity level (i.e., smaller $\theta$) and signal strength. It is known that the extreme eigenvalues of the noise random matrix $\bW$ under the rescaling in \eqref{scaling_eq} are of order $O(1)$; see Lemma \ref{lem_opbound} for details. For strong non-degenerate signals $|d_k|\gg 1$ under the rescaling in \eqref{scaling_eq}, we can choose contour $\cal C_k$ such that $|z|\sim |d_k|$ for each $z\in \cal C_k$. Under the assumptions that the signal-to-noise ratio is at least $n^{c}$ (i.e., $|d_k|\ge n^c$) and the sparsity level is at least $\theta\ge n^{-1+c}$ for some small constant $c>0$, \cite{ATE} employed the idea of series expansion to investigate $\bx^\top \bG(z) \by$ through
\beq\label{series_W} \bx^\top(\bW-z \bI)^{-1}\by = - \sum_{l=0}^\infty z^{-(l+1)}\bx^\top\bW^{l}\by. \eeq
Using $|z|^{-1}\|\bW\| \sim |d_k|^{-1}\|\bW\| =\OO_p(n^{-c})$, \cite{ATE} truncated the above series at $l=L$ for some large but finite order $L$ so that the resulting error is of order $\OO_p(n^{-C})$ for some large constant $C>1$. Then, by exploiting the concentration of $\bx^\top\bW^{l}\by$ for each $1\le l\le L$,  the asymptotic limit of $\bx^\top(\bW-z\bI)^{-1}\by$ along with a convergence rate of $n^\e/q$ for an arbitrarily small constant $\e>0$ can be derived. Here, it is worth mentioning that $q^{-1}=(n\theta)^{-1/2}$ is the order of the optimal central limit theorem (CLT) error. 

In contrast, in this paper we will consider much weaker signals with the signal-to-noise ratio as low as $\sqrt{\log n}$ and much sparser networks with parameter $\theta$ as small as $(\log n)^8/n$. Hence, to obtain a small enough error of order $\OO_p(n^{-C})$, we need to truncate the series \eqref{series_W} at $l=C(\log n)/\log\log n$. We would like to mention that characterizing the concentration of $\bx^\top\bW^{l}\by$ for a diverging $l$ is much more challenging. Moreover, even if we can derive the convergence rate of $n^\e/q$ for an arbitrarily small constant $\e > 0$, it is useless for very sparse networks with $q\ll n^{\e}$. To address these challenging issues, a much finer and more delicate combinatorial argument for evaluating huge products of random matrices will be needed, which is a rather challenging task. Instead of using the series expansion \eqref{series_W}, our proof of the anisotropic local law is based on another method in RMT \citep{AjaErdKru2015,EKYY_ER1}, that is, we derive a self-consistent quadratic vector equation (QVE) satisfied by the diagonal entries of $\bG$ (see equation \eqref{1Gii} in the Supplementary Material and the corresponding deterministic QVE \eqref{QVE} for details). By analyzing the diagonal entries of $\bG$ through the QVE and bounding the off-diagonal entries of $\bG$  through %standard 
classical concentration inequalities, we can prove the \emph{entrywise local law} 
\beq\label{entry_intro}
\max_{i,j \in [n]}|G_{ij}(z)-\Pie_{ij}(z)| \lesssim \frac{1}{q|z|^2}
\eeq
w.h.p., where $\Pii(z)$ is the matrix limit of $\bG(z)$ as defined in \eqref{defn_pi} that is a diagonal matrix with entries satisfying the QVE (see Theorem \ref{lem_locallaw1} in Section \ref{new.Sec4.2} for the precise statement).

The arguments for the proof of \eqref{entry_intro} above have a similar flavor to those in \cite{AjaErdKru2015,EKYY_ER1}. However, the major technical challenge is to establish a stronger \emph{anisotropic local law} (see Theorem \ref{lem_locallaw2} in Section \ref{new.Sec4.2} for the precise statement): for any deterministic unit vectors $\bx,\by\in \R^n$, %we have  
\beq\label{aniso_intro}
\left|\bx^\top [\bG(z)-\Pii(z)]\by\right| \lesssim \frac{\log n}{q|z|^2}
\eeq
w.h.p. If we ignore the $\log n$ factor, \eqref{entry_intro} is a special case of \eqref{aniso_intro} with $\bx,\by$ being the standard basis unit vectors, but not vice versa since 
$$ \left|\bx^\top [\bG(z)-\Pii(z)]\by\right| \le \max_{  i,j \in [n]}|G_{ij}(z)-\Pie_{ij}(z)| \cdot \sum_{i,j \in [n]}|x(i)||y(j)| \lesssim \frac{n}{q|z|^2}$$
w.h.p., where we have used the facts $\|\bx\|_1\le \sqrt{n}\|\bx\|_2$ and $\|\by\|_1\le \sqrt{n}\|\by\|_2$. To get rid of the extra $n$ factor in the above bound, one needs to exploit the cancellation effect in the summation $\sum_{  i,j \in [n]}x(i)y(j)[G_{ij}(z)-\Pie_{ij}(z)]$ due to the random fluctuations of the resolvent entries. With the aid of the Markov inequality, proving \eqref{aniso_intro} above amounts to showing the high moment bound
\beq\label{highm_aniso2}
\E\left|\bx^\top [\bG(z)-\Pii(z)]\by\right|^{2r} \le \left(\frac{\log n}{q|z|^2}\right)^{2r}
\eeq
for $r\in \N$ as large as $\log n$. To further prove \eqref{highm_aniso2}, we will adopt an idea in \cite{HKR2018} based on cumulant expansions. More precisely, using the simple 
identity $\bG-\Pii=-\Pii(\bW-z-\Pii^{-1})\bG$, we write the left-hand side (LHS) of \eqref{highm_aniso2} as
$$\E\left|\bx^\top [\bG(z)-\Pii(z)]\by\right|^{2r} = - \E \left[\bx^\top \Pii(\bW-z-\Pii^{-1})\bG\by^\top Y^{r-1} \overline Y^{r-1}\right], $$
where $Y:=\bx^\top [\bG(z)-\Pii(z)]\by$. Then, the RHS above can be estimated by applying the cumulant expansions with respect to the entries of $\bbW$ and bounding each term through the entrywise local law in  \eqref{entry_intro}. 

%stopped here
 %In fact, to get the bound \eqref{highm_aniso2}, we will need to control some huge products of the resolvent entries and $r$-dependent combinatorial factors in a careful way, which is one main technical contribution of our proofs; see Section \ref{new.secB.claim1} of the Supplementary Material for the full details. 
Utilizing the anisotropic local law established in \eqref{aniso_intro}, we can calculate the contour integral in \eqref{eq:Cauchy2} and further derive the asymptotic expansion of the empirical spiked eigenvector $\wh \bv_k$ in Theorem \ref{asymp_evector} in Section \ref{new.Sec4.3}. Such expansion enables us to derive the limiting distribution of each eigenvector entry $ \wh v_k(i)$ with $i \in [n]$. Furthermore, the asymptotic expansion holds with high probability $1-\OO(n^{-D})$ for any large constant $D>0$, and thus holds simultaneously for all empirical spiked eigenvector entries $\wh v_k(i)$ with $i\in [n]$ and $1\le k\le K_{\max}$ by taking a union bound. Such a uniform result is crucial for constructing and analyzing the random coupling test statistics in the group testing.

To the best of our knowledge, most existing proofs of anisotropic local laws in the literature (see, e.g., \cite{isotropic,HKR2018,KY_isotropic,Anisotropic}) %XYY,yang2018,XYY_VESD
require the stronger assumption of $q\ge n^\e$ and prove much weaker bounds than \eqref{highm_aniso2} for a large but \textit{fixed} $r$ and with $\log n$ replaced by an $n^\e$ factor. Thus, establishing \eqref{aniso_intro} is one of our main technical innovations. 

There is a growing literature on the asymptotic behaviors of eigenvectors for large random matrices. Besides the comparisons with \cite{SIMPLE,ATE} as discussed before, we compare our results with some additional existing works that are most related to our paper, and the list is far from being comprehensive. In \cite{AFWZ2020,FWZ2018}, a tight bound is  provided for the difference between the sample eigenvector and some linear transformation of the population eigenvector through a delicate entrywise eigenvector analysis for the first-order approximation under the $\ell_\infty$-norm. The spiked sample eigenvectors of spiked Wigner matrices and spiked sample covariance matrices have been studied on the level of first-order limits (see e.g., \cite{bgn2011,principal,DY2021,Nadler2008,DP_spike}). %\cite{bgn2012,bgn2012,Ding_BEJ,KX2016}, i.e., the limit of the projection of the sample eigenvector onto a given unit vector $\bx$ and some concentrations bounds have been derived in these references. 
However, these works have assumed either Gaussian distribution or finite high moments for random matrix entries, while none of them covered the sparse settings. 
\iffalse
In particular, \cite{DP_spike} established the asymptotic distribution of the spiked sample eigenvectors for the 
%Johnstone's 
spiked covariance model \citep{Johnstone_spike} with Gaussian entries and population covariance matrix $\bSig = \bI+\bA$, where $\bA$ is a low-rank matrix. 
%This result was extended to the non-Gaussian setting for a special case with diagonal $\bA$ in \cite{Dai2013}. 
\fi

Moreover, under the Gaussian assumption, Koltchinskii and Lounici \cite{KL_AIHP2016} considered the spiked covariance model %with a more general $\bSig$ 
and derived the asymptotic expansion of the bilinear form  $\bx^\top \wh\bv_k \wh\bv_k^\top \by$. %, where $\wh v_k$ is a spiked sample eigenvector and $\bx$ and $\by$ are two deterministic unit vectors. 
Wang and Fan \cite{WF2017} derived the asymptotic distribution of the linear form $\bv_i^\top \wh\bv_k$ for a general spiked covariance model with sub-Gaussian entries and strong signals (i.e., diverging signal-to-noise ratio), where $\bv_i$ and $\wh \bv_k$ are the spiked population and sample eigenvectors, respectively. 
% In \cite{WF2017} considered the spiked covariance model with general non-diagonal population covariance matrices, sub-Gaussian entries, and strong signals (i.e., diverging signal-to-noise ratio). They proved the asymptotic distribution of the linear form $\bv_i^\top \wh\bv_k$, where $\bv_i$ and $\wh \bv_k$ are the spiked population and sample eigenvectors, respectively.
Recently, the asymptotic distribution of the bilinear form  $\bx^\top \wh\bv_k \wh\bv^\top \by$ is derived for the spiked Wigner matrices \cite{Capitaine2021} and spiked covariance matrices \cite{BDWW2022,BDW2021}, respectively, in the more challenging setting with moderate signals, i.e., the signal-to-noise ratio is of constant order. Tang and Priebe \cite{TP2018} established the CLT for the entries of sample eigenvectors for a random adjacency matrix, but they assumed a prior distribution on the mean adjacency matrix. We would like to emphasize that the above works %\citep{DP_spike,Dai2013,KL_AIHP2016,WF2017,Capitaine2021,BDW20%21,BDWW2022,TP2018} 
cannot be applied or extended to our setting due to several key reasons. First, they considered different random matrix models than ours. Second, all of them except \cite{TP2018} did not cover the sparse setting. Third, they showed only the convergence of certain linear or bilinear forms in distribution, whereas our technical analysis requires an asymptotic expansion that holds with high probability $1-\OO(n^{-D})$ in order to deal with the challenging group testing case.

% Finally, in the current paper, we are interested in the asymptotic expansions and limiting distributions of signal (spiked) sample eigenvectors. In the past decade, there have also been great progresses in showing the limiting distributions of noise (bulk) sample eigenvectors \citep{TaoVu_evector,KnoYin2013,BouYau2017,BouHuaYau2017,PartI,Lucas_AIHP,GES_ETH1,GES_ETH2}. This could also be an interesting problem for our setting and requires a new theory, but we will not pursue this direction in the current paper.

%\newpage
\section{Simulation studies}
\label{new.Sec5}
%In this section, we will investigate the finite-sample performance of the suggested SIMPLE-RC tests and verify the theoretical results established in Section \ref{new.Sec3} through four simulation examples. We will focus on both perspectives of the size and power by considering two popular network model settings without or with degree heterogeneity.
\subsection{The size of SIMPLE-RC} \label{new.Sec5.1}
%We begin with examining the sizes of the two forms of the SIMPLE-RC test for the scenarios without and with degree heterogeneity based on the mixed membership model and the degree-corrected mixed membership model, respectively. 
We begin with examining the sizes of the two forms of the SIMPLE-RC test with the significance level $\alpha$ setting to be 0.05. Our simulation example 1 considers the mixed membership model (\ref{MM}) with a similar setting to that investigated in \cite{SIMPLE}. In contrast to \cite{SIMPLE}, we will conduct group network testing as opposed to testing a pair of given nodes, allow for a larger number of communities $K = 5$ instead of $K = 3$, and consider non-sharp nulls under weak signals as opposed to sharp nulls under relatively stronger signals. Specifically, we assume that the network size is $n = 3000$ and there are $K = 5$ communities, each of which has $n_0 = 300$ pure nodes. For each pure node in the $k$th community with $1 \leq k \leq K$, the associated community membership probability vector $\bpi$ is simply the $k$th basis vector $\be_k \in \mathbb{R}^K$. We further divide the other $n - K n_0$ nodes into four groups of equal size. Each mixed (i.e., non-pure) node from the $l$th group with $1 \leq l \leq 4$ has a community membership probability vector $\bpi$ given by $\ba_l$, where $\ba_1 = (0.1, 0.6, 0.1, 0.1, 0.1)^T$, $\ba_2 = (0.6, 0.1, 0.1, 0.1, 0.1)^T$, $\ba_3 = (0.1, 0.1, 0.6, 0.1, 0.1)^T$, and $\ba_4 = (1/K, \cdots, 1/K)^T$. Thus, we now have a complete specification of the $n \times K$ matrix of community membership probability vectors $\bPi$ in (\ref{MM}). 

We choose the matrix $\bP$ in (\ref{MM}) as a $K \times K$ nonsingular matrix with diagonal entries being one and $(i,j)$th entries being $\rho/|i - j|$ for each $1 \leq i \neq j \leq n$ with $\rho=0.2$. Finally, we let the sparsity parameter $\theta$ in (\ref{MM}) vary in $\{0.1, 0.2, \cdots, 0.8\}$, with smaller value leading to lower average node degree and thus weaker signal strength. In particular, the model setting (\ref{MM}) specified above indeed involves weaker signals compared to that in \cite{SIMPLE}. %Recall that we have four non-pure membership profile groups that are given by community membership probability vectors $\ba_l$ with $1 \leq l \leq 4$. 
For the null hypothesis $H_0$ in (\ref{eq: hypothesis}), we choose a representative group $\mathcal{M}$ of $m = |\mathcal{M}| = 10$ or $20$ nodes from the non-pure membership profile group with community membership probability vector $\ba_1$. To evaluate the performance of the SIMPLE-RC test  (\ref{eq: T-test_group}), we generate $500$ data sets for each model setting and apply the SIMPLE-RC test with parameter $K_0$ chosen to be in $\{3, 4, 5\}$, which is an important parameter for dealing with the issue of weak signals and determining the center of the asymptotic null distribution as revealed in Theorems \ref{thm:group-null} and \ref{thm:pg-sample}.

Our simulation example 2 considers the DCMM model (\ref{new.DCMM2}). The setting is the same as that of simulation example 1 above except that the  $\theta$ in (\ref{MM}) is now replaced with the degree heterogeneity matrix $\bTheta$ in (\ref{new.DCMM2}). Specifically, we first generate an $n \times n$ diagonal matrix with independent entries from the uniform distribution on $[0.5, 1]$ and then define $\bTheta$ as such a diagonal matrix rescaled by a scalar network sparsity parameter $r$ that varies in $\{0.1^{1/2}, 0.2^{1/2}, \cdots, 0.8^{1/2}\}$. From (\ref{new.DCMM2}), we see that parameter $r^2$ plays a similar role as $\theta$ with a smaller value indicating lower average node degree and weaker signal strength. The representative group $\mathcal{M}$ is defined similarly as in simulation example 1. For each simulation replication under each model setting, we apply the SIMPLE-RC test in (\ref{eq: T-test DCMM}), where the parameter $K_0$ is chosen to vary in $\{3, 4, 5\}$. %Recall that parameter $K_0 - 1$ determines the centering of the asymptotic null distribution as shown in Theorems \ref{thm:group-het} and \ref{thm:group-het-sample} under weak signals. 

\begin{table}
\centering
\begin{tabular}[t]{llllllllll}
\toprule
$m$ & $K_0$ & \multicolumn{8}{c}{$\theta$}\\
\cline{3-10}
 & & 0.1 & 0.2 & 0.3 & 0.4 & 0.5 & 0.6 & 0.7 & 0.8\\
\midrule	
10 & 3 & 0.052 & 0.028 & 0.034 & 0.04 & 0.032 & 0.02 & 0.022 & 0.028\\
 & 4 & 0.148 & 0.094 & 0.086 & 0.078 & 0.086 & 0.06 & 0.056 & 0.076\\
 & 5 & 0.328 & 0.188 & 0.204 & 0.182 & 0.194 & 0.142 & 0.132 & 0.138\\
\midrule
20 & 3 & 0.038 & 0.026 & 0.028 & 0.018 & 0.024 & 0.032 & 0.018 & 0.024\\
 & 4 & 0.108 & 0.064 & 0.06 & 0.048 & 0.044 & 0.056 & 0.04 & 0.064\\
 & 5 & 0.246 & 0.15 & 0.13 & 0.116 & 0.116 & 0.104 & 0.09 & 0.104\\
\bottomrule
\end{tabular}
\caption{The empirical sizes of the SIMPLE-RC test with test statistic $T$ under different values of $(m, K_0, \theta)$ and with nominal level $\alpha = 0.05$ for simulation example 1 in Section \ref{new.Sec5.1}.} \label{tab1}
\end{table}

\begin{table}
\centering
\begin{tabular}[t]{llllllllll}
\toprule
$m$ & $K_0$ & \multicolumn{8}{c}{$r^2$}\\
\cline{3-10}
 & & 0.1 & 0.2 & 0.3 & 0.4 & 0.5 & 0.6 & 0.7 & 0.8\\
\midrule	
10 & 3 & 0.102 & 0.072 & 0.044 & 0.036 & 0.046 & 0.038 & 0.034 & 0.036\\
 & 4 & 0.198 & 0.128 & 0.098 & 0.09 & 0.1 & 0.086 & 0.09 & 0.072\\
 & 5 & 0.43 & 0.246 & 0.196 & 0.19 & 0.16 & 0.174 & 0.162 & 0.136\\
\midrule
20 & 3 & 0.118 & 0.044 & 0.05 & 0.05 & 0.052 & 0.044 & 0.03 & 0.042\\
 & 4 & 0.212 & 0.088 & 0.094 & 0.084 & 0.092 & 0.058 & 0.058 & 0.078\\
 & 5 & 0.372 & 0.192 & 0.174 & 0.15 & 0.162 & 0.12 & 0.116 & 0.124\\
\bottomrule
\end{tabular}
\caption{The empirical sizes of the SIMPLE-RC test with test statistic $\mathcal{T}$ under different values of $(m, K_0, r^2)$ and with nominal level $\alpha = 0.05$ for simulation example 2 in Section \ref{new.Sec5.1}.} \label{tab2}
\end{table}

We present some representative empirical null distribution plots of both forms of the SIMPLE-RC test with test statistics $T$ and $\mathcal{T}$  in  Section \ref{Sec.supp.sim} of the Supplementary Material. In addition, Tables \ref{tab1} and \ref{tab2} present the empirical sizes of both forms of the SIMPLE-RC test with test statistics $T$ and $\mathcal{T}$ across different model settings for simulation examples 1 and 2, respectively. In particular, we see from Tables \ref{tab1} and \ref{tab2} that the choice of $K_0 = 3$ generally results in sizes that are around or below the nominal level $\alpha = 0.05$. We notice that the setting of rather weak signals (i.e., very small values of the network sparsity parameter $r^2$) for simulation example 2 under degree heterogeneity can be more challenging, which suggests that a lower value of parameter $K_0$ may be needed to alleviate such issue. In contrast, the choice of $K_0$ as the true value of $K = 5$ renders the sizes much inflated, which becomes more severe as the signal strength becomes weaker (i.e., the network becomes more sparse). These observations can be explained by the relatively small spiked eigenvalues $d_k$, $k=4,5$  and support our theoretical results obtained in Section \ref{new.Sec3}.

\subsection{The power of SIMPLE-RC} \label{new.Sec5.2}

We further investigate the power performance for both forms of the SIMPLE-RC test with significance level $\alpha = 0.05$. To this end, we will introduce two additional simulation examples. Our simulation example 3 is similar to simulation example 1 in Section \ref{new.Sec5.1} except that the second non-pure membership profile group with community membership probability vector $\ba_2$ is now defined through $\ba_2 = (0.1 + \delta, 0.6 - \delta, 0.1, 0.1, 0.1)^T$, where the additional parameter $\delta$ varies in $\{0.5, 0.4, \cdots, 0\}$. Observe that parameter $\delta$ measures the overall distance between $\ba_1$ and $\ba_2$. In particular, the nodes in the first two non-pure membership profile groups will share more similar (but non-identical) membership profiles as $\delta$ stays positive and approaches zero. To specify the alternative hypothesis $H_a$ in (\ref{eq: hypothesis-a}), we choose a representative group $\mathcal{M}$ of $m  = 10$ or $20$ nodes with half corresponding to $\ba_1$ and the other half corresponding to $\ba_2$. For each model setting of simulation example 3, we apply the SIMPLE-RC test without degree heterogeneity. Our simulation example 4 is also similar to simulation example 2 in Section \ref{new.Sec5.1}, but with the same modifications made to both $\ba_2$ and the representative group $\mathcal{M}$ as above. We apply the SIMPLE-RC test with degree heterogeneity for simulation example 4. Simulation examples 3 and 4 introduced above will showcase the empirical powers for both forms of the SIMPLE-RC test, respectively. As demonstrated in Section \ref{new.Sec5.1}, we will focus our attention on the choice of parameter $K_0 = 3$ for the power investigations.

\begin{table}
\centering
\begin{tabular}[t]{llllllllll}
\toprule
$m$ & $\delta$ & \multicolumn{8}{c}{$\theta$}\\
\cline{3-10}
 & & 0.1 & 0.2 & 0.3 & 0.4 & 0.5 & 0.6 & 0.7 & 0.8\\
\midrule	
10 & 0.5 & 0.922 & 0.998 & 1 & 1 & 1 & 1 & 1 & 1\\
 & 0.4 & 0.496 & 0.756 & 0.888 & 0.966 & 0.992 & 0.994 & 1 & 0.998\\
 & 0.3 & 0.132 & 0.18 & 0.218 & 0.288 & 0.42 & 0.56 & 0.598 & 0.722\\
 & 0.2 & 0.098 & 0.088 & 0.088 & 0.1 & 0.116 & 0.176 & 0.202 & 0.254\\
 & 0.1 & 0.088 & 0.05 & 0.058 & 0.05 & 0.066 & 0.082 & 0.08 & 0.082\\
 & 0 & 0.09 & 0.044 & 0.05 & 0.044 & 0.042 & 0.058 & 0.04 & 0.028\\
\midrule
20 & 0.5 & 0.978 & 0.998 & 1 & 0.998 & 1 & 0.998 & 0.998 & 1\\
 & 0.4 & 0.596 & 0.884 & 0.976 & 0.996 & 1 & 0.998 & 0.998 & 1\\
 & 0.3 & 0.146 & 0.166 & 0.252 & 0.384 & 0.462 & 0.604 & 0.678 & 0.794\\
 & 0.2 & 0.106 & 0.09 & 0.076 & 0.116 & 0.116 & 0.172 & 0.216 & 0.248\\
 & 0.1 & 0.09 & 0.066 & 0.054 & 0.064 & 0.044 & 0.07 & 0.06 & 0.094\\
 & 0 & 0.07 & 0.06 & 0.038 & 0.036 & 0.028 & 0.032 & 0.022 & 0.042\\
\bottomrule
\end{tabular}
\caption{The empirical powers of the SIMPLE-RC test with test statistic $T$ under different values of $(m, \delta, \theta)$ and with nominal level $\alpha = 0.05$ for simulation example 3 in Section \ref{new.Sec5.2}, where parameter $K_0$ is chosen as $3$.} \label{tab3}
\end{table}

\begin{table}
\centering
\begin{tabular}[t]{llllllllll}
\toprule
$m$ & $\delta$ & \multicolumn{8}{c}{$r^2$}\\
\cline{3-10}
 & & 0.1 & 0.2 & 0.3 & 0.4 & 0.5 & 0.6 & 0.7 & 0.8\\
\midrule	
10 & 0.5 & 0.804 & 0.96 & 0.986 & 1 & 1 & 1 & 1 & 1\\
 & 0.4 & 0.326 & 0.478 & 0.666 & 0.788 & 0.872 & 0.944 & 0.952 & 0.97\\
 & 0.3 & 0.138 & 0.122 & 0.148 & 0.178 & 0.186 & 0.216 & 0.296 & 0.282\\
 & 0.2 & 0.134 & 0.078 & 0.064 & 0.066 & 0.068 & 0.068 & 0.076 & 0.066\\
 & 0.1 & 0.132 & 0.068 & 0.056 & 0.05 & 0.058 & 0.06 & 0.072 & 0.05\\
 & 0 & 0.132 & 0.056 & 0.052 & 0.044 & 0.044 & 0.06 & 0.07 & 0.042\\
\midrule
20 & 0.5 & 0.908 & 0.994 & 0.998 & 1 & 1 & 0.996 & 1 & 1\\
 & 0.4 & 0.48 & 0.596 & 0.766 & 0.888 & 0.966 & 0.984 & 0.992 & 1\\
 & 0.3 & 0.224 & 0.098 & 0.16 & 0.168 & 0.238 & 0.272 & 0.292 & 0.354\\
 & 0.2 & 0.19 & 0.062 & 0.092 & 0.066 & 0.062 & 0.064 & 0.058 & 0.092\\
 & 0.1 & 0.202 & 0.07 & 0.058 & 0.048 & 0.046 & 0.044 & 0.042 & 0.046\\
 & 0 & 0.186 & 0.054 & 0.07 & 0.056 & 0.044 & 0.042 & 0.042 & 0.052\\
\bottomrule
\end{tabular}
\caption{The empirical powers of the SIMPLE-RC test with test statistic $\mathcal{T}$ under different values of $(m, \delta, r^2)$ and with nominal level $\alpha = 0.05$ for simulation example 4 in Section \ref{new.Sec5.2}, where parameter $K_0$ is chosen as $3$.} \label{tab4}
\end{table}

We present some representative empirical distribution plots of both forms of the SIMPLE-RC test for simulation examples 3 and 4, respectively, in Section \ref{Sec.supp.sim} of the Supplementary Material. It is worth mentioning that those plots provide part of the insights into the power of the SIMPLE-RC test, because suitably small values of parameter $\delta$ make the non-sharp null hypothesis $H_0$ in (\ref{eq: hypothesis}) satisfied. Tables \ref{tab3} and \ref{tab4} further provide a more complete picture on the empirical powers of both forms of the SIMPLE-RC test across different model settings for simulation examples 3 and 4, respectively. From Tables \ref{tab3} and \ref{tab4}, we see that the power of the SIMPLE-RC test generally approaches one as parameter $\delta$ increases from $0$ to $0.5$. Moreover, the power enhances as the signal strength becomes stronger (i.e., as parameter $\theta$ or $r^2$ increases). We also observe that a larger value of $m$ can boost the power of group network inference with SIMPLE-RC particularly under weaker signals, which is natural and sensible. These empirical results confirm our asymptotic theory on the power analysis established in Section \ref{new.Sec3}. % (see Theorems \ref{thm:group-alt}, \ref{thm:pg-sample}, \ref{thm:group-het}, and \ref{thm:group-het-sample}).

\section{Real data application} \label{new.Sec6}

We further demonstrate the practical utilities of the SIMPLE-RC for group network inference with a financial application. As in \cite{SIMPLE}, we consider the network of stocks in the Standard and Poor $500$ (S\&P $500$) list, which index tracks the stock performance of $500$ large companies listed on exchanges in the United States. Each node of the network represents the time series of a stock. Specifically, we look at a three-year period of January 3, 2017 to December 30, 2019. The main reason for choosing a three-year period instead of a longer time horizon is that the underlying network structure may change when the time horizon expands due to various economic factors. For each stock in the S\&P $500$ list, the daily closing prices over the specified time period are converted into a time series of the daily log returns. We further remove any stocks with missing values, which yields a total of $n = 495$ stocks. It is well-known from finance  that all the individual stock excess returns (i.e., returns minus the risk-free interest rate) are correlated globally through some common factors such as the Fama--French factors. To better understand the intrinsic network structure, we regress the time series of excess returns for each stock on the Fama--French three factors and treat the resulting residual vector as a new time series for the stock, which corresponds to the idiosyncratic components of the factor model.

We are now ready to construct the $n \times n$ adjacency matrix $\bX$ for the group network inference. To this end, let us first calculate the correlation matrix based on the new time series above and then apply a simple hard-thresholding with threshold $0.5$ to each entry of the absolute correlation matrix, which gives rise to an $n \times n$ binary data matrix $\bX$. Since the stock network is known to be of node degree heterogeneity, we will apply the SIMPLE-RC test introduced in Section \ref{new.Sec3.4} with test statistic $\mathcal{T}$ given in (\ref{eq: T-test DCMM}). We choose parameter $K_0$ as $3$ following the analysis in \cite{SIMPLE}. It remains to specify the groups out of the above list of $n$ stocks. Specifically, we consider a total of five groups labelled as \texttt{Technology}, \texttt{Healthcare}, \texttt{Financial Services}, \texttt{Energy}, and \texttt{Communication Services}, which correspond to five sectors of the stock market. For the technology sector, we select a list of four stocks: Apple (\texttt{AAPL}), IBM (\texttt{IBM}), Intel (\texttt{INTC}), and NVIDIA (\texttt{NVDA}). For the healthcare sector, we select a list of four stocks: Abbott Laboratories (\texttt{ABT}), Amgen (\texttt{AMGN}), Eli Lilly (\texttt{LLY}), and UnitedHealth Group (\texttt{UNH}). For the financial services sector, we select a list of four stocks: Bank of America (\texttt{BAC}), Citigroup (\texttt{C}), Goldman Sachs (\texttt{GS}), and JPMorgan Chase (\texttt{JPM}). For the energy sector, we select a list of four stocks: Chevron (\texttt{CVX}), Devon Energy (\texttt{DVN}), EOG Resources (\texttt{EOG}), and Exxon Mobil (\texttt{XOM}). Finally, for the communication services sector, we select a list of four stocks: Activision Blizzard (\texttt{ATVI}), Comcast (\texttt{CMCSA}), DISH Network (\texttt{DISH}), and Netflix (\texttt{NFLX}).

\begin{table}
\centering
\begin{tabular}[t]{llllll}
\toprule
 & Technology & Healthcare & Financial %Services 
 & Energy & Communication %Services
 \\
\midrule	
Technology & 5.420 & 8.760 & 25.036 & 19.225 & 39.324 \\
Healthcare & 8.760 & 6.762 & 8.514 & 8.132 & 39.324 \\
Financial %Services 
&  25.036 & 8.514 & 0.601 & 17.050 & 39.324 \\
Energy &  19.225 & 8.132 & 17.050 & 0.414 & 39.324 \\
Communication %Services 
&  39.324 & 39.324 & 39.324 & 39.324 & 0.892 \\
\bottomrule
\end{tabular}
\caption{The values of the SIMPLE-RC test statistic $\mathcal{T}$ for different groups of selected stocks within and across the five sectors for the stock data example in Section \ref{new.Sec6}.} \label{tab5}
\end{table}

\begin{table}
\centering
\begin{tabular}[t]{llllll}
\toprule
 & Technology & Healthcare & Financial %Services 
 & Energy & Communication %Services
 \\
\midrule	
Technology &  0.1246 & 0.0247 & 0.0000 & 0.0001 & 0.0000 \\
Healthcare &  0.0247 & 0.0658 & 0.0279 & 0.0337 & 0.0000 \\
Financial %Services 
&  0.0000 & 0.0279 & 0.7726 & 0.0004 & 0.0000 \\
Energy &  0.0001 & 0.0337 & 0.0004 & 0.8033 & 0.0000 \\
Communication %Services 
&  0.0000 & 0.0000 & 0.0000 & 0.0000 & 0.7220 \\
\bottomrule
\end{tabular}
\caption{The corresponding p-values of the SIMPLE-RC test with test statistic $\mathcal{T}$ for different groups of selected stocks within and across the five sectors for the stock data example in Section \ref{new.Sec6}.} \label{tab6}
\end{table}

For each group and each pair of groups, we calculate the values of the SIMPLE-RC test statistic $\mathcal{T}$ and the associated p-values as in simulation examples 2 and 4 from Section \ref{new.Sec5}, with the choice of group size $m = |\mathcal{M}| = 4$ (when conducting the between-group tests, we randomly sample two stocks from the pool of four in each group). %It is worth mentioning that the choice of a smaller group size $m$ here is due to the relatively smaller network size $n=495$ in this application.
%{\color{red}Q:  For two different sectors, do we involve with 8 stocks?   Is the number of nodes $5*4$ or 495?} {\color{blue}: doesn't the pair of group has size $m=8$ with 4 from each group?}
 Table \ref{tab5} presents the values of the SIMPLE-RC test statistic $\mathcal{T}$ for the above group network inference, while Table \ref{tab6} provides the corresponding network p-values, both calculated using the adjacency matrix of 495 stocks constructed above. From Table \ref{tab5}, we see that the values of the SIMPLE-RC test statistic for groups of selected stocks within the five sectors are uniformly dominated by those of the SIMPLE-RC test statistic for groups of selected stocks across the five sectors. Such results indicate that the group of selected stocks within each sector tend to have similar (but possibly non-identical) membership profiles. In contrast, the group of selected stocks across the sectors tend to have more distinct membership profiles. These observations are made more precise in view of the SIMPLE-RC test p-values for group network inference listed in Table \ref{tab6}. Such empirical findings are consistent with the stock classifications by sector, showcasing the practical usage of the SIMPLE-RC test for group network inference with non-sharp nulls and weak signals.

	\bibliographystyle{imsart-number} % Style BST file (imsart-number.bst or imsart-nameyear.bst)
	\bibliography{references}       % Bibliography file (usually '*.bib')

%%%%%%%%%%%%%%%%%%%%%%%%%%%%%%%%%%%%%%%%%%%%
	
\newpage
\appendix
\setcounter{page}{1}
\setcounter{section}{0}
\renewcommand{\theequation}{A.\arabic{equation}}
\renewcommand{\thesubsection}{A.\arabic{subsection}}
\setcounter{equation}{0}
	
\begin{center}{\bf \Large Supplementary Material to ``SIMPLE-RC: Group Network Inference with Non-Sharp Nulls and Weak Signals''}
		
\bigskip
		
Jianqing Fan, Yingying Fan, Jinchi Lv and Fan Yang
\end{center}
	
\noindent %This Supplementary Material contains all RMT theory, the proofs of Theorems \ref{thm:pair-null}--\ref{asymp_evector}, Propositions \ref{lem_locallawv}--\ref{lem_locallawv2}, and some key lemmas as well as additional technical details. All the notations are the same as defined in the main body of the paper.

This Supplementary Material contains all the RMT results, all the proofs, and some additional simulation results. All the notation is the same as defined in the main body of the paper.

\renewcommand{\thesubsection}{A.\arabic{subsection}}

\section{Asymptotic expansions of eigenvalues and eigenvectors} \label{Sec.new-new-A}

\subsection{Local laws} \label{new.Sec4.2}

Given the variances $s_{ij}$ of the entries of the noise random matrix $\bbW$ (recall \eqref{support-W}) and a complex number $z$, we define $\bbM=(M_1, \cdots ,M_n)^T$ as the $z$-dependent solution to the QVE
\beq\label{QVE} \frac1{M_i}= - z - \sum_{  j \in [n]} s_{ij}M_j
\eeq
with $ i \in [n]$ such that $\im M_i \in \C_+$ whenever $z\in \C_+$, where $\C_+$ denotes the upper half of the complex plane $\mathbb{C}$. Then it is known that 1) $$\langle \bbM (z)\rangle: = \frac1n\sum_{i \in [n]} M_i(z)$$
is the Stieltjes transform of some probability measure, say $\mu_c$, on the real line $\mathbb{R}$; 2) $\mu_c$ is absolutely continuous with respect to the Lebesgue measure on $\mathbb{R}$ and its density $\rho_c$ is determined by
$$ \rho_c(x)= \frac{1}{\pi}\lim_{\eta\rightarrow 0+} \im \langle \bbM (x+\ii \eta)\rangle ;$$
3) the measure $\mu_{c}$ is compactly supported with $\supp (\mu_c) \subseteq [-2\sqrt{\mathfrak M}, 2\sqrt{\mathfrak M}]$, where $\mathfrak M:= \max_{i\in [n]}\sum_{j\in [n]} s_{ij}$; 4) each $M_i(z)$ is the Stieltjes transform of some finite measure that has the same support as $\mu_c$ and is uniformly bounded with \smash{$\max_{z\in \C_+} |M_i(z)|\lesssim 1$}. See, e.g., Corollary 1.3 of \cite{AjaErdKru2015} for more details. In fact, the measure $\mu_c$ is known to be the asymptotic empirical spectral distribution (ESD) of the noise random matrix $\bbW$ \citep{AjaErdKru2015}.
%where $\lambda_+$ is a positive value of order 1 and determined by the variances $s_{ij}$ of the entries of $\bbW$.
% We also define the resolvent (or Green's function) of $\bbW$ as 
% \beq\label{eqn_defG}\bG(z):= (\bbW-z)^{-1}.\eeq
Let us further introduce the deterministic matrix
\beq\label{defn_pi} \Pii(z):=\diag\{\bbM(z)\},\eeq
which turns out to be the asymptotic limit of $\bG(z)$ as formally shown later. 

To facilitate the technical presentation, we will use the \emph{minors} of matrices $\bbW$ and $\bG$ as defined below. % throughout the paper.

\begin{definition}[Minors] \label{defminor}
	%For any $ \cal J \times \cal J$ matrix $\cal A$ and $\mathbb T \subseteq \mathcal J$, where $\cal J$ and $\mathbb T$ are some index sets, 
	Given a subset of indices $\mathbb T$, we define the minor $\bbW^{(\mathbb T)}:=(W_{ij}:i,j\notin  \mathbb T)$ as the $ (n- |\mathbb T|)\times (n- |\mathbb T|)$ matrix obtained by removing all rows and columns indexed by $\mathbb T$. We keep the names of indices when defining $\bbW^{(\mathbb T)}$, i.e., $W^{(\mathbb{T})}_{ij}= W_{ij}$ for $i,j \notin \mathbb{{T}}$.  Correspondingly, we define the resolvent minor as  {$	\bG^{(\mathbb T)}(z):=(\bbW^{(\mathbb T)}- z\bI )^{-1}.$}
	For convenience, we will adopt the convention that %for any minor $\cal A^{(T)}$, 
	$W^{(\mathbb T)}_{ij} = 0$ and $G^{(\mathbb T)}_{ij} = 0$ if $i \in \mathbb T$ or $j \in \mathbb T$. We will abbreviate $(\{i\})\equiv (i)$,  $(\{i, j\})\equiv (ij)$, and $\sum_{i}^{(\mathbb T)} := \sum_{i \in [n]\setminus \mathbb T} .$ %$[\{\mu\}]\equiv [\mu] $ and  $[\{\mu, \nu \}]=[\mu\nu]$ 
\end{definition}

With an application of classical moment methods, we can show in the lemma below that the operator norms of the noise random matrix $\bbW$ and its minors are of order $O_p(1)$ under assumption \eqref{support-W}. Let $\xi\equiv \xi_n$ be a sequence of deterministic quantities satisfying that
\beq\label{defn_xin} 
\log n \ll \xi  \le (\log n)^{\log \log n}.
\eeq
Hereafter, we say that an event $\Omega$ holds \emph{with $(a,\xi)$-high probability} if we have 
$\P(\Omega^c)\le e^{-a\xi}$
for large enough $n$. 

\begin{lemma}\label{lem_opbound}
Assume that condition \eqref{support-W} holds, $\xi$ satisfies \eqref{defn_xin}, and $q\gg \xi^2$. %For any small constant $\tau$, 
Then there exists some constant $c_0>0$ such that with $(c_0,\xi)$-high probability,
\beq\label{bound_OP}   
\max\left\{\|\bbW\| ,\ \max_{i\in [n]} \|\bbW^{(i)}\|,\ \max_{i,j\in [n]} \|\bbW^{(ij)} \|\right\}\le 2\sqrt{\mathfrak M} +\xi/\sqrt{q}.
\eeq 
Consequently, it holds for each $C_0>2\sqrt{\mathfrak M} +\tau$ with some constant $\tau>0$ that with $(c_0,\xi)$-high probability,
\beq\label{bound_G}   
\sup_{z\in S(C_0)} (|z|-2\sqrt{\mathfrak M})\cdot \max\left\{ \|\bG(z)\| ,\ \max_{i\in [n]} \|\bG^{(i)}(z)\|,\ \max_{i,j\in [n]}  \|\bG^{(ij)}(z)\| \right\}\le 1, 
\eeq
 where we define the spectral domain
$S(C_0):=\{z=E+\ii \eta: |E|> C_0,\, \eta\ge 0\}$.
\end{lemma}

\begin{remark}
Denote by $\lambda_+$ the rightmost edge of the compactly supported measure $\mu_c$, which is the asymptotic limit of the ESD of the noise random matrix $\bbW$. Then we expect that \eqref{bound_OP} also holds with $2\sqrt{\mathfrak M} $ replaced by $\lambda_+$. However, we will not pursue such improvement here since it is not the main focus of the current paper.
\end{remark}

We are now ready to claim a sharp \emph{entrywise local law} for $\bG(z)$ when $z\in S(C_0)$ in the theorem below. 

\begin{theorem}[Entrywise local law]\label{lem_locallaw1}
%Under the assumptions of Lemma \ref{lem_opbound}, 
Assume that condition \eqref{support-W} holds, $\xi$ satisfies \eqref{defn_xin}, and $q\gg \xi^2$. Then for each $C_0 \ge 2\sqrt{\mathfrak M} +\tau $ with some constant $\tau>0$, 
%Assume that 
%$$ q=\sqrt{n\theta} \ge (\log n)^{C_1\xi}.$$
% \beq\label{bound_q}
% {\cor \text{need to check! }} q=\sqrt{n\theta}\ge \xi^{3}.
% \eeq
%{\cor (Check whether we can relax it to $ q \ge (\log n)^{C_1}$?)} 
there exist some constants $ c_1,C_1>0$ such that both events
\beq\label{entry_law}
\bigcap_{z\in S(C_0)} \left\{ \max_{ i\in [n]}\left|G_{ii}(z)-M_{i}(z)\right| \le \frac{C_1}{|z|^2  }\left(\frac{1}{q} +\frac{\xi^{1/2}}{q|z|} + \frac{\xi^2}{\sqrt{n}|z|}\right)\right\}
%\frac{C_1}{q} + \frac{C_1\xi^2}{\sqrt{n}} \right\} 
\eeq
and
\beq\label{entry_law2}
\bigcap_{z\in S(C_0)} \left\{\max_{ i\ne j\in [n]}\left|G_{ij}(z)\right| \le \frac{C_1}{|z|^2}\left(\frac{1}{q} + \frac{\xi^2}{\sqrt{n}|z|}\right)\right\}
\eeq
hold with $(c_1,\xi)$-high probability.
\end{theorem}

Theorem \ref{lem_locallaw1} above reveals that the entries of $\bG$ are sufficiently close to those of $\Pii$ given in (\ref{defn_pi}). Observe that the leading term in the bounds \eqref{entry_law} and \eqref{entry_law2} is of order $(q|z|^2)^{-1}$, which shows that the entrywise local law is weaker for sparser networks. 
%The proof of Theorem \ref{lem_locallaw1} is similar to that for \cite[Theorem 3.1]{EKYY_ER1} and will be given in Section \ref{appd_local1}. 
For the relatively denser case when $q\ge (\log n)^C$ with $C > 0$ some large constant, the weighted average of the diagonal resolvent entries satisfies a better bound which is provided in the theorem below.

\begin{theorem}[Averaged local law]\label{lem_averlaw}
Assume that condition \eqref{support-W} holds, $q\gg \xi^2$, and 
\beq\label{bound_xi}
\xi\gg (\log n)^{3} . %,\quad q=\sqrt{n\theta}\gg \xi^{2}.
\eeq
%{\cor (Check whether we can relax it to $ q \ge (\log n)^{C_1}$?)} 
Then for each $C_0 \ge 2\sqrt{\mathfrak M} +\tau $ with some constant $\tau>0$, there exist some constants $ c_2,C_2>0$ such that the event 
\beq\label{aver_law}
\bigcap_{z\in S(C_0)} \bigg\{ %|z|^2 \cdot  \left|\langle G (z)-\Pi (z)\rangle \right| +
\max_{ i \in [n]} \Big|\sum_{  k\in [n]} s_{ik}\left[G_{kk} (z)-M_k (z)\right]\Big|    \le \frac{C_2}{ |z|^2} \left(\frac{\xi^{1/2}}{n}+\frac{p_0^8}{q^2|z|}+\frac{p_0^8\xi^4}{n|z|^3}\right)\bigg\}
\eeq
holds with $(c_2,p)$-high probability for any deterministic parameter $p$ satisfying that 
\beq\label{p_constraint}
\log n\ll p\ll p_0:=(\xi/\log n)\wedge q^{1/3},
\eeq
where $\wedge$ represents the minimum of two given numbers.
\end{theorem}

We see from Theorem \ref{lem_averlaw} above that the bound \eqref{aver_law} from the \emph{averaged local law} for $\bG(z)$ has a leading term of order $p_0^8/(q^2|z|^3)$, which is better than the leading order of the entrywise estimate in \eqref{entry_law} for each individual $|G_{kk}-M_k|$ when $q|z|\gg p_0^8$. 
%The proof of Theorem \ref{lem_locallaw1} is similar to that for \cite[Theorem 2.8]{EKYY_ER1} and will be given in Section \ref{appd_local2}. 

By assuming Condition \ref{main_assm}(i), we can establish the \emph{anisotropic local law} for $\bG(z)$ in the theorem below, which provides an estimate of the bilinear form $\bu^\top \bG \bv$ for arbitrary deterministic unit vectors $\bu$ and $\bv$ and thus generalizes the entrywise local law in Theorem \ref{lem_locallaw1}.

\begin{theorem}[Anisotropic local law]\label{lem_locallaw2}
Assume that condition \eqref{support-W} holds and $q\gg (\log n)^{4}$.
% 	Under the assumptions of Theorem \ref{lem_averlaw}, assume further 
% 	\beq\label{bound_q}
% {\cor \text{need to check! }}q\gg \xi^{2} (\log n)^5.
% \eeq
For each $C_0 \ge 2\sqrt{\mathfrak M} +\tau $ for some constant $\tau>0$ and each constant $D>0$, there exists some constant $C_3>0$ such that for any deterministic unit vectors $\bu,\bv\in \R^n$, the event
\beq\label{aniso_law}
\bigcap_{z\in S(C_0)} \left\{\left|\bu^\top \left[\bG(z)-\Pii(z)\right]\bv\right| \le \frac{C_3\log n}{q|z|^2 }\right\}
\eeq
holds with probability at least $1-n^{-D}$.
\end{theorem}

In particular, when $\bu$ is a standard basis unit vector, we have a better bound when $\|\bv\|_\infty$ is small as shown in the proposition below.
%on $\bu^\top (\bG(z)-\Pii(z))\bv$. 

\begin{proposition}\label{lem_locallawv}
Under the conditions of Theorem \ref{lem_locallaw2}, for each constant $D>0$ there exists some constant $C_4>0$ such that for any deterministic unit vector $\bv\in \R^n$, the event
\beq\label{aniso_law222}
\bigcap_{z\in S(C_0)} \left\{ \max_{ i\in [n]} \left|\be_i^\top \left[\bG(z)-\Pii(z)\right] \bv\right| \le \frac{C_4}{|z|^{2}} \left(\sqrt{\frac{\log n}{ n}} +\frac{\log n}{q}\|\bv\|_\infty \right) \right\}
\eeq
holds with probability at least $1-n^{-D}$.
\end{proposition}

We have a similar anisotropic local law for $\bW\bG(z)$ as stated in the proposition below. 

\begin{proposition}\label{lem_locallawv2}
	Under the conditions of Theorem \ref{lem_locallaw2}, for each constant $D>0$ there exists some constant $C_5>0$ such that for any deterministic unit vector $\bv\in \R^n$, the event
\beq\label{aniso_law333}
\begin{split}
\bigcap_{z\in S(C_0)} &\left\{\max_{i\in [n]} \left|\be_i^\top \bW\left[\bG(z)-\Pii(z)\right] \bv\right| \le  \frac{C_5}{|z|^{2} }  \left(\sqrt{\frac{\log n}{ n}}  +\|\bv\|_\infty\right) \right\} %+\left(\frac{\log n}{q|z|}+\frac{(\log n)^2}{q^2}\right)\|\bv\|_\infty
\end{split}
\eeq
holds with probability at least $1-n^{-D}$.
\end{proposition}

The almost sharp anisotropic local laws established in Theorem \ref{lem_locallaw2} and Propositions \ref{lem_locallawv} and \ref{lem_locallawv2} above provide the main technical tools for deriving the asymptotic expansion of the empirical spiked eigenvectors. Specifically, the proof of Theorem \ref{lem_locallaw2} will be rooted on the entrywise local law in Theorem \ref{lem_locallaw1} and the averaged local law in Theorem \ref{lem_averlaw}. 
%We will prove Theorem \ref{lem_locallaw2} based on the entrywise local law, Theorem \ref{lem_locallaw1}, and the averaged local law, Theorem \ref{lem_averlaw}, in Section \ref{sec_moments}. 
Propositions \ref{lem_locallawv} and \ref{lem_locallawv2} are direct consequences of Theorem \ref{lem_locallaw2}. 
%and their proofs will be presented in Sections \ref{Sec.newA.6} and \ref{Sec.newA.7}, respectively. 
In addition, note that Theorem \ref{lem_locallaw2} and Propositions \ref{lem_locallawv} and \ref{lem_locallawv2} %requires a stronger assumption on $q$ and 
hold with a slightly weaker probability bound than Theorem \ref{lem_locallaw1}.
 
\iffalse
Most existing proofs of anisotropic local laws in the literature \citep{isotropic,ATE,SIMPLE,HKR2018,KY_isotropic,Anisotropic,yang2018,XYY,XYY_VESD}  require that $q\ge n^{\e}$ and give slightly weaker bounds with the $\log n$ factor in \eqref{aniso_law} replaced by an $n^\e$ factor. One mathematical contribution of this paper is that we prove better anisotropic local law bounds under a relaxed sparsity condition $q\gg (\log n)^{4}$. As discussed in the introduction, our proof amounts to bounding high moments 
$\E\left|\bu^\top [\bG(z)-\Pii(z)]\bv\right|^{2r} $ for $r$ as large as $\log n$. To address this challenge, a cumulant expansion method and some careful 
combinatorial arguments are needed; we refer the reader to Section \ref{new.secB.claim1} for more details. 
\fi

\subsection{Spiked eigenvectors and eigenvalues} \label{new.Sec4.3}

With the new theoretical results on the local laws for $\bG(z)$ established in Section \ref{new.Sec4.2}, we are now ready to investigate the asymptotic expansions of the empirical spiked eigenvectors and spiked eigenvalues. In view of the spectral decomposition of the mean matrix $\bbH$, we can rewrite the rescaled random matrix $\bbX$ given in \eqref{eq: model.general} as
\beq\label{RMT_X2}
\bbX = \sum_{k=1}^n \wh d_k \widehat\bbv_k \widehat\bbv_k^T= \bbV\bbD\bbV^\top + \bbW = \sum_{k=1}^K d_k \bv_k\bv_k^\top + \bbW. 
\eeq
We aim at providing the asymptotic expansions for the empirical spiked eigenvectors $\wh\bv_k$ and spiked eigenvalues $\wh d_k$ with $1\le k \le K_{\max}$ by utilizing the anisotropic local laws established in Theorem \ref{lem_locallaw2} and Propositions \ref{lem_locallawv} and \ref{lem_locallawv2}. 
%We still assume non-degenerate $d_k$'s.
% We are interested in the eigenvalue $\lambda_k$ and eigenvector $\hat \bv_k$ corresponding to the $k$-th signal $d_k \bv_k\bv_k^\top$, $1\le k \le K$.  
% We assume that $\lambda_k$ is an outlier, i.e., 
% $$\lambda_k > C_0 + \tau,$$
% for $C_0$ in \eqref{bound_OP} and an arbitrary constant $\tau>0$. 
Observe that under the rescaling in \eqref{scaling_eq}, it holds that $|d_k|\gg 1$ for all $1\le k \le K_{\max}$. 
%Under the condition $|d_k|\gg 1$ (which corresponds to Condition \ref{main_assm} (ii) with rescaling \eqref{scaling_eq}), 
With an application of Weyl's inequality \citep{Weyl} and Lemma \ref{lem_opbound}, we can obtain that $\wh d_k=[1+\oo(1)]d_k$ with $(c_0,\xi)$-high probability. Using the anisotropic local laws mentioned above, we can find a better deterministic approximation $t_k$ of $\wh d_k$, which we will formally introduce next.

Denote by $ \bV_{-k}$ an $n\times (K-1)$ matrix obtained by removing the $k$th column of matrix $\bV$ and $\bbD_{-k}$ a $(K-1)\times (K-1)$ matrix obtained by removing the $k$th row and column of matrix $\bbD$. Let us define 
$$\mathcal I_k:=\left\{x\in \R:\frac{|d_{k}|}{1+\e_0/2}\le |x| \le (1+\e_0/2)|d_{k}|\right\}.$$ 
Then for each $1\le k \le K_{\max}$, we  define $t_k \in \cal I_k$ as the solution to the equation 
%Using Lemma \ref{lem_locallaw2}, we can rewrite this equation as
\beq\label{eq_evalue}
1 + d_k \bv_k^\top \Pii(x) \bv_k - d_k \bv_k^\top \Pii(x) \bV_{-k} \frac{1}{(\bbD_{-k})^{-1} + \bV_{-k}^\top \Pii(x)\bV_{-k} } \bV_{-k}^\top \Pii(x) \bv_{k}  =0
\eeq
with respect to $x$. Using a similar argument as for the proof of Lemma 3 in \cite{ATE}, we can show that under Condition \ref{main_assm}(ii)-(iv), equation \eqref{eq_evalue} above has a unique solution in $\mathcal I_k$ and thus the population quantity $t_k$ is well-defined. 
%with high probability.  

Moreover, it follows from \eqref{QVE} that for $|z|\gg 1$, 
$$M_i(z) = - {z}^{-1} -  z^{-3} \sum_{ j\in [n]} s_{ij} + \OO(z^{-5}) \ \text{ and } \ M_i'(z) = {z}^{-2} + \OO(z^{-4}).$$
Then we can deduce that 
\beq\label{asymp_Pi}\Pii(z) = -z^{-1}\bI + \bcE_1(z) \ \text{ and } \ \Pii'(z) = z^{-2}\bI + \bcE_2(z),\eeq
where $\bcE_1(z)$ and $\bcE_2(z)$ are diagonal matrices satisfying that  $\|\bcE_1(z)\| =\OO(z^{-3})$ and $\|\bcE_2(z)\| =\OO(z^{-4})$. With the aid of the eigengap assumption in \eqref{eq_eigengap} and the representations in \eqref{asymp_Pi}, we can obtain that  
\beq\label{denom_lower}
\max_{z\in \cal I_k}\left\| \left[d_k(\bbD_{-k})^{-1} + d_k\bV_{-k}^\top \Pii(z)\bV_{-k}\right]^{-1}\right\|\le C
\eeq
for some constant $C>0$ depending only on $\e_0$. Furthermore, in light of \eqref{asymp_Pi} and $\bV_{-k}^\top\bv_k=0$, there exists some constant $C'>0$ depending only on $\e_0$ such that for each $z\in \cal I_k$,
\beq\label{num_lower}
\left|d_k \bv_k^\top \Pii(z) \bv_k + \frac{d_k}{z}\right| \le C'd_k^{-2} \ \text{ and } \   \left\|d_k \bv_k^\top \Pii(z) \bV_{-k}\right\| \le C'd_k^{-2}.
\eeq

Finally, combining \eqref{eq_evalue}, \eqref{denom_lower}, and \eqref{num_lower} yields that 
\beq\label{eq_tk}
t_k=d_k + \OO\left(|d_k|^{-1}\right),
\eeq
which reveals that the population quantity $t_k$ introduced in (\ref{eq_evalue}) above indeed provides a close approximation of the population spiked eigenvalue $d_k$. We will characterize the asymptotic expansion for its empirical counterpart $\wh d_k$ in the theorem below. 
% {\color{red}Q: The difference between $t_k$ and $d_k$ is so mall, according to (92).  Can we replace $t_k$ by $d_k$ there?} {\color{blue} A: When $d_k$ is close to constant order, the $\OO\left(|d_k|^{-1}\right)$ term is much larger than the rate in \eqref{eq_evalue_conv}. Hence, we have chosen to keep $t_k$ so that Theorem \ref{asymp_evalue} holds in the general setting with $d_k\gg 1$, although we indeed replace $t_k$ with $d_k$ when proving Theorem \ref{thm:pair-null} (where $d_k\gg \sqrt{\log n}$) and other main results.}

\begin{theorem}\label{asymp_evalue}
Assume that parts (i)--(iv) of Condition \ref{main_assm} hold with $d_k\to d_k/q$ under the rescaling in \eqref{scaling_eq}, and %$|d_k| \gg 1$ and 
$K\log n \ll q|d_k|$ for each $1\le k \le K_{\max}$. Then it holds that w.h.p., 
\beq\label{eq_evalue_conv}
|\wh d_k - t_k| = \OO\left\{\frac{\log n}{q} +\frac{K\log n}{q|d_k|^4} \right
\}
\eeq
for each $1\le k \le K_{\max}$.
\end{theorem}

%The proof of this theorem will be presented in Section \ref{Sec.newA.55}. 
The asymptotic expansion \eqref{eq_evalue_conv} in Theorem \ref{asymp_evalue} above shows that the empirical spiked eigenvalue $\wh d_k$ is rather close to the population quantity $t_k$ with an approximation error of order $q^{-1}$ up to some $\log n$ and $K$ factors. Combining this theorem with the eigengap assumption \eqref{eq_eigengap}, we see that any circle $\cal C_k$ in the complex plane $\mathbb{C}$ and centered around $t_k$ will enclose only $\wh d_k$ and no other empirical eigenvalues with high probability as long as its radius is much larger than $ (\log n)/q +(K\log n)/(q|d_k|^4)$ and much smaller than $|d_k|$, which entails that the representation in \eqref{eq:Cauchy2} becomes applicable. Hence, an application of the anisotropic local laws for bounding the RHS of \eqref{eq:Cauchy2} will give an asymptotic expansion of the bilinear form $\bx^\top \wh\bv_k \wh\bv_k^\top \by$. With proper choices of deterministic unit vectors $\bx$ and $\by$, we can further characterize the asymptotic expansion of the empirical spiked eigenvectors in the theorem below.  

\begin{theorem}\label{asymp_evector}
Under the conditions of Theorem \ref{asymp_evalue}, %assume further that {\cor $K^2\log n \ll q|d_k|^5$}. Then, 
it holds that w.h.p.,   
%Condition \ref{main_assm} (where the scaling \eqref{scaling_eq} is used with $d_k\to d_k/q$), given any $1\le k \le K_0$, suppose $|d_k|\gg 1$ and $K\log n \ll q|d_k|$. Then, we have that
\beq\label{expansion_evector2}
\begin{split}
\wh v_k(i) &=v_k(i) + \frac{1}{t_k}\sum_{ l \in [n]} W_{il} v_k(l)  \\
&+\OO\left\{\frac{1}{|d_k|^2} \sqrt{\frac{\log n}{n}}  + \left(\frac{\sqrt{K}}{|d_k|} + \frac{K\log n}{q}\right) \left(\frac{\|\bV\|_{\max}}{|d_k|}+\frac{ 1}{d_k^2}\sqrt{\frac{\log n}{n}}\right)\right\}
\end{split}
\eeq
for each $1\le k \le K_{\max}$, where we have chosen the direction of $\wh\bv_k$ such that $ \wh\bv_k^\top \bv_k \ge 0$.
% \beq\label{expansion_evector2}
% \begin{split}
% \wh\bv_k(i) &=\bv_k(i) + \frac{1}{t_k}\sum_{l} \bW_{il} \bv_k(l) \\
% &+  \OO\left[\frac{\|\bv_k\|_\infty}{|d_k|}  \left(\frac{K\log n}{q} +\frac{K}{|d_k|}\right) \left(1+\frac{\sqrt{{\log n} }}{|d_k|}  \right)+ \frac{\sqrt{\log n}}{|d_k|^2} \|\bv_k\|_\infty\right] .
% \end{split}
% \eeq
% \beq\label{expansion_evector2}
% \wh\bv_k(i) =\bv_k(i) + \frac{1}{t_k}\sum_{l} \bW_{il} \bv_k(l) + \frac{1}{\sqrt{n}|d_k|} \cdot \OO\left(  \left(\frac{\sqrt{ \log n }}{|d_k| }   +  \frac{\log n}{q}\right)\right)\quad \text{w.h.p.}
% \eeq
\end{theorem}

Theorem \ref{asymp_evector} above is the major RMT result of our paper and lays the foundation for all the theoretical results on the SIMPLE-RC established in Section \ref{new.Sec3}. 
%Its proof will be given in Section \ref{Sec.newA.thm10}. %Theorem \ref{thm:pair-null} and Theorem \ref{thm:group-null}. 
Compared to Theorem 2 in \cite{ATE} and the main results in \cite{BDWW2022,BDW2021,Capitaine2021}, a key feature of our theoretical result in \eqref{expansion_evector2} is that it holds with high probability $1-\OO(n^{-D})$ instead of just $1-\oo(1)$. An important consequence is that by taking a union bound, \eqref{expansion_evector2} in fact holds simultaneously for all empirical spiked eigenvector entries $\wh v_k(i)$ with $ i\in [n]$ and $1\le k\le K_{\max}$. Such result is one of the main reasons why we can deal with the group testing in the current paper. Moreover, in comparison to \cite{SIMPLE,ATE}, our results on the asymptotic expansion of empirical spiked eigenvectors hold with tighter error bounds under weaker conditions on the network sparsity and the signal-to-noise ratio. 
%It is worth mentioning that it is the new asymptotic theory for spiked random matrices built in this section that has allowed us to prove stronger results in Theorems \ref{thm:pair-null} and \ref{thm:pair-null-het} than those in \cite{SIMPLE} regarding the testing for a given pair of network nodes.

\renewcommand{\thesubsection}{B.\arabic{subsection}}
	
\section{Additional simulation results corresponding to Sections \ref{new.Sec5.1} and \ref{new.Sec5.2}} \label{Sec.supp.sim}

%\subsection{Empirical distribution plots corresponding to Sections \ref{new.Sec5.1} and \ref{new.Sec5.2}} \label{supp.sim.empirical.null}

Figures \ref{fig1} and \ref{fig2} depict some representative empirical null distributions of both forms of the SIMPLE-RC test with test statistics $T$ and $\mathcal{T}$ for simulation examples 1 and 2, respectively. We see that even under weak signals (i.e., small values of $\theta$), the empirical null distributions of the SIMPLE-RC test closely match the theoretical asymptotic null distributions established in Theorems \ref{thm:group-null}, \ref{thm:pg-sample}, \ref{thm:group-het}, and \ref{thm:group-het-sample} under the choice of $K_0 = 3$. In contrast, when the parameter $K_0$ increases from $3$ to the true value of $K = 5$, the discrepancy between the empirical and theoretical null distributions becomes more pronounced. Such a phenomenon is indeed in line with our theoretical findings presented in Section \ref{new.Sec3}, reflecting the impact of weak signals on group network inference with SIMPLE-RC through the choice of parameter $K_0$.

\begin{figure}[thp]
\centering
\includegraphics[scale=0.6,width=4in,height=3.5in]{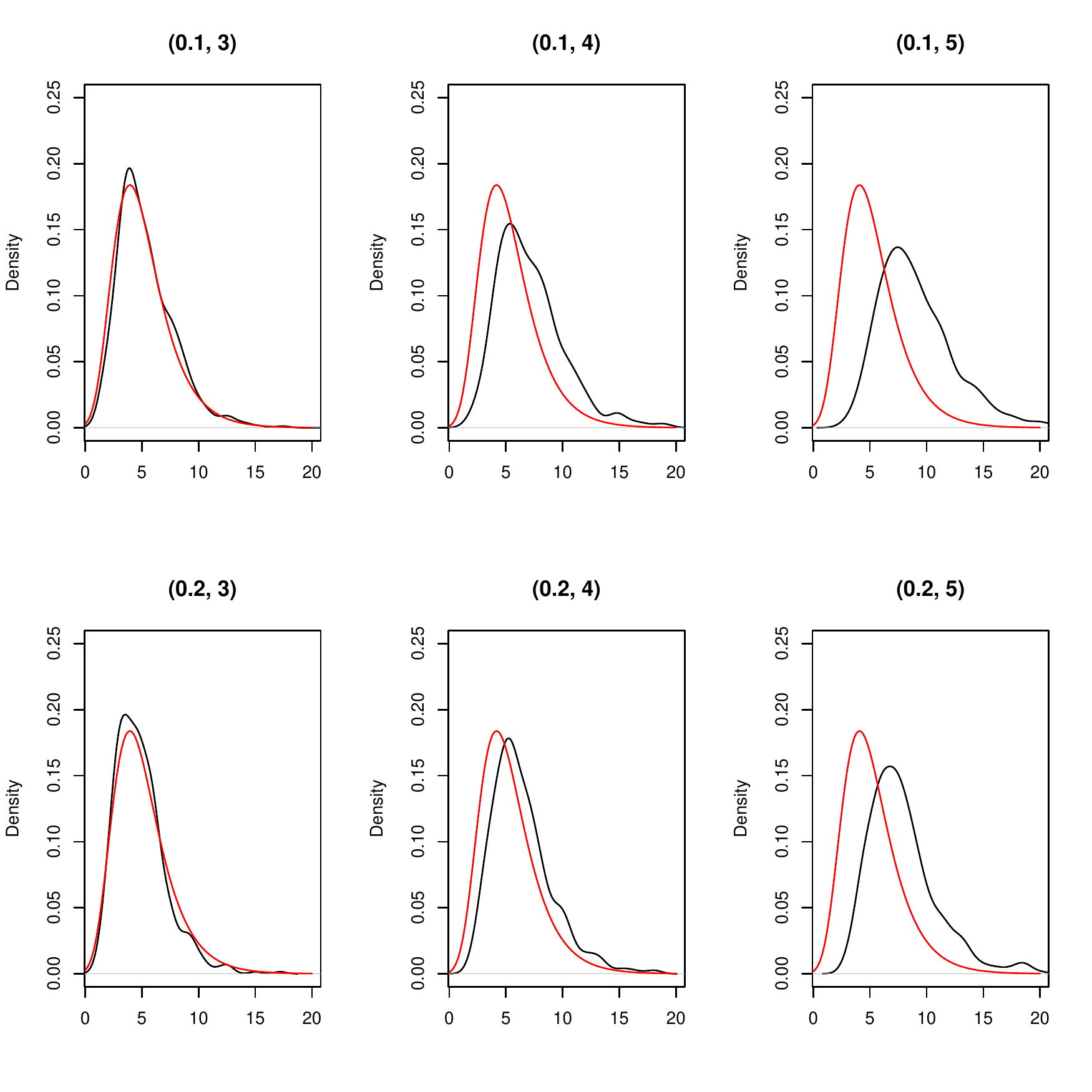}
\caption{The representative empirical null distributions (the black curves with kernel smoothing) of the SIMPLE-RC test statistic $T$ under different values of $(\theta, K_0)$ and with $m = 10$ for simulation example 1 in Section \ref{new.Sec5.1}. The red curves represent the asymptotic null distributions specified in Theorems \ref{thm:group-null} and \ref{thm:pg-sample}.}
\label{fig1}
\end{figure}

\begin{figure}[thp]
\centering
\includegraphics[scale=0.6,width=4in,height=3.5in]{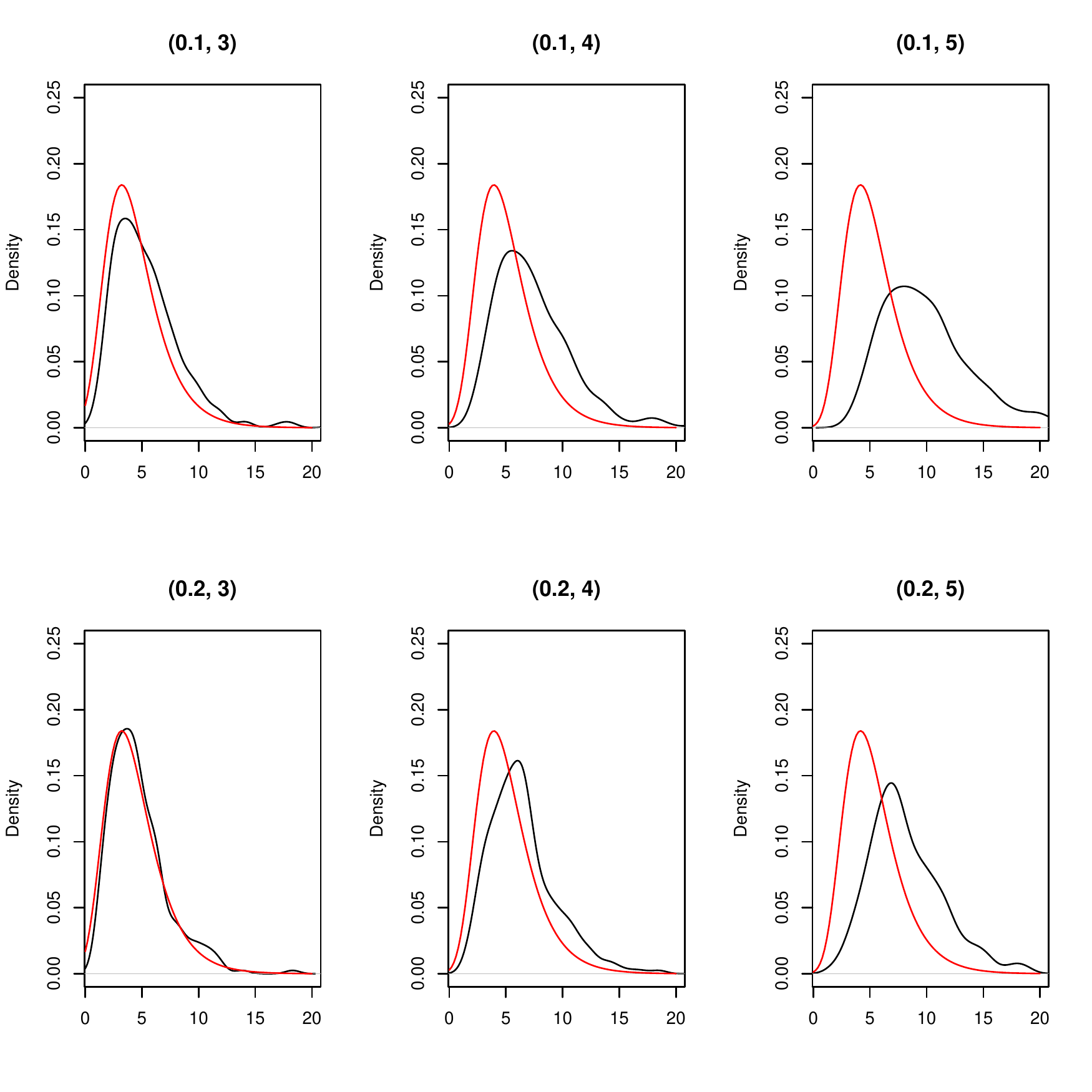}
\caption{The representative empirical null distributions (the black curves with kernel smoothing) of the SIMPLE-RC test statistic $\mathcal{T}$ under different values of $(r^2, K_0)$ and with $m = 10$ for simulation example 2 in Section \ref{new.Sec5.1}. The red curves represent the asymptotic null distributions specified in Theorems \ref{thm:group-het} and \ref{thm:group-het-sample}.}
\label{fig2}
\end{figure}

\begin{figure}[thp]
\centering
\includegraphics[scale=0.6]{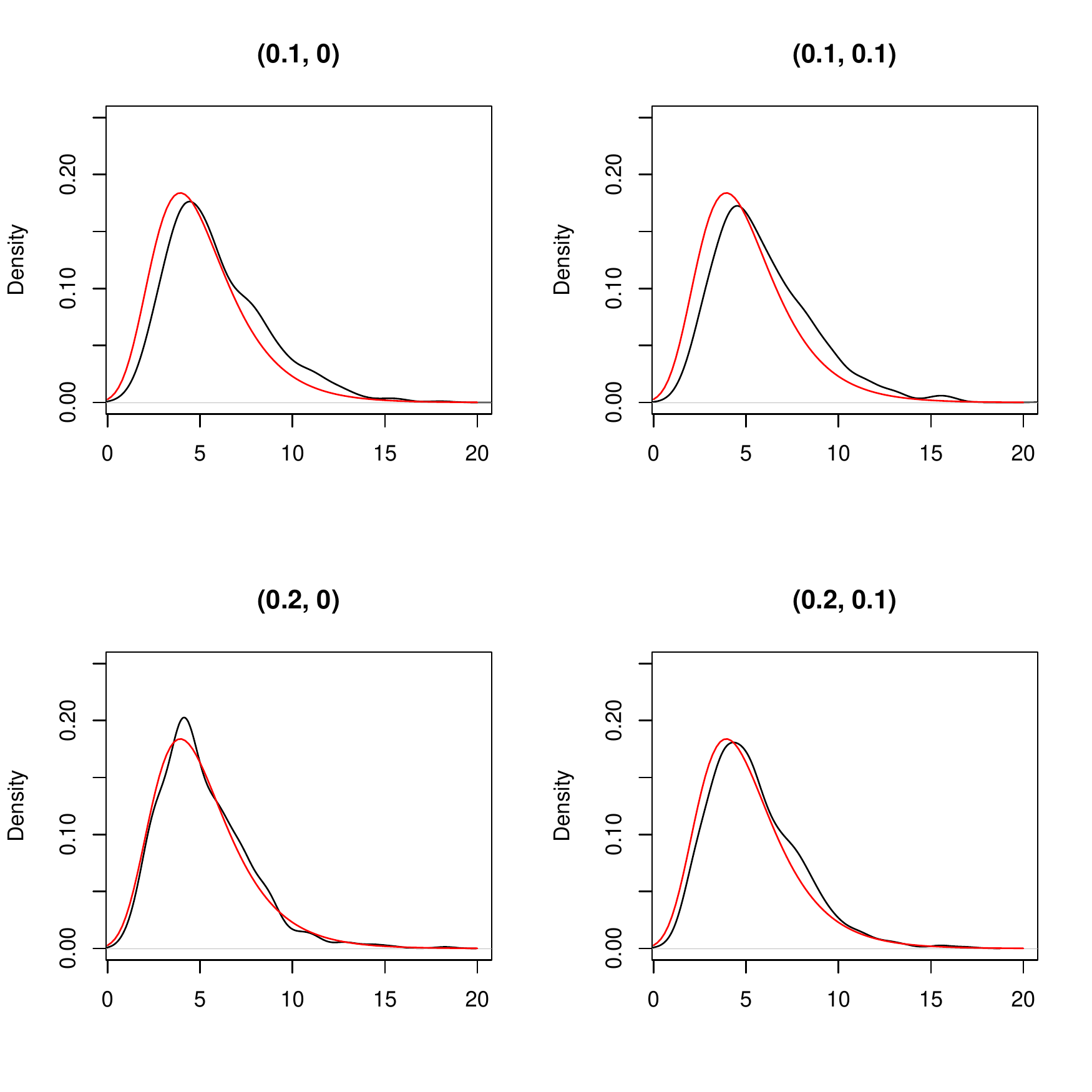}
\caption{The representative empirical distributions (the black curves with kernel smoothing) of the SIMPLE-RC test statistic $T$ under different values of $(\theta, \delta)$ and with $(m, K_0) = (10, 3)$ for simulation example 3 in Section \ref{new.Sec5.2}. The red curves represent the asymptotic null distributions specified in Theorems \ref{thm:group-null} and \ref{thm:pg-sample}.}
\label{fig3}
\end{figure}

\begin{figure}[thp]
\centering
\includegraphics[scale=0.6]{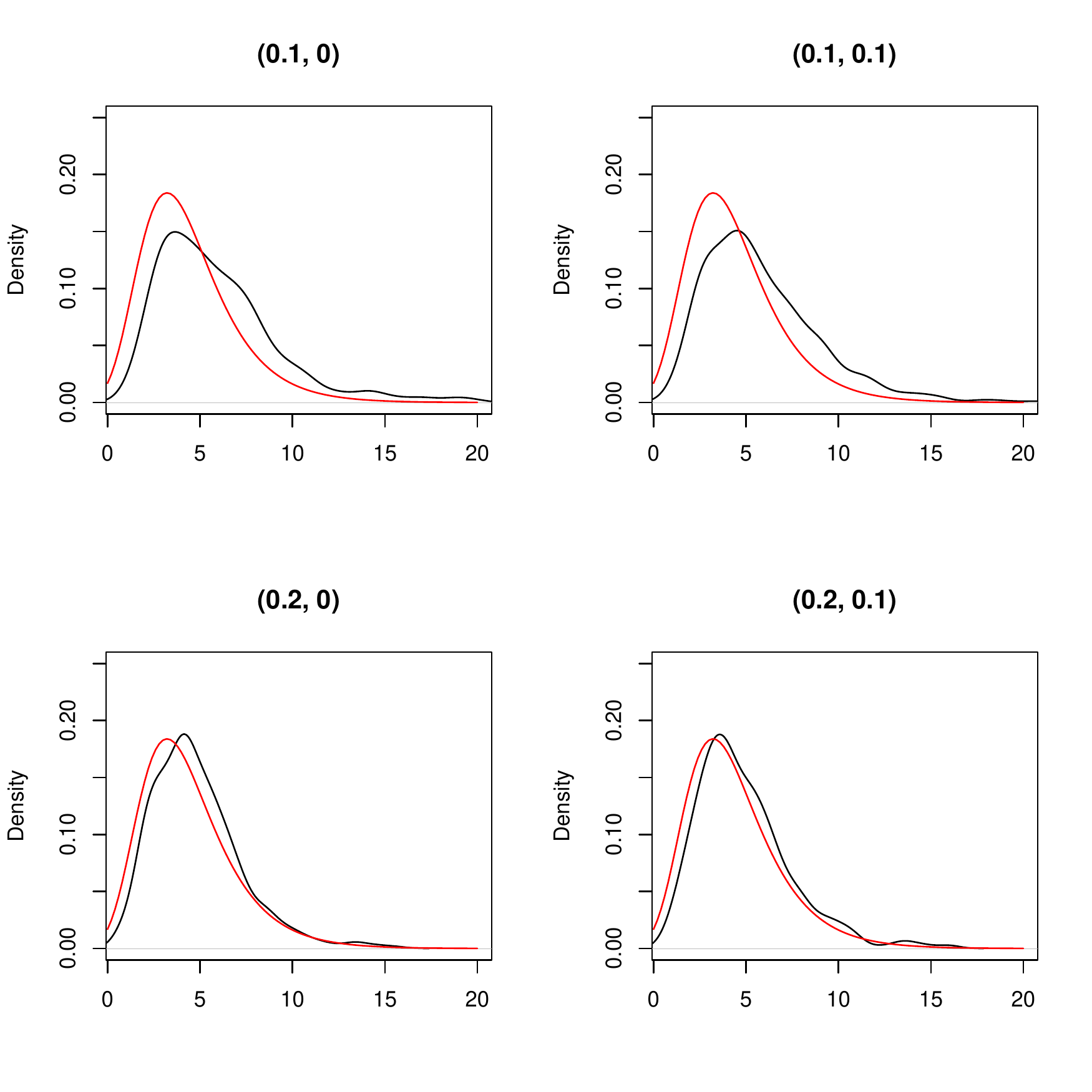}
\caption{The representative empirical distributions (the black curves with kernel smoothing) of the SIMPLE-RC test statistic $\mathcal{T}$ under different values of $(r^2, \delta)$ and with $(m, K_0) = (10, 3)$ for simulation example 4 in Section \ref{new.Sec5.2}. The red curves represent the asymptotic null distributions specified in Theorems \ref{thm:group-het} and \ref{thm:group-het-sample}.}
\label{fig4}
\end{figure}

Figures \ref{fig3} and \ref{fig4} depict some representative empirical distributions of both forms of the SIMPLE-RC test with test statistics $T$ and $\mathcal{T}$ for simulation examples 3 and 4, respectively. In particular, the signal strengths are rather weak in view of the small values of parameter $\theta$ in Figure \ref{fig3} and parameter $r^2$ in Figure \ref{fig4}. As mentioned above, the value of $\delta = 0$ corresponds to the scenario when the data is generated according to the sharp null hypothesis, while that of $\delta = 0.1$ corresponds to the scenario when the data is generated according to the non-sharp null hypothesis. We see from Figures \ref{fig3} and \ref{fig4} that the empirical distributions of the SIMPLE-RC test for $\delta =0$ and $0.1$ closely match the asymptotic null distributions revealed in Theorems \ref{thm:group-null}, \ref{thm:pg-sample}, \ref{thm:group-het}, and \ref{thm:group-het-sample} under non-sharp nulls and weak signals. Similarly as in simulation examples 1 and 2, when the signal strength becomes too weak (e.g., the value of $0.1$ for network sparsity parameters $\theta$ or $r^2$), a lower value of parameter $K_0$ may improve the distributional fits in Figures \ref{fig3} and \ref{fig4}, which can in turn lead to more reasonable (i.e., non-inflated) empirical sizes as parameter $\delta \rightarrow 0+$ (see the cases of $\theta = 0.1$ in Figure \ref{fig3} and $r^2 = 0.1$ in Figure \ref{fig4}). 

\renewcommand{\thesubsection}{C.\arabic{subsection}}
	
\section{Proofs of Theorems \ref{thm:pair-null}--\ref{asymp_evector}} \label{Sec.newA}

\subsection{Technical preparations} \label{new.Sec4.1}

Hereafter, we will frequently resort to some classical concentration inequalities for independent random variables. Let us recall the commonly used Bernstein's inequality.

\begin{lemma}[Bernstein's inequality \citep{vershynin2018}]\label{lemma_bern}
Let $(x_i)_{ i \in [n]}$ be a family of centered independent random variables satisfying that  $\max_{i \in [n]}|x_{i}|\le \phi_n$ for some ($n$-dependent) parameter $\phi_n>0$. Then it holds that for each $t>0$, 
$$ \P \Big(\sum_{ i \in [n]}x_i >t \Big)\le 2\exp\left(- \frac{ct^2}{\sum_{i \in [n]}\E x_i^2 +\phi_n t}\right)$$
with $c>0$ some absolute constant.
\end{lemma}

%\begin{remark}
Using Lemma \ref{lemma_bern} above, we can obtain that for some absolute constant $a>0$, 
\beq\label{Ber_xi} 
\Big|\sum_{i \in [n]} x_i\Big|\le  \Big(\sum_{i \in [n]}\E x_i^2\Big)^{1/2}\xi^{1/2} +\phi_n \xi
\eeq
with $(a,\xi)$-high probability, and for each constant $D>0$, there exists some constant $C>0$ such that
\beq\label{Ber_logn} 
\mathbb P\bigg\{\Big|\sum_{i \in [n]} x_i\Big|\le  C\Big[\Big(\sum_{i \in [n]}\E x_i^2\Big)^{1/2}(\log n)^{1/2} +  \phi_n \log n\Big]\bigg\}\ge 1-n^{-D}. \eeq
We will employ the bounds in (\ref{Ber_xi}) and (\ref{Ber_logn}) frequently in our proofs. For simplicity, we will often restate \eqref{Ber_logn} as 
$$ \Big|\sum_{i \in [n]} x_i\Big| \lesssim \Big(\sum_{i \in [n]}\E x_i^2\Big)^{1/2}(\log n)^{1/2} + \phi_n \log n$$
w.h.p. without specifying constants $C$ and $D$ explicitly.
%\end{remark}

Moreover, the quadratic and bilinear forms of independent random variables also satisfy certain large deviation bounds as given in the lemma below. 

\begin{lemma}[Lemma 3.8 of \cite{EKYY_ER1}]\label{largedeviation}
Let $(x_i)_{i \in [n]}$ and $(y_i)_{i \in [n]}$ be independent families of centered independent complex-valued random variables, and \smash{$(\cal B_{ij})_{i, j \in [n]}$} a family of deterministic complex numbers. Assume that the components  $x_i$ and $y_j$ have variances at most $n^{-1}$ and satisfy that $\max_{i \in [n]}|x_{i}|\le \phi_n$ and $\max_{i \in [n]}|y_{i}|\le \phi_n$ for some ($n$-dependent) parameter $\phi_n\ge n^{-1/2}$. Then %the following large deviation bounds hold 
it holds with $(a,\xi)$-high probability that 
%Then we have the following large deviation bounds:
	%for any fixed $\xi>0$, the following bounds hold with $\xi$-high probability:
	\begin{align}
		 %\Big\vert \sum_i \cal A_i x_i \Big\vert \prec  \phi_n \max_{i} \left\vert \cal A_i \right\vert+ \frac{1}{\sqrt{n}}\Big(\sum_i |\cal A_i|^2 \Big)^{1/2} , \quad
		&\Big\vert \sum_{ i,j\in [n]} x_i \cal B_{ij} y_j \Big\vert \le \xi^2 \bigg[\phi_n^2 \cal B_d  + \phi_n \cal B_o + \frac{1}{n}\Big(\sum_{ i\ne j \in [n]} |\cal B_{ij}|^2\Big)^{{1}/{2}} \bigg], \label{eq_concentr_1}\\
		&\Big\vert \sum_{i \in [n]} \overline x_i \cal B_{ii} x_i - \sum_{i \in [n]} (\mathbb E|x_i|^2) \cal B_{ii}  \Big\vert  \le \left(\xi^{1/2} \phi_n +\xi \phi_n^2\right) \cal B_d  ,\label{eq_concentr_2}\\ 
		& \Big\vert \sum_{ i\ne j \in [n]} \overline x_i \cal B_{ij} x_j \Big\vert  \le \xi^2\bigg[ \phi_n\cal B_o + \frac{1}{n}\Big(\sum_{ i\ne j\in [n]} |\cal B_{ij}|^2\Big)^{{1}/{2}}\bigg] ,\label{eq_concentr_3}
	\end{align}
	where $\cal B_d:=\max_{i \in [n]} |\cal B_{ii} |$, $\cal B_o:= \max_{i\ne j\in [n]} |\cal B_{ij}|$, and $a>0$ is some absolute constant.
\end{lemma} 
	
\subsection{Proof of Theorem \ref{thm:pair-null}} \label{subsec:cal-thm-pair}

% Some simple calculations related to Theorem \ref{thm:pair-null}}. 

% {\color{red} 
% For each proof in the supplementary part (i.e. each subsection), it would be very helpful to start with a high-level description of the main ideas and key ingredients/steps of that proof. That way it is easier for readers to appreciate the complicated math behind the proofs at a high level. It's also helpful to highlight the key challenges and innovations in the mathematical arguments (in the first paragraph of each proof).
% }
The proof of Theorem \ref{thm:pair-null} is a direct consequence of the expansion \eqref{expansion_evector2}. By the classical central limit theorem (CLT), the second term on the right-hand side (RHS) of \eqref{expansion_evector2} converges to a normal random variable for any $1\le k \le K_0$. Hence, we can write $\widehat\bbV_{K_0}(i)-\widehat\bbV_{K_0}(j)$ as $\bV_{K_0}(i)-\bV_{K_0}(j)$ plus an asymptotic multivariate normal random vector with covariance $\bSig_{i,j}$ and a random error. If we can show that $\bV_{K_0}(i)-\bV_{K_0}(j)$ and the error term are asymptotically negligible under $c_{1n}\ll (d_{1}\lambda_1(\bP))^{-1/2}$, the normal random vector leads to the asymptotic $\chi_{K_0}^2$ distribution in part (i). On the other hand, we will show that the deterministic term $\bV_{K_0}(i)-\bV_{K_0}(j)$ dominates in part (ii), which leads to \eqref{eq_power_pair}.

{Notice that it suffices to prove Theorem \ref{thm:pair-null} for a deterministic $K_0$. In fact, let  $\Omega(K_0)$ be the event that \eqref{expansion_evector} and \eqref{eq_tkdk} below hold for all $1\le k \le K_0$.
%appears in the following proof {\color{red}Is $\Omega(k)$ the event where (A.6) holds?}. 
Then we can write that
$$ \P\left\{\Omega(K_0)\right\}  = \sum_{l=0}^{K_{\max}\wedge C_0}\P\left\{\Omega(l), K_0=l\right\}= \sum_{l=0}^{K_{\max}\wedge C_0}\P\left\{\Omega(l)\right\} \P\left\{K_0=l\right\}, $$
where we have used the trivial fact that event $\Omega(l)$ is independent of $K_0$. Hence, we can obtain the desired results for $T_{ij}(K_0)$ from those for $T_{ij}(l)$. We will also use this simple fact tacitly in the proofs of Theorems \ref{thm:group-null}--\ref{thm:group-het-sample} later.}

Observe that $K\log n \ll q|d_{K_0}|$ is entailed by the conditions of Theorem \ref{thm:pair-null} and thus Theorem \ref{asymp_evector} from the general theoretical foundation established in Section \ref{new.Sec4.3} can be applied for any $1\le k \le K_0$. Specifically, by resorting to \eqref{expansion_evector2} in Theorem \ref{asymp_evector} with the rescalings $\bbW\to \bbW/q$, $t_k\to t_k/q$, and $d_k\to d_k/q$, we can obtain that with high probability (w.h.p.),
%{\cor revisions} With the expansions of eigenvectors in \eqref{expansion_evector2}, we have that for $1\le k \le K,$
\beq \label{expansion_evector}
\begin{split}
& \wh v_k(i) = v_k(i) + \frac{1}{t_k}\sum_{l=1}^n  W_{il}  v_k(l) \\
& +  \OO\left\{ \frac{q}{\sqrt{n}|d_k|} \left[\frac{q \sqrt{ \log n }}{|d_k|}+ \left(\frac{q\sqrt{K}}{|d_k|} + \frac{K\log n}{q}\right) \left( \sqrt{n}\|\bV\|_{\max} +\frac{q\sqrt{\log n }}{|d_k|}\right)\right]\right\}
\end{split}
\eeq
for each $1\le k \le K_{\max}$ and $i \in [n]$.
%where $\|\bbA\|_{\max}:=\max_{j,\ell}|A_{j\ell}|$ stands for the matrix maximum norm of a given matrix $\bbA = (A_{j\ell})$. 
% \beq \label{expansion_evector}
% \begin{split}
% \wh\bv_k(i) &=\bv_k(i) + \frac{1}{t_k}\sum_{l} \bW_{il} \bv_k(l) \\
% &+  \OO\left\{\frac{q}{\sqrt{n}|d_k|}  \left[\left(\frac{K\log n}{q} +\frac{qK}{|d_k|}\right) \left(1+\frac{q\sqrt{{\log n}}}{|d_k|}  \right)+ \frac{q\sqrt{\log n}}{|d_k|} \right]\right\} .
% \end{split}
% \eeq
% \beq\label{expansion_evector}
% \wh\bv_k(i) =\bv_k(i) + \frac{1}{t_k}\sum_{l} \bW_{il} \bv_k(l) + \OO\left( \frac{q}{|d_k|\sqrt{n}} \left(\frac{q\sqrt{ \log n }}{|d_k| }   +  \frac{\log n}{q}\right)\right)
% \eeq
Note that under the rescaling $t_k\to t_k/q$, \eqref{eq_tk} becomes  
\beq \label{eq_tkdk} 
t_k = d_k + \OO\left({q^2}/{|d_k|}\right) = d_k \left\{1+\oo[(\log n)^{-1}]\right\}, 
\eeq
where we have used the assumption $|d_k|\gg q\sqrt{\log n}$ in the second step. 

Let us denote by $\mathbf T:=\diag\{t_1,\cdots, t_K\}$ and $\mathbf T_{K_0}:=\diag\{t_1,\cdots, t_{K_0}\}$. Then in view of \eqref{expansion_evector}, we can deduce that for each $ i \in [n]$,
\beq\label{expansion_evectorV}
\begin{split}
\wh\bV_{K_0}(i) & =(\widehat  v_1(i),\cdots,\widehat  v_{K_0}(i))^T\\
&=\bV_{K_0}(i) +   \bbT_{K_0}^{-1} \sum_{l=1}^n  W_{il}\bV_{K_0}(l) + \frac{q}{\sqrt{n}}\bbD_{K_0}^{-1}\bcE_i ,
\end{split}
\eeq
where $\bcE_i\in \R^{K_0}$ is a random vector satisfying that w.h.p.,
\beq\label{eq_Eik}
\begin{split}
\cal E_i(k)&=\OO\left\{\frac{q \sqrt{ \log n }}{|d_k|}+ \left(\frac{q\sqrt{K}}{|d_k|} + \frac{K\log n}{q}\right) \left( \sqrt{n}\|\bV\|_{\max} +\frac{q\sqrt{\log n }}{|d_k|}\right) \right\} =o(1)
\end{split}
\eeq
under the assumptions of $|d_k|\gg q\sqrt{\log n}$ and \eqref{eq: condK}. It follows from \eqref{expansion_evectorV} that 
\beq\label{eq_hatV}
\begin{split}
\wh\bV_{K_0}(i) - \wh\bV_{K_0}(j)&=\bV_{K_0}(i)-\bV_{K_0}(j) + \bbT_{K_0}^{-1} \sum_{l=1}^n ( W_{il}- W_{jl})\bV_{K_0}(l)\\
&\quad +{\sqrt{\theta}}\bbD_{K_0}^{-1}\left({\bcE_i}-{\bcE_j}\right)
\end{split}
\eeq
for each pair of nodes $\{i, j\}$ with $ i \neq j \in [n]$. Recall that $\bSig_{i,j} =  \cov[(\be_i-\be_j)^T\bW\bV_{K_0}\bbD_{K_0}^{-1}]$. Hence, using \eqref{eq_tkdk} and the classical multidimensional CLT (see e.g., \cite{MCLT}) or the Berry--Esseen inequality (see, e.g.,  \cite{Raic2019}), we can show that 
\begin{align}\label{eq_hatV1}
\sup_{x\in \R}\Big[\P\Big(\Big\|\bSig_{i,j}^{-1/2}\cdot \bbT_{K_0}^{-1} \sum_{l=1}^n (W_{il}-W_{jl}) \bV_{K_0}(l)\Big\|^2\le x\Big) - F_{K_0}(x)\Big]\to 0.
\end{align}
% \begin{align}\label{eq_hatV1}
% \bSig_{i,j}^{-1/2}\cdot \bbT_{K_0}^{-1} \sum_{l=1}^n (\bW_{il}-\bW_{jl}) \bV_{K_0}(l) \stackrel{\mathscr D}{\lto} \cal N(0,I_{K_0})
% \end{align}
%as the network size $n$ increases, where $I_m$ denotes the identity matrix of size $m$.

With the aid of part (v) of Condition \ref{main_assm}, \eqref{eq_Eik}, and $|d_1| \lesssim n\theta $, we can deduce that 
\beq\label{eq_hatV2} 
\left\| \bSig_{i,j}^{-1/2}\cdot  {\sqrt{\theta}}\bbD_{K_0}^{-1}\left({\bcE_i}-{\bcE_j}\right)\right\| \lesssim \left\| {\bcE_i}-{\bcE_j}\right\| =o(1)\quad \text{w.h.p.},
\eeq
and
\begin{align}
\left\| \bSig_{i,j}^{-1/2} \left(\bV_{K_0}(i)-\bV_{K_0}(j)\right)\right\| &=\left\| \left(\bSig_{i,j}^{-1/2}\bD_{K_0}^{-1}\right) \cdot \bD_{K_0}\left(\bV_{K_0}(i)-\bV_{K_0}(j)\right)\right\| \nonumber\\
&\lesssim \theta^{-1/2} \left\|  \bD_{K_0} \left(\bV_{K_0}(i)-\bV_{K_0}(j)\right)\right\| .\label{ineq_DV}
\end{align}
Furthermore, if $c_{1n} \ll (d_{1}\lambda_1(\bP))^{-1/2}$, then it holds under \eqref{null_hyp} that 
\beq\label{eq_hatV3}
\left\| \bSig_{i,j}^{-1/2} \left(\bV_{K_0}(i)-\bV_{K_0}(j)\right)\right\| \lesssim  c_{1n}\sqrt{d_{1}\lambda_1(\bP)} = o(1).
\eeq
Thus, combining \eqref{eq_hatV}--\eqref{eq_hatV3}, we see that the term \eqref{eq_hatV1} dominates asymptotically under the null hypothesis $H_0$ in (\ref{eq: hypothesis}), which leads to \eqref{eq_null}.
% \beq\label{eq_null}
% T_{ij}=\left\| \bSig_{i,j}^{-1/2} \left(\wh\bV_{K_0}(i)-\wh\bV_{K_0}(j)\right)\right\|_2^2 \stackrel{\mathscr D}{\lto}  \chi_{K_0}^2
% \eeq
% \beq\label{eq_null}
% T_{ij}=\left\| \bSig_{i,j}^{-1/2} \left(\wh\bV_{K_0}(i)-\wh\bV_{K_0}(j)\right)\right\|_2^2 \stackrel{\mathscr D}{\lto}  \chi_{K_0}^2
% \eeq
%as the network size $n$ increases. This completes the proof for part (i) of Theorem \ref{thm:pair-null}.

We now proceed with proving part (ii) of Theorem \ref{thm:pair-null}. Using \eqref{alt_hyp} and part (v) of Condition \ref{main_assm},  an application of similar arguments to those for \eqref{ineq_DV} gives 
\begin{align}
\left\| \bSig_{i,j}^{-1/2} \left(\bV_{K_0}(i)-\bV_{K_0}(j)\right)\right\| \gtrsim 
c_{2n}\sqrt{d_{K}\lambda_K(\bP)} \gg 1, \label{ineq_DV2}
%\left(\lambda_{K_0}(\bbP) n\right)^{1/2}   \left\|  \bD_{K_0}^{1/2} \left(\bV_{K_0}(i)-\bV_{K_0}(j)\right)\right\|_2 
\end{align}
under $c_{2n} \gg (d_{K}\lambda_K(\bP))^{-1/2}$.
Therefore, combining \eqref{eq_hatV}--\eqref{eq_hatV2} and \eqref{ineq_DV2}, we see that 
% if $\sqrt{\theta }c_{1n} \ll n^{-1/2}$, then \eqref{eq_hatV1} dominates under $H_0$, i.e.,
% \beq\label{eq_null}
% T_{ij}=\left\| \bSig_{i,j}^{-1/2} \left(\wh\bV(i)-\wh\bV(j)\right)\right\|^2 \to \chi_K^2,
% \eeq while 
%if \eqref{eq_c2n} holds, then 
\eqref{ineq_DV2} dominates asymptotically under the alternative hypothesis $H_a$ in (\ref{eq: hypothesis-a}), which leads to \eqref{eq_power_pair}.
\iffalse
that is, 
\beq\label{eq_alt}
T_{ij}=\left\| \bSig_{i,j}^{-1/2} \left(\wh\bV(i)-\wh\bV(j)\right)\right\|_2^2 \to \infty
\eeq
as the network size $n$ increases. This concludes the proof for part (ii) of Theorem \ref{thm:pair-null}.
\fi

%Finally, the idea in Section \ref{sec:group-test} can be extended to the non-sharp-null setting.

% We consider a slight more general situation, where $\bP$ can have some small eigenvalues. Then, we have that
%  \begin{align*}
%  \theta_{\min}\|\bpi_i-\bpi_j\|^2 \leq (\bV(i)-\bV(j))^T\bD(\bV(i)-\bV(j))\leq C\theta\|\bpi_i-\bpi_j\|^2
%  \end{align*}
% for some $\theta_{\min}$ that can be much less than $\theta$. In this case, if $\sqrt{\theta }c_{1n} \ll n^{-1/2}$, then \eqref{eq_null} holds, while if $\frac{\theta_{\min}}{\sqrt{\theta}}c_{2n}\gg n^{-1/2}$, then \eqref{eq_alt} holds.  %{\color{red} A question: if $\mathbf P$ has vanishing smallest eigenvalue,  how will this affect the eigenvalues of $\mathbf D$?  Does the second displayed equation right above (3.1) still hold?}

\subsection{Proof of Theorem \ref{thm:group-null}} \label{Sec.newA.2}

The proof of Theorem \ref{thm:group-null} also relies on the asymptotic expansion for the spiked eigenvectors in \eqref{expansion_evector}. From the proof of Theorem \ref{thm:pair-null} in Section \ref{subsec:cal-thm-pair}, we have seen that $T_{ij}$ are asymptotic   $\chi_{K_0}^2$ random variables. Since the maximum of $m/2$ i.i.d. $\chi_{K_0}^2$ random variables converges in distribution to the Gumbel distribution as $m\to \infty$ under the scaling $1/2$ and shift $b_m$ in \eqref{eq_ambm} \citep{Extremal_Events}, we immediately get the conclusion if (a) $T_{ij}$'s are independent of each other and (b) they satisfy the exact $\chi_{K_0}^2$ distribution. However, both (a) and (b) only hold approximately and asymptotically. To deal with (a), we notice that the correlation between two eigenvector entries $\wh v_k(i)$ and $\wh v_{k'}(j)$, $i,j\in \cal M$, comes from the sum over $l\in \cal M$ in the second term in \eqref{expansion_evector}. Thus, removing the sum over $l\in \cal M$ from $T_{ij}$ and $T_{i'j'}$ ensures the asymptotic independence between two different pairs $\{i,j\}, \{i',j'\}\in \cal P$. Furthermore, with the second condition in \eqref{cond_m}, we can show that the above operation leads to an asymptotically negligible error (see \eqref{Bern_vw} below). To deal with (b), we need to control the Kolmogorov--Smirnov distance between the distribution of $T_{ij}$ and the $\chi_{K_0}^2$ distribution. We apply the Berry--Esseen inequality to get a bound \smash{$ {\|\bV_{K_0}\|_{\max}}/{\sqrt{\theta}}$} (see \eqref{BerryEsseen} below). To ensure that this error does not affect the asymptotic distribution of $T$, we need the first condition in \eqref{cond_m}. Finally, we remark that it is crucial that \eqref{expansion_evector} holds with high probability $1-\OO(n^{-D})$ for $D>1$. By taking a union bound, it holds simultaneously for all $T_{ij}$, $\{i,j\}\in \cal P$, with probability $1-\oo(1)$.  

As discussed in Section \ref{subsec:cal-thm-pair}, we take $K_0$ to be deterministic in the following proof. Recall that the variances of the entries of $\bbW$ are denoted as $\sigma^2_{ij}= \var(W_{ij})\lesssim \theta$. Then an application of \eqref{cond_m}, \eqref{upper-Vmax}, and \eqref{Ber_logn} yields that w.h.p.,
\beq\label{Bern_vw}
\begin{split}
\sum_{l\in \cal M} W_{il} v_k(l) & \lesssim  \|\bbV\|_{\max} {\log n} + \sqrt{\log n}\Big\{\sum_{l\in \cal M}\sigma^2_{il} [v_k(l)]^2\Big\}^{1/2}\\
&\ll \sqrt{{\theta}/{\log n}},
\end{split}
\eeq
for each $1 \leq k \leq K_0$ and $ i \in [n]$. Combining (\ref{Bern_vw}) with \eqref{expansion_evectorV}--\eqref{eq_Eik} yields that for $ i \neq j \in [n]$,
\begin{align}\label{Vij}
	&  \widehat\bbV_{K_0}(i)-\widehat\bbV_{K_0}(j) =\bbV_{K_0}(i)-\bbV_{K_0}(j)+\bbT_{K_0}^{-1}\wt\bU_{i,j}+ \sqrt{\theta}\bbD_{K_0}^{-1}{\bm\e}_{i,j}, %\OO_\prec\left(\frac{\al_n^2}{\sqrt{n}|t_k|}+ \frac1{\sqrt{n}}+ \sqrt{\frac{m\theta}{n}}\right),
\end{align}
where $\wt\bU_{i,j}:=\sum_{ l \in [n]\setminus \cal M} (W_{il}-W_{jl})  \bV_{K_0}(l) $ and $\bm\e_{i,j}\in \R^{K_0}$ is a random vector satisfying that {w.h.p.},
\beq\label{Eij}
\begin{split}
| \e_{i,j}(k)|&\lesssim \frac{q \sqrt{ \log n }}{|d_k|}+ \left(\frac{q\sqrt{K}}{|d_k|} + \frac{K\log n}{q}\right) \left( \sqrt{n}\|\bV\|_{\max} +\frac{q\sqrt{\log n }}{|d_k|}\right) \\
&
 \quad + \oo\left(\frac{1}{\sqrt{\log n}}\right) \ll  \frac{1}{\sqrt{\log n}}
\end{split}
\eeq
under the assumptions of %$m\ll {\sqrt{\theta}}/{\|\bbV_{K_0}\|_{\max}} \le q$, 
$|d_k|\gg q {\log n}$ and \eqref{eq: condK2}. With the aid of \eqref{upper-Vmax} and \eqref{Ber_logn}, we can further deduce that \text{w.h.p.},
\beq\label{Uij}
 \|\wt\bU_{i,j} \|  \lesssim  \|\bbV\|_{\max} {\log n}  + \sqrt{{\theta \log n}}  \le 2\sqrt{{\theta \log n}}.
\eeq

Let us introduce another covariance matrix $\wt\bSig_{i,j} =  \cov(\bbT_{K_0}^{-1} \wt\bU_{i,j}).$ Through a direct calculation, we can show that for each $1 \leq a, b \leq K_0$,
\begin{align*} 
&\Big(  \bD_{K_0}\bSig_{i,j}\bD_{K_0} - \bbT_{K_0}\wt\bSig_{i,j}\bbT_{K_0}\Big)_{ab} \\
= &\, \sum_{l\in \cal M} (\sigma^2_{il} + \sigma^2_{jl}) v_a(l)v_{b}(l)  - \sigma_{ij}^2\left[v_a(i)v_b(j)+ v_a(j)v_b(i)\right]  \\
\ll & \, \theta/(\log n)^2,
\end{align*}
where we have used the assumption \eqref{cond_m} in the second step above. This together with part (v) of Condition \ref{main_assm} and \eqref{eq_tkdk} entails that 
\beq\label{Vij0} 
\bbT_{K_0}\wt\bSig_{i,j}\bbT_{K_0} \sim  \theta,\ \ \bbD_{K_0} \wt\bSig_{i,j}\bbD_{K_0}  \sim  \theta, \ \ \bbT_{K_0} (  \bSig_{i,j}- \wt\bSig_{i,j} )\bbT_{K_0} =\oo\left\{ \frac{\theta}{(\log n)^2}\right\},
\eeq
% and
% \beq\label{Vij05}  
% \bbT\left(  \bSig_{i,j}- \wt\bSig_{i,j}\right)\bbT =\oo\left( \frac{\theta}{(\log n)^2}\right),
% \eeq
where the notations of asymptotic equivalence $\sim$ and asymptotic order $\oo(\cdot)$ are used in the sense of matrix eigenvalues. 
Then a combination of \eqref{Eij}--\eqref{Vij0} %under the conditions $|d_k|\gg (\log n) q$ and $1\ll m \ll q$, 
gives that w.h.p.,
\beq\label{Vij1}
\left\|\wt\bSig_{i,j}^{-1/2} \sqrt{\theta}\bbD_{K_0}^{-1}\bm\e_{i,j}\right\| \lesssim \|\bm\e_{i,j}\| \ll \frac{1}{\sqrt{\log n}},\quad 
%\eeq
%\beq\label{Vij2}
\left\|\wt\bSig_{i,j}^{-1/2} \bbT_{K_0}^{-1}\wt\bU_{i,j}\right\|\lesssim  \sqrt{\log n} .
\eeq
Moreover, in view of \eqref{ineq_DV} and \eqref{null_hyp}, if $c_{1n}\ll (d_{1}\lambda_1(\bP))^{-1/2}(\log n)^{-1/2}$, then it holds that 
\beq\label{Vij3}
\left\| \bSig_{i,j}^{-1/2} \left(\bV_{K_0}(i)-\bV_{K_0}(j)\right)\right\|  \lesssim c_{1n} \sqrt{d_{1}\lambda_1(\bP)} \ll \frac{1}{\sqrt{\log n}}.
\eeq

Now inserting \eqref{Vij} into equation \eqref{eq: T-test} and using results in \eqref{Vij0}--\eqref{Vij3}, we can deduce that \text{w.h.p.},
\begin{align*}
T_{ij}&=\wt \bU_{i,j}^\top \left(\bbT_{K_0}\bSig_{i,j}\bbT_{K_0}\right)^{-1} \wt\bU_{i,j}  + \oo(1)\\ &=\wt \bU_{i,j}^\top \left(\bbT_{K_0}\wt\bSig_{i,j}\bbT_{K_0}\right)^{-1} \wt\bU_{i,j}  + \oo(1).
\end{align*}
Then it follows from a simple union bound argument that \text{w.h.p.},
\beq\label{reduce_max}
T= \max_{\{i,j\}\in {\cal P} } \wt \bU_{i,j}^\top \left(\bbT_{K_0}\wt\bSig_{i,j}\bbT_{K_0}\right)^{-1} \wt\bU_{i,j}  + \oo(1).
\eeq
%for some constant $\delta>0$ depending on the model assumption. 
We can further apply the Berry--Esseen inequality (see, e.g.,  Theorem 1.1 of \cite{Raic2019}) to $ (\bbT_{K_0}\wt\bSig_{i,j}\bbT_{K_0})^{-1/2} \wt\bU_{i,j}  $ 
%e.g., \href{https://en.wikipedia.org/wiki/Berry\%E2\%80\%93Esseen_theorem}{Berry-Esseen theorem}), 
and obtain that uniformly in $x\ge 0$,  
\begin{align} 
&\left|\P\left\{ \wt \bU_{i,j}^\top \left(\bbT_{K_0}\wt\bSig_{i,j}\bbT_{K_0}\right)^{-1} \wt\bU_{i,j} \le x\right\} - \P\{\chi_{K_0}^2 \le x\}\right| \nonumber\\
 \lesssim &\, \sum_{ l \in [n]\setminus \cal M} \E\left[ \left\|\left(\bbT_{K_0}\wt\bSig_{i,j}\bbT_{K_0}\right)^{-1/2} (W_{il}-W_{jl})  \bV_{K_0}(l)\right\|^3\right]\nonumber\\ 
 \lesssim  &\, {\|\bV_{K_0}\|_{\max}}/{\sqrt{\theta}}. \label{BerryEsseen}
\end{align}
%where $\chi_{K_0}^2$ denotes the chi-square distribution with $K_0$ degrees of freedom. 
%where we also used \eqref{linfty_vk} $\chi_K^2$ denotes a chi-squared distribution with $K$ degrees of freedom.  
% Now we write the statistic $T$ as
% \beq\label{reduce_max}T= \max_{\{i,j\}\in \cal P} \wt \bU_{i,j}^\top\left(\bbT\wt\bSig_{i,j}\bbT\right)^{-1} \wt\bU_{i,j}  + \OO_\prec(n^{-\delta}),\eeq

Observe that random variables $\wt \bU_{i,j}^\top (\bbT_{K_0}\wt\bSig_{i,j}\bbT_{K_0})^{-1} \wt\bU_{i,j}$ in \eqref{reduce_max} with $\{i,j\}\in \cal P$ are independent since we have removed the entries with column indices $l\in \cal M$, and are in fact asymptotically distributed as $\chi_{K_0}^2$. In addition, it is well-known that the maximum of independent and identically distributed (i.i.d.) chi-square random variables converges weakly to the Gumbel distribution under some proper centering and rescaling as the sample size increases (see, e.g., Table 3.4.4 of \cite{Extremal_Events}), that is, 
%(see e.g., \href{https://math.stackexchange.com/questions/450139/asymptotics-of-maxima-of-i-i-d-chi-square-random-variables}{asymptotics of maxima of i.i.d. chi-square random variables}): 
\beq\label{max_Gumbel}
\frac{\max_{1\le i \le m/2}Y_i - b_m}{2}\toD \mathcal G
\eeq
as $m \rightarrow \infty$, where $Y_i$'s are i.i.d. $\chi_{K_0}^2$ random variables, $\cal G$ represents the Gumbel distribution with the cumulative distribution function (CDF) $\exp(-e^{-x})$ for $x \in \mathbb{R}$, and the centering parameter $b_m$ %and rescaling parameters $a_m$ 
is as given in \eqref{eq_ambm}. Therefore, combining \eqref{BerryEsseen} and (\ref{max_Gumbel}), we can obtain that for any $x\in \R$,
\begin{align*}
& \P\left\{  \max_{\{i,j\}\in \cal P} \wt \bU_{i,j}^\top\left(\bbT\wt\bSig_{i,j}\bbT\right)^{-1} \wt\bU_{i,j}  \le 2 x  + b_m\right\} \\
&= \left[ \P(\chi_{K_0}^2 \le 2 x + b_m) + \OO\left({\|\bV_{K_0}\|_{\max}}/{\sqrt{\theta}}\right)\right]^{m/2} \\
&= \left[ \P(\chi_{K_0}^2 \le 2 x  + b_m) \right]^{m/2} +   \OO\left(m {\|\bV_{K_0}\|_{\max}}/{\sqrt{\theta}}\right) \to  \exp({-e^{-x}})
\end{align*}
as $m \rightarrow \infty$, where we have used assumption \eqref{cond_m} and \eqref{max_Gumbel} in the last step and the convergence in the last step is uniform in $x$. Since the small-order $ \oo(1)$ term in \eqref{reduce_max} does not affect the asymptotic distribution, we see that \eqref{eq:conv_Gumbel} holds for the SIMPLE-RC test statistic $(T-b_m)/{2}$. 
%converges weakly to the Gumbel distribution as $m \rightarrow \infty$, which completes the proof of Theorem \ref{thm:group-null}.

\subsection{Proof of Theorem \ref{thm:group-alt}}\label{Sec.newA.3}

With \eqref{alt_hyp}, under \eqref{DV-Vgp},  we get that 
$$\max_{\{i, j\} \subset \cal M} \left\|\bD_{K_0} \left[\bV_{K_0}(i)-\bV_{K_0}(j)\right]\right\|  \gtrsim c_{2n}\sqrt{d_{K}\theta_{\min}}. $$
Together with part (v) of Condition \ref{main_assm} and Lemma \ref{lem-signal}, it implies that with probability $1-\oo(1)$,
\begin{align*}
    &  \max_{\{i,j\}\in \cal P}  \left\| \bSig_{i,j}^{-1/2} \left(\bV_{K_0}(i)-\bV_{K_0}(j)\right)\right\| \\
    \gtrsim &\, \theta^{-1/2} \max_{\{i,j\}\in \cal P}\|\bD_{K_0}(\bV_{K_0}(i)-\bV_{K_0}(j))\| \\
    \ge &\, \frac {\theta^{-1/2}}{3}   \max_{ \{i,j\} \subset \cal M}\|\bD_{K_0}(\bV_{K_0}(i)-\bV_{K_0}(j))\| \\
 \gtrsim &\, c_{2n}\sqrt{d_{K}\lambda_{K}(\bbP) } \gg \sqrt{\log n}. 
\end{align*} 
Moreover, an application of similar arguments to those in the proof of Theorem \ref{thm:group-null} in Section \ref{Sec.newA.2} results in 
$$\max_{\{i,j\}\in \cal P}\left\| \bSig_{i,j}^{-1/2} \left(\bbT_{K_0}^{-1}\wt\bU_{i,j}+  {\sqrt{\theta}}\bbD_{K_0}^{-1}\bm\e_{i,j}\right)\right\| \lesssim \sqrt{\log n}$$
with probability $1-\oo(1)$. Thus, combining the above two bounds with \eqref{Vij}, we obtain that 
$$T^{1/2}=\max_{\{i,j\}\in \cal P} \left\| \bSig_{i,j}^{-1/2} \left(\wh\bV_{K_0}(i)-\wh\bV_{K_0}(j)\right)\right\| \gg \sqrt{\log n}$$
with probability $1-\oo(1)$. This concludes Theorem \ref{thm:group-alt}.

\subsection{Proof of Theorem \ref{thm_consist_Sigma}} \label{Sec.newA.4}

For each pair of nodes $ i \neq j \in [n]$, we will aim at controlling the entrywise differences %between $\wh\bSig_{i,j}$ and $\bSig_{i,j}$ 
$$\theta^{-1}\left[\bbD_{K_0}(\wh\bSig_{i,j}(K_0)-\bSig_{i,j}(K_0))\bbD_{K_0}\right]_{ab},\quad 1\le a,b\le K_0.$$
Specifically, it follows from the definitions that 
\begin{align*}
(\bbD_{K_0}\bSig_{i,j}\bbD_{K_0})_{ab}&=\sum_{l=1}^n (\sigma^2_{il}+\sigma^2_{jl}) v_a(l)v_b(l)  -  \sigma^2_{ij}[v_a(i)v_b(j)+v_a(j)v_b(i)],\\
(\bbD_{K_0}\wh \bSig_{i,j}\bbD_{K_0})_{ab}&=\frac{d_ad_b}{\wh d_a\wh d_b } \sum_{l=1}^n (\wh w^2_{il}+\wh w^2_{jl}) \wh v_a(l)\wh v_b(l) -  \frac{d_ad_b}{\wh d_a\wh d_b }\wh w^2_{ij} [ \wh v_a(i)\wh v_b(j)+ \wh v_a(j)\wh v_b(i)].
\end{align*}
Using the above representations, we can deduce that for $1 \leq a ,b \leq K_0$,
\begin{align}
&\left[\bbD_{K_0}(\wh \bSig_{i,j}-\bSig_{i,j})\bbD_{K_0}\right]_{ab}\nonumber\\
&=\left(\frac{d_ad_b}{\wh d_a\wh d_b }-1\right) \left\{\sum_{l=1}^n (\sigma^2_{il}+\sigma^2_{jl}) v_a(l)v_b(l)  -  \sigma^2_{ij}[v_a(i)v_b(j)+v_a(j)v_b(i)]\right\} \nonumber\\
&+ \frac{d_ad_b}{\wh d_a\wh d_b } \left\{\sum_{l=1}^n (\wh w^2_{il}+\wh w^2_{jl}) \wh v_a(l) \wh v_b(l) -\sum_{l=1}^n (\sigma^2_{il}+\sigma^2_{jl}) v_a(l) v_b(l)   \right\}  \nonumber\\
&+\frac{d_ad_b}{\wh d_a\wh d_b } \left\{\sigma^2_{ij}[v_a(i)v_b(j)+v_a(j)v_b(i)]- \wh w^2_{ij} [\wh v_a(i)\wh v_b(j)+\wh v_a(j)\wh v_b(i)]\right\}. \label{diff_sigma}
\end{align}
Here, the first term on the RHS can be bounded using \eqref{eq_evalue_conv}, while the third term is simply bounded by $\|\bbV\|_{\max}^2$ because we will show that the entries of $\wh \bv_a$ are bounded by $\|\bbV\|_{\max}$. We mainly need to control the second term on the RHS, which can be rewritten as $(d_ad_b)/(\wh d_a \wh d_b)$ times  
\begin{align*}
    & \sum_{l=1}^n [(\wh w^2_{il}+\wh w^2_{jl})-(w^2_{il}+w^2_{jl})] \wh v_a(l) \wh v_b(l)  \\
    +& \sum_{l=1}^n  (w^2_{il}+w^2_{jl}) [ \wh v_a(l) \wh v_b(l)- v_a(l) v_b(l)]  \\
    +& \sum_{l=1}^n[(w^2_{il}+w^2_{jl})- (\sigma^2_{il}+\sigma^2_{jl})] v_a(l) v_b(l). 
\end{align*}
The second term can be bounded using \eqref{expansion_evector}, which shows that every $\wh v_a(l)$ is close to $v_a(l)$ (see \eqref{linfty_hatvk} below), and the third term can be bounded using the Bernstein's inequality in Lemma \ref{lemma_bern} with respect to the centered random variables $w_{il}^2-\sigma_{il}^2$. For the first term, we need to control the differences $\wh w_{ij}-w_{ij}$. Using the definition \eqref{whW}, we can show that %w.h.p.
\begin{align}
\wh w_{ij} - w_{ij}  = & \sum_{k=1}^{K_0} d_k  \left[ v_k(i)  v_k (j)-\wh v_k(i) \wh v_k (j)\right] + \sum_{k=1}^{K_0} (d_k - \wh d_k) \wh v_k(i) \wh  v_k(j) \nonumber\\
& + \sum_{k=K_0+1}^{K} d_k  v_k(i)  v_k(j) .\label{w-w}
\end{align}
The first term is again bounded using \eqref{expansion_evector}, the second term is bounded using \eqref{eq_evalue_conv}, and the last term leads to an error $\cal E(K_0)$.

%{\color{red} Add a high-level summary of the main ideas/ingredients/challenges/innovations of the proof here.}

For simplicity of presentation, in the following proof, we only show the argument for bounding \eqref{diff_sigma} with $a=b$, while the differences between off-diagonal entries can be bounded in a similar fashion except for some minor changes of notations. Observe that with the aid of \eqref{eq_evalue_conv} using the rescalings $t_k\to t_k/q$, $d_k\to d_k/q$, and $\wh d_k\to \wh d_k/q$, we can obtain that w.h.p.,
\beq\label{hatd-d} 
\begin{split}
|\wh d_k - t_k| & = \OO\left( {\log n}  +\frac{q^4}{|d_{k}|^4}{K\log n}\right) = \OO(\log n),
\end{split}
\eeq
where we have used assumption \eqref{eq: condK} in the second step above. Then a combination of (\ref{hatd-d}) and \eqref{eq_tkdk} entails that w.h.p.,
\beq\label{eq_dk_conv} \frac{|\wh d_k - d_k|}{|d_k|} =\OO\{\cal E_1(K_0)\} ,
\eeq
where we denote  
$$\cal E_1(K_0):=\frac{\log n}{|d_{K_0}|} + \frac{q^2}{d_{K_0}^2} \ll \frac{1}{\log n}. $$
In addition, using the fact of $\sigma^2_{kl}=\OO(\theta)$, it holds that 
$$\sum_{l=1}^n (\sigma^2_{il}+\sigma^2_{jl})[v_a(l)]^2 = \OO(\theta),\quad  \sigma^2_{ij}|v_a(i)v_a(j)| = \OO(\theta).$$
Hence, in light of \eqref{eq_dk_conv}, we can bound the first term on the RHS of \eqref{diff_sigma} with $a=b$ as w.h.p.,
\beq\label{eq_first_diff1}
\left(\frac{d_a^2}{\wh d_a^2 }-1\right) \left\{\sum_{l=1}^n (\sigma^2_{il}+\sigma^2_{jl})[v_a(l)]^2  - 2\sigma^2_{ij}v_a(i)v_a(j)\right\} =\OO\{\theta \cal E_1(K_0)\} .
\eeq

It remains to bound the second and third terms on the RHS of \eqref{diff_sigma}. To this end, we will need to analyze the term $|\wh w_{kl}-w_{kl}|$ by controlling \eqref{w-w}. 
In view of \eqref{defn_En}, the third term on the RHS of \eqref{w-w} can be simply bounded as
\beq\label{w-w1}
\sum_{k=K_0+1}^{K} d_k  v_k(i)  v_k (j) =  \OO\{\theta\cal E(K_0)\}.
\eeq
Using \eqref{Ber_logn}, we obtain that w.h.p.,
\beq\label{sub_leadingv}
\begin{split}
\frac{1}{t_k}\sum_{l = 1}^n W_{il} v_k(l)  & =\OO\left(  \frac{\|\bv_{k}\|_\infty }{ |d_k|}\log n + \frac{\sqrt{ \theta \log n }}{|d_k|}\right)\\
&=\OO\left( \frac{\sqrt{ \theta \log n }}{|d_k|}\right) \ll n^{-1/2} ,
\end{split}
\eeq
where we have used the fact \eqref{upper-Vmax0} in the second step and the assumption $|d_k|\gg q\sqrt{\log n}$ in the third step above. Since the third term on the RHS of \eqref{expansion_evector} is of asymptotic order $\oo(\sqrt{\theta}/|d_{k}|)$, it follows that w.h.p.,
\beq\label{linfty_hatvk}
\wh v_k(i)=v_k(i) + \OO\left( \frac{\sqrt{ \theta \log n }}{|d_k|}\right)  .%\max_{1\le k \le K_0}\|\wh \bv_k\|_\infty =\OO(n^{-1/2}).
\eeq
Combining \eqref{eq_tkdk}, \eqref{hatd-d} and \eqref{linfty_hatvk}, we can bound the second term on the RHS of \eqref{w-w} as 
\beq\label{w-w2}
\begin{split}
\sum_{k=1}^{K_0} (d_k - \wh d_k) \wh v_k(i) \wh  v_k (j) & = \OO\left\{  \sum_{k=1}^{K_0}\left(\log n + \frac{q^2}{|d_k|}\right)\|\bv_k\|_\infty^2 \right\}   = \OO\left\{ \theta \cal E_2(K_0) \right\}
\end{split}
\eeq
w.h.p., and bound the first term on the RHS of \eqref{w-w} as 
\begin{align}\label{w-w3}
    \sum_{k=1}^{K_0} d_k  \left[ v_k(i)  v_k(j)-\wh v_k(i) \wh v_k(j)\right] =\OO\left\{ \theta \cal E_2(K_0) \right\}
\end{align}
w.h.p., where we define $\cal E_2(K_0)$ as 
$$\cal E_2(K_0):=\theta^{-1}{\|\bbV\|_{\max}^2 \log n} + \sqrt{\theta^{-1}{\|\bbV\|_{\max}^2 \log n}} + \frac{n}{|d_{K_0}|} \|\bbV\|_{\max}^2.$$ 
Furthermore, it follows from  \eqref{upper-Vmax0} that 
\beq\label{E2n-est}
\begin{split}
\cal E_2(K_0) & =(1+\oo(1))\sqrt{\theta^{-1}{\|\bbV\|_{\max}^2 \log n}}   \ll (\log n)^{-1/2} .%+\frac{n}{|d_{K_0}|}\frac{\sqrt{\theta}}{\log n}  \frac{|d_{K_0}|}{q\sqrt{n}} \le 2(\log n)^{-1/2}.
\end{split}
\eeq
%and we used $|d_k|\gg q\sqrt{\log n}$ and $K_0=\OO(1)$ in the second step. 
% Next, using \eqref{linfty_hatvk}, we can calculate the first term on the RHS of \eqref{w-w} as 
% \begin{align}\label{w-w3}
%     \sum_{k=1}^{K_0} d_k  \left( \bv_k(i)  \bv_k^\top(j)-\wh \bv_k(i) \wh \bv_k^\top(j)\right) =\OO\left(  \|\bbV\|_{\max} \sqrt{\theta \log n } \right) \quad \text{w.h.p.}
% \end{align}
Thus, plugging \eqref{w-w1}--\eqref{w-w3} into \eqref{w-w} yields that w.h.p.,
\beq\label{w-w4}
|\wh w_{ij} - w_{ij}|=\OO\left\{\theta\cal E(K_0)+\theta\cal E_2(K_0)\right\}.
\eeq

Observe that $w_{ij}\le 1$ and $\wh w_{ij}=\OO(1)$ w.h.p. due to \eqref{w-w4}. This along with \eqref{eq_dk_conv} and \eqref{linfty_hatvk} enables us to bound the third term on the RHS of \eqref{diff_sigma} with $a=b$ as w.h.p.,
\beq\label{eq_third_diff}
\begin{split}
\frac{d_a^2}{\wh d_a^2 } & \left[2\sigma^2_{ij}v_a(i)v_a(j)- 2\wh w^2_{ij} \wh v_a(i)\wh v_a(j)\right]  =\OO\left(\theta\frac{\|\bbV\|_{\max}^2}{\theta} \right) =\oo\left\{\theta\cal E_2(K_0) \right\}.
\end{split}
\eeq
To further bound the second term on the RHS of \eqref{diff_sigma}, an application of \eqref{w-w4} gives that 
\begin{align}
    &\, \sum_{l=1}^n (\wh w^2_{il}+\wh w^2_{jl}) [\wh v_a(l)]^2 -\sum_{l=1}^n (\sigma^2_{il}+\sigma^2_{jl}) [v_a(l)]^2 \nonumber\\
      =&\, \sum_{l=1}^n (w^2_{il}+w^2_{jl}) [\wh v_a(l)]^2 -\sum_{l=1}^n (\sigma^2_{il}+\sigma^2_{jl}) [v_a(l)]^2  + \OO\left\{\theta\cal E(K_0)+\theta\cal E_2(K_0)\right\}\label{eq_second_diff1}
\end{align}
w.h.p. We can decompose the first two terms on the RHS of (\ref{eq_second_diff1}) as
\begin{align}
  \sum_{l=1}^n (w^2_{il}+w^2_{jl})\left\{[\wh v_a(l)]^2 - [v_a(l)]^2\right\}+\sum_{l=1}^n \left[(w^2_{il}+w^2_{jl})-(\sigma^2_{il}+\sigma^2_{jl})\right] [v_a(l)]^2 .\label{eq_second_diff2}
\end{align}

In light of \eqref{Ber_logn}, we can deduce that w.h.p.,
\beq\label{CLT_w000}
%\begin{split}
\sum_{l=1}^n \left[(w^2_{il}+w^2_{jl})-(\sigma^2_{il}+\sigma^2_{jl})\right] =\OO\left( \sqrt{n\theta \log n} + \log n\right)   \ll n\theta,
%\end{split}
\eeq
\beq\label{CLT_w111}
\sum_{l=1}^n \left[(w^2_{il}+w^2_{jl})-(\sigma^2_{il}+\sigma^2_{jl})\right]|v_a(l)| =\OO\left( \sqrt{\theta \log n} +\|\bbV\|_{\max} \log n\right) ,
\eeq
%and
\beq\label{CLT_w}
\begin{split}
\sum_{l=1}^n \left[(w^2_{il}+w^2_{jl})-(\sigma^2_{il}+\sigma^2_{jl})\right] [v_a(l)]^2 &=\OO \left(  \sqrt{\|\bbV\|_{\max}^2\theta \log n} + \|\bbV\|_{\max}^2\log n \right) \\
&=\OO \left\{ \theta\cal E_2(K_0)\right\},
\end{split}
\eeq
where \eqref{CLT_w} provides a bound on the second term on the RHS of \eqref{eq_second_diff2}, while \eqref{CLT_w000} and \eqref{CLT_w111} will be used to analyze the first term on the RHS of \eqref{eq_second_diff2}. With the aid of \eqref{linfty_hatvk}, we can obtain that w.h.p.,
\begin{align} 
\sum_{l=1}^n  & (w^2_{il}+w^2_{jl})\left|[\wh v_a(l)]^2 - [v_a(l)]^2\right| \lesssim \sum_{l=1}^n (w^2_{il}+w^2_{jl})\left[ \frac{\sqrt{ \theta \log n }}{|d_a|} |v_a(l)|+ \frac{  \theta \log n  }{|d_a|^2}\right] \nonumber\\
& \lesssim  \left( \sqrt{n}\theta + \sqrt{\theta \log n} +\|\bbV\|_{\max} \log n\right)  \frac{\sqrt{ \theta \log n }}{|d_a|}   + n\theta\cdot \frac{  \theta \log n  }{|d_a|^2}  \nonumber\\
&\le  \theta \cdot 2\frac{q\sqrt{\log n}}{|d_a|}
,\label{eq_second_diff3}
\end{align}
where we have used \eqref{CLT_w000}, \eqref{CLT_w111}, and the facts that $\sum_{l=1}^n (\sigma^2_{il}+\sigma^2_{jl})=\OO(n\theta)$ and $\sum_l(\sigma^2_{il}+\sigma^2_{jl}) |\bv_a(l)|\le \sqrt{n}\theta$ in the second step, and used \eqref{upper-Vmax0} and the assumption $|d_k|\gg q\sqrt{\log n}$ in the third step above.

% where in the second step we used $\sum_{l=1}^n (\sigma^2_{il}+\sigma^2_{jl})=\OO(n\theta)$ and that by \eqref{Ber_logn}, 
% $$ \sum_{l=1}^n \left[(w^2_{il}+w^2_{jl})-\E(w^2_{il}+w^2_{jl})\right] =\OO\left( \sqrt{n\theta \log n} + \log n\right) \ll n\theta\quad \text{w.h.p.}$$
Therefore, combining \eqref{eq_first_diff1}, \eqref{eq_third_diff}, \eqref{eq_second_diff1}, \eqref{CLT_w}, and \eqref{eq_second_diff3}, we can finally bound \eqref{diff_sigma} as  (recall the definition \eqref{wtEn})
\begin{align}
\left|[\bbD_{K_0}(\wh \bSig_{i,j}-\bSig_{i,j})\bbD_{K_0}]_{aa}\right| &\lesssim \theta\left\{\cal E(K_0) + \cal E_1(K_0)+ \cal E_2(K_0)+ \frac{q\sqrt{ \log n }}{|d_{K_0}|}\right\} \nonumber\\
& \lesssim \theta\left\{ \cal E(K_0)+\wt{\cal E}(K_0)\right\}, \label{diff_sigma_conc}
\end{align}
where we have used \eqref{E2n-est} and the fact that $\cal E_1(K_0)\ll  {q\sqrt{ \log n }}/{|d_{K_0}|}$ in the second step above. Using a similar argument, we can obtain the same bound for the off-diagonal entries of \smash{$\bbD_{K_0}(\wh \bSig_{i,j}-\bSig_{i,j})\bbD_{K_0}$}.  
This concludes the proof of Theorem \ref{thm_consist_Sigma}.

\subsection{Proof of Theorem \ref{thm:pg-sample}} \label{Sec.newA.thm5}

By \eqref{Ber_logn}, we indeed have that \eqref{checkq_q} holds w.h.p. 
Then for $\wh K_0$ defined in \eqref{whK}, an application of Theorem \ref{asymp_evalue} yields that $\wh K_0 \le K_1 \le K_{\max}\wedge C_0$ w.h.p. Moreover, under \eqref{cond:Pi}, we have 
\beq\label{pf-Vmax}
\|\bV\|_{\max}\lesssim \sqrt{K/n}
\eeq
by \eqref{Vmax0}. With \eqref{pf-Vmax}, it is easy to check the condition \eqref{eq: condK} via \eqref{eq: condK-app}, so Theorem \ref{thm:pair-null} holds for $K_0=\wh K_0$. Then combining Theorem \ref{thm:pair-null}, Theorem \ref{thm_consist_Sigma} and \eqref{eq:small_E} concludes part (i). 

Similarly, for $\wh K_0$ defined in \eqref{whK_group}, an application of Theorem \ref{asymp_evalue} yields that $\wh K_0 \le K_2 \le K_1 \le K_{\max}\wedge C_0$ w.h.p. Using \eqref{pf-Vmax}, \eqref{eq: condK2-app} and that $1\ll m \ll q/\sqrt{K}$, it is easy to check that \eqref{eq: condK2} and \eqref{cond_m} hold, so Theorems \ref{thm:group-null} and \ref{thm:group-alt} hold for $K_0=\wh K_0$. Furthermore, with the aid of Theorem \ref{thm_consist_Sigma} and \eqref{eq:small_E2}, it is easy to check that $\wh T(\wh K_0)=(1+\oo(1))T(\wh K_0)$ w.h.p. We omit the full technical details here for simplicity since it follows directly from the arguments in Section \ref{Sec.newA.2}. This completes the proof of part (ii) of Theorem \ref{thm:pg-sample}.

\iffalse
The arguments for the proof of Theorem \ref{thm:pg-sample} are similar to those for the proofs of Theorems \ref{thm:pair-null}--\ref{thm:group-alt} in Sections \ref{subsec:cal-thm-pair}--\ref{Sec.newA.3}, respectively. The only difference is that we will need to control an extra term $|\wt T_{ij}-T_{ij}|$ using Theorem \ref{thm_consist_Sigma} and part (v) of Condition \ref{main_assm}. We omit the full technical details here for simplicity, which completes the proof of Theorem \ref{thm:pg-sample}.
\fi

\subsection{Proof of Theorem \ref{thm:pair-null-het}} \label{appd_pair_het}

Using \eqref{expansion_evector}, we can express $Y_i(k)-Y_j(k)$ as a deterministic term ${v_k(i)}/{v_1(i)} - {v_k(j)}/{v_1(j)}$ plus \smash{$f_{k}^{(i,j)}$} and some small enough errors. With the classical CLT, $ \bbf^{(i,j)} $ is an asymptotic multivariate normal random vector with covariance \smash{$ \bSig^{(2)}_{i,j}$}. Then, the proof for Theorem \ref{thm:pair-null-het} is similar to those of the proof for Theorem \ref{thm:pair-null} in Section \ref{subsec:cal-thm-pair}: under $H_0$ with $c_{1n} \ll (Kd_1\lambda_1(\bP))^{-1/2}$, the vector $ \bbf^{(i,j)} $ dominates and leads to the asymptotic $\chi_{K_0-1}^2$ distribution; the deterministic part dominates under $H_a$ when \smash{$c_{2n}\gg (d_{K}\lambda_{K}(\bbP))^{-1/2} $}, which leads to \eqref{eq_power_pair_DCMM}.

We begin with proving part (i) of Theorem \ref{thm:pair-null-het}. From the asymptotic expansion  \eqref{expansion_evector} and the assumptions, we can deduce that for each $2 \leq k \leq K_0$ and $i\in [n]$, 
$$\wh v_k(i)=v_k(i) +  \frac{\be_i^\top \bW\bv_{k}}{t_k} + \frac{\sqrt{\theta}}{|d_k|} \cdot \oo(1)$$
and
\begin{align*}
\wh v_1(i)&=v_1(i) +  \frac{\be_i^\top \bW\bv_{1}}{t_1}  +  \OO\left\{ \frac{\sqrt{\theta}}{|d_1|} \left[\frac{q \sqrt{ \log n }}{|d_1|}+ \left(\frac{q\sqrt{K}}{|d_1|} + \frac{K\log n}{q}\right) \left( \sqrt{K}  +\frac{q\sqrt{\log n }}{|d_1|}\right)\right]\right\} \\
&=v_1(i) +  \frac{\be_i^\top \bW\bv_{1}}{t_1}  +  \OO\left( \frac{\sqrt{\theta}}{|d_1|} \cdot \frac{K^{3/2}\log n}{q} \right) \\
&=v_1(i) +  \frac{\be_i^\top \bW\bv_{1}}{t_1}  + \frac{\sqrt{\theta}}{|d_1|\sqrt{K}}\cdot \oo(1)
\end{align*}
w.h.p., where we have used the facts of $\sqrt{n}\|\bV\|_{\max}\lesssim \sqrt{K}$ by \eqref{Vmax0} and part (iii) of Condition \ref{main_assm_DCMM}, $d_1\gtrsim n\theta=q^2$, and assumption \eqref{eq: condK02}. Then it follows from the above two asymptotic expansions that w.h.p.,
\beq\label{Yik-Yjk}
\begin{split}
  Y_i(k)-Y_j(k) & = \frac{v_k(i) +  t_k^{-1}\be_i^\top \bW\bv_{k} + \sqrt{\theta}|d_k|^{-1} \cdot \oo(1)}{v_1(i) +  t_1^{-1}\be_i^\top \bW\bv_{1} + \sqrt{\theta}|d_1|^{-1} K^{-1/2} \cdot \oo(1)} \\
  &\quad- \frac{v_k(j) +  t_k^{-1}\be_j^\top \bW\bv_{k} + \sqrt{\theta}|d_k|^{-1} \cdot \oo(1)}{v_1(j) +  t_1^{-1}\be_j^\top \bW\bv_{1} + \sqrt{\theta}|d_1|^{-1}K^{-1/2} \cdot \oo(1)}
\end{split}
\eeq
for each $2 \leq k \leq K_0$ and each pair of nodes $i \neq j \in [n]$.

Note that by \eqref{sub_leadingv}, it holds that w.h.p.,
\beq\label{Wvt} 
t_k^{-1}\be_i^\top \bW\bv_{k} =\OO\left( \frac{\sqrt{ \theta \log n }}{|d_k|}\right)\ll n^{-1/2}  
%,\quad t_1^{-1}\be_i^\top \bW\bv_{1} =\OO\left( \frac{\sqrt{ \theta \log n }}{|d_1|}\right) \ll n^{-1/2} .     
\eeq
for each $1 \leq k \leq K_0$.
%and $\sqrt{\theta}|d_k|^{-1} \ll (n\log n)^{-1/2}$ under the condition $|d_k|\gg q\sqrt{\log n}$. 
Then, in view of \eqref{Wvt} and part (iv) of Condition \ref{main_assm_DCMM}, expanding \eqref{Yik-Yjk} yields that w.h.p.,
\begin{align}
  Y_i(k) & -Y_j(k) = \frac{v_k(i)}{v_1(i)} - \frac{v_k(j)}{v_1(j)} + f_k^{(i,j)}  -\frac{ (\be_i^\top \bW\bv_{k})(\be_i^\top \bW\bv_{1})}{t_k t_1 v_1^2(i)} +\frac{ (\be_j^\top \bW\bv_{k})(\be_j^\top \bW\bv_{1})}{t_k t_1 v_1^2(j)} \nonumber\\
   & \quad + \left\{\frac{ \sqrt{\theta} }{|d_k| v_1(i)} +  \frac{ |v_k(i)|\sqrt{\theta} }{|d_1|\sqrt{K} [v_1(i)]^2}+\frac{\sqrt{\theta} }{|d_k| v_1(j)} +  \frac{ |v_k(j)|\sqrt{\theta} }{|d_1| \sqrt{K} [v_1(j)]^2}\right\} \oo(1) \nonumber\\
   &= \frac{v_k(i)}{v_1(i)} - \frac{v_k(j)}{v_1(j)} + f_k^{(i,j)}  + \cal E_{ij}(k)\label{Yijk}
\end{align}
for each $2 \leq k \leq K_0$ and each pair of nodes $i \neq j \in [n]$, where %we used $\sqrt{n}\|\bV\|_{\max}\lesssim \sqrt{K}$ in the second step and 
$f_k^{(i,j)}$ is given in (\ref{asymp-variance2}) and $\bcE_{ij} = (\cal E_{ij}(k))_{2 \leq k \leq K_0} \in \R^{K_0-1}$ is a random vector satisfying that w.h.p.,
\begin{equation}\label{eq:addYijk}
\begin{split}
\cal E_{ij}(k) & =\OO\left( \frac{q^2\log n}{|d_k||d_1|}\right) + \oo\left(\frac{ q }{|d_k| } \right) = \oo\left(\frac{ q }{|d_k| } \right).
\end{split}
\end{equation}
%with $2\le k \le K_0$.

In light of part (ii) of Condition \ref{main_assm_DCMM} and \eqref{eq:addYijk}, we can show that w.h.p.,
\beq\label{eq_hatV2_ratio}
\begin{split}
\left\|\left(\bSig^{(2)}_{i,j}\right)^{-1/2} \bcE_{ij}\right\| & = \left\|\left(\bSig^{(2)}_{i,j}\right)^{-1/2} \wt \bD_{K_0}^{-1} \cdot \wt \bD_{K_0}\bcE_{ij}\right\|\ll 1
\end{split}
\eeq
and that under the assumption of  $c_{1n} \ll  (Kd_1\lambda_1(\bP))^{-1/2}$,
\begin{align}
   & \left\|\left(\bSig^{(2)}_{i,j}\right)^{-1/2} \left[\frac{\wt\bV_{K_0-1}(i)}{v_1(i)} - \frac{\wt\bV_{K_0-1}(j)}{v_1(j)}\right]\right\| \nonumber\\
    =& \left\|\left(\bSig^{(2)}_{i,j}\right)^{-1/2} \wt \bD_{K_0}^{-1} \cdot \wt \bD_{K_0} \left[\frac{\wt\bV_{K_0-1}(i)}{v_1(i)} - \frac{\wt\bV_{K_0-1}(j)}{v_1(j)}\right]\right\|  \nonumber\\
    \ll &\, c_{1n}\sqrt{Kd_1\lambda_1(\bP)} \ll 1 ,\label{ineq-DV_ratio}
\end{align}
where $\wt\bV_{K_0-1}$ denotes the $n\times (K_0-1)$ submatrix formed by the second to the $K_0$th columns of matrix $\bV$, and we have used \eqref{diff_ratios_null} and the simple identity 
$$ \left\|\bD_{K_0}\left[\frac{\bV_{K_0}(i)}{v_1(i)} - \frac{\bV_{K_0}(j)}{v_1(j)}\right] \right\|=\left\|\wt\bD_{K_0}\left[\frac{\wt\bV_{K_0-1}(i)}{v_1(i)} - \frac{\wt\bV_{K_0-1}(j)}{v_1(j)}\right] \right\|.$$
In addition, as shown in Section A.4 of \cite{SIMPLE}, it holds that 
\begin{align}\label{eq_hatV1_ratio}
\sup_{x\in \R}\Big[\P\Big(\left\| \big(\bSig^{(2)}_{i,j}\big)^{-1/2} \bbf^{(i,j)}\right\|^2\le x\Big) - F_{K_0-1}(x)\Big]\to 0.
\end{align}
% \begin{align}\label{eq_hatV1_ratio}
%  \left(\bSig^{(2)}_{i,j}\right)^{-1/2} \bbf^{(i,j)} \toD \cal N(0,I_{K_0-1})
% \end{align}
%as network size $n \rightarrow \infty$. 
Hence, combining \eqref{eq_hatV2_ratio}, \eqref{ineq-DV_ratio} and \eqref{eq_hatV1_ratio} concludes part (i) of Theorem \ref{thm:pair-null-het}.

We now proceed with establishing part (ii) of Theorem \ref{thm:pair-null-het}. From \eqref{diff_ratios_alt}, \eqref{DV-Vgp_het}, and part (ii) of Condition \ref{main_assm_DCMM}, we can obtain that under $H_a$ and $c_{2n} \gg (d_{K}\lambda_K(\bP))^{-1/2}$,
%assumption \eqref{hypothesis1_DCMM},
\begin{align}
   \left\|\left(\bSig^{(2)}_{i,j}\right)^{-1/2} \left[\frac{\wt\bV_{K_0-1}(i)}{v_1(i)} - \frac{\wt\bV_{K_0-1}(j)}{v_1(j)}\right]\right\| & \gtrsim
   c_{2n} \sqrt{d_{K}\lambda_K(\bP)} \gg 1, \label{ineq-DV_ratio_alt}
   %\sqrt{d_{K_0}} \lambda_{\min}^{1/2}\left\{ \left(\bpi_i , \bpi_j\right)^\top  \left(\bpi_i , \bpi_j\right)\right\} \nonumber\\
   %&\gg 1, \label{ineq-DV_ratio_alt}
\end{align}
which dominates the other terms \eqref{eq_hatV2_ratio} and \eqref{eq_hatV1_ratio} asymptotically. Therefore, the desired conclusion in \eqref{eq_power_pair_DCMM} follows from such result, which concludes part (ii) of Theorem \ref{thm:pair-null-het}.

\subsection{Proof of Theorem \ref{thm:group-het}} \label{appd_group_het}

The main ingredients for the proof of Theorem \ref{thm:group-het} are similar to those for the proofs of Theorems \ref{thm:group-null} and \ref{thm:group-alt} in Sections \ref{Sec.newA.2} and \ref{Sec.newA.3}, respectively. To simplify the technical presentation, we will provide here an outline of the arguments instead of the full details. Observe that similar to \eqref{Vij}, we can write \eqref{Yijk} as 
\beq\label{Yijk2}
Y_i(k)-Y_j(k)= \frac{v_k(i)}{v_1(i)} - \frac{v_k(j)}{v_1(j)} + \wt f_k^{(i,j)}  + \wt{\cal E}_{ij}(k)
\eeq
for each $2 \leq k \leq K_0$ and each pair of nodes $ i \neq j \in [n]$. Here, $\wt\bcE_{ij} = (\wt{\cal E}_{ij}(k))_{2 \leq k \leq K_0} \in \R^{K_0-1}$ is a random vector satisfying that w.h.p.,  
$$\wt{\cal E}_{ij}(k)=\oo\left(\frac{ q }{|d_k| \sqrt{\log n}} \right)$$
with $2\le k \le K_0$, and $\wt \bbf^{(i,j)}=(\wt f_2^{(i,j)}, \cdots, \wt f_{K_0}^{(i,j)})^T\in \R^{K_0-1}$ is defined as % removing the $W$ entries with indices in $\cal M$:
\beq\label{asymp-variance2-tilde}
\wt f_k^{(i,j)}:= \frac{1}{t_k}  \sum_{ l \in [n]\setminus \cal M}W_{il}y_{k}^{(i)}(l) - \frac{1}{t_k}  \sum_{l \in [n]\setminus \cal M}W_{jl}y_{k}^{(j)}(l)
\eeq
with  
$$\by_{k}^{(i)}:=\frac{\bv_{k}}{v_1(i)}-  \frac{t_k v_k(i) \bv_{1}}{t_1 [v_1(i)]^2} .$$
From parts (iii) and (iv) of Condition \ref{main_assm_DCMM} and $\sqrt{n}\|\bV\|_{\max}\lesssim \sqrt{K}$ by \eqref{Vmax0}, we deduce that  
\beq\label{ykinfty}\max_{i \in [n]}\|\by_{k}^{(i)}\|_\infty \lesssim \sqrt{K}. \eeq

Now similar to \eqref{reduce_max}, we can show that if $c_{1n} \ll (Kd_1\lambda_1(\bP))^{-1/2}(\log n)^{-1/2}$, it holds that under the null hypothesis $H_0$ in \eqref{eq: hypothesis},  
\beq\label{reduce_max-het}
\cal T= \max_{\{i,j\}\in {\cal P} } (\wt \bbf^{(i,j)})^\top \big( \wt\bSig^{(2)}_{i,j} \big)^{-1} \wt \bbf^{(i,j)}  + \oo(1)
\eeq
w.h.p. Furthermore, an application of \eqref{ykinfty} and the Berry--Esseen inequality as in \eqref{BerryEsseen} yields that for each $x\ge 0$,  
\begin{align} 
&\left|\P\left\{ (\wt \bbf^{(i,j)})^\top \big( \wt\bSig^{(2)}_{i,j} \big)^{-1} \wt \bbf^{(i,j)} \le x\right\} - \P\{\chi_{K_0-1}^2 \le x\}\right| \lesssim \frac{K^{3/2}}{q}. \label{BerryEsseen2}
\end{align}
Thus, in view of $m \rightarrow \infty$, combining (\ref{reduce_max-het}), (\ref{BerryEsseen2}), and \eqref{max_Gumbel} leads to the desired conclusion in part (i) of Theorem \ref{thm:group-het}. 

We now move on to establish part (ii) of Theorem \ref{thm:group-het}. Similar to Lemma \ref{lem-signal}, it holds that with probability $1-\oo(1)$,
\begin{align*}
& \max_{\{i,j\}\in \cal P}\left\|\left( \wt\bSig^{(2)}_{i,j} \right)^{-1/2} \left[\frac{\wt\bV_{K_0-1}(i)}{v_1(i)} - \frac{\wt\bV_{K_0-1}(j)}{v_1(j)}\right] \right\| \\
& \ge \frac13 \max_{\{i,j\}\subset \cal M}\left\|\left( \wt\bSig^{(2)}_{i,j} \right)^{-1} \left[\frac{\wt\bV_{K_0-1}(i)}{v_1(i)} - \frac{\wt\bV_{K_0-1}(j)}{v_1(j)}\right] \right\|.
\end{align*}
Therefore, it follows from \eqref{ineq-DV_ratio_alt} and the condition $c_{2n}\gg (d_{K}\lambda_{K}(\bbP))^{-1/2}\sqrt{\log n}$ that
%assumption \eqref{hypothesis1_DCMM_alt} that 
\beq\nonumber %\label{group_dominw}
\begin{split}
    & \max_{\{i,j\}\in \cal P}\left\|\left( \wt\bSig^{(2)}_{i,j} \right)^{-1/2}\left[\frac{\wt \bV_{K_0-1}(i)}{v_1(i)} - \frac{\wt \bV_{K_0-1}(j)}{v_1(j)}\right] \right\| \gtrsim c_{2n} \sqrt{d_{K}\lambda_K(\bP)} \gg \sqrt{\log n},
    % \\
    % & \gtrsim \sqrt{d_{K_0}} \max_{1 \leq i\ne j \leq n}\lambda_{\min}^{1/2}\left\{ \left(\bpi_i , \bpi_j\right)^\top  \left(\bpi_i , \bpi_j\right)\right\} \\
    % & \gg \sqrt{\log n},
\end{split}
\eeq
which completes the proof for part (ii) of Theorem \ref{thm:group-het}.

\subsection{Proof of Theorem \ref{thm:group-het-sample}} \label{appd_group_het_sample}

%{\color{red} Add a high-level summary of the main ideas/ingredients/challenges/innovations of the proof here.}

Using similar arguments as in the proof of Theorem \ref{thm_consist_Sigma} in Section \ref{Sec.newA.4}, we can establish a similar probability bound as in \eqref{consist_Sigma} that for each large constant $C>1$, 
\beq\label{consist_Sigma2}
\P\left( \max_{ i\neq j \in {\cal M}}  q^{-2}\left\|\wt \bbD_{\wh K_0} \left(\wh \bSig_{i,j}^{(2)}(\wh K_0)-\bSig_{i,j}^{(2)}( \wh K_0)\right)\wt\bbD_{\wh K_0}\right\| >  \mathcal E'(\wh K_0)  \right) \le n^{-C} ,
\eeq
where $\cal E'(\wh K_0)$ is a deterministic parameter satisfying \eqref{eq:small_E} or \eqref{eq:small_E2} (depending on whether we prove part (i) or part (ii)). 
%that  $$0<\cal E(\wh K_0)\ll (\log n)^{-1/2}.$$
We omit the full technical details here for simplicity. Then, in light of  \eqref{consist_Sigma2}, an application of the arguments for the proofs of Theorems \ref{thm:pg-sample}, \ref{thm:pair-null-het} and  \ref{thm:group-het} in Sections \ref{Sec.newA.thm5}, \ref{appd_pair_het} and \ref{appd_group_het} yields the desired conclusions of Theorem \ref{thm:group-het-sample}. 
%This concludes the proof of Theorem \ref{thm:group-het-sample}.

\subsection{Proof of Theorem \ref{lem_locallaw1}} \label{appd_local1}

Recall that we will always work with the rescaling given in \eqref{scaling_eq} throughout this proof. To facilitate the technical analysis, let us first list some useful resolvent identities in the lemma below, which can be proved directly using the Schur complement formula; see, e.g., 
%Lemma 4.5 in \cite{Semicircle} and 
Lemma 3.4 in \cite{EKYY_ER1}. Recall the resolvent minor and the notation $\sum_i^{(\mathbb T)}$ defined in Definition \ref{defminor}.

\begin{lemma}[Lemma 3.4 in \cite{EKYY_ER1}]\label{lemm_resolvent}
We have the following resolvent identities.
\begin{itemize}
\item[(i)]
For each $ i \in [n]$, it holds that %{\color{red} (To do: we need to explain the notation $(i)$ used in the summation here)}
\begin{equation}
\frac{1}{{G_{ii} }} =  - z - W_{ii} - \sum_{  k,l }^{(i)} W_{ik}W_{il} G^{\left( i \right)}_{kl} .\label{resolvent1}
\end{equation}

\item[(ii)]
For each $ i\ne j \in [n]$, it holds that %{\color{red} (To do: we need to explain the notation $(i)$ and $(ij)$ used in the summation here)}
\begin{equation}\label{resolvent2}
G_{ij}= - G_{ii} \sum_{ k }^{(i)} W_{ik}G^{(i)}_{kj}  = G_{ii} G_{jj}^{\left( i \right)} \Big(- W_{ij} +\sum_{ k,l }^{(ij)} W_{ik}W_{jl} G^{\left( {ij} \right)}_{kl}\Big). 
\end{equation}

 \item[(iii)]
 For each $k \in [n] \setminus \{i,j\}$, it holds that
\begin{equation}\label{resolvent3}
G_{ij}^{\left( k\right)}  = G_{ij}  - \frac{G_{ik} G_{kj}}{G_{kk}}. 
\end{equation}

 \item[(iv)]
All of the above identities in parts (i)--(iii) also hold for $G^{(\mathbb T)}$ instead of $G$ for every subset $\mathbb T$ of indices.
\end{itemize}

\end{lemma}

With these resolvent identities, the proof of Theorem \ref{lem_locallaw1} is similar to that for \cite[Theorem 3.1]{EKYY_ER1}. Roughly speaking, in \eqref{resolvent1} and \eqref{resolvent2}, the resolvent entries \smash{$G^{\left( {ij} \right)}_{kl}$} are independent of $W_{ik}$ and $W_{jl}$, so we can apply the concentration inequalities in Lemma \ref{largedeviation} to show that {$\sum_{k,l}^{(ij)} W_{ik}W_{jl} G^{\left( {ij} \right)}_{kl}$} is small, which yields  \eqref{entry_law2}, and that \smash{$\sum_{k,l}^{(i)} W_{ik}W_{il} G^{\left( i \right)}_{kl}$} concentrates around the partial expectation over the entries $W_{ik}$, i.e., \smash{$\sum_{k}^{(ij)} s_{ik} G^{\left( {ij} \right)}_{kk}$}. Intuitively, $G^{\left( {ij} \right)}_{kk}$ is approximately equal to $G_{kk}$, because out of all $n^2$ entries, $\bbW$ and $\bbW^{(ij)}$ are only different in at most $2n$ many of them. Hence, the equation \eqref{resolvent1} can be approximately written as
\begin{equation*} 
\frac{1}{G_{ii}} \approx - z  - \sum_{ k  \in [n]} s_{ik} G_{kk} . 
\end{equation*}
This shows that the vector formed by the diagonal $G$ entries satisfies the QVE \eqref{QVE} approximately, which, together with the stability of the QVE established in \cite{AjaErdKru2015}, yields that $G_{kk}$ is close to $M_k$. Our following proof gives rigorous control of all the errors in the above argument.

For each $1 \leq i, j \leq n$, we define some complex-valued random variables
\begin{equation*}
  Z_{i}(z):=\sum_{ k,l \in [n]} \left(W_{ik}W_{il} -\delta_{kl}s_{ik}\right)G^{\left( i \right)}_{kl}(z)
\end{equation*}
and 
\begin{equation*}
  Z_{ij}(z):=\sum_{k,l \in [n]} W_{ik}W_{jl} G^{\left( {ij} \right)}_{kl}(z),
\end{equation*}
where $z \in \mathbb{C}$, $s_{ik}$ denotes the variance of  $W_{ik}$, and $\delta_{kl}$ is the Kronecker delta notation with $\delta_{kl}=1$ if $k=l$ and $\delta_{kl}=0$ otherwise. With the aid of Lemma \ref{lem_opbound} and Lemma \ref{largedeviation}, we can deduce that for each $z \in S(C_0)$, 
\begin{align}
\left| Z_{i}(z)\right| & \le \Big|\sum_{k} \left(W_{ik}^2 -s_{ik}\right)G^{\left( i \right)}_{kk}(z)\Big|+\Big|\sum_{ k\ne l } W_{ik}W_{il}G^{\left( i \right)}_{kl}(z)\Big| \nonumber\\
& \lesssim  \frac{\xi^{1/2}}{q} \max_{k}\Big|G^{\left( i \right)}_{kk}(z)\Big|+ \xi^2 \bigg[\frac{1}{q} \max_{ k\ne l } \Big|G^{\left( i \right)}_{kl}(z)\Big| + \frac{1}{n}\Big( \sum_{ k\ne l } \Big|G^{\left( i \right)}_{kl}(z)\Big|^2\Big)\bigg] \nonumber\\
& \lesssim \frac{\xi^{1/2}}{q |z|} + \frac{ \xi^2 }{q} \max_{  k\ne l } \Big|G^{\left( i \right)}_{kl}(z)\Big| + \frac{\xi^2 }{\sqrt{n} |z|} \label{estimate_Zi}
\end{align}
with $(c,\xi)$-high probability for some constant $c>0$. Here, we have used \eqref{bound_G} and the fact of $|z|-2\sqrt{\mathfrak M}\sim |z|$ for $z\in S(C_0)$ to obtain that 
\beq\label{simpleG}
\max_{k}\big|G^{\left( i \right)}_{kk}(z)\big| \lesssim |z|^{-1}, \quad \sum_{ k\ne l } \big|G^{\left( i \right)}_{kl}(z)\big|^2 \le \sum_{k} \big[\bG^{\left( i \right)} (\bG^{\left( i \right)})^*\big]_{kk}(z)\lesssim \frac{n}{|z|^2}
\eeq
in the third step above, where $*$ stands for the conjugate transpose of a given complex-valued matrix. Similarly, we can show that for each $z \in S(C_0)$, 
\begin{align}
\left| Z_{ij}(z)\right| &  \lesssim \frac{\xi^2}{q^2|z|} + \frac{ \xi^2 }{q} \max_{ k\ne l } \big|G^{\left( ij \right)}_{kl}(z)\big| + \frac{\xi^2 }{\sqrt{n} |z|} \label{estimate_Zij}
\end{align}
with $(c,\xi)$-high probability.

We now introduce the diagonal error 
\[ \Lambda_d(z):=\max_{i\in [n]} |G_{ii}(z)-M_i(z)| \]
and the off-diagonal error 
\[ \Lambda_o(z):=\max_{ i\ne j \in [n]} |G_{ij}(z)|
\]
for each $z \in S(C_0)$. Then it remains to bound both terms $\Lambda_d(z)$ and $\Lambda_o(z)$ defined above. Using \eqref{estimate_Zi}, \eqref{estimate_Zij}, and the resolvent identities given in Lemma \ref{lemm_resolvent}, we will show the following bounds for $z\in S(C_0)$: there exist some constants $c'_1,C'_1>0$ such that 
\beq\label{lambda_estimate1}
\Lambda_o(z) \le C_1'\left( \frac{1}{q|z|^2} + \frac{\xi^2}{\sqrt{n}|z|^3}\right)
\eeq
and
\beq\label{lambda_estimate2}
\Lambda_d(z)\le C_1'\left( \frac{1}{q|z|^2} +\frac{\xi^{1/2}}{q|z|^3} + \frac{\xi^2}{\sqrt{n}|z|^3}\right)
\eeq
with $(c_1',\xi)$-high probability. %The proof of Lemma \ref{lem_lambda0} is provided in Section \ref{Sec.newB.2}. 
%With Lemma \ref{lem_lambda0}, we can easily conclude Theorem \ref{lem_locallaw1}. 
%\begin{proof}[Proof of Theorem \ref{lem_locallaw1}] 
The upper bounds \eqref{lambda_estimate1} and \eqref{lambda_estimate2}  provide the entrywise local laws \eqref{entry_law} and \eqref{entry_law2} for each fixed $z\in S(C_0)$. Then, we can resort to a standard $\epsilon$-net (e.g., with $\epsilon=n^{-2}$) method along with a simple union bound argument to strengthen the results to a uniform bound over $z\in S(C_0)$, which leads to the desired conclusions in \eqref{entry_law} and \eqref{entry_law2}. 
%We omit the details. 
This completes the proof of Theorem \ref{lem_locallaw1}.

% \begin{lemma}\label{lem_lambda0}
% Under the conditions of Theorem \ref{lem_locallaw1}, 
% %Define the $z$-dependent event $\Xi(z):=\{\Lambda_d(z) \le (\log N)^{-1}\}$. 
% % Let $c_0>0$ be a sufficiently small constant and fix $C_0, \epsilon >0$.
% there exist some constants $c'_1,C'_1>0$ such that %the following estimates hold with $(c'_1,\xi)$-high probability for any $z\in S(C_0)$:
% %uniformly for all $a\in \mathcal I$ and $z\in S(c_0,C_0,\epsilon)$:
% \beq\label{lambda_estimate1}
% \Lambda_o(z) \le C_1'\left( \frac{1}{q|z|^2} + \frac{\xi^2}{\sqrt{n}|z|^3}\right)
% \eeq
% and
% \beq\label{lambda_estimate2}
% \Lambda_d(z)\le C_1'\left( \frac{1}{q|z|^2} +\frac{\xi^{1/2}}{q|z|^3} + \frac{\xi^2}{\sqrt{n}|z|^3}\right)
% \eeq
% with $(c_1',\xi)$-high probability.
% \end{lemma}

%\subsection{Proof of Lemma \ref{lem_lambda0}} \label{Sec.newB.2}

%{\color{red} Add a high-level summary of the main ideas/ingredients/challenges/innovations of the proof here.}
It remains to show \eqref{lambda_estimate1} and \eqref{lambda_estimate2}. 
To simplify the notation, we will suppress all the dependence on $z \in \mathbb{C}$ whenever there is no confusion hereafter. 
From \eqref{resolvent2} and \eqref{resolvent3}, we can deduce that for all $k\ne l\in [n]\setminus \{i\}$,
\begin{align} 
|G_{kl}^{(i)}| &\le \left|G_{kl}\right| + \left| \frac{G_{ki}G_{il}}{G_{ii}}\right| \le \left|G_{kl}\right| + \Lambda_o \left| G_{ll}^{(i)}(-W_{il} + Z_{il}) \right|\nonumber \\
& \le \left|G_{kl}\right| + C\left( \frac{ \xi^2 }{q|z|} + \frac{\xi^2 }{\sqrt{n} |z|}\right) \Lambda_o  \le \left[1+\oo(1)\right]\Lambda_o \label{lambdao1}
\end{align}
with $(c,\xi)$-high probability, where  we have used \eqref{bound_G} and \eqref{estimate_Zi} in the third step above, and applied $q\gg \xi^2$ in the last step. Similarly, we can also show that for all $k\ne l\in [n]\setminus \{i,j\}$,
\begin{align} \label{lambdao2}
|G_{kl}^{(ij)}| &\le  \left[1+\oo(1)\right]\Lambda_o
\end{align}
with $(c,\xi)$-high probability. Then combining \eqref{bound_G}, \eqref{resolvent2}, \eqref{estimate_Zij}, and \eqref{lambdao2}, we can obtain that for all $ i\ne j \in [n]$,
$$ |G_{ij}| \lesssim |z|^{-2} \left(|W_{ij}| + |Z_{ij}|\right) \lesssim \frac{1}{q|z|^2} + \frac{ \xi^2 }{q|z|^2}\Lambda_o + \frac{\xi^2 }{\sqrt{n} |z|^3} $$
with $(c,\xi)$-high probability. Taking a union bound over $  i\ne j \in [n]$, it follows that 
\beq\nonumber
\Lambda_o\lesssim \frac{1}{q|z|^2} + \frac{ \xi^2 }{q|z|^2}\Lambda_o + \frac{\xi^2 }{\sqrt{n} |z|^3}
\eeq
and thus 
\beq\nonumber
\Lambda_o\lesssim \frac{1}{q|z|^2}  + \frac{\xi^2 }{\sqrt{n} |z|^3} 
\eeq
with $(c-\oo(1),\xi)$-high probability, which gives the desired conclusion in \eqref{lambda_estimate1}.

It remains to establish the bound on $\Lambda_d$. With the aid of \eqref{resolvent1} and \eqref{resolvent3}, we can write that for each $i\in [n]$,
\begin{align}\label{1Gii}
    \frac{1}{{G_{ii} }} &=  - z - W_{ii} - \sum_{k\in [n]} s_{ik} G^{\left( i \right)}_{kk} - Z_i \nonumber\\
    &=  - z  - \sum_{k\in [n]} s_{ik} G_{kk} -\epsilon_i,
\end{align}
where $\epsilon_i$ is a random error defined as
$$\epsilon_i:=W_{ii} -  \sum_{k\in [n]} s_{ik} \frac{G_{ki}G_{ik}}{G_{ii}} + Z_i.$$
Then it follows from \eqref{estimate_Zi}, \eqref{lambdao1}, and the off-diagonal estimate established in \eqref{lambda_estimate1} that 
\beq\label{epsiloni} |\epsilon_i |\lesssim \frac1q +\frac{\xi^{1/2}}{q|z|} + \frac{\xi^2}{\sqrt{n}|z|}
\eeq
with $(c-\oo(1),\xi)$-high probability, where we have bounded ${G_{ki}G_{ik}}/{G_{ii}}$ in a similar way as for ${G_{ki}G_{il}}/{G_{ii}}$ in \eqref{lambdao1} previously, 
\beq\label{GikGik}
\begin{split}
\left|\frac{G_{ki}G_{ik}}{G_{ii}}\right|&\lesssim \Lambda_o \left(\frac{1}{q|z|} + \frac{\xi^2}{q^2|z|^2} + \frac{ \xi^2 }{q|z|} \Lambda_o + \frac{\xi^2 }{\sqrt{n} |z|^2}\right) \\
&\lesssim \frac{1}{q^2 |z|^3} + \frac{ \xi^4 }{q^4 |z|^5} + \frac{\xi^4 }{n|z|^5}
\end{split}
\eeq
with $(c-\oo(1),\xi)$-high probability.

Observe that subtracting \eqref{1Gii} from \eqref{QVE}, we can deduce that for each $i\in [n]$,
\begin{align}\label{2Gii}
    \frac{G_{ii}-M_i}{G_{ii}M_i}=\sum_{k\in [n]} s_{ik} \left(G_{kk}-M_k\right) + \epsilon_i,
\end{align}
%In view of \eqref{2Gii} from the proof of Lemma \ref{lem_lambda0} in Section \ref{Sec.newB.2}, With \eqref{2Gii}, we can deduce that 
which yields that for each $i\in [n]$ and $z \in \mathbb{C}$, 
\beq\label{G-Mii}
\begin{split}
   G_{ii}-M_i  &=M_i^2 \sum_{k\in [n]} s_{ik} \left(G_{kk}-M_k\right) + M_i^2 \epsilon_i + M_i (G_{ii}-M_i) \epsilon_i  \\
   &\quad+\left(G_{ii}-M_i\right)M_i\sum_{k\in [n]} s_{ik} \left(G_{kk}-M_k\right).
\end{split}
\eeq
Observe that \eqref{G-Mii} above defines a system of linear equations  $\mathbf h:=(G_{ii}-M_{ii})_{i \in [n]}$ as 
\begin{align} \label{neq.eq.FL000}
   \left[(1-\Pii^2 \bS)\mathbf h\right]_i  &=\OO\left( |M_iG_{ii}||\epsilon_i| + |M_i| \Lambda_d^2\right),
\end{align}
where  $\bS:=(s_{ij})_{1 \leq i, j \leq n}$ denotes the error variance matrix that corresponds to the entries of matrix $\bW$. It is worth mentioning that outside the support of $\mu_c$, the stability of equation (\ref{neq.eq.FL000}) above is known in that 
\begin{align}\label{G-mii02}
   \Lambda_d=\max_{i\in [n]}|h_i| \lesssim  \max_{i\in [n]}\left(|M_iG_{ii}||\epsilon_i| + |M_i| \Lambda_d^2\right);
\end{align}
see, e.g., Corollary 3.4 in \cite{AjaErdKru2015}. Recall that $M_i(z)$ is the Stieltjes transform 
\beq\label{resol_Mi}
M_i(z)=\int_{\mathbb{R}} \frac{\mu_i(\dd x)}{x-z}
\eeq
of a finite measure $\mu_i$ on the real line $\mathbb{R}$ with support $\supp \, \mu_i \subseteq [-2\sqrt{\mathfrak M}, 2\sqrt{\mathfrak M}]$. From \eqref{resol_Mi}, we deduce that
%Furthermore, in light of \eqref{resol_Mi}, it holds that 
\beq\label{boundMz}
\begin{split}
|M_i(z)| & \le \int_{-2\sqrt{\mathfrak M}}^{2\sqrt{\mathfrak M}} \frac{1}{|x-z|} \mu_i(\dd x) \lesssim \frac{1}{|z|-2\sqrt{\mathfrak M}} \sim |z|^{-1}
\end{split}
\eeq
for each $z\in S(C_0)$. Therefore, plugging the bound (\ref{boundMz}) above and \eqref{bound_G} from Lemma \ref{lem_opbound} into \eqref{G-mii02}, it follows from \eqref{epsiloni} that  
\beq\label{ineq_lambdad}
\Lambda_d\lesssim \frac1{q|z|^2} +\frac{\xi^{1/2}}{q|z|^3} + \frac{\xi^2}{\sqrt{n}|z|^3} + \frac{\Lambda_d^2}{|z|}
\eeq
with $(c-\oo(1),\xi)$-high probability. If $\Lambda_d/|z|\ll 1$ with $(c-\oo(1),\xi)$-high probability, then the above estimate immediately leads to the desired conclusion in \eqref{lambda_estimate2}.
However,  a priori, we do not know that $\Lambda_d/|z|\ll 1$ holds. To deal with this issue, we can use the dichotomy argument and the continuity argument in Sections 3.5 and 3.6 of \cite{EKYY_ER1} to show that  $\Lambda_d/|z|\ll 1$ actually holds for all $z\in S(C_0)$. This concludes \eqref{lambda_estimate2} together with \eqref{ineq_lambdad}.

\subsection{Proof of Theorem \ref{lem_averlaw}} \label{appd_local2}

To facilitate the technical presentation, let us define 
$$ g_i(z):= \sum_{k\in [n]} s_{ik} \left[G_{kk} (z)-M_k (z)\right] $$
for each $i\in [n]$ and $z \in \mathbb{C}$. It follows from \eqref{G-Mii} that for each $j \in [n]$ and $z \in S(C_0)$,
\begin{align}\label{g-mii}
   g_j(z)  &= \sum_{i\in [n]} s_{ji} [M_i(z)]^2 g_i(z) + \sum_{i\in [n]}s_{ji} [M_i(z)]^2 \epsilon_i(z) \nonumber\\
   & \quad + \OO\left( \frac{1}{q^2 |z|^3} + \frac{ \xi }{q^2|z|^5} + \frac{\xi^4 }{n|z|^5}\right)
\end{align}
with $(c,\xi)$-high probability for some constant $c>0$, where we have used \eqref{entry_law}, \eqref{epsiloni}, and \eqref{boundMz} to bound the third and fourth terms on the RHS of \eqref{G-Mii}. Taking a union bound over $j$ implies that \eqref{g-mii} holds uniformaly for all $j\in [n]$ with $(c-\oo(1),\xi)$-high probability. Similar to \eqref{neq.eq.FL000}, \eqref{g-mii} above also defines a system of linear equations of $\mathbf g=(g_j)_{j \in [n]}$ as 
\begin{align} \label{neq.eq.FL001}
   \left[(1-\bS\Pii^2 )\mathbf g\right]_j  &=   \sum_{i\in [n]}s_{ji} M_i^2 \epsilon_i  + \OO\left( \frac{1}{q^2 |z|^3} + \frac{ \xi }{q^2|z|^5} + \frac{\xi^4 }{n|z|^5}\right).
\end{align}
%where  $S:=(s_{ij})_{1 \leq i, j \leq n}$ denotes the error variance matrix that corresponds to the entries of matrix $\bW$. It is worth mentioning that 
Outside the support of $\mu_c$, the stability of equation (\ref{neq.eq.FL001}) above is also proved in Corollary 3.4 of \cite{AjaErdKru2015} that 
\begin{align}\label{g-mii2}
   \max_{j\in [n]}|g_j| \lesssim  \max_{j\in [n]} \Big|\sum_{i\in [n]}s_{ji} M_i^2  \epsilon_i\Big|  + \OO\left( \frac{1}{q^2 |z|^3} + \frac{ \xi }{q^2|z|^5} + \frac{\xi^4 }{n|z|^5}\right)
\end{align}
with $(c-\oo(1),\xi)$-high probability. Thus, to conclude the proof, we need to show that $\sum_{i}s_{ji} M_i^2  \epsilon_i$ can be bounded as in \eqref{aver_law}. The term $\sum_{i}s_{ji} M_i^2 W_{ii}$ can be bounded using the Bernstein's inequality, and the term $\sum_{k} s_{ik} {G_{ki}G_{ik}}/{G_{ii}} $ in $\epsilon_i$ can be bounded using Lemma \ref{lem_opbound}. To control the term $\sum_{i}s_{ji} M_i^2 Z_i$, we need to exploit a \emph{flucatuation averaging argument} as in Section 5 of \cite{EKYY_ER1}, which we discuss in more detail now. 

Let us now bound the first term on the RHS of \eqref{g-mii2} above. It follows from \eqref{Ber_xi} and \eqref{boundMz} that for each $j\in [n]$,
\beq\label{sum_Wq}
\sum_{i\in [n]}s_{ji} M_i^2 W_{ii} \le \frac{\xi^{1/2}}{n|z|^2} + \frac{\xi}{nq|z|^2} \le 2 \frac{\xi^{1/2}}{n|z|^2}
\eeq
with $(a,\xi)$-high probability for some constant $a>0$. In light of the Stieltjes transform $M_i(z)$ as given in \eqref{resol_Mi}, we can obtain that for $z\in S(C_0)$,
$$ |\re M_i(z)|\gtrsim \frac{|\re z| - 2\sqrt{\mathfrak M}}{(|z|+2\sqrt{\mathfrak M})^2} \ \ \text{ and }\ \ \im M_i(z)\gtrsim \frac{\im z}{(|z|+2\sqrt{\mathfrak M})^2},$$
where $\re$ and $\im$ represent the real and imaginary parts of a given complex number, respectively. This implies that 
\[ |M_i(z)|\gtrsim |z|^{-1} \]
for $z\in S(C_0)$, which together with \eqref{entry_law} from Theorem \ref{lem_locallaw1} yields that \begin{equation}
    \label{diagonal_Gii}
    |G_{ii}(z)|\gtrsim |z|^{-1}
\end{equation} 
with $(c_1,\xi)$-high probability. Combining (\ref{diagonal_Gii}), \eqref{bound_G} from Lemma \ref{lem_opbound}, and \eqref{boundMz} gives that for each $i\in [n]$, 
\beq\label{sum_Gik}
\left|M_i^2\right| \sum_{k\in [n]} s_{ik} \left|\frac{G_{ki}G_{ik}}{G_{ii}}\right| \lesssim \frac{1}{n|z|} \left(\bG\bG^*\right)_{ii}\lesssim \frac{1}{n|z|^3}
\eeq
with $(c_0\wedge c_1,\xi)$-high probability. With the aid of \eqref{sum_Wq} and \eqref{sum_Gik}, we can deduce that
\beq\label{decompose_ei} 
\max_{j\in [n]} \Big|\sum_{i\in [n]}s_{ji} M_i^2  \epsilon_i\Big| \lesssim  \max_{j\in [n]}\Big|\sum_{i\in [n]}s_{ji} M_i^2 Z_i\Big|  + \OO\left(  \frac{\xi^{1/2}}{n|z|^2}\right) 
\eeq
with $(c'',\xi)$-high probability for some constant $c''>0$. 

It remains to bound the first term on the RHS of \eqref{decompose_ei} above. For each fixed $j\in [n]$, denote by $a_i := n s_{ji} M_i^2 \lesssim |z|^{-2}$. Then we can decompose $n^{-1}\sum_{i\in [n]} a_i Z_i$ as 
$$\frac1n\sum_{i\in [n]} a_iZ_i:=\frac{1}{n}\sum_{i} \sum_{k}^{(i)}a_i \left(W_{ik}^2 -s_{ik}\right)G^{\left( i \right)}_{kk} + \frac{1}{n}\sum_{i} a_i \cal Z_i,
$$
where $\cal Z_i:= \sum_{ k\ne l } W_{ik}W_{il}  G^{\left( i \right)}_{kl}$. From \eqref{resolvent3}, it holds that 
\begin{align*}
     & \frac{1}{n}\sum_{i} \sum_{k}^{(i)}a_i \left(W_{ik}^2 -s_{ik}\right)G^{\left( i \right)}_{kk} =\    \frac{1}{n}\sum_{ i\ne k }a_i \left(W_{ik}^2 -s_{ik}\right)  M_{k} \\
    &\quad+  \frac{1}{n} \sum_{ k } \sum_{i}^{(k)} a_i \left(W_{ik}^2 -s_{ik}\right) \left(G_{kk}-M_k\right)  \\
   & \quad - \frac{1}{n}\sum_{i} \sum_{k}^{(i)}a_i \left(W_{ik}^2 -s_{ik}\right)\frac{G_{ki}G_{ik}}{G_{ii}}.
\end{align*}
Using similar arguments as in \eqref{sum_Gik}, we can bound the last term in the expression above as
\beq\label{eq_fluc1}
\begin{split}
\Big| \frac{1}{n}\sum_{i} \sum_{k}^{(i)}a_i \left(W_{ik}^2 -s_{ik}\right)\frac{G_{ki}G_{ik}}{G_{ii}} \Big| & \lesssim \frac{1}{nq^2 |z|^2}\sum_{i} \sum_{k} \left|\frac{G_{ki}G_{ik}}{G_{ii}}\right|  \lesssim \frac{1}{q^2|z|^3}
\end{split}
\eeq
with $(c_0\wedge c_1,\xi)$-high probability. 

Observe that $W_{ik}^2 -s_{ik}$ with $i<k$ are independent centered random variables such that 
$$\max_{ i< k } |W_{ik}^2 -s_{ik}|\lesssim q^{-2} \ \text{ and } \ \E |W_{ik}^2 -s_{ik}|^2 \lesssim q^{-2} n^{-1}.$$
Then using \eqref{Ber_xi}, we can deduce that
\begin{align}\label{eq_fluc2}
    \bigg|\frac{1}{n}\sum_{ i< k  }a_i \left(W_{ik}^2 -s_{ik}\right) M_{k}\bigg| + \bigg|\frac{1}{n}\sum_{  k < i }a_i \left(W_{ik}^2 -s_{ik}\right) M_{k}\bigg|
    \le \frac{\xi^{1/2}}{q\sqrt{n}|z|^3}
\end{align}
and
\begin{align}\label{eq_fluc2.5}
    \Big|\sum_{i}^{(k)} a_i \left(W_{ik}^2 -s_{ik}\right)\Big|
    \le \frac{\xi^{1/2}}{q|z|^2}
\end{align}
with $(a,\xi)$-high probability for some constant $a>0$. With the aid of \eqref{eq_fluc2.5} and \eqref{entry_law}, we can obtain that
\begin{align}\label{eq_fluc3}
    & \bigg|\frac{1}{n} \sum_{k} \sum_{i}^{(k)} a_i \left(W_{ik}^2 -s_{ik}\right) \left(G_{kk}-M_k\right) \bigg|  \lesssim \frac{ \xi^{1/2}}{q|z|^4}\left(\frac{1}{q} +\frac{\xi^{1/2}}{q|z|} + \frac{\xi^2}{\sqrt{n}|z|}\right)
\end{align}
with $( a\wedge c_1-\oo(1),\xi)$-high probability. Thus, combining \eqref{eq_fluc1}, \eqref{eq_fluc2}, and \eqref{eq_fluc3} yields that
\begin{align}\label{aver_Zi1}
    & \bigg| \frac{1}{n}\sum_{i } \sum_{k}^{(i)}a_i \left(W_{ik}^2 -s_{ik}\right)G^{\left( i \right)}_{kk} \bigg|  \lesssim  \frac{1}{q^2|z|^3}\left( 1 + \frac{\xi}{|z|^2}+ \frac{q\xi^{1/2}}{\sqrt{n}} + \frac{q\xi^{5/2}}{\sqrt{n}|z|^2}\right)
\end{align}
with $( c,\xi)$-high probability for some constant $c>0$.

To analyze the term $ {n}^{-1}\sum_{i\in [n]} a_i \cal Z_i$, we will exploit a similar argument as in Section 5 of \cite{EKYY_ER1}. Given any subset of indices $\mathbb V$, $\mathbb U$, and $\mathbb S$ with $\mathbb U\subset \mathbb S$, let us define random variables
$$\cal Z_i^{[\mathbb V]}:= \mathbf 1(i\notin \mathbb V)\sum_{  k\ne l } W_{ik}W_{il}  G^{\left( i \mathbb V\right)}_{kl}$$
and
$$\cal Z_i^{\mathbb S,\mathbb U}:= (-1)^{|\mathbb S\setminus \mathbb U|}\sum_{\mathbb V:\, \mathbb S\setminus \mathbb U\subset \mathbb V \subset \mathbb S} (-1)^{|\mathbb V|}\cal Z_i^{[\mathbb V]}, $$ %\sum_{k\ne l} W_{ik}W_{il}  G^{\left( i \mathbb V\right)}_{kl},$$
where $\mathbf 1(\cdot)$ denotes the indicator function, $|\cdot|$ stands for the cardinality of a given set, and $G^{\left( i \mathbb V\right)}:=G^{\left( \{i\} \cup \mathbb V\right)}$ is the resolvent minor defined in Definition \ref{defminor}. 
%{\color{red} (To do: it would be helpful to explain the notation $G^{\left( i \mathbb V\right)}_{kl}$ here)}. 
As a convention, we denote by $\cal Z_i^{[\emptyset]}:=\cal Z_i$.
Using the notation introduced above, for each given subset of indices $\mathbb S$ we can decompose $\cal Z_i$ as
\beq\label{resolution_of_dep}
\cal Z_i = \sum_{\mathbb U\subset \mathbb S}\cal Z_i^{\mathbb S,\mathbb U}.
\eeq
The decomposition in (\ref{resolution_of_dep}) follows directly from an inclusion-exclusion argument. Then we have the \emph{abstract decoupling} lemma below which was proved as Theorem 5.6 of \cite{EKYY_ER1}.

\begin{lemma}[Abstract decoupling] %\cite{EKYY_ER1}]
\label{lem_abs_decouple}
   Let $\Xi$ be an event and $p\in 2\N$ an even integer. Assume that conditions (i)--(iv) below are satisfied for the family of random variables $(\cal Z_i^{[\mathbb U]})_{i,\mathbb U}$ with some constants $\wt c, \wt C>0$. 
   \begin{enumerate}
       \item For each $i\notin \mathbb U$, it holds that $\cal Z_i^{[\mathbb U]}$ is independent of the entries in the rows of $\bbW$ with row indices belonging to $\mathbb U$, and 
       $$\E_i \cal Z_i^{[\mathbb U]} =0,$$
       where $\mathbb E_{i}$ is the %partial 
       expectation taken with respect to the randomness of the $i$th row of $\bbW$.
       
       \item Given any subsets of indices $\mathbb U$ and $\mathbb S$ with $\mathbb U\subset \mathbb S$, $|\mathbb S|\le p$, and $i\notin \mathbb S$, it holds that 
       \beq\label{rth_moment}
       \E\left[ \mathbf 1(\Xi) |\cal Z_i^{\mathbb S,\mathbb U}|^r\right] \le [Y(\wt CXu)^u]^r
       \eeq
       with $u:=|\mathbb U|+1$ for each $r\le p$, where $X$ and $Y$ are deterministic parameters with $X$ satisfying that 
       \beq\label{constraint_X}
       X\ll p^{-3}.
       \eeq
       
       \item Given any subset of indices $\mathbb V$, it holds that almost surely,
        \beq\label{rough_Z}
       |\cal Z_i^{[\mathbb V]}|\le n^{\wt C}. %\quad |\cal Z_i^{\mathbb S,\mathbb U}|\le n^{\wt C p}.
       \eeq
       
       \item There exists some deterministic parameter $\zeta_n\gg \log n$ such that
        \beq\label{highp_Xi}
       \P(\Xi)\ge 1 - e^{-\zeta_n p}.
       \eeq
       
   \end{enumerate}
Then there exists some constant $C>0$ such that
\beq\label{eq_abs_decouple}
       \P\bigg\{\mathbf 1(\Xi) |z|^2 \cdot \bigg|\frac{1}{n}\sum_{i\in [n]} a_i \cal Z_i\bigg| \ge Cp^6 Y(X^2+n^{-1})\bigg\} \le Ce^{-2p}.
\eeq
    %for large enough $n$.
\end{lemma}

A sketch of the proof for Lemma \ref{lem_abs_decouple} is provided in Section \ref{Sec.newB.3}. Now we aim to show that the family of random variables \smash{$(\cal Z_i^{[\mathbb U]})_{i,\mathbb U}$} indeed satisfy conditions (i)--(iv) of Lemma \ref{lem_abs_decouple} above for some suitably chosen $\Xi$, $X$, $Y$, and $p$. First, note that condition (i) above follows directly from the definition of $\cal Z_i^{[\mathbb U]}$.
Second, let us choose a subset of $S(C_0)$ as 
\beq\label{wtSC0}
\wt S(C_0):=\left\{z\in S(C_0):  \im z\ge n^{-4}\right\}.
\eeq
Then we immediately obtain that for each $z\in \wt S(C_0)$,
\beq\label{bdd_wtSC0}
\|\bG^{\left( i\mathbb V\right)}(z)\|=\big\| (\bW^{\left( i\mathbb V\right)}-z)^{-1}\big\|\le (\im z)^{-1} \le n^{4},
\eeq
which by definition entails that $|\cal Z_i^{[\mathbb U]}|\le n^{6}$. This verifies condition (iii) above. Next we define event $\Xi$ as 
\begin{align*}
   \Xi:= & \bigcap_{z\in \wt S(C_0)} \left\{|z|^2 \cdot \max_{i\in [n]}\left|G_{ii}(z)-M_{i}(z)\right| \le C_1\left(\frac{1}{q} +\frac{\xi^{1/2}}{q|z|} + \frac{\xi^2}{\sqrt{n}|z|}\right)\right\}\\
   & \quad \cap \bigcap_{z\in \wt S(C_0)} \left\{|z|^2 \cdot \max_{  i\ne j \in [n]}\left|G_{ij}(z)\right| \le C_1\left(\frac{1}{q} + \frac{\xi^2}{\sqrt{n}|z|}\right)\right\}.
\end{align*}
From Theorem \ref{lem_locallaw1}, we see that event $\Xi$ holds with probability
$$\P(\Xi)\ge 1- 2e^{-c_1\xi},$$
which establishes condition (iv) above with parameter $\zeta_n = c_1\xi/(2p)$ as long as $p$ is chosen such that $p \ll \xi/\log n$.

It remains to verify condition (ii) of Lemma \ref{lem_abs_decouple} above. To this end, let us choose
\beq\label{chooseXY}
X:=\frac{1}{q} + \frac{\xi^2}{\sqrt{n}|z|} \ \text{ and } \ Y:=\frac{p_0^2}{|z|},
\eeq
where recall that $p_0$ was defined in \eqref{p_constraint}. Then we see that \eqref{constraint_X} of condition (ii) holds under assumption \eqref{p_constraint}. Using \eqref{resolvent3}, we can derive the identity
\beq\label{resolvent4}
\frac{1}{{G_{ii} }} = \frac{1}{{G_{ii}^{(k)} }} - \frac{{G_{ik} G_{ki} }}{{G_{ii} G_{ii}^{(k)} G_{kk} }},
\eeq
which also holds for $G^{(\mathbb T)}$ instead of $G$ for each subset $\mathbb T$ of indices. We will exploit the arguments in the proof from Section 5.2 of \cite{EKYY_ER1} with identities \eqref{resolvent3} and \eqref{resolvent4} as the main tools. Specifically, we can obtain a similar result as in Lemma 5.11 of \cite{EKYY_ER1} that given any subsets of indices $\mathbb U\subset \mathbb S$ with $|\mathbb S|\le p$, there exists some constant $C>0$ such that 
\beq\label{bound_GU}
\mathbf 1(\Xi)|G^{\mathbb S,\emptyset}_{ij}(z) -M_i(z) \delta_{ij}| \le \frac{C}{|z|^2}\left(\frac{1}{q} +\frac{\xi^{1/2}}{q|z|} \delta_{ij}+ \frac{\xi^2}{\sqrt{n}|z|}\right),
\eeq
and that if $\mathbb U\ne \emptyset$ and $i,j\notin \mathbb S$, then %we have 
\beq\label{bound_GU2}
\mathbf 1(\Xi)|G^{\mathbb S,\mathbb U}_{ij}(z)| \le |z|^{-1}\left(\frac{C}{|z|}|\mathbb U| X\right)^{|\mathbb U|+1}.
\eeq
Using the bounds in (\ref{bound_GU}) and (\ref{bound_GU2}) above, we can apply the arguments used in the proof of Lemma 5.13 in \cite{EKYY_ER1} to show that \eqref{rth_moment} of condition (ii) holds for $X$ and $Y$ as given in \eqref{chooseXY}.

We are now ready to apply Lemma \ref{lem_abs_decouple} and show that there exists some constant $C>0$ such that for each $1\ll p\ll p_0$,
\beq\label{aver_Z_bound}
 \bigg|\frac{1}{n}\sum_{i\in [n]} a_i \cal Z_i\bigg|\le Cp_0^8\left(\frac{1}{q^2|z|^3}+\frac{\xi^4}{n|z|^5}\right)
\eeq
with $(1,p)$-high probability. Therefore, combining \eqref{g-mii2}, \eqref{decompose_ei}, \eqref{aver_Zi1}, and \eqref{aver_Z_bound} yields that
\begin{align*}
   \max_{i\in [n]}|g_i| \le C \left(\frac{\xi^{1/2}}{n|z|^2}+\frac{p_0^8}{q^2|z|^3}+\frac{p_0^8\xi^4}{n|z|^5}\right)
\end{align*}
with $(c,p)$-high probability for some constants $c, C>0$, which leads to the desired conclusion in \eqref{aver_law} for each fixed \smash{$z\in \wt S(C_0)$}. Furthermore, invoking a standard $\epsilon$-net method along with a simple union bound argument can improve the result to a uniform bound in \smash{$z\in \wt S(C_0)$}. Finally, with an application of a simple perturbation argument, we can extend the result to a uniform bound over all $z\in S(C_0)$, which concludes the proof of Theorem \ref{lem_averlaw}.

\subsection{Proof of Theorem \ref{lem_locallaw2}} 
% anisotropic local laws} 
\label{sec_moments}

The main idea for the proof of Theorem \ref{lem_locallaw2} is to bound the high moments of 
$$Y(z):=q|z|^2 \bu^\top (\bG(z)-\Pii(z))\bv $$ 
for any deterministic unit vectors $\bu, \bv \in \mathbb{R}^n$. In particular, we make the following claim.

\begin{claim}\label{claim_moment_aniso}
Under the conditions of Theorem \ref{lem_locallaw2}, it holds that for each $z\in \wt S(C_0)$ given in \eqref{wtSC0} and $r:=\lfloor\log n\rfloor$,
\beq\label{moment_aniso}
\E \left|Y(z)\right|^{2r}\le (\wt Cr)^{2r}
\eeq
for some large enough constant $\wt C>0$ that does not depend on $z$. 
\end{claim}

The proof of Claim \ref{claim_moment_aniso} above is provided in Section \ref{new.secB.claim1} with full details. For each constant $D>0$, it follows from \eqref{moment_aniso} and the Markov inequality that 
$$\P\left\{\left|\bu^\top [\bG(z)-\Pii(z)]\bv\right| \ge \frac{C\wt C \log n}{q|z|^2} \right\} \le C^{-2\lfloor\log n\rfloor} \le n^{-D}$$
as long as the positive constant $C$ is taken to be large enough. Then by resorting to a standard $\epsilon$-net method together with a simple union bound argument, we can strengthen the result to a uniform bound in $z\in \wt S(C_0)$. Finally, an application of a simple perturbation argument yields the desired conclusion in \eqref{aniso_law} that provides a uniform bound over all $z\in S(C_0)$. This completes the proof of Theorem \ref{lem_locallaw2}.

\subsection{Proof of Theorem \ref{asymp_evalue}} \label{Sec.newA.55}

There are several key steps in the proofs of Theorem \ref{asymp_evalue} for the asymptotic expansion of the empirical spiked eigenvalues \smash{$\wh d_k$}. To establish the desired conclusion in \eqref{eq_evalue_conv}, we will first rewrite the determinantal eigenvalue equation into a master equation of \smash{$\wh d_k$} in terms of certain linear forms of $\bG$; see \eqref{masterx} below for details. Then these linear forms can be estimated using the anisotropic local law established in Theorem \ref{lem_locallaw2}, which will reveal that the master equation is almost identical to the deterministic equation \eqref{eq_evalue} with solution $t_k \in \cal I_k$. Through analyzing the small difference between \eqref{masterx} and \eqref{eq_evalue}, we will see that \smash{$\wh d_k$} is indeed close to $t_k$ up to a small-order error asymptotically.

It follows from Weyl's inequality \citep{Weyl} and Lemma \ref{lem_opbound} that w.h.p., $\wh d_k\in \cal I_k$ for each $1\le k \le K_{\max}$. We will make a useful claim that w.h.p., $\wh d_k$ satisfies equation
\beq\label{masterx}
1 + d_k \bv_k^\top \bG(\wh d_k) \bv_k - d_k \bv_k^\top \bG(\wh d_k) \bV_{-k} \frac{1}{\bbD_{-k}^{-1} + \bV_{-k}^\top \bG(\wh d_k)\bV_{-k}} \bV_{-k}^\top \bG(\wh d_k) \bv_{k} = 0 . 
\eeq
In fact, it holds by definition that
$$\det(\bbX-\wh d_k \bI)=0 \ \Leftrightarrow \ \det(\bbW-\wh d_k \bI + \bbV \bbD \bbV^\top)=0,$$
where $\det(\cdot)$ denotes the determinant of a given matrix. If $\wh d_k$ is not an eigenvalue of matrix $\bbW$ (which holds w.h.p. by Lemma \ref{lem_opbound}), then we can rewrite this equation as
\begin{equation*}
\begin{split}
& \det\left(\bbD^{-1}+ \bbV^{\top} \bG(\wh d_k) \bbV\right)=0 \ \\
  \Leftrightarrow \ &\det\begin{pmatrix} d_k^{-1} + \bv_k^\top \bG(\wh d_k) \bv_k & \bv_k^\top \bG(\wh d_k) \bV_{-k} \\ \bV_{-k}^\top \bG(\wh d_k) \bv_{k} & \bbD_{-k}^{-1} + \bV_{-k}^\top \bG(\wh d_k)\bV_{-k}\end{pmatrix}=0,
\end{split}
\end{equation*} 
where we have used the identity $\det(\bI+\bA\bB)=\det(\bI+\bB\bA)$ for any two matrices $\bA$ and $\bB$ of conformable dimensions. Using the Schur complement formula for determinants of block matrices, we can show that the above equation is in fact equivalent to equation \eqref{masterx} or
\begin{align*}
   &\det \left(d_k\bbD_{-k}^{-1} + d_k\bV_{-k}^\top \bG(\wh d_k)\bV_{-k}\right)=0.
\end{align*} 
\iffalse
By Theorem \ref{lem_locallaw2}, we have that for $\wh d_k\in \cal I_k$,
\beq\label{eq_use_aniso10}
\left\| d_k\bV_{-k}^\top \left[\bG(\wh d_k)-\Pii(\wh d_k)\right]\bV_{-k}\right\| \lesssim \frac{K\log n}{q|d_k|} \quad \text{w.h.p.}
\eeq
\fi
But, it follows from \eqref{denom_lower} and the results in \eqref{eq_use_aniso1} below that $d_k\bbD_{-k}^{-1} + d_k\bV_{-k}^\top \bG(\wh d_k)\bV_{-k}$ is nonsingular w.h.p. as long as $K\log n \ll q|d_k|$. Hence, we see that equation \eqref{masterx} indeed holds w.h.p. 

We are now ready to work with the representation given in equation \eqref{masterx}. From \eqref{asymp_Pi} and Theorem \ref{lem_locallaw2}, it holds with high probability that for all $z\in \cal  I_k$,
\begin{align}
& \bv_{k}^\top \Pii(z)\bv_{k}=- z^{-1} +\OO\left(|d_k|^{-3}\right), \ \ \left|  \bv_{k}^\top \left[\bG(z)-\Pii(z)\right]\bv_{k}\right| \lesssim \frac{\log n}{q|d_k|^2},\label{eq_use_aniso0.5}\\
& \bV_{-k}^\top \Pii(z)\bV_{-k}=-z^{-1}\bI + \OO\left(|d_k|^{-3}\right), \ \  \left\| \bV_{-k}^\top \left[\bG(z)-\Pii(z)\right]\bV_{-k}\right\| \lesssim \frac{K\log n}{q|d_k|^2} , \label{eq_use_aniso1}\\
& \bv_{k}^\top \Pii(z)\bV_{-k}=\OO\left(|d_k|^{-3}\right), \ \ \left\| \bv_{k}^\top \left[\bG(z)-\Pii(z)\right]\bV_{-k}\right\| \lesssim \frac{\sqrt{K}\log n}{q|d_k|^2}, \label{eq_use_aniso1.5}
\end{align}
where $\bbA=\OO(a_n)$ means that $\|\bbA\|=\OO(a_n)$ for a given vector or matrix $\bbA$. Then in view of \eqref{masterx}--\eqref{eq_use_aniso1.5} and  \eqref{denom_lower}, we can deduce that w.h.p., 
\begin{align*}
& 1 + d_k \bv_k^\top \Pii(\wh d_k) \bv_k - d_k \bv_k^\top \Pii(\wh d_k) \bV_{-k} \frac{1}{\bbD_{-k}^{-1} + \bV_{-k}^\top \Pii(\wh d_k)\bV_{-k} } \bV_{-k}^\top \Pii(\wh d_k) \bv_{k} \\
& = \OO\left\{\frac{\log n}{q|d_k|} +\frac{K\log n}{q|d_k|} \left(\frac{\sqrt{K}\log n}{q|d_k|}+ \frac{1}{d_k^2}\right)^2 + \frac{\sqrt{K}\log n}{q|d_k|}\left(\frac{\sqrt{K}\log n}{q|d_k|}+ \frac{1}{d_k^2}\right) \right\}\\
&= \OO\left(\frac{\log n}{q|d_k|} +\frac{K\log n}{q|d_k|^5} \right).
\end{align*}
This together with the definition of $t_k \in \cal I_k$ as the solution to equation \eqref{eq_evalue} entails that
\begin{align}
  d_k \bv_k^\top \Pii(\wh d_k) \bv_k  - & d_k \bv_k^\top \Pii(t_k) \bv_k = d_k \bv_k^\top \Pii(\wh d_k) \bV_{-k} \frac{1}{\bbD_{-k}^{-1} + \bV_{-k}^\top \Pii(\wh d_k)\bV_{-k} } \bV_{-k}^\top \Pii(\wh d_k) \bv_{k} \nonumber\\
 & - d_k \bv_k^\top \Pii(t_k) \bV_{-k} \frac{1}{\bbD_{-k}^{-1} + \bV_{-k}^\top \Pii(t_k)\bV_{-k} } \bV_{-k}^\top \Pii(t_k) \bv_{k} \nonumber\\
 &+ \OO\left(\frac{\log n}{q|d_k|} +\frac{K\log n}{q|d_k|^5} \right) \label{subtract_eq}
\end{align}
 w.h.p. Using \eqref{asymp_Pi}, we can bound the RHS of (\ref{subtract_eq}) as 
\begin{align}\label{subtract_eq1}
\OO\left(\frac{|\wh d_k-t_k|}{|d_k|^5}+\frac{\log n}{q|d_k|} +\frac{K\log n}{q|d_k|^5}\right)
\end{align}
for all $\wh d_k \in \cal I_k$. Moreover, it follows from \eqref{resol_Mi} that function $\bv_k^\top \Pii(\lambda) \bv_k$ is strictly increasing in $\lambda$ on $[-2\sqrt{\mathfrak M}, 2\sqrt{\mathfrak M}]^c$, where $c$ represents the set complement. This fact along with \eqref{asymp_Pi} yields that
\begin{align}\label{subtract_eq2}
 \left|d_k \bv_k^\top \Pii(\wh d_k) \bv_k  -  d_k \bv_k^\top \Pii(t_k) \bv_k\right| \gtrsim \frac{|\wh d_k -t_k|}{|d_k|}. 
\end{align}
Therefore, applying \eqref{subtract_eq1} and \eqref{subtract_eq2} to \eqref{subtract_eq} yields that w.h.p.,
$$ |\wh d_k-t_k|=\OO \left(\frac{\log n}{q} +\frac{K\log n}{q|d_k|^4}\right),
$$
 which gives the desired conclusion in \eqref{eq_evalue_conv}. This concludes the proof of Theorem \ref{asymp_evalue}.

\subsection{Proof of Theorem \ref{asymp_evector}} \label{Sec.newA.thm10}

We now aim to establish the asymptotic expansion for the empirical spiked eigenvectors $\hat \bv_k$. To this end, let us define a small contour $\cal C_k$ around the population quantity $t_k$ as
\beq\label{defn_Ck}\cal C_k:=\{z\in \C: |z-t_k| = c t_k\},\eeq
where $c>0$ is some small constant depending on $\e_0$ such that both $(1-c)t_k$ and $(1+c)t_k$ belong to $\cal I_k$. From Theorem \ref{asymp_evalue}, we see that w.h.p., contour $\cal C_k$ encloses only the $k$th spiked eigenvalue $\wh d_k$ and no other eigenvalues of matrix $\bbX$. Then in view of the Cauchy integral formula given in \eqref{eq:Cauchy1}, it holds w.h.p. that
\beq\label{eq_evector0} \bx^\top \hat \bv_k  \hat \bv_k^\top \by = -\frac{1}{2\pi \ii}\oint_{\cal C_k} \bx^\top \frac{1}{\bbX-z} \by \dd z
\eeq
for any deterministic unit vectors $\bx,\by\in \R^n$, where $\ii = \sqrt{-1}$ denotes the imaginary unit in the complex plane $\mathbb{C}$. To calculate the RHS of (\ref{eq_evector0}), we will need to introduce a new class of resolvents 
\begin{align}
\bR_k(z):=&\left( \bbH + \bV_{-k}\bbD_{-k}\bV_{-k}^\top - z \bI \right)^{-1}\nonumber\\
=&\, \bG(z)- \bG(z)  \bV_{-k} \frac{1}{\bbD_{-k}^{-1} + \bV_{-k}^\top \bG(z)\bV_{-k} } \bV_{-k}^\top \bG(z)\label{def_Rk}
\end{align}
with $z \in \mathbb{C}$, where we have used the Woodbury formula in the second step above. 

With $\bR_k(z)$ and the Woodbury formula, we can rewrite matrix $(\bbX-z )^{-1}$ as
\begin{align*}
(\bbX-z )^{-1} & =\left[ \bR_k^{-1}(z) + d_k \bv_k\bv_k^\top \right]^{-1}\\
&=\bR_k(z)- \bR_k(z)  \bv_{k} \frac{1}{d_k^{-1} + \bv_{k}^\top \bR_k(z)\bv_{k} } \bv_{k}^\top \bR_k(z).
\end{align*}
%where $z \in \mathbb{C}$. 
An application of Weyl's inequality and Lemma \ref{lem_opbound} shows that with high probability, contour $\cal C_k$ does not enclose any eigenvalue of matrix $\bbH + \bV_{-k}\bbD_{-k}\bV_{-k}^\top$. This entails that w.h.p.,
$$ \oint_{\cal C_k} \bx^\top \bR_k(z) \by \dd z=0, $$
and thus the representation in \eqref{eq_evector0} becomes 
\beq\label{eq_evector} 
\bx^\top \hat \bv_k  \hat \bv_k^\top \by = \frac{1}{2\pi \ii}\oint_{\cal C_k}\frac{ \bx^\top \bR_k(z) \bv_k  \bv_k^\top \bR_k(z) \by  }{d_k^{-1} + \bv_k^\top \bR_k(z)\bv_k}\dd z
\eeq
w.h.p. Note that evaluating the RHS of \eqref{eq_evector} yields an asymptotic expansion of the bilinear form  $\bx^\top \hat \bv_k  \hat \bv_k^\top \by$ for any deterministic vectors $\bx,\by \in \R^n$. In particular, with the choice of $\bx=\by=\bv_k$, we will obtain an asymptotic expansion of the quadratic form $|\bv_k^\top \hat \bv_k|^2$, whose square root gives an asymptotic expansion of $\bv_k^\top \hat \bv_k$ around one; see \eqref{evector_main0.5} below for details. On the other hand, taking $\bx=\mathbf e_i$ and $\by=\bv_k$ yields an asymptotic expansion of $(\bv_k^\top \hat \bv_k) \hat v_k(i)$, and further dividing it by $\bv_k^\top \hat \bv_k$ will result in the desired asymptotic expansion in \eqref{expansion_evector2}.

As mentioned above, the major goal is to evaluate the RHS of \eqref{eq_evector}. To do so, we will need to estimate the term $\bx^\top \bR_k(z) \bv_k$ with the choices of $\bx=\bv_k$ or $\bx=\be_i$, which requires an anisotropic local law for $\bR_k$. This can be easily achieved by combining the identity \eqref{def_Rk} with the anisotropic local laws for $\bG(z)$ established in Theorem \ref{lem_locallaw2} and Proposition \ref{lem_locallawv}--\ref{lem_locallawv2}; see \eqref{aniso_lawk}, \eqref{eq_use_aniso2.0}, and \eqref{eq_use_aniso2} below for details. With the aid of the anisotropic local laws, we can replace $\bR_k$ with its deterministic asymptotic limit, i.e., $\Pii_k$, up to some random errors. In particular, the resulting deterministic part can be estimated easily using the Cauchy residue theorem with a simple pole at $z=t_k$. For the random errors, most of them are asymptotically negligible and included into the third term on the RHS of \eqref{expansion_evector2}, while the leading term is given by 
\begin{align*}
- t_k\mathbf e_i^\top \bG(t_k) \bv_k & = \mathbf e_i^\top \frac{t_k}{t_k-\bW} \bv_k  = v_k(i) +\frac{1}{t_k} \mathbf e_i^\top \bW \bv_k+  \mathbf e_i^\top \left(\frac{\bW}{t_k-\bW }-\frac{\bW}{t_k}\right) \bv_k;
\end{align*}
see \eqref{evector_main2} and \eqref{eq_expvi} below for details. Here, the first two terms on the RHS of the expression above lead to the first two terms on the RHS of \eqref{expansion_evector2}, while the last term can be controlled using Proposition \ref{lem_locallawv2}.

With the above ideas and the anisotropic local laws, we proceed with the proof of Theorem \ref{asymp_evector} that will involve delicate calculations, where we need to carefully track the error for each step. As mentioned above, we will need to calculate \eqref{eq_evector} for two cases: (i) $\bx=\by=\bv_k$ and (ii) $\bx=\mathbf e_i$ and $\by=\bv_k$. %Our main tools are the anisotropic local laws, Theorem \ref{lem_locallaw2} and Propositions \ref{lem_locallawv} and \ref{lem_locallawv2}. 
It follows from \eqref{asymp_Pi} and Theorem \ref{lem_locallaw2} that all the estimates given in  \eqref{eq_use_aniso0.5}--\eqref{eq_use_aniso1.5} hold for $z\in \cal C_k$ with high probability.
\iffalse
for $z\in \cal  I_k\cup \cal C_k$, with high probability,
\begin{align}
\bv_{k}^\top \Pii(z)\bv_{k}=- z^{-1} +\OO\left(|d_k|^{-3}\right),\quad &\left|  \bv_{k}^\top \left[\bG(z)-\Pii(z)\right]\bv_{k}\right| \lesssim \frac{\log n}{q|d_k|^2},\label{eq_use_aniso0.5}\\
\bV_{-k}^\top \Pii(z)\bV_{-k}=-z^{-1}I_{K-1} + \OO\left(|d_k|^{-3}\right),\quad &\left\| \bV_{-k}^\top \left[\bG(z)-\Pii(z)\right]\bV_{-k}\right\| \lesssim \frac{K\log n}{q|d_k|^2} , \label{eq_use_aniso1}\\
\bv_{k}^\top \Pii(z)\bV_{-k}=\OO\left(|d_k|^{-3}\right),\quad &\left\| \bv_{k}^\top \left[\bG(z)-\Pii(z)\right]\bV_{-k}\right\|_2 \lesssim \frac{\sqrt{K}\log n}{q|d_k|^2}  .\label{eq_use_aniso1.5}
\end{align}
\fi
We can also obtain a similar estimate as in \eqref{denom_lower} that there exists some constant $C>0$ such that %which, together with \eqref{eq_use_aniso1}, gives that 
\beq\label{denom_lower2}
\max_{z\in \cal C_k}\left\| \left[d_k\bbD_{-k}^{-1} + d_k\bV_{-k}^\top \Pii(z)\bV_{-k}\right]^{-1}\right\|\le C
\eeq
w.h.p. for $z\in \cal C_k$. Hence, combining \eqref{eq_use_aniso0.5}--\eqref{eq_use_aniso1.5}, \eqref{def_Rk}, \eqref{denom_lower2}, and Theorem \ref{lem_locallaw2} yields that %for any deterministic unit vectors $\bu,\bv\in \R^n$, 
\beq\label{aniso_lawk}
\sup_{z\in \cal C_k}  \left|\bv_k^\top \left[\bR_k(z)-\Pii_k(z)\right]\bv_k\right| \lesssim \frac{\log n}{q d_k^2}\left( 1+\frac{K}{d_k^4}\right)
\eeq
w.h.p., where $\Pii_k$ is defined as
$$\Pii_k(z):=\Pii(z)-\Pii(z)  \bV_{-k} \frac{1}{\bbD_{-k}^{-1} + \bV_{-k}^\top \Pii(z)\bV_{-k} } \bV_{-k}^\top \Pii(z). $$
%Hence, by \eqref{asymp_Pi} and 

 In light of \eqref{asymp_Pi} and Proposition \ref{lem_locallawv}, it holds that 
\begin{align}
&\mathbf e_{i}^\top \Pii(z)\bv_{k}=\OO\left\{|d_k|^{-1}|v_k(i)| \right\}, \ \ \mathbf e_{i}^\top \Pii(z)\bV_{-k}=\OO\left\{ |d_k|^{-1}\| \bV_{-k}(i)\|\right\},\label{eq_use_aniso1.9}
\end{align}
and w.h.p.,
\begin{align}
 & \left| \mathbf e_{i}^\top \left[\bG(z)-\Pii(z)\right]\bv_{k}\right| \lesssim \frac{1}{|d_k|^2}\left(\sqrt{\frac{\log n}{n}} +\frac{\log n}{q}\|\bv_k\|_\infty  \right) ,\label{eq_use_aniso2.0}\\
 & \left\| \mathbf e_{i}^\top \left[\bG(z)-\Pii(z)\right]\bV_{-k}\right\| \lesssim \frac{\sqrt{K}}{|d_k|^2}\left(\sqrt{\frac{\log n}{n}} +\frac{\log n}{q}\|\bV_{-k}\|_{\max}\right) . \label{eq_use_aniso2}
\end{align}
Then combining \eqref{eq_use_aniso0.5}--\eqref{eq_use_aniso1.5}, \eqref{denom_lower2}, and \eqref{eq_use_aniso1.9}--\eqref{eq_use_aniso2}, we can deduce that w.h.p.,
\begin{align}
&\sup_{z\in \cal C_k}  \left|\be_i^\top \left[\bR_k(z)-\Pii_k(z)\right]\bv_k\right|\nonumber\\
&\lesssim \frac{1}{d_k^2} \left(\sqrt{\frac{\log n}{n}} +\frac{\log n}{q}\|\bv_k\|_\infty  \right) +   \frac{\| \bV_{-k}(i)\|}{|d_k|^3} \frac{K\log n}{q|d_k|}  + \frac{\| \bV_{-k}(i)\|}{|d_k| }\frac{\sqrt{K}\log n}{q|d_k|} \nonumber\\
& \quad  +  \frac{\sqrt{K}}{d_k^2} \left(\sqrt{\frac{\log n}{n}} +\frac{\log n}{q}\|\bV_{-k}\|_{\max}\right)\left(\frac{1}{d_k^2}+\frac{\sqrt{K}\log n}{q|d_k|} \right)  \nonumber\\
 &  \lesssim \frac{1}{d_k^2} \left(1+\frac{\sqrt{K}}{d_k^2}\right)\left(\sqrt{\frac{\log n}{ n}}+ \frac{K {\log n}}{q} \|\bV\|_{\max}\right).\label{aniso_lawk0}
 %& \lesssim  \frac{1}{d_k^2}  \sqrt{\frac{\log n}{ n}}\left(1+\frac{K}{d_k^2}\right)\left(1+\frac{K\sqrt{\log n}}{q}\right) . \quad \label{aniso_lawk0}
\end{align}

% \beq\label{aniso_lawk0}
% \sup_{z\in \cal C_k}  \left|\be_i^\top (\bR_k(z)-\Pii_k(z))\bv_k\right| \lesssim \frac{1}{d_k^2} \left(\sqrt{\frac{\log n}{ n}} +\frac{\log n}{q \sqrt{n}}  \right) \le \frac{2}{d_k^2}  \sqrt{\frac{\log n}{ n}} , \quad \text{}
% \eeq

Let us first calculate \eqref{eq_evector} for the case of $\bx=\by=\bv_k$. It follows from \eqref{asymp_Pi} and \eqref{denom_lower2} that for $z\in \mathcal C_k$,
$$ \bv_k^\top \Pii_k(z) \bv_k = - z^{-1} + \OO(|z|^{-3}),$$
% Combining it with \eqref{aniso_lawk}, we get that 
% $$ \bv_k^\top \bR_k(z) \bv_k = - z^{-1} + \OO\left(\frac{1}{|d_k|^3}+ \frac{\log n}{q d_k^2}\right)\quad \text{w.h.p.},$$
which entails that
\beq\label{denom_Rkvk}
\max_{z\in \cal C_k}\left|1 + d_k\bv_k^\top \Pii_k(z)\bv_k\right| \ge \frac{c}{1+c} +\oo(1)
\eeq
with $c$ the constant given in \eqref{defn_Ck}.
Then from \eqref{aniso_lawk} and \eqref{denom_Rkvk}, we can obtain that
\begin{align}
 \bv_k^\top \hat \bv_k  \hat \bv_k^\top \bv_k &= \frac{1}{2\pi \ii}\oint_{\cal C_k}\frac{ \left[\bv_k^\top \bR_k(z) \bv_k\right]^2     }{d_k^{-1} + \bv_k^\top \bR_k(z)\bv_k}\dd z   = \frac{1}{2\pi \ii \,d_k}\oint_{\cal C_k}\frac{ 1}{1 + d_k\bv_k^\top \bR_k(z)\bv_k}\dd z \nonumber\\
&= \frac{1}{2\pi \ii\, d_k}\oint_{\cal C_k}\frac{ 1}{1 + d_k\bv_k^\top \Pii_k(z)\bv_k}\dd z  + \OO\left\{  \frac{\log n }{q |d_k|} \left( 1+\frac{K}{|d_k|^4}\right)\right\},   %\label{xvvy1}
\end{align}
{w.h.p.}, where we have applied the Cauchy residue theorem at the pole $z=\wh d_k$ in the second step above (recall \eqref{masterx}). A further application of the Cauchy residue theorem at the pole $z=t_k$ yields that w.h.p.,
\begin{align}\label{evector_main}
\bv_k^\top \hat \bv_k  \hat \bv_k^\top \bv_k &= \frac{ 1}{  d_k^2\bv_k^\top \Pii_k'(t_k)\bv_k}   + \OO\left\{  \frac{\log n }{q|d_k|} \left( 1+\frac{K}{|d_k|^4}\right)\right\}.
\end{align}
Observe that by \eqref{asymp_Pi} and \eqref{eq_tk}, we have 
\[ d_k^2\bv_k^\top \Pii_k'(t_k)\bv_k = d_k^2/t_k^{2}+\OO(d_k^2t_k^{-4})=1+\OO(|d_k|^{-2}). \] 
Thus, taking the square root of \eqref{evector_main} gives that w.h.p.,
\begin{align} \label{evector_main0.5}
\bv_k^\top \hat \bv_k  = |d_k|^{-1} \left[\bv_k^\top \Pii_k'(t_k)\bv_k\right]^{-1/2}   + \OO\left\{  \frac{\log n }{q|d_k|} \left( 1+\frac{K}{|d_k|^4}\right)\right\}.
\end{align}

We next estimate \eqref{eq_evector} for the case of $\bx=\be_i$ and $\by=\bv_k$. 
% Hence using $\|\bv\|_\infty=\OO(n^{-1/2})$, we get from \eqref{aniso_lawk0} that w.h.p.,
% \beq\label{aniso_lawk2}
% \sup_{z\in S(\wt C)}  \left[\left|\bx^\top (\bG(z)-\Pii(z))\by\right|  + \left|\bx^\top (\bR_k(z)-\Pii_k(z))\by\right| \right] \le \frac{C}{d_k^2}  \sqrt{\frac{\log n}{ n}} , \quad \text{w.h.p.}
% \eeq
From \eqref{asymp_Pi} and \eqref{aniso_lawk0}, we can deduce that for $z\in \cal C_k$,
\begin{align}
\mathbf e_i^\top \bR_k(z) \bv_k &= \mathbf e_i^\top \Pii_k(z) \bv_k + \OO\left\{\frac{1}{d_k^2} \left(1+\frac{\sqrt{K}}{d_k^2}\right)\left(\sqrt{\frac{\log n}{ n}}+ \frac{K {\log n}}{q} \|\bV\|_{\max}\right)  \right\} \nonumber\\
&= \OO\left\{\frac{\|\bV\|_{\max}}{|d_k|}\left(1+\frac{\sqrt{K}}{d_k^2}\right) +\frac{1}{d_k^2} \left(1+\frac{\sqrt{K}}{d_k^2}\right)\left(\sqrt{\frac{\log n}{ n}}+ \frac{K {\log n}}{q} \|\bV\|_{\max}\right)\right\}\nonumber\\
&=\OO\left\{\left(1+\frac{\sqrt{K}}{d_k^2}\right) \left(\frac{\|\bV\|_{\max}}{|d_k|}+\frac{ 1}{d_k^2}\sqrt{\frac{\log n}{n}}\right) \right\}\label{aniso_lawk2}
\end{align}
% \begin{align}
% \mathbf e_i^\top \bR_k(z) \bv_k &= \mathbf e_i^\top \Pii_k(z) \bv_k + \OO\left(\frac{1}{d_k^2}  \sqrt{\frac{\log n}{ n}}\left(1+\frac{K}{d_k^2}\right)\left(1+\frac{K\sqrt{\log n}}{q}\right)  \right) \nonumber\\
% &= \OO\left(\frac{1}{\sqrt{n}|d_k|}\left(1+\frac{K}{d_k^2}\right) +\frac{1}{d_k^2}  \sqrt{\frac{\log n}{ n}}\left(1+\frac{K}{d_k^2}\right)\left(1+\frac{K\sqrt{\log n}}{q}\right) \right)\nonumber\\
% &=\OO\left(\frac{1}{\sqrt{n}|d_k|}\left(1+\frac{K}{d_k^2}\right) \left(1+\frac{ \sqrt{\log n}}{|d_k|}\right) \right).\label{aniso_lawk2}
% \end{align}
w.h.p. Thus, in view of \eqref{eq_evector}, we can obtain that w.h.p.,
\begin{align}
 &\mathbf e_i^\top \hat \bv_k  \hat \bv_k^\top \bv_k  = \frac{1}{2\pi \ii}\oint_{\cal C_k}\frac{ \be_i^\top \bR_k(z) \bv_k  \bv_k^\top \bR_k(z) \bv_k  }{d_k^{-1} + \bv_k^\top \bR_k(z)\bv_k}\dd z \nonumber\\
 & = -\frac{1}{2\pi \ii  }\oint_{\cal C_k}\frac{ \mathbf e_i^\top \bR_k(z) \bv_k  }{1 + d_k\bv_k^\top \bR_k(z)\bv_k}\dd z = -\frac{1}{2\pi \ii  }\oint_{\cal C_k}\frac{ \mathbf e_i^\top \bR_k(z) \bv_k  }{1 + d_k\bv_k^\top \Pii_k(z)\bv_k}\dd z \nonumber\\ 
 & \quad+  \OO\left\{\frac{\log n}{q }\left(1+\frac{\sqrt{K}}{d_k^2}\right)^3 \left(\frac{\|\bV\|_{\max}}{|d_k|}+\frac{ 1}{d_k^2}\sqrt{\frac{\log n}{n}}\right)\right\}\nonumber\\
 &=  \frac{  -\mathbf e_i^\top \bR_k(t_k) \bv_k  }{d_k\bv_k^\top \Pii_k'(t_k)\bv_k}  +  \OO\left\{\frac{\log n}{q }\left(1+\frac{\sqrt{K}}{d_k^2}\right)^3 \left(\frac{\|\bV\|_{\max}}{|d_k|}+\frac{ 1}{d_k^2}\sqrt{\frac{\log n}{n}}\right)\right\},
 \label{xvvy2}
\end{align}
where we have used \eqref{aniso_lawk}, \eqref{denom_Rkvk}, and \eqref{aniso_lawk2} in the third step above, and applied the Cauchy residue theorem at the pole $z=t_k$ in the last step.

Dividing equation \eqref{xvvy2} by \eqref{evector_main0.5}, 
we can deduce that w.h.p.,
\begin{align}
  \hat v_k (i)  = -   \sgn&(d_k)  \frac{  \mathbf e_i^\top \bR_k(t_k) \bv_k  }{\left[\bv_k^\top \Pii_k'(t_k)\bv_k\right]^{1/2}}+ \OO\left\{\frac{\log n}{q }\left(1+\frac{\sqrt{K}}{d_k^2}\right)^3 \left(\frac{\|\bV\|_{\max}}{|d_k|}+\frac{ 1}{d_k^2}\sqrt{\frac{\log n}{n}}\right)\right\}\nonumber\\
& =   - t_k\mathbf e_i^\top \bR_k(t_k) \bv_k   + \OO\left\{\left[\frac{\log n}{q} \left(1+\frac{K}{d_k^4}\right) +\frac{1}{|d_k|}\right] \left(1+\frac{\sqrt{K}}{d_k^2}\right) \right. \nonumber\\
&\qquad \qquad \qquad \qquad \qquad  \left. \times \left(\frac{\|\bV\|_{\max}}{|d_k|}+\frac{ 1}{d_k^2}\sqrt{\frac{\log n}{n}}\right)\right\} \nonumber\\ 
&  = - t_k\mathbf e_i^\top \bG(t_k) \bv_k  + \OO\left\{\left(\frac{\sqrt{K}}{|d_k|} + \frac{K\log n}{q}\right) \left(\frac{\|\bV\|_{\max}}{|d_k|}+\frac{ 1}{d_k^2}\sqrt{\frac{\log n}{n}}\right)\right\}, \label{evector_main2}
\end{align}
% \begin{align}
%  &\hat \bv_k (i) = - \sgn(d_k)  \frac{  \mathbf e_i^\top \bR_k(t_k) \bv_k  }{(\bv_k^\top \Pii_k'(t_k)\bv_k)^{1/2}}+ \OO\left(\frac{1}{\sqrt{n}|d_k|}\frac{\log n}{q}\left(1+\frac{K}{d_k^2}\right)^3 \left(1+\frac{\sqrt{{\log n} }}{|d_k|}  \right)\right)\nonumber\\
% &=   - t_k\mathbf e_i^\top \bR_k(t_k) \bv_k  + \OO\left[\frac{1}{\sqrt{n}|d_k|}\left(\frac{\log n}{q} \left(1+\frac{K^2}{d_k^4}\right) +\frac{1}{|d_k|}\right)\left(1+\frac{K}{d_k^2}\right)\left(1+\frac{\sqrt{{\log n} }}{|d_k|}  \right)\right] \nonumber\\ 
% & = - t_k\mathbf e_i^\top \bG(t_k) \bv_k  + \OO\left[\frac{1}{\sqrt{n}|d_k|}\left(\frac{K\log n}{q} +\frac{K}{|d_k|}\right) \left(1+\frac{\sqrt{{\log n} }}{|d_k|}  \right)\right] , \label{evector_main2}
% \end{align}
where $\sgn(\cdot)$ denotes the sign of a given real number, we have used \eqref{aniso_lawk2}, $ \bv_k^\top \Pii_k'(t_k)\bv_k = t_k^{-2}+\OO(t_k^{-4})$, and $|t_k|\sgn(d_k) =t_k$ in the second step above, and we have applied \eqref{eq_use_aniso1.5}, \eqref{denom_lower2}, and \eqref{eq_use_aniso1.9}--\eqref{eq_use_aniso2} in the third step to bound $\mathbf e_i^\top [\bG(t_k)-\bR_k(t_k)] \bv_k  $ as w.h.p.,
\begin{align*}
    & \mathbf e_i^\top [\bG(t_k)-\bR_k(t_k)] \bv_k = \mathbf e_i^\top \bG(t_k)  \bV_{-k} \frac{1}{\bbD_{-k}^{-1} + \bV_{-k}^\top \bG(t_k)\bV_{-k} } \bV_{-k}^\top \bG(t_k) \bv_k\\
&= \OO\left\{\left(\frac{\sqrt{K}\| \bV\|_{\max}}{|d_k|} +\frac{\sqrt{K}}{|d_k|^2}\sqrt{\frac{\log n}{n}}\right) \left(\frac{1}{|d_k|^2}+\frac{\sqrt{K}\log n}{q|d_k|} \right)\right\}.
\end{align*}

%It remains to estimate the first term on the RHS of \eqref{evector_main2}.
%derive an asymptotic expansion of $\bx^\top \hat \bv_k$ under a diverging spike $d_k\to \infty$. 
\iffalse
 Since $\bV_{-k}^\top\bv_k=0$, by \eqref{asymp_Pi} and Lemma \ref{lem_locallaw2}, we have that
$$\bV_{-k}^\top \bG(t_k)\bv_k =\OO\left(\frac1{d_k^3} + \frac{\log n}{qd_k^2}\right) \quad \text{w.h.p.}$$
Using \eqref{eq_use_aniso0.5}--\eqref{eq_use_aniso1.5}, we get that .... 
Plugging it into \eqref{evector_main2} yields that
\begin{align*}
\be_i^\top \bR_k(t_k) \bv_k &= \be_i^\top \bG(t_k)\bv_k - \be_i^\top \bG(t_k)  \bV_{-k} \frac{1}{\bbD_{-k}^{-1} + \bV_{-k}^\top \bG(t_k)\bV_{-k} } \bV_{-k}^\top \bG(t_k) \bv_k \\
&= \be_i^\top \bG(t_k)\bv_k  + \OO\left(  \frac{1}{d_k^2\sqrt{n}}\left(\frac{\log n}{q} +\frac{1}{|d_k|}\right)\left(1+\frac{\sqrt{{\log n} }}{|d_k|}  \right)\right) \quad \text{w.h.p.}
\end{align*}
%Moreover, using \eqref{asymp_Pi}, we obtain that
%$$ \bv_k^\top \Pii_k'(t_k) \bv_k= t_k^{-2} + \OO(t_k^{-4}). $$

Plugging the above two equations into \eqref{evector_main2}, we obtain that
\begin{align}\label{evector_main3}
\be_i^\top \hat \bv_k =   -t_k \be_i^\top \bG(t_k) \bv_k  + \OO\left(  \frac{1}{d_k\sqrt{n}}\left(\frac{\sqrt{{\log n}} }{d_k} + 1\right) \left(\frac1{d_k} + \frac{\log n}{q}\right)\right)\quad \text{w.h.p.}
\end{align}
\fi

It remains to estimate the first term on the RHS of \eqref{evector_main2}. In light of the definition \eqref{eqn_defG}, we can write that 
\begin{align}
	& -t_k \be_i^\top \bG(t_k) \bv_k = \be_i^\top \frac{t_k}{t_k-\bW} \bv_k=  v_k(i) + \be_i^\top \frac{\bW}{t_k-\bW} \bv_k \nonumber\\
	&\quad= v_k(i) - \be_i^\top \bW\Pii(t_k)\bv_k - \be_i^\top \bW\left[\bG(t_k)-\Pii(t_k) \right] \bv_k \nonumber\\
	&\quad= v_k(i) +t_k^{-1} \be_i^\top \bW \bv_k - \be_i^\top \bW\bcE_1(t_k)\bv_k - \be_i^\top \bW\left[\bG(t_k)-\Pii(t_k) \right] \bv_k , \label{eq_expvi}
\end{align}
where $\bcE_1$ is as given in \eqref{asymp_Pi}. It follows from \eqref{Ber_logn} that w.h.p.,
\beq\label{eq_expvi_derv1}
\begin{split}
 \be_i^\top \bW\bcE_1(t_k)\bv_k &= \sum_{j\in [n]} W_{ij}[\bcE_1(t_k)]_{jj} v_k(j) \\
& = \OO\left\{\frac{1}{ |d_k|^3}\left(\sqrt{\frac{\log n}{n}}+\frac{\log n}{q}\|\bv_k\|_{\infty}\right)\right\}.
\end{split}
\eeq
 Moreover,  an application of Proposition \ref{lem_locallawv2} gives that 
\beq\label{eq_expvi_derv2}
\be_i^\top \bW\left[\bG(t_k)-\Pii(t_k) \right] \bv_k = \OO\left\{ \frac{1}{|d_k|^2}  \left(\sqrt{\frac{\log n}{n}}+\|\bv_k\|_\infty\right)\right\}
\eeq
w.h.p. Therefore, plugging \eqref{eq_expvi_derv1} and \eqref{eq_expvi_derv2} into \eqref{eq_expvi}, we can obtain that
\beq\nonumber
\begin{split}
& -t_k \be_i^\top \bG(t_k) \bv_k =v_k(i) + \frac{1}{t_k}\sum_{ l \in [n]} W_{il} v_k(l)  + \OO\left\{ \frac{1}{|d_k|^2}  \left(\sqrt{\frac{\log n}{n}}+\|\bv_k\|_\infty\right)\right\}
\end{split}
\eeq
w.h.p., which along with \eqref{evector_main2} leads to the desired conclusion in \eqref{expansion_evector2}. This completes the proof of Theorem \ref{asymp_evector}.

%\numberwithin{equation}{section}
%\renewcommand{\theequation}{\thesection.\arabic{equation}}
\renewcommand{\thesubsection}{D.\arabic{subsection}}
	
\section{Proofs of Propositions \ref{lem_locallawv}--\ref{lem_locallawv2} and some key lemmas as well as additional technical details
} \label{Sec.newB}

\subsection{Proof of Proposition \ref{lem_locallawv}} \label{Sec.newA.6}

To facilitate the technical presentation, let us introduce the inner product notation $\langle\bv,\bw\rangle:=\bv^*\bw$ 
for any complex-valued vectors $\mathbf v,\mathbf w \in \mathbb C^{n}$ and the notion of generalized entries % in the following proof:  and $i\in \{1,\cdots, n\}$, we denote
\begin{equation}
G_{\mathbf{vw}}:=\langle \mathbf v,\bG\mathbf w\rangle, \quad G_{\mathbf{v}i}:=\langle \mathbf v,\bG\mathbf e_i\rangle, \quad G_{i\mathbf{w}}:=\langle \mathbf e_i,\bG\mathbf w\rangle.
\end{equation}
%where $\mathbf e_i$ represents the standard unit vector along the $i$th coordinate.
For each deterministic unit vector $\bv\in \C^n$, let us define
$\Lambda_{\bv}(z):=\max_{i\in [n]} \left| G_{i\bv}(z)-\Pie_{i\bv}(z) \right|$
for $z \in \mathbb{C}$. It follows from \eqref{bound_G} in Lemma \ref{lem_opbound} and \eqref{resolvent2} that 
\beq\label{schur_off2} 
\begin{split}
G_{i\bv} -\Pie_{i\bv}&= - G_{ii} \sum_{k}^{(i)} W_{ik} \cdot \sum_{j}^{(i)}G^{(i)}_{kj} v_j  + v_i (G_{ii}-M_i)\\
&=  - G_{ii} \sum_{k}^{(i)} W_{ik}G^{(i)}_{k\bv^{(i)}}   + \OO\left(\frac{\log n}{q|z|^2}\|\bv\|_\infty \right)
\end{split}
\eeq
w.h.p., where $\bv^{(i)}$ denotes the vector obtained by setting the $i$th entry of $\bv$ as zero and we have used \eqref{aniso_law} from Theorem \ref{lem_locallaw2} to bound the term $|G_{ii}-M_i|$ above. Then in view of \eqref{Ber_logn}, \eqref{bound_G},  \eqref{simpleG}, and \eqref{schur_off2}, we can obtain that w.h.p., for each $i\in [n]$, 
%for a large enough constant $C>0$,
\beq\label{eq_pf_off2}
 \left|G_{i\bv} -\Pie_{i\bv}\right| \lesssim \frac{1}{|z|^2} \sqrt{\frac{\log n}{n}} + \frac{\log n}{q|z|^2}\|\bv\|_\infty  +  \frac{\log n}{q|z|}\max_{k\in [n]\setminus \{i\}} \left|G^{(i)}_{k\bv^{(i)}}\right|.
\eeq

Furthermore, from \eqref{resolvent3} we can deduce that w.h.p.,
\begin{align*}
\left|G^{(i)}_{k\bv^{(i)}}\right| &\le |G_{k\bv^{(i)}}| + \left|\frac{G_{ki}}{G_{ii}}\right| |G_{i\bv^{(i)}}|\le |G_{k\bv^{(i)}}| + |G_{i\bv^{(i)}}| \\
&\le |G_{k\bv}-\Pie_{k\bv}| + |G_{i\bv}-\Pie_{i\bv}|+  \left({C}{|z|}^{-1}\|\bv\|_\infty + 2\|\Pii\bv\|_\infty \right) \\
&\le 2\Lambda_{\bv} +  {C}{|z|}^{-1}\|\bv\|_\infty,
\end{align*}
where we have used \eqref{aniso_law} and \eqref{diagonal_Gii} in the second step above and applied \eqref{boundMz} in the last step. Therefore, plugging it into \eqref{eq_pf_off2} above and taking the maximum over $i \in [n]$, we can obtain that w.h.p.,
$$\Lambda_{\bv}(z)\lesssim \frac{1}{|z|^2} \sqrt{\frac{\log n}{n}} + \frac{\log n}{q|z|^2}\|\bv\|_\infty + \frac{\log n}{q|z|}\Lambda_{\bv} ,$$ 
%\ \Rightarrow \ \Lambda_{\bv}(z)\le \frac{C}{|z|^2} \sqrt{\frac{\log n}{n}} + \frac{C\log n}{q|z|^2}\|\bv\|_\infty .$$
which entails the desired conclusion in \eqref{aniso_law222} for each fixed $z$. Finally, an application of a standard $\epsilon$-net method with a simple union bound argument can enable us to improve the result to a uniform bound in $z\in S(C_0)$, which concludes the proof of Proposition \ref{lem_locallawv}.

\subsection{Proof of Proposition \ref{lem_locallawv2}} \label{Sec.newA.7}

For each deterministic unit vector $\bv\in \R^n$, we write that 
	\begin{align}
		&  \be_i^\top \bW\left[\bG(z)-\Pii(z)\right] \bv  
		= \sum_{j} W_{ij}\left(G_{j\bv}- \Pie_{j\bv}\right) \nonumber\\
		%&= \sum_{j} W_{ij}\left(G_{j\bv}^{(i)}- \Pie_{j\bv}^{(i)}\right)  + \sum_{j} W_{ij} \frac{G_{ji}G_{i\bv}}{G_{ii}}-v(i)W_{ii}M_{i} \nonumber\\
		&=  \sum_{j}^{(i)} W_{ij}\left(G_{j\bv}^{(i)}- \Pie_{j\bv}\right)  + \sum_{j}^{(i)} W_{ij} \frac{G_{ji}G_{i\bv}}{G_{ii}}+ W_{ii} \left(G_{i\bv}-\Pie_{i\bv}\right), \label{eq_localGPi0}
% 		\nonumber\\
% 		&= \sum_{j} W_{ij}\left(G_{j\bv}^{(i)}- \Pie_{j\bv}^{(i)}\right)  + \sum_{j}^{(i)} W_{ij} \frac{G_{ji}G_{i\bv}}{G_{ii}}+ \OO\left( \frac{1}{q|z|^{2}} \left(\sqrt{\frac{\log n}{ n}} +\frac{\log n}{q}\|\bv\|_\infty \right)\right), 
	\end{align}
	where %$\Pii^{(i)}$ denotes the minor of matrix $\Pii$ corresponding to subset $\{i\}$ in the sense of Definition \ref{defminor} and 
	we have used \eqref{resolvent3} in the second step. 
The third term on the RHS of \eqref{eq_localGPi0} can be controlled using Proposition \ref{lem_locallawv}.
% \iffalse
% For each deterministic unit vector $\bv\in \R^n$, we write that 
% 	\begin{align}
% 		  \be_i^\top \bW\left[\bG(z)-\Pii(z)\right] \bv  %= \sum_{ j\in [n]} W_{ij}(G_{j\bv}- \Pie_{j\bv}) \nonumber\\ &
% 		&= \sum_{j} W_{ij}\left(G_{j\bv^{(i)}}- \Pie_{j\bv^{(i)}}\right) + v(i) \sum_{j} W_{ij}(G_{ji}-\Pie_{ji}) \nonumber\\
% 		&= \sum_{j}^{(i)} W_{ij}\left(G_{j\bv^{(i)}}- \Pie_{j\bv^{(i)}}\right)  + \OO\left(\|\bv\|_\infty\frac{\log n}{q^2|z|^2}\right) \label{eq_localGPi0}
% 	\end{align}
% w.h.p., where we have exploited Theorem \ref{lem_locallaw2} and \eqref{support-W} in the third step above. Observe that the first term on the RHS of (\ref{eq_localGPi0}) satisfies that 
% 	\begin{align}\label{eq_localGPi0.5}
% & \sum_{j}^{(i)} W_{ij}\left(G_{j\bv^{(i)}}- \Pie_{j\bv^{(i)}}\right) =\sum_{j}^{(i)} W_{ij}\left(\bG^{(i)} -\Pii^{(i)}\right)_{j\bv^{(i)}}  + \sum_{j}^{(i)} W_{ij} \frac{G_{ji}G_{i\bv^{(i)}}}{G_{ii}} ,
% \end{align}
% where $\Pii^{(i)}$ denotes the minor of matrix $\Pii$ corresponding to subset $\{i\}$ in the sense of Definition \ref{defminor} and we have used \eqref{resolvent3}.
% \fi
Now we bound the first term on the RHS of  \eqref{eq_localGPi0}. 
It follows from \eqref{QVE}, \eqref{bound_OP}, and \eqref{boundMz} that 
\begin{align*}
\left\| \bG^{(i)}-\Pii^{(i)} \right\| \le \left\| \frac{\bW^{(i)}}{z(\bW^{(i)}-z)} \right\| + \left\| \Pii^{(i)} +z^{-1}\bI \right\| \lesssim z^{-2}
\end{align*}
w.h.p., where $\Pii^{(i)}$ denotes the minor of matrix $\Pii$ corresponding to subset $\{i\}$ in the sense of Definition \ref{defminor}.  Furthermore, by applying Proposition \ref{lem_locallawv} to $\bG^{(i)}$, we have that w.h.p.,
\beq\label{eq_temp1} 
\max_{j \in [n]} \big|G_{j\bv}^{(i)}- \Pie_{j\bv}^{(i)}\big| \lesssim  \frac{1}{|z|^2} \left(\sqrt{\frac{\log n}{ n}} +\frac{\log n}{q}\|\bv\|_\infty \right).
\eeq
Since  $W_{ij}$ is independent of $G_{j\bv}^{(i)}- \Pie_{j\bv}^{(i)}$, an application of \eqref{Ber_logn} then yields that w.h.p.,
\beq\label{eq_localGPi1}
\begin{split}
 & \sum_{j}  W_{ij}\left(G_{j\bv}^{(i)}- \Pie_{j\bv}^{(i)}\right) = \OO\left\{ \frac1{|z|^2}\sqrt{\frac{\log n}{n}}\right.\\
 &\quad\quad\left.+\frac{\log n}{q |z|^2} \left(\sqrt{\frac{\log n}{ n}} +\frac{\log n}{q}\|\bv\|_\infty \right)\right\}.
 \end{split}
\eeq

It remains to bound the second term on the RHS of \eqref{eq_localGPi0}. In light of \eqref{resolvent2}, we have 
\begin{align}
	\sum_{j }^{(i)} W_{ij} \frac{G_{ji}G_{i\bv}}{G_{ii}} %&= v(i) \sum_{j }^{(i)} W_{ij} G_{ji} +\sum_{j }^{(i)} W_{ij} \frac{G_{ji}G_{i\bv^{(1)}}}{G_{ii}} \nonumber\\
	&=- G_{i\bv} \sum_{j,k}^{(i)} W_{ij} W_{ik} G^{(i)}_{jk}.\label{eq:Giv}
\end{align}
Then an application of Lemma \ref{largedeviation} with $\xi=(\log n)^2$ yields that w.h.p., 
$$\bigg|\sum_{j,k}^{(i)} W_{ij}W_{ik}G_{jk}^{(i)} - \sum_{k}^{(i)}s_{ik}G_{kk}^{(i)}\bigg| \lesssim \frac{(\log n)^4}{q|z|} \ll |z|^{-1}, $$
which entails that  $\sum_{j,k}^{(i)} W_{ij}W_{ik}G_{jk}^{(i)}=\OO(|z|^{-1})$ w.h.p.
This together with \eqref{eq:Giv} and Proposition \ref{lem_locallawv} for $G_{i\bv}$ yields that
\beq\label{eq_localGPi2}
	\sum_{j}^{(i)} W_{ij} \frac{G_{ji}G_{i\bv}}{G_{ii}}= \OO\left( \frac{\|\bv\|_\infty}{|z|^2} + \frac{1}{|z|^3}\sqrt{\frac{\log n}{ n}} \right).
\eeq
Therefore, plugging \eqref{eq_localGPi1} and \eqref{eq_localGPi2} into \eqref{eq_localGPi0}, we can obtain the desired conclusion in \eqref{aniso_law333} for each fixed $z$. Finally, using a standard $\epsilon$-net argument and taking a union bound, we can further strengthen the result to a uniform bound in $z\in S(C_0)$. This completes the proof of Proposition \ref{lem_locallawv2}.

\subsection{Proof of Lemma \ref{lem-signal}}\label{Sec.new.RPpf}

To show \eqref{eq:signal-after-pairing}, we denote \eqref{eq:max_sig} as $\ell$ and assume that the maximum is achieved at the node pair $\{i_0,j_0\}$. Let us define two subsets 
$$A_1:=\{i\in \cal M: \left\|\bD_{K_0}\left[\bV_{K_0}(i)-\bV_{K_0}(i_0)\right]\right\|\le \ell/3\}$$ 
and 
$$A_2:=\{i\in \cal M: \left\|\bD_{K_0}\left[\bV_{K_0}(i)-\bV_{K_0}(j_0)\right]\right\|\le \ell/3\}.$$ That is, $A_1$ and $A_2$ are the sets of network nodes that are ``close" to nodes $i_0$ and $j_0$, respectively. It is easy to see that $A_1\cap A_2=\emptyset$.  If $|A_1|/m=\oo(1)$, then with probability $1-\oo(1)$, node $i_0$ is coupled with some node $l\notin A_1$ and thus 
\begin{align*}
\max_{\{i,j\}\in \cal P} \left\|\bD_{K_0}\left[\bV_{K_0}(i)-\bV_{K_0}(j)\right]\right\| & \ge   \left\|\bD_{K_0} \left[\bV_{K_0}(i_0)-\bV_{K_0}(l)\right]\right\| \ge  {\ell}/{3}.
\end{align*}
Similar argument also works when $|A_2|/m=\oo(1)$. 

Now let us assume that $|A_1|/m\ge c$ and $|A_2|/m\ge c$ for some constant $c>0$. Then with probability $1-\oo(1)$\footnote{Assume that the nodes in $A_1$ are $a_1, \cdots, a_{|A_1|}$. Then node $a_1$ is not coupled with a node in $A_2$ with probability at most $1-|A_2|/m$. Conditional on this event, node $a_2$ is not coupled with a node in $A_2$ with probability at most $1-|A_2|/(m-2) \le 1-|A_2|/m$. Hence, the probability that no element in $A_1$ is coupled with elements in $A_2$ is at most $ \left(1-{|A_2|}/{m}\right)^{|A_1|} \le (1-c)^{|A_1|}\to 0$ as $m \rightarrow \infty$.
}, %{\color{blue} can you elaborate this more?} 
there is at least one element $l_1\in A_1$ that is coupled with an element in $l_2 \in A_2$ and hence  
\beq\label{power_pair}
\max_{\{i,j\}\in \cal P} \left\|\bD_{K_0} \left[\bV_{K_0}(i)-\bV_{K_0}(j)\right]\right\|_2  \ge   \left\|\bD_{K_0}\left[\bV_{K_0}(l_1)-\bV_{K_0}(l_2) \right ] \right\|_2 
\ge {\ell}/{3}.
\eeq
This concludes the proof of Lemma \ref{lem-signal}.

\subsection{Proof of \eqref{diff_ratios_null} and \eqref{diff_ratios_alt} }\label{sec:proof-diff_ratios_null}

\iffalse
Let us first note that expression \eqref{diff_ratios} holds because
%can be further written as
\begin{equation*}
    \begin{split}
& \frac{\bV_{K_0}(i)}{v_1(i)} - \frac{\bV_{K_0}(j)}{v_1(j)} = \frac{\bB_{K_0}^\top \bpi_i }{\bpi_i^{\top} \bB \be_1}- \frac{\bB_{K_0}^\top \bpi_j}{\bpi_j^{\top} \bB\be_1}\\
& \quad = \frac{\vartheta_i\vartheta_j}{v_1(i)v_1(j)}\left[\bB_{K_0}^\top \left(\bpi_i \bpi_j^\top- \bpi_j \bpi_i^\top\right) \bB\be_1 \right],
\end{split}
\end{equation*}
where %$\be_1$ is the standard unit vector along the first coordinate axis in $\R^{K_0}$ and $\bI_{K\times K_0} $ and 
$\bB_{K_0}\in \R^{K\times K_0}$ is defined as
$$
 \bB_{K_0}:=\bB\bI_{K\times K_0}= \bP \bPi^\top \bTheta \bV \bD^{-1}\bI_{K\times K_0}= \bP \bPi^\top \bTheta \bV_{K_0}\bD_{K_0}^{-1}$$
with $\bI_{K, K_0}:=(\bI_{K_0} , \mathbf 0 )^\top \in \R^{K\times K_0}$. 
where the node degree parameters $\vartheta_i$ and $\vart_j$ are cancelled by taking the ratios. Then 
\fi

From \eqref{diff_ratios}, we can obtain that under the null hypothesis $H_0$ in \eqref{eq: hypothesis}, 
\begin{align*}
\left\|\bD_{K_0}\left[\frac{\bV_{K_0}(i)}{v_1(i)} - \frac{\bV_{K_0}(j)}{v_1(j)}\right] \right\| 
& \lesssim \frac{\vartheta_i\vartheta_j \left\|\bpi_i-\bpi_j\right\|}{\left|v_1(i)\right|\left|v_1(j)\right|}  \Big[ \left\|\bB \be_1\right\| \left\| \bD_{K_0}\bB_{K_0}^\top \bpi_j\right\| \\
& \quad + \left|\bpi_j^{\top} \bB\be_1\right| \lambda_{\max}^{1/2} \left\{\bD_{K_0} \bB_{K_0}^\top \bB_{K_0} \bD_{K_0}\right\} \Big] \nonumber\\
& \lesssim q\sqrt{ Kd_1\lambda_1(\bP)} \left\|\bpi_i-\bpi_j\right\| \nonumber\\
& \le  q \sqrt{K d_1 \lambda_1(\bP)} c_{1n},
\end{align*}
where in the second step above, we have utilized the facts that $\sqrt{\theta}\sim \vartheta_i\sim \vartheta_j $ by part (i) of Condition \ref{main_assm_DCMM}, $v_1(i)\sim v_1(j)\sim n^{-1/2}$ by part (iv) of Condition \ref{main_assm_DCMM}, 
$$ \left\| \bD_{K_0}\bB_{K_0}^\top \bpi_j\right\|^2 \le \lambda_{\max} \left\{\bD_{K_0} \bB_{K_0}^\top \bB_{K_0} \bD_{K_0}\right\} \le \lambda_1(\bP)\lambda_{\max} \left(\bH\right) = d_1\lambda_1(\bP),$$
and that $\bB^\top \bPi^\top \bTheta^2 \bPi\bB=\bV^\top \bV=\bI_K$, which entails that 
$$ \left\|\bB \be_1\right\| \lesssim \big\|\left(\bPi^\top \bTheta^2 \bPi\right)^{-1}\big\|^{1/2} \lesssim \sqrt{K}/q
$$
by part (iii) of Condition \ref{main_assm_DCMM}.
%and in the third step we used $H_0$ in \eqref{eq: hypothesis} and $\lambda_{\max}\left(\bH\right) \le \lambda_1(\bP) \lambda_{\max}(\bPi^\top \bPi) \vartheta_{\max}^2 \lesssim n\theta=q^2 $. 

On the other hand, 
% we have another representation 
% \beq\label{diff_ratios2}
% \frac{\bV_{K_0}(i)}{v_1(i)} - \frac{\bV_{K_0}(j)}{v_1(j)} =\bB_{K_0}^\top \left(\bpi_i , \bpi_j\right)\left( \frac{\vartheta_i}{v_1(i)}, -\frac{\vartheta_j}{v_1(j)}\right)^\top.
% \eeq
with the aid of (\ref{diff_ratios2}), we can deduce that 
\begin{align}
 & \left\|\bD\left[\frac{\bV(i)}{v_1(i)} - \frac{\bV(j)}{v_1(j)}\right] \right\| \gtrsim \sqrt{n\theta}\lambda_{\min}^{1/2}\left\{ \left(\bpi_i , \bpi_j\right)^\top  \left(\bpi_i , \bpi_j\right)\right\}\lambda_{\min}^{1/2}\left\{\bB \bD^2 \bB^\top \right\} \nonumber\\
&\quad \ge q\sqrt{ \lambda_K(\bP)}\lambda_{\min}^{1/2}\left\{ \left(\bpi_i , \bpi_j\right)^\top  \left(\bpi_i , \bpi_j\right)\right\} \lambda_{\min}^{1/2}\left\{\bV^\top \bH\bV \right\} \nonumber\\
&\quad \ge q \sqrt{d_{K}\lambda_K(\bP)} \lambda_{\min}^{1/2}\left\{ \left(\bpi_i , \bpi_j\right)^\top  \left(\bpi_i , \bpi_j\right)\right\},
\end{align}
where we have again used parts (i) and (iv) of Condition \ref{main_assm_DCMM} in the above derivation. This completes the proof of both \eqref{diff_ratios_null} and \eqref{diff_ratios_alt}.

\subsection{Proof of Lemma \ref{lem_opbound}} \label{Sec.newB.1}

The proof of Lemma \ref{lem_opbound} involves a standard application of the moment method used in random matrix theory. Specifically, for each $k\in \N$, it holds that
\beq\label{moment_W}\E \|\bW\|^{2k} \le \E \, \tr (\bW^{2k}).\eeq
Using similar arguments as in the proof
%s of Lemma 7.2 of \cite{ErdYauYin2012Univ} and 
Lemma 4.3 of \cite{EKYY_ER1}, we can show that for each $k\le c\sqrt{q}$,
$$\E  \, \tr (\bW^{2k}) \le C n k (4\mathfrak M)^{k}, $$
where $c,C>0$ are some constants. In fact, \cite{EKYY_ER1} considered the case with $\sum_{j\in [n]} s_{ij}\equiv 1$, which can be replaced with the assumption $\sum_{j\in [n]}s_{ij}\le \mathfrak M$ in our technical analysis. 

Then it follows from \eqref{moment_W} and the Markov's inequality that 
$$\|\bbW\| \le 2\sqrt{\mathfrak M} +\xi/\sqrt{q}
$$
with $(c',\xi)$-high probability for some small constant $c'>0$. Moreover, we can establish similar bounds for each $\|\bW^{(i)}\|$ and $\|\bW^{(ij)}\|$. Hence, taking a union bound yields the desired conclusion in \eqref{bound_OP}. Finally, the desired bound in \eqref{bound_G} follows immediately from  that 
$$\|\bG(z)\|\le (|z|-\|\bbW\|)^{-1}$$ and similar bounds for $\|\bG^{(i)}(z)\|$ and $\|\bG^{(ij)}(z)\|$. This concludes the proof of Lemma \ref{lem_opbound}.

\subsection{Proof of Lemma \ref{lem_abs_decouple}} \label{Sec.newB.3}

Using similar arguments as in the proof of Theorem 5.6 in \cite{EKYY_ER1}, we can establish that 
\begin{align}\label{improve_abs_decouple}
   \E\bigg[\mathbf 1(\Xi)  \bigg||z|^2\cdot \frac{1}{n}\sum_{i\in [n]} a_i \cal Z_i\bigg|^p\bigg] \le \left[Cp^6 Y(X^2+n^{-1})\right]^p
\end{align}
for some constant $C>0$ and all large enough $n$, where $p\in 2\N$ is an even integer. In fact, a slightly weaker bound 
\begin{align*}
   \E\bigg[\mathbf 1(\Xi)  \bigg||z|^2\cdot \frac{1}{n}\sum_{i\in [n]} a_i \cal Z_i\bigg|^p\bigg] \le \left[Cp^{11} Y(X^2+n^{-1})\right]^p
\end{align*}
was shown in \cite{EKYY_ER1}. 
We can improve the above bound to \eqref{improve_abs_decouple} through tightening the argument below equation (5.21) of \cite{EKYY_ER1}. Since such improvement is straightforward to derive, we do not provide the technical details here for simplicity. Now combining \eqref{improve_abs_decouple} with the Markov's inequality yields the desired conclusion in (\ref{eq_abs_decouple}). This concludes the proof of Lemma \ref{lem_abs_decouple}.

\subsection{Proof of Claim \ref{claim_moment_aniso}} \label{new.secB.claim1}

We now focus on establishing the desired bound \eqref{moment_aniso} in Claim \ref{claim_moment_aniso} from the proof of Theorem \ref{lem_locallaw2} in Section \ref{sec_moments} using the cumulant expansion formula listed in the lemma below, which was proved in Proposition 3.1 of \cite{Cumulant1} and Section II of \cite{Cumulant2}.
	
\begin{lemma}%[\cite{Cumulant1,Cumulant2}] 
\label{cumulant lem}
		%Fix any $l\in \N$ and let $f\in \cal C^{l+1}(\R)$. 
Let $h$ be a real-valued random variable bounded by $|h|\le a$ for some $a > 0$. Then for each $l\in \N$ and $f\in \cal C^{l+1}(\R)$, the class of $(l+1)$-th order differentiable functions on $\R$, it holds that 
		$$\mathbb E [f(h) h]=\sum_{k=0}^{l}\frac1{k!}\kappa_{k+1}(h)\mathbb Ef^{(k)}(h)+ R_{l+1},$$
		where $\kappa_{k}(h)$ is the $k$th cumulant of $h$, $f^{(k)}(\cdot)$ denotes the $k$th derivative of function $f(\cdot)$, and $R_{l+1}$ satisfies that 
		$$|R_{l+1}|\le \frac{1+(3+2l)^{l+2}}{(l+1)!}\mathbb E\left| h\right|^{l+2}\cdot \sup_{|x|\le a}|f^{(l+1)}(x)|.$$
	\end{lemma}

It follows from the definitions of $\bG$ and $\Pii$ given in \eqref{eqn_defG} and \eqref{defn_pi}, respectively, and the QVE given in \eqref{QVE} that 
\begin{align*}
	\bG(z)-\Pii(z)&=\Pii(z)\left[\Pii^{-1}(z)-\bG^{-1}(z)\right]\bG(z) \\
	&=-\Pii(z)\left[ \diag(\bbS \bbM) + \bbW\right]\bG(z).
\end{align*}
With the aid of the above identity, we can write that
\beq\label{EG0}
\begin{split}
	\mathbb E|Y(z)|^{2r} & = - q|z|^2 \cdot \mathbb E \left\{ \left\langle \bu, \Pii(z) \left[ \diag(\bbS \bbM) + \bbW\right]\bG(z) \bv\right\rangle Y^{r-1}\overline Y^{r}\right\},
\end{split}
\eeq
where $\overline{z}$ stands for the complex conjugate of a given complex number $z$.

For each $k\ge 2$, let us define 
$$\kappa_k(i,j):= nq^{k-2}\kappa_k(W_{ij})$$
with $ i,j \in [n]$. In view of \eqref{support-W}, there exists some constant $C>0$ such that 
$$\E |W_{ij}|^k \le \frac{C^k}{nq^{k-2}}.$$
%\href{https://mathoverflow.net/questions/368761/bounds-on-cumulants-in-terms-of-moments}{Bounding cumulants in terms of moments}, 
Then from the inequality $|\kappa_k(W_{ij})|\le k^k \E |W_{ij}|^k$, we can deduce that 
\beq\label{bdd_cumu}
|\kappa_k(i,j)| \le (Ck)^k
\eeq
with $k\ge 2$. The main ingredient for the proof of Claim \ref{claim_moment_aniso} is to exploit Lemma \ref{cumulant lem}, \eqref{bdd_cumu}, and the local laws established in Theorems \ref{lem_locallaw1} and \ref{lem_averlaw} to bound the term $\mathbb E|Y(z)|^{2r}$. 
%We now describe an outline of the proof. 
First, we will apply Lemma \ref{cumulant lem} with $l=4r$ to 
$$\mathbb E \left[ \left\langle \bu, \Pii   \bbW \bG \bv\right\rangle Y^{r-1}\overline Y^{r}\right]=\sum_{ i,j }\mathbb E \left\{ (\Pii\bu)_i W_{ij} G_{j\bv}  Y^{r-1}\overline Y^{r} \right\}$$
with respect to each $W_{ij}$ and write it as a sum of terms containing $k$-th order derivatives of $G_{j\bv}  Y^{r-1}\overline Y^{r}$ with respect to $W_{ij}$ for $1\le k \le l$; see \eqref{G123} below. Next, we will calculate the derivatives using the identity \eqref{dervG} below, which will produce some terms expressed as polynomials of generalized resolvent entries. We then use Lemma \ref{lem_opbound}, the entrywise local law, Theorem \ref{lem_locallaw1}, and the averaged local law, Theorem \ref{lem_averlaw}, to estimate each term. Roughly speaking, our goal is to establish an estimate 
\beq \label{eq:EY2r} \mathbb E|Y(z)|^{2r}   \le   1+ \sum_{k=1}^{2r}  (Cr)^{k} \cdot  \E |Y|^{2r-k} , 
\eeq
which bounds a high moment of $Y(z)$ with lower moments. With the aid of \eqref{eq:EY2r}, applying the H{\"o}lder's and Young's inequalities leads to the desired conclusion \eqref{moment_aniso}. The main technical parts of the proof are some intricate combinatorial arguments that count the number of terms in very high order derivatives of $G_{j\bv}  Y^{r-1}\overline Y^{r}$ with respect to $W_{ij}$.

Hereafter, we will make use of the following simple fact implicitly. Assume that $\mathfrak X$ is a random variable satisfying that 
$|\mathfrak X|\le \Psi$
with $(c,p)$-high probability and $ |\mathfrak X|\le n^{C\log n}$ almost surely, where $\Psi\ge 0$ is a deterministic parameter and $c,C>0$ are some constants. Then if $p\ge \wt C (\log n)^2$ for some constant $\wt C>C/c$, it holds that 
\beq\label{control_smallp} |\E \mathfrak X| \le \Psi + e^{-c'p}\eeq
for some constant $0<c'<c$. Later, we will apply such estimate to bound polynomials of generalized resolvent entries. In particular, for each $z\in \wt S(C_0)$, we have the deterministic bound 
$ \|\bG(z)\|\le \eta^{-1}\le n^4. $ 
Hence, if $\mathfrak X$ is a product of no more than $Cr$ many generalized resolvent entries, it holds that 
$ |\mathfrak X|\le n^{4C\log n}, $
which entails that the bound \eqref{control_smallp} can be applied.

Let us begin with establishing the desired bound \eqref{moment_aniso} for the relatively dense case with $q\gg (\log n)^{16}$, where we will need to resort to Theorems \ref{lem_locallaw1} and \ref{lem_averlaw}. To this end, we will choose parameters $\xi$ and $p$ such that
\beq\label{choose_xiq}
(\log n)^3 \ll \xi \ll (q^{1/8}\log n) \wedge (\log n)^{\log \log n}, \ \ (\log n)^2 \ll p\ll (\xi/\log n)\wedge q^{1/3}. 
\eeq
Then an application of Lemma \ref{cumulant lem} with $h=W_{ij}$ and $a= C/q$ yields that
\begin{align}
	 &\, - q|z|^2 \cdot \mathbb E  \left\{\left\langle \bu, \Pii  \bbW \bG \bv\right\rangle Y^{r-1}\overline Y^{r}\right\} \nonumber \\
	 =&\, - q|z|^2 \cdot \sum_{ i,j \in [n]}\mathbb E \left\{ (\Pii\bu)_i W_{ij} G_{j\bv}  Y^{r-1}\overline Y^{r} \right\} =  \sum_{k=1}^l \mathfrak G_k+ \cal E,\label{eq_cumuW}
\end{align}
where $\mathfrak G_k$ and $\cal E$ are defined as 
\beq\label{G123}
\begin{split}
	\mathfrak G_k :=-\frac{q|z|^2}{k! nq^{k-1}}  \sum_{ i,j \in [n]} \mathcal \kappa_{k+1}(i,j)  (\Pii\bu)_i \mathbb E \frac{\partial^k  (G_{j\bv} Y^{r-1}\overline Y^{r}  )}{\partial (W_{ij})^k } \end{split}
\eeq
and 
\beq\label{GE}\mathcal E: =-q|z|^2  \sum_{ i,j \in [n]}  (\Pii\bu)_i R_{l+1}(i,j),\eeq
respectively, with $R_{l+1}(i,j)$ satisfying that %the bound (using the fact that $|W_{ij}|\le C/q$)
\begin{align}\label{Rl+1}
	|R_{l+1}(i,j)|\le \frac{1+(3+2l)^{l+2}}{(l+1)!} \mathbb E\left| W_{ij}\right|^{l+2}\cdot \mathbb E\sup_{|x|\le C/q}\left| \partial_{ij}^{l+1}  f_{j} (\bW^{[ij]}+x\bDelta_{ij})\right| . 
\end{align}
Here, we have used the shorthand notation that $f_{j}:=G_{j\bv}(z)  Y^{r-1}\overline Y^{r}$, $\partial_{ij}:=\partial/\partial W_{ij}$, 
$$\bDelta_{ij}:= \mathbf e_i   \mathbf e_j^\top + \mathbf 1_{j\ne i} \mathbf e_j   \mathbf e_i^\top , \  \text{ and } \ \bW^{[ij]}= \bW - W_{ij}\bDelta_{ij},$$
that is, $\bW^{[ij]}$ is obtained by setting the $(i,j)$th and $(j,i)$th entries of matrix $\bW$ as zero. % such that $H^{(ij)}$ is independent of $X_{i\mu}$. 
For our technical analysis, we will work with the choice of $l=4r$. 

We next estimate the RHS of \eqref{eq_cumuW} above term by term using the identity
\beq\label{dervG}\frac{\partial^k \bG }{\partial (W_{ij})^k} = (-1)^k k! \bG (\bDelta_{ij}\bG)^k.\eeq
The above identity can be derived from the resolvent expansion that for $x,x'\in \mathbb R$ and $k\in \mathbb N$,
\begin{equation} \label{eq_comp_expansion}
\begin{split}
	\bG_{(ij)}^{x'} & = \bG_{(ij)}^{x}+\sum_{r=1}^{k}  (x-x')^r \bG_{(ij)}^{x}\big[ \bDelta_{ij} \bG_{(ij)}^{x}\big]^r \\
	& \quad +(x-x')^{k+1} \bG_{(ij)}^{x'}\big[\bDelta_{ij} \bG_{(ij)}^{x}\big]^{k+1},
\end{split}
\end{equation}
where we have used the shorthand notation  
$ \bG_{(ij)}^{x}:=  (\bW^{[ij]}+x\bDelta_{ij}-z)^{-1}.$
For simplicity, hereafter we will always assume that the diagonal entries of matrix $\bW$ are zero. The technical analysis without such assumption is almost the same except for some minor differences in the notation regarding the terms that contain $\kappa_k(i,i)$. 

%The following local laws and eigenvalue rigidity have been proved in the literature under the assumption that $\theta \ge n^{-1+c}$ for a constant $c>0$. 

In light of \eqref{dervG}, we can expand term $\mathfrak G_1$ as
  \beq\label{G1}
 \begin{split}
 & \mathfrak G_1 =\frac{q|z|^2}{ n }   \mathbb E  \sum_{ i,j \in [n]}\kappa_2({i,j}) (\Pii\bu)_i  (G_{ji}G_{j\bv} + G_{jj}G_{i\bv}) Y^{r-1}\overline Y^{r}  \\
&\quad+ (r -1) \frac{(q|z|^2)^2}{ n }   \mathbb E  \sum_{ i,j \in [n]} \kappa_2({i,j}) (\Pii\bu)_i  G_{j\bv} Y^{r-2} \overline Y^{r} (G_{\bu i} G_{j\bv}+ G_{\bu j} G_{i\bv}) \\
&\quad+ r\frac{(q|z|^2)^2}{ n }   \mathbb E  \sum_{ i,j \in [n]} \kappa_2({i,j})(\Pii\bu)_i  G_{j\bv} Y^{r-1} \overline Y^{r-1} (\overline G_{\bu i} \overline G_{j\bv}+ \overline G_{\bu j} \overline G_{i\bv}) . 
\end{split}
\eeq
It follows from \eqref{bound_G}, \eqref{boundMz}, and the Cauchy--Schwarz inequality that
\begin{align*}
&  \sum_{ i,j \in [n]} \kappa_2({i,j}) \left|(\Pii\bu)_i\right| \left| G_{j\bv} \right|^2 \left| G_{\bu i}\right| \lesssim \sum_{ i \in [n]} \left|(\Pii\bu)_i\right|  \left| G_{\bu i}\right| \sum_{ j\in [n]} \left| G_{j\bv} \right|^2 \\
&\qquad \lesssim |z|^{-2}\Big(\sum_{i\in [n]} \left|(\Pii\bu)_i\right|^2\Big)^{1/2} \Big(\sum_{i\in [n]} \left| G_{\bu i}\right|^2\Big)^{1/2}   \lesssim |z|^{-4}
\end{align*}
 with $(c_0,\xi)$-high probability. Using a similar argument, we can deduce that
\begin{align*}
    &\sum_{i,j\in [n]} \kappa_2({i,j}) |(\Pii\bu)_i| | G_{j\bv}| |G_{\bu j}| |G_{i\bv}| \lesssim |z|^{-4},\\
    &\sum_{i,j\in [n]}\kappa_2({i,j}) \left|(\Pii\bu)_i \right| \left|G_{ji}\right| \left|G_{j\bv}\right| \lesssim  {n^{1/2}}/{|z|^3},
\end{align*}
with $(c_0,\xi)$-high probability.
Then, with the aid of \eqref{control_smallp} and H{\"o}lder's inequality, we can estimate \eqref{G1} above as
\begin{align*}
	\mathfrak G_1 &= q|z|^2  \mathbb E  \bigg\{\sum_{i\in [n]}  (\Pii\bu)_i G_{i\bv} \cdot \sum_{j\in [n]} s_{ij}G_{jj}\cdot Y^{r-1}\overline Y^{r}\bigg\}   \\
	&\quad+ \OO\left\{ \frac{q}{\sqrt{n}|z|} (\E |Y|^{2r})^{\frac{2r-1}{2r}}+ r\frac{q^2}{n}(\E |Y|^{2r})^{\frac{2r-2}{2r}} +e^{-cp}\right\}
\end{align*}
for some constant $c>0$, where we have used the fact that $s_{ij}=\kappa_2({i,j})/n$. Plugging the above representation into \eqref{eq_cumuW} and further into \eqref{EG0}, it holds that 
\begin{align}
	 & \mathbb E|Y(z)|^{2r} =q|z|^2  \mathbb E  \bigg\{\sum_{i\in [n]}  (\Pii\bu)_i G_{i\bv} \cdot \sum_{j\in [n]} s_{ij}(G_{jj}-M_j)\cdot Y^{r-1}\overline Y^{r}\bigg\}  \nonumber\\
	& \quad +\sum_{k=2}^l \mathfrak G_k+ \cal E+ \OO\left( \frac{q}{\sqrt{n}|z|} (\E |Y|^{2r})^{\frac{2r-1}{2r}} + r\frac{q^2}{n}(\E |Y|^{2r})^{\frac{2r-2}{2r}} +e^{-cp}\right)  \nonumber\\
	&= \sum_{k=2}^l \mathfrak G_k+ \cal E  + \OO\left\{ \left(\frac{\xi^{1/2}}{\sqrt{n}|z|^2}+\frac{p_0^8}{q |z|^3}+\frac{p_0^8 \xi^4}{\sqrt{n}|z|^5}\right) (\E |Y|^{2r})^{\frac{2r-1}{2r}} \right\}  \nonumber\\
	&\quad + \OO\left\{ \frac{q}{\sqrt{n}|z|}(\E |Y|^{2r})^{\frac{2r-1}{2r}}+r\frac{q^2}{n}(\E |Y|^{2r})^{\frac{2r-2}{2r}} +e^{-cp}\right\},\label{G1_est1.0}
	\end{align}
where we have used \eqref{aver_law}, \smash{$\sum_{i\in [n]}  |(\Pii\bu)_i| |G_{i\bv}|\lesssim |z|^{-2} $} with $(c_0,\xi)$-high probability, and $q=\sqrt{n\theta}\le \sqrt{n}$ in the second step above. Moreover, under the choice of parameters given in \eqref{choose_xiq} with $q\gg (\log n)^{16}$, we can further rewrite \eqref{G1_est1.0} as
\beq\label{G1_est1}
	 \mathbb E|Y(z)|^{2r}  = \sum_{k=2}^l \mathfrak G_k+ \cal E  + \OO\left\{ (\E |Y|^{2r})^{\frac{2r-1}{2r}}+r(\E |Y|^{2r})^{\frac{2r-2}{2r}} +e^{-cp}\right\}.
\eeq

Now it remains to estimate the terms $\mathfrak G_k$ and $\cal E$ on the RHS of (\ref{G1_est1}). In order to exploit the structures of the derivatives of resolvent entries in a systematic fashion, let us introduce the algebraic objects below.

\begin{definition}[Words] \label{def_comp_words}
For each given pair of indices $1\le i < j \le n$, we define $\sW$ as the set of words of even lengths in two letters $\{\mathbf i, \mathbf j\}$. Denote by $2{\bm \ell}(w)$ with ${\bm \ell}(w)\in \mathbb N$ the length of each word $w\in\sW$. We use bold symbols to denote the letters of words; for example, $w=\mathbf a_1\mathbf b_2\mathbf a_2\mathbf b_3\cdots\mathbf a_r\mathbf b_{r+1}$ represents a word of length $2r$. Let $\sW_r:=\{w\in \mathcal W: {\bm \ell}(w)=r\}$ be the set of words of lengths $2r$ such that each word $w\in \sW_r$ satisfies that $\mathbf a_l\mathbf b_{l+1}\in\{\mathbf i\mathbf j,\mathbf j\mathbf i\}$ for all $1\le l\le r$.
We assign to each letter a value $[\cdot]$ through $[\mathbf i]:=i$ and $[\mathbf j]:=j$. %where  where $\mathbf u_i$ and $\bv_\mu$ are defined in Lemma \ref{lem_comp_gbound} and are regarded as summation indices. 
It is important to distinguish the abstract letter from its value, which is an index. To each word $w \in \sW_r$, we assign two types of random variables $A_{i, j}(w)$ and $\wt A_{i, j}(w)$ as specified below. If ${\bm \ell}(w)=0$, we define
$$A_{i, j}(w):=G_{\mathbf u\mathbf v}-\Pie_{\mathbf u \mathbf v}, \ \  \wt A_{i, j}(w):=G_{j\mathbf v}.$$
Further, if ${\bm \ell}(w)\ge 1$ with $w=\mathbf a_1\mathbf b_2\mathbf a_2\mathbf b_3\cdots\mathbf a_r\mathbf b_{r+1}$, we define
\begin{equation}\label{eq_comp_A(W)}
\begin{split}
& A_{i, j}(w):=G_{\bu[\mathbf a_1]} G_{[\mathbf b_2][\mathbf a_2]}\cdots G_{[\mathbf b_r][\mathbf a_r]} G_{[\mathbf b_{r+1}]\bv},\\
&\wt A_{i, j}(w):=G_{j[\mathbf a_1]} \overline G_{[\mathbf b_2][\mathbf a_2]}\cdots \overline G_{[\mathbf b_r][\mathbf a_r]} \overline G_{[\mathbf b_{r+1}]\bv}.
\end{split}
\end{equation}
%Finally, for $w=\mathbf a_1\mathbf b_2\mathbf a_2\mathbf b_3\cdots\mathbf a_r\mathbf b_{r+1}$ we define another type of word as
%\begin{equation}\label{eq_comp_A(W)2}
%\wt A_{i, \mu}(w):=G^{(1)}_{\bu_1[\mathbf a_1]} G^{(1)}_{[\mathbf b_2][\mathbf a_2]}\cdots G^{(1)}_{[\mathbf b_r][\mathbf a_r]} G^{(1)}_{[\mathbf b_{r+1}]\mu}.
%\end{equation}
\end{definition}

Observe that words introduced in Definition \ref{def_comp_words} above are constructed in such a way that by \eqref{dervG}, we have that for each $k\in \mathbb N$,
	\[\left(\frac{\partial}{\partial W_{ij}}\right)^k Y=(-1)^k k! q|z|^2 \cdot \sum_{w\in \mathcal W_k} A_{i, j}(w) .\]
Similarly, we see that $\wt A_{i, j}(w)$ is related to the derivatives of $G_{j \bv}$. It follows from \eqref{bound_G} that for each word $w$ with ${\bm \ell}(w)\ge 1$, 
	\begin{align}
		\left|A_{i, j}(w)\right|\le \left( {C}/{|z|}\right)^{{\bm \ell}(w)+1}, & \ \ |\wt A_{i, j}(w)|\le \left( {C}/{|z|}\right)^{{\bm \ell}(w)+1}, \label{Rimu1}\\
		\sum_{i,j\in [n]} \left|A_{i, j}(w)\right|^2 \le n\left( {C}/{|z|}\right)^{2{\bm \ell}(w)+2}, & \ \  \sum_{i,j\in [n]}|\wt A_{i, j}(w)|^2 \le n\left({C}/{|z|}\right)^{2{\bm \ell}(w)+2}, \label{Rimu2}
	\end{align}
	with $(c_0,\xi)$-high probability. In fact, the fourth estimate above also holds for the case of ${\bm \ell}(w)=0$. Moreover, if word $w$ has length ${\bm \ell}(w)= 1$, we have a better bound 
	\begin{align}
		\sum_{i,j\in [n]} \left|A_{i, j}(w)\right|^2 \le \left(  {C}/{|z|}\right)^{2{\bm \ell}(w)+2} \label{Rimu3}
	\end{align}
	with $(c_0,\xi)$-high probability.

	With the above notations in mind, we can deduce that
	\beq\label{dervWord}
	\begin{split}
	& \frac{\partial^k  (G_{j\bv} Y^{r-1}\overline Y^{r}  )}{\partial (W_{ij})^k } = (-1)^k (q|z|^2)^{2r-1}  \sum_{l_1, \cdots, l_{2r},\,l_1+\cdots+l_{2r}=k} \bigg[l_1! \sum_{w_1\in \cal W_{l_1}} \wt A_{i, j}(w_1)\bigg] \\
		&\quad\quad \times \prod_{s=2}^{r} \bigg[l_s! \sum_{w_s\in\sW_{l_s}} A_{i, j}(w_s)\bigg]\prod_{s=r+1}^{2r} \bigg[l_s! \sum_{w_s\in\sW_{l_s}} \overline A_{i, j}(w_s)\bigg].
	\end{split}
	\eeq
	Let us define 
	$$a:=a_1+a_2, \ \ a_1:= \#\{2\le s\le 2r:l_s=1\}, \ \ a_2:=  \#\{2\le s\le 2r:l_s\ge 2\}.$$
	%Without loss of generality, we assume that the words with nonzero length are $w_{s_1},\cdots, w_{s_a}$, and the words with length 1 are $w_{s_1},\cdots, w_{s_{a_1}}$. Then we have 
	Then by definition, it holds that
	\beq\label{sumS}a_1 + 2a_2\le k-l_1 .\eeq
	%Let us also define an $(a+1)$-tuple $\bm l=(l_1, \cdots, l_{a+1})$ and 
	Let us denote the subset of $a$ nonempty words by  
	$\cal S:=\{2\le s\le 2r: l_s\ge 1\} \subset \{2,\cdots, 2r\}.$  
	The empty words will contribute to the $|Y(z)|^{2r-1-a}$ factor. Moreover, we have defined $a_1$ because depending on the number of words of length 1, the RHS of \eqref{dervWord} will behave differently. It is worth mentioning that the set $\cal S$ introduced above does not contain $s=1$. 
	%{\color{red} (To do: it would be helpful to provide some quick intuition on ``the subset of $a$ nonempty words" here)}

Inserting \eqref{dervWord} into \eqref{G123}, we can obtain that 
\begin{align}
	& \left|\mathfrak G_k\right|     \le   \sum_{l_1=0}^k \sum_{a=0}^{(2r-1)\wedge (k-l_1)}\frac{(q|z|^2)^{a+1} (Ck)^k}{k! nq^{k-1}}   \nonumber\\
	&\quad  \times \sum_{\cal S=\{s_1, \cdots, s_a\}\subset \{2,\cdots, 2r\}} \sum_{l_{s_1},\cdots,l_{s_a}:\,  l_{s_1}+ \cdots+l_{s_a}=k-l_1} \left( l_1!\prod_{s\in \cal S}  l_s!\right)\nonumber\\
	&\quad \times \E \bigg\{\sum_{\substack{w_1\in \cal W_{l_1},\,w_{s_1}\in\sW_{l_{s_1}}, \\
	\cdots,w_{s_a}\in\sW_{l_{s_a}} } } \sum_{i,j\in [n]}  |(\Pii\bu)_i| | \wt A_{i, j}(w_1)| \prod_{s\in \cal S} |A_{i, j}(w_s)| \cdot |Y|^{2r-1-a}\bigg\}, \label{abs_Gk}
\end{align}
where we have used \eqref{bdd_cumu} and the fact that $q|z|^2 A_{i,j}(w_s)=Y$ if $\bm\ell(w_s)=0$. Then it follows from \eqref{Rimu1}--\eqref{Rimu3} and the Cauchy--Schwarz inequality that 
\begin{align*}
 & \sum_{i,j\in [n]}  |(\Pii\bu)_i| | \wt A_{i, j}(w_1)| \prod_{s\in \cal S} |A_{i, j}(w_s)| \\
 & \le \left[\mathbf 1(a_1=0) n   + \mathbf 1(a_1=1) \sqrt{n} + \mathbf 1(a_1\ge 2) \right]\left( {C}/{|z|}\right)^{a+2+k}
\end{align*}
with $(c_0,\xi)$-high probability. 
Furthermore, an application of some simple combinatorial arguments yields that for each fixed $l_1$, $a_1$, and $a_2$, %{\cor entropy arument .....}
$$ \left|\{\cal S\subset \{2,\cdots, 2r\}, \, l_{s_1}+ \cdots+l_{s_a}=k-l_1 \}\right|\le  \frac{(2r)^a}{a!}k^a $$
and
$$ \frac1{k!}\cdot l_1!\prod_{s\in \cal S}  l_s!\le \frac{(k-a_1)!}{k!}, \ \ \ \  |\cal W_{l_s}|\le 2^{l_s}.$$
With the aid of the above estimates, we can further bound the RHS of \eqref{abs_Gk} as 
\begin{align}
	\left|\mathfrak G_k\right|   \le  & \sum_{l_1=0}^k \sum_{a=0}^{(2r-1)\wedge (k-l_1)} \sum_{a_1=0}^a\frac{(Cq)^{a+1}\cdot  (Ck)^k}{|z|^{k-a}\cdot  nq^{k-1}}  \frac{ (2rk)^a }{a!} \frac{(k-a_1)!}{k!} \nonumber\\
	& \times  \left[\mathbf 1(a_1=0) n   + \mathbf 1(a_1=1) \sqrt{n} + \mathbf 1(a_1\ge 2) \right]  \cdot \left(\E |Y|^{2r}\right)^{\frac{2r-1-a}{2r}}.\label{boundGk}
\end{align}
We will aim to bound the RHS of \eqref{boundGk} case by case for $3\le k\le l=4r$. 

\medskip
\textit{Case 1: $a_1=0$.} If $3\le k< 20$, it holds that 
\begin{align*}
    \frac{(Cq)^{a+1}\cdot  (Ck)^k}{|z|^{k-a}\cdot  nq^{k-1}}  \frac{(2rk)^a}{a!} {n} &\le (Cr)^{a} \left(\frac{C}{q}\right)^{k-a-2} \le \left({Cr}\right)^{a}, 
%    &\le \min\left\{(Cr)^a, (Cr)^{a} \left(\frac{C}{q}\right)^{a-2}\right\} \le \left({Cr}\right)^{\min\{2, a\}}, 
\end{align*}
where we have used \eqref{sumS} and $a_1=0$ to obtain that 
$k\ge 2a$ and thus 
$$k-a-2 \ge 0.$$ If $k\ge 20$, it follows from $k-a-2 \ge (k+a)/4 + (k-16)/8$ and $q\gg  (\log n)^{4} \gtrsim k^{4}$ that  
%$$\frac{k-a-2}{k+a} \ge \frac{k-4}{3k} \ge \frac{3}{15}.$$
%Together with the facts $q\gg (\log n)^{16}$ and $2a\le k \le 4\log n$, it implies that
\begin{align*}
& \frac{(Cq)^{a+1}\cdot  (Ck)^k}{|z|^{k-a}\cdot  nq^{k-1}}\frac{(2rk)^a}{a!} {n} \le (Cr)^a \frac{(Ck)^{k+a}}{q^{k-a-2}}  \le (Cr)^a \left( \frac{1}{\sqrt{\log n}}\right)^{ k-16 }.
\end{align*}

\medskip
\textit{Case 2: $a_1=1$.} If $3\le k <20$, it holds that 
\begin{align*}
\frac{(Cq)^{a+1}\cdot  (Ck)^k}{ |z|^{k-a}\cdot  nq^{k-1}}   \frac{(2rk)^a}{a!} \sqrt{n} &\le (Cr)^{a} \left(\frac{C}{q}\right)^{k-a-1} \le \left({Cr}\right)^{a},
%&\le \min\left\{(Cr)^a , (Cr)^{a}\left(\frac{C}{q}\right)^{a-2}\right\} \le \left(Cr\right)^{\min\{2, a\}},
\end{align*}
where we have used \eqref{sumS} and $ a_1=1$ to obtain that 
$k\ge 2a-1$ and thus 
\[ k-a-1\ge \frac{k-3}{2} \ge 0. \]
If $k\ge 20$, an application of a similar argument as in Case 1 above yields that
$$\frac{(Cq)^{a+1}\cdot  (Ck)^k}{ |z|^{k-a}\cdot  nq^{k-1}} \frac{ (2rk)^a}{a!} \sqrt{n}  \le (Cr)^a \left( \frac{1}{\sqrt{\log n}}\right)^{ k-13 }.$$

\medskip
\textit{Case 3: $a_1\ge 2$.} It follows that 
\begin{equation*} 
    \frac{(Cq)^{a+1}\cdot  (Ck)^k}{|z|^{k-a}\cdot  nq^{k-1}}   \frac{(2rk)^a}{a!}\frac{(k-a_1)!}{k!} \le (Cr)^{a}  \left(\frac{Ck}{q}\right)^{k-a} \frac{k^a}{a!}\frac{k^a (k-a_1)!}{k!}.
\end{equation*}
When $k>2a$, we have that 
\begin{align*}
(Cr)^{a}  \left(\frac{Ck}{q}\right)^{k-a} \frac{k^a}{a!}\frac{k^a (k-a_1)!}{k!} \le (Cr)^{a}  \left(\frac{Ck^3}{q}\right)^{k-a}  . 
\end{align*}
When $k\le 2a$, we have that
\begin{align*}
 (Cr)^{a}  \left(\frac{Ck}{q}\right)^{k-a} \frac{k^a}{a!}\frac{k^a (k-a_1)!}{k!}  & \le (Cr)^{a}  \left(\frac{Ck}{q}\right)^{k-a} k^{a-a_1}   \le (Cr)^{a}  \left(\frac{Ck^2}{q}\right)^{k-a}   ,
\end{align*}
where we have utilized $ k-a\ge a-a_1$ by \eqref{sumS} in the second step above. 

% First, suppose $q\le n^{1/2-\epsilon}$, we have that
% $$\frac{(Cq)^{a+1}\cdot  (Ck)^k}{|z|^{k-a}\cdot  nq^{k-1}}  \cdot (2r)^a \le \frac{q^2}{n} (Crk)^{a}  \left(\frac{Ck}{q|z|^2}\right)^{k-a} \le 1,$$
% where we used that $k\ge a$ and $(\log n)^{C\log n}\ll n^{-2\e}$ for any constant $C>0$. Second, suppose $q>n^{1/2-\epsilon}$ and $k>a$. Then, we have that
% $$\frac{(Cq)^{a+1}\cdot  (Ck)^k}{|z|^{k-a}\cdot  nq^{k-1}}  \cdot (2r)^a \le (Crk)^{a}  \left(\frac{Ck}{q|z|^2}\right)^{k-a} \le (Crk)^{a} \cdot  \frac{Ck}{q|z|^2} \le 1.$$
% Finally, suppose $q>n^{1/2-\epsilon}$ and $k=a$. Then, 
% $$\frac{(Cq)^{a+1}\cdot  (Ck)^k}{|z|^{k-a}\cdot  nq^{k-1}}  \cdot {2r \choose a} \le (Ca)^a  {2r \choose a}  \le (Ca)^a  \frac{(2r)^a}{a!} \le (Cr)^a,$$
% where we used Stirling's approximation in the last step.

\medskip
Applying the estimates established for the three cases above to \eqref{boundGk} and summing over $a_1$ and $k$, we can deduce that 
\begin{align}
	\sum_{k=3}^{4r}\left|\mathfrak G_k\right|  & \le \sum_{k=3}^{19} \sum_{a=0}^{(2r-1)\wedge k} k(Cr)^a  \cdot \left(\E |Y|^{2r}\right)^{\frac{2r-1-a}{2r}} \nonumber\\
	&\quad+ \sum_{k=20}^{4r}\sum_{a=0}^{(2r-1)\wedge k}  \frac{k(Cr)^a}{(\sqrt{\log n})^{(k-a)\wedge (k-16)}} \cdot \left(\E |Y|^{2r}\right)^{\frac{2r-1-a}{2r}} \nonumber\\
	&\le  \sum_{a=0}^{2r-1} (Cr)^{a+1} \cdot \left(\E |Y|^{2r}\right)^{\frac{2r-1-a}{2r}}  .\label{finalG3}
\end{align}
Thus, it now remains to investigate the case of $k=2$ with 
\begin{align*}
	\mathfrak G_2 :=-\frac{|z|^2}{2n}  \sum_{i,j\in [n]} \mathcal \kappa_{3}(i,j)  (\Pii\bu)_i \mathbb E \frac{\partial^2  (G_{j\bv} Y^{r-1}\overline Y^{r}  )}{\partial (W_{ij})^2 } .
\end{align*}
Observe that under the notations in \eqref{dervWord}, if $l_1\ge 1$, it holds that 
\[ a=a_1=1 \ \text{ or } \ a=a_1=0. \] 
For either case, \eqref{abs_Gk} can be bounded similarly as above. We still need to consider the case of $l_1=0$. It follows from \eqref{Rimu1}--\eqref{Rimu3} and the Cauchy--Schwarz inequality that 
\begin{align*}
	&\, \sum_{i,j\in [n]}  \left|(\Pii\bu)_i\right|\left|G_{j\bv}\right| \prod_{s\in \cal S} \left|A_{i, j}(w_s)\right| \\
	  \le &\, \left[\mathbf 1(a_1=0) \sqrt{n}   + \mathbf 1(a_1\ge 1) \right]\left( \frac{C}{|z|}\right)^{a+4}
\end{align*}
with $(c_0,\xi)$-high probability. With the aid of the above estimate, we can obtain that 
\begin{align}\label{finalG2}
	\left|\mathfrak G_2\right|   \le  &  \sum_{a=0}^2 (Cr)^{a+1} \cdot \left(\E |Y|^{2r}\right)^{\frac{2r-1-a}{2r}}.
\end{align}

We now move on to bound the error term $\cal E$ given in \eqref{GE} for $l=4r$. Observe that 
\begin{align*}
\bG_{(ij)}^{x}=\left(\bW^{[ij]}+x\bDelta_{ij}-z\right)^{-1} & = \left[\bW- z + (x-\bW_{ij})\bDelta_{ij}\right]^{-1} \\
& = \left[\bG(z)^{-1} + (x-\bW_{ij})\bDelta_{ij}\right]^{-1}.
\end{align*}
Together with \eqref{bound_OP} of Lemma \ref{lem_opbound} and the bound $|x- W_{ij}|\lesssim q^{-1}$, it implies that 
\beq\label{supbdd}
\sup_{|x|\le C/q} \left\|\bG_{(ij)}^{x} \right\| \lesssim {|z|}^{-1}
\eeq
with $(c_0,\xi)$-high probability. Then an application of a similar argument as above yields that 
\begin{align*}
	R_{l+1}(i,j)\le  \frac{(Clr)^l}{nq^l} \frac{(q|z|^2)^{2r-1}}{|z|^{2r+l}}. 
\end{align*}
Hence, when $l=4r$, we can bound \eqref{GE} as
\beq\label{finalG5}
|\mathcal E|\le \frac{(Cr^2)^{4r}}{nq^{4r}} \frac{(q|z|^2)^{2r}}{|z|^{6r}} \frac{n^{3/2}}{|z|} \ll 1.
\eeq

Finally, we are ready to combine \eqref{G1_est1}, \eqref{finalG3}, \eqref{finalG2}, and \eqref{finalG5} to deduce that  
\begin{align*}
	\mathbb E|Y(z)|^{2r}   & \le   1+ \sum_{a=0}^{2r-1}  (Cr)^{a+1} \cdot  \left(\E |Y|^{2r}\right)^{\frac{2r-1-a}{2r}}.
\end{align*}
Then an application of Young's inequality to the terms inside the summation yields that 
\begin{align}
	\mathbb E|Y(z)|^{2r}   %& \le   1+ \sum_{a=0}^{2r-1}  (Cr)^{a+1} \cdot  \left(\E |Y|^{2r}\right)^{\frac{2r-1-a}{2r}} \nonumber\\
&	\le   1+ \sum_{a=0}^{2r-1} \left[\frac{a+1}{2r} 4^{2r-1-a} (Cr)^{2r} + \frac{2r-1-a}{2r}4^{-(a+1)} \E |Y|^{2r} \right] \nonumber\\
& \le \frac12 \mathbb E|Y(z)|^{2r} + (Cr)^{2r},\label{final_aniso}
\end{align}
which gives the desired conclusion \eqref{moment_aniso} in Claim \ref{claim_moment_aniso}. This completes the proof of Claim \ref{claim_moment_aniso} for the relatively dense case with $q\gg (\log n)^{16}$.

It now remains to examine the relatively sparse case with $(\log n)^{4}\ll q\le (\log n)^{20}$. The proof of the desired bound \eqref{moment_aniso} for such case is similar to that for the relatively dense case with $q\gg (\log n)^{16}$ considered above, except that we will need to use the entrywise local law in Theorem \ref{lem_locallaw1}, but not the averaged local law in Theorem \ref{lem_averlaw}. Specifically, to exploit Theorem \ref{lem_locallaw1}, let us choose 
\[ \xi=\wt C(\log n)^2\ll q^{1/2} 
\]
so that \eqref{bound_G} from Lemma \ref{lem_opbound} holds with $(c_0,\xi)$-high probability, and \eqref{entry_law} and \eqref{entry_law2} from Theorem \ref{lem_locallaw1} hold with $(c_1,\xi)$-high probability. Moreover, note that the proof of \eqref{moment_aniso} involves products of at most $6r+1$ many resolvent entries. Hence, we will choose a sufficiently large constant \smash{$\wt C>6/c_0 + 6/c_1$} so that \eqref{control_smallp} can be applied in the technical analysis. Then we repeat the arguments between \eqref{G1} and \eqref{G1_est1.0}, except that instead of Theorem \ref{lem_averlaw}, we can resort to Theorem \ref{lem_locallaw1} to obtain that for each $i\in [n]$, 
$$q|z|^2\cdot \sum_{j\in [n]} s_{ij}(G_{jj}-M_j) =\OO \left( \xi^{1/2}  + \frac{q\xi^2}{\sqrt{n}}\right) =\OO(r).$$
Thus, the representation given in  \eqref{G1_est1} becomes 
\beq\label{G1_est1_add}
	 \mathbb E|Y(z)|^{2r}  = \sum_{k=2}^l \mathfrak G_k+ \cal E  + \OO\left\{ r (\E |Y|^{2r})^{\frac{2r-1}{2r}}+r(\E |Y|^{2r})^{\frac{2r-2}{2r}} +e^{-c\xi}\right\}.
\eeq
In fact, the terms $\mathfrak G_k$ and $\cal E$ in (\ref{G1_est1_add}) above can be bounded similarly as before, which leads to \eqref{final_aniso} again and finally yields the desired bound \eqref{moment_aniso} for the sparse case with $(\log n)^{4}\ll q\le (\log n)^{20}$. This concludes the proof of Claim \ref{claim_moment_aniso}.

\end{document}

%% file: macro_fan.tex
\newcommand{\beq}{\begin{equation}}
\newcommand{\eeq}{\end{equation}}

\newcommand{\e}{{\varepsilon}}

\renewcommand{\cal}{\mathcal}

\newcommand{\ii}{\mathrm{i}} %\newcommand{\mi}{\mathrm{i}}
\newcommand{\dd}{\mathrm{d}}

\renewcommand{\le}{\leq}
\renewcommand{\ge}{\geq}

\renewcommand{\P}{\mathbb{P}}
\newcommand{\E}{\mathbb{E}}
\newcommand{\R}{\mathbb{R}}
\newcommand{\C}{\mathbb{C}}
\newcommand{\N}{\mathbb{N}}

\DeclareMathOperator{\re}{Re}
\DeclareMathOperator{\im}{Im}

\newcommand{\OO}{O}
\newcommand{\oo}{o}

\DeclareMathOperator{\sW}{\mathcal{W}}